%% file: main.tex
\documentclass[lettersize,journal]{IEEEtran}
\usepackage{amsmath,amsfonts,amssymb,amsopn,amsthm}
\usepackage{array}
\usepackage{subfig}
\usepackage{textcomp}
\usepackage{stfloats}
\usepackage{url}
\usepackage{verbatim}
\usepackage{graphicx}
\usepackage{cite}
\usepackage{tcolorbox}
\usepackage{enumerate}
\usepackage{prettyref}

\usepackage{xcolor}
\usepackage{adjustbox}
\usepackage{colortbl}
\usepackage{tabularx}
\usepackage{booktabs} 
\usepackage[linesnumbered, vlined, ruled]{algorithm2e}
\SetArgSty{textup}
\usepackage{color, colortbl}
\usepackage{braket}

\usepackage{bbding}
\usepackage{pifont}
\usepackage{wasysym}

\usepackage[hidelinks]{hyperref}

\newcommand\tabblue[1]{{\color[HTML]{1f77b4}{#1}}}
\newcommand\taborange[1]{{\color[HTML]{ff7f0e}{#1}}}
\newcommand\tabgreen[1]{{\color[HTML]{2ca02c}{#1}}}

\newcommand\tabpink[1]{{\color[HTML]{e377c2}{#1}}}

\DeclareMathOperator*{\argmax}{arg\,max}
\DeclareMathOperator*{\argmin}{arg\,min}
\newcommand{\pref}{\prettyref}

\newtheorem{remark}{Remark}

\newrefformat{fig}{Fig.~\ref{#1}}
\newrefformat{tab}{Table~\ref{#1}}
\newrefformat{sec}{Section~\ref{#1}}
\newrefformat{app}{Appendix~\ref{#1}}
\newrefformat{alg}{Algorithm~\ref{#1}}
\newrefformat{property}{Property~\ref{#1}}
\newrefformat{theorem}{Theorem~\ref{#1}}
\newrefformat{corollary}{Corollary~\ref{#1}}
\newrefformat{proposition}{Proposition~\ref{#1}}
\newrefformat{def}{Definition~\ref{#1}}
\newrefformat{eq}{equation~(\ref{#1})}

\definecolor{mygray}{RGB}{220,220,220}

\begin{document}

\title{Quality Indicators for Preference-based Evolutionary Multi-objective Optimization Using a Reference Point: A Review and Analysis}


\author{
Ryoji~Tanabe,~\IEEEmembership{Member,~IEEE}~and~Ke~Li,~\IEEEmembership{Senior Member,~IEEE}
\thanks{R. Tanabe is with Faculty of Environment and Information Sciences, Yokohama National University, Yokohama, Japan. (e-mail: rt.ryoji.tanabe@gmail.com).}
\thanks{K. Li is with Department of Computer Science, University of Exeter, Exeter, EX4 4QF, UK (e-mail: k.li@exeter.ac.uk).}
}



\maketitle

\begin{abstract}
    Some quality indicators have been proposed for benchmarking preference-based evolutionary multi-objective optimization algorithms using a reference point. Although a systematic review and analysis of the quality indicators are helpful for both benchmarking and practical decision-making, neither has been conducted.
    In this context, first, this paper reviews existing regions of interest and quality indicators for preference-based evolutionary multi-objective optimization using the reference point.
    We point out that each quality indicator was designed for a different region of interest.
    Then, this paper investigates the properties of the quality indicators.
    We demonstrate that an achievement scalarizing function value is not always consistent with the distance from a solution to the reference point in the objective space.
    We observe that the regions of interest can be significantly different depending on the position of the reference point and the shape of the Pareto front. 
    We identify undesirable properties of some quality indicators.
    We also show that the ranking of preference-based evolutionary multi-objective optimization algorithms depends on the choice of quality indicators.
   
\end{abstract}


\input{introduction}

\input{preliminary}

\input{roi}

\input{review}

\input{setup}
\input{results}

\input{conclusion}

\section*{Acknowledgment}

Tanabe was supported by JSPS KAKENHI Grant Number 21K17824 and LEADER, MEXT, Japan.
Li was supported by UKRI Future Leaders Fellowship (MR/S017062/1, MR/X011135/1), NSFC (62076056), EPSRC (2404317), Royal Society (IES/R2/212077) and Amazon Research Award.

\bibliography{reference}
\bibliographystyle{IEEEtran}

\vfill

\begin{IEEEbiography}[{\includegraphics[width=1in,height=1.25in,keepaspectratio]{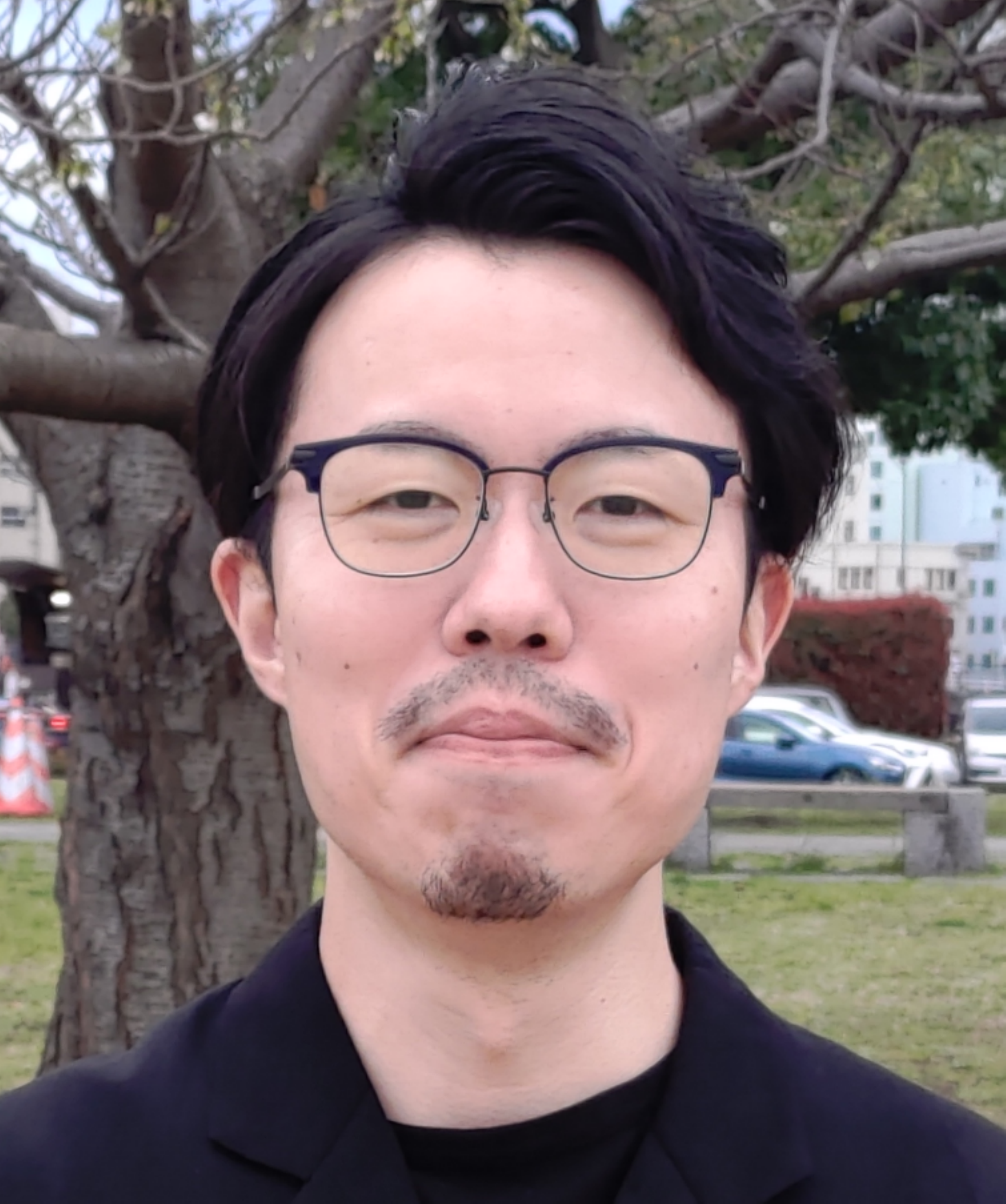}}]{Ryoji Tanabe}
is an Assistant Professor at Yokohama National University, Japan (2019--).
He received the Ph.D. degree in Science from The University of Tokyo, Japan, in 2016.
His research interests include single- and multi-objective black-box optimization, benchmarking, and automatic algorithm configuration and selection.
  \end{IEEEbiography}

\begin{IEEEbiography}[{\includegraphics[width=1in,height=1.25in,clip,keepaspectratio]{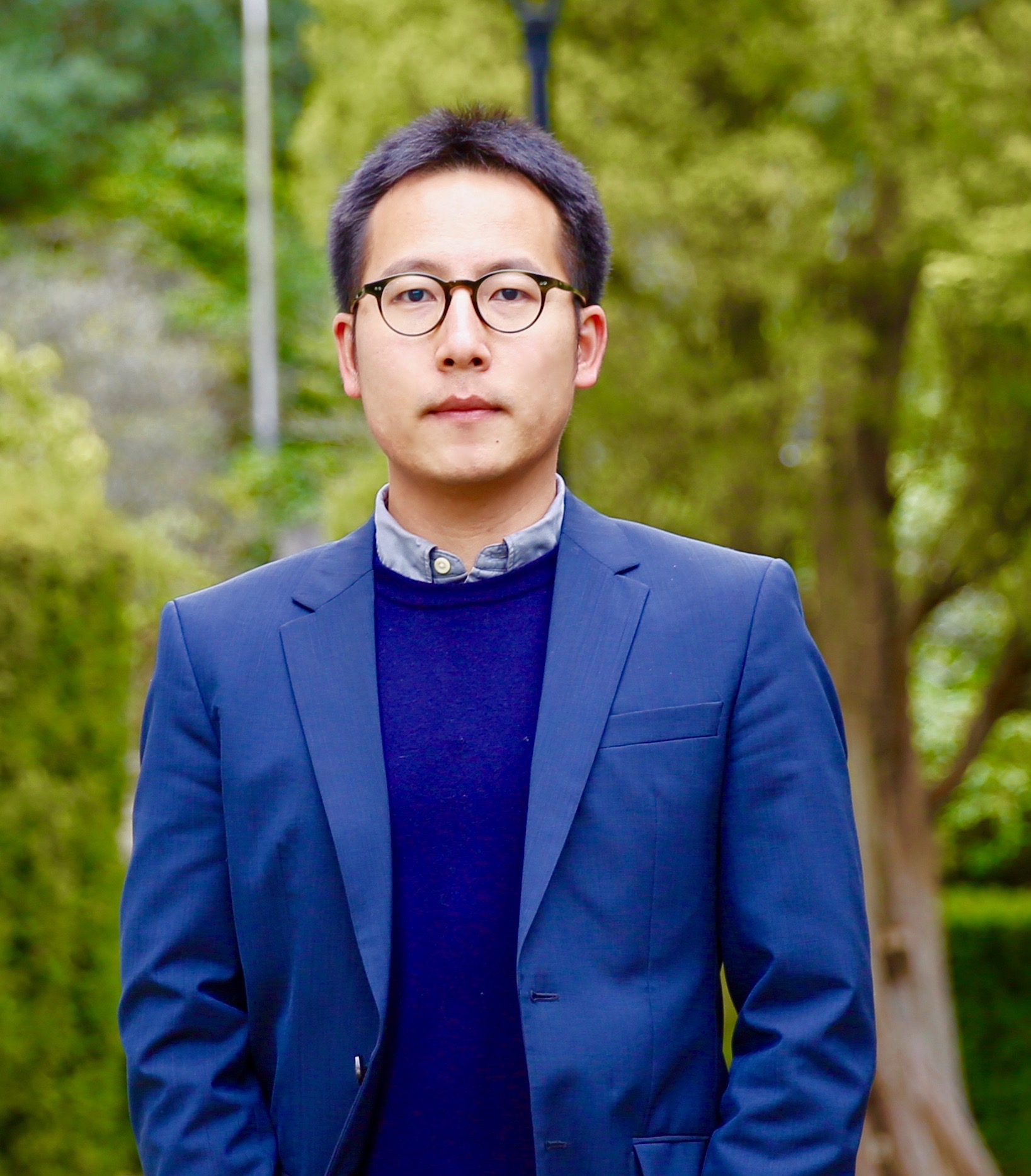}}]{Ke Li} (M'17) received the B.Sc. and M.Sc. in computer science and technology from Xiangtan University, China, in 2007 and 2010, respectively, and the Ph.D. in computer science from City University of Hong Kong, Hong Kong, in 2015. His current research interests include the evolutionary multi-objective optimization, multi-criterion decision-making, machine learning and applications in search-based software engineering, scientific discovery, and healthcare.

He has been founding chair of IEEE Computational Intelligence Society Task Force on Decomposition-based Techniques in EC with the EC Technical Committee. He is currently serving as an Associate Editor for \textsc{IEEE Transactions on Evolutionary Computation} and five other academic journals.

\end{IEEEbiography}

\clearpage

\begin{figure*}
\centering
\fontsize{30pt}{100pt}\selectfont{\textbf{Supplement}}
\end{figure*}


\setcounter{figure}{0}
\setcounter{table}{0}

\input{supplement.tex}

\end{document}

%% file: introduction.tex

\section{Introduction}
\label{sec:introduction}

Partially due to the population-based property, evolutionary algorithms (EAs) have been widely recognized as an effective approach for multi-objective optimization, as known as evolutionary multi-objective optimization (EMO). Conventional EMO algorithms, such as NSGA-II~\cite{DebAPM02}, IBEA~\cite{ZitzlerK04}, and MOEA/D~\cite{ZhangL07}, are designed to search for a set of trade-off alternatives that approximate the Pareto-optimal front (PF) without considering any preference information~\cite{Deb01}. Thereafter, this solution set is handed over to the decision maker (DM) for an \textit{a posteriori} multi-criterion decision-making (MCDM) to choose the solution(s) of interest (SOI). On the other hand, if the DM's preference information is available \textit{a priori}, it can be used to navigate an EMO algorithm, also known as preference-based EMO (PBEMO) algorithm~\cite{PurshouseDMMW14,CoelloCoello00,BechikhKSG15}, to search for a set of \lq\lq preferred\rq\rq \ trade-off solutions lying in a region of interest (ROI), i.e., a subregion of the PF specified according to the DM's preference information~\cite{RuizSL15}. From the perspective of EMO, approximating an ROI can be easier than approximating the complete PF, especially when having many objectives.
From the perspective of MCDM, the property of PBEMO is beneficial for the DM.
Only a few number of potentially preferred solutions are shown to the DM when using a PBEMO algorithm.
This can reduce the DM's cognitive load.



The use of a reference point, also known as an aspiration level vector~\cite{Wierzbicki80}, which consists of desirable objective values specified by the DM, is one of the most popular approaches for expressing the preference information in the EMO literature~\cite{LiLDMY20,AfsarMR21}.
Note that this paper does not consider a reservation point~\cite{MiettinenM02}, which consists of objective values to be achieved at least.
Many conventional EMO algorithms have been extended to PBEMO using a reference point, such as R-NSGA-II~\cite{DebSBC06}, PBEA~\cite{ThieleMKL09}, and MOEA/D-NUMS~\cite{LiCMY18}.



Quality indicators play a vital role in quantitatively benchmarking EMO algorithms for approximating the whole PF~\cite{ZitzlerTLFF03,LiY19}. 
Since an indicator-based EMO algorithm aims to optimize a quality indicator value of the population~\cite{Falcon-CardonaC20}, studying quality indicators is also helpful for algorithm development.
Representative quality indicators are the hypervolume (HV)~\cite{ZitzlerT98}, the additive $\epsilon$-indicator ($I_{\epsilon+}$)~\cite{ZitzlerTLFF03}, and the inverted generational distance (IGD)~\cite{CoelloS04}. It is worth noting that none of these quality indicators take any preference information into account in quality assessment. Thus, they are not suitable for evaluating the performance of PBEMO algorithms for approximating the ROI(s). In fact, quality assessment on PBEMO algorithms have not received significant attention in the EMO community until~\cite{WickramasingheCL10}. Early studies mainly relied on visual comparisons which are neither reliable nor scalable to many objectives~\cite{DebSBC06,MolinaSHCC09}. On the other hand, some studies around 2010 (e.g., \cite{SaidBG10,MohammadiOL12}) directly applied conventional quality indicators and thus are likely to lead to some misleading conclusions~\cite{LiDY18}. To the best of our knowledge, the first quality indicator for PBEMO was proposed in~\cite{WickramasingheCL10}. Although it has several technical flaws, this quality indicator had a significant impact on the quality assessment for PBEMO as discussed in~\cite{LiDY18} and~\cite{MohammadiOL13}.

\noindent \textbf{Motivation for a review.}
Although there have been a number of preference-based quality indicators proposed since~\cite{WickramasingheCL10}, there is no systematic survey along this line of research. Some survey papers on quality indicators~\cite{ZitzlerTLFF03,LiY19} are available, but they are hardly about preference-based ones.
Afsar et al.~\cite{AfsarMR21} conducted a survey on how to evaluate the performance of interactive preference-based multi-objective optimizers, but they focused on experimental conditions rather than quality indicators.
This is because no quality indicator had been specifically developed for interactive preference-based multi-objective optimization as of 2021.
Bechikh et al. \cite{BechikhKSG15} presented an exhaustive review of PBEMO algorithms yet on quality indicators. In addition, some previous studies \textit{implicitly} proposed quality indicators. For example, Ruiz et al. \cite{RuizSL15} proposed WASF-GA. In \cite{RuizSL15}, they also designed a new quality indicator called HV$_{\mathbf{z}}$ to evaluate the performance of WASF-GA. However, they did not clearly state that the design of HV$_{\mathbf{z}}$ was their contribution. For this reason, most previous studies on preference-based quality indicators (e.g.,~\cite{LiDY18,HouYZZYR18,BandaruS19}) overlooked HV$_{\mathbf{z}}$.

\noindent \textbf{Motivation for analysis.}
The properties of quality indicators are not obvious, including which point set a quality indicator prefers and which quality indicators are consistent with each other.
Thus, it is likely to incorrectly evaluate the performance of EMO algorithms when using a particular quality indicator.
To address this issue, some previous studies analyzed quality indicators \cite{TanabeI20b}.
%
Nevertheless, little is known about the properties of  quality indicators for PBEMO.
Although some previous studies (e.g., \cite{YuZL15,BandaruS19}) analyzed a few quality indicators for PBEMO, the scale of their experiments is relatively small.


Apart from this issue of quality indicators, Li et. al. \cite{LiLDMY20} reported the pathological behavior of some PBEMO algorithms when setting the reference point far from the PF.
They showed that R-NSGA-II \cite{DebSBC06}, r-NSGA-II \cite{SaidBG10}, and R-MEAD2 \cite{MohammadiOLD14} unexpectedly obtain points on the edge of the PF, which are far from the reference point.
They also showed that only MOEA/D-NUMS \cite{LiCMY18} works expectedly even in this case.
Since the DM does not know any information about the PF in real-world applications, these undesirable behavior can be observed in practice.
However, the previous study \cite{LiLDMY20} could not determine what caused these undesirable behavior.

Note that the true ideal and nadir points are known on some real-world problems~\cite{HeITWNS21}.
Well-approximated ideal and nadir points can possibly be available a priori, e.g., based on the DM's experience.
If so, the DM can set the reference point near the PF, and the above-discussed issue does not occur.
However, the true (or well-approximated) ideal and nadir points are not always available a priori~\cite{HeITWNS21}.
In any case, it is important to understand the impact on the behavior of PBEMO algorithms when setting the reference point incorrectly.



\noindent \textbf{Contributions.}
Motivated by the above discussion, first, we review ROIs and preference-based quality indicators. We clarify the quality indicators based on their target ROIs.
We believe that the quality indicators can be used for the performance assessment of non-evolutionary algorithms, including Bayesian optimization and numerical PF approximation.
Then, we analyze the quality indicators. Through an analysis, we address the following four research questions:
\begin{enumerate}[RQ1:]
    \item Does a Pareto-optimal point with the minimum achievement scalarizing function (ASF) value always minimize the distance from the reference point?
    \item What are the differences of the definitions of ROIs considered in previous studies? How do these differences influence the behavior of EMO algorithms?
    \item What are the properties of existing quality indicators for PBEMO?
    \item How does the choice of quality indicator affect the ranking of PBEMO algorithms?
\end{enumerate}










\noindent \textbf{Outline.}
The rest of this paper is organized as follows. \pref{sec:preliminaries} provides some preliminary knowledge pertinent to this paper. \pref{sec:review_roi} reviews and analyzes three ROIs considered in previous studies. \pref{sec:review_qi} reviews 14 preference-based quality indicators developed in the literature. Our experimental settings are provided in~\pref{sec:setting} while the results are analyzed in~\pref{sec:results}. \pref{sec:conclusion} concludes this paper.

\noindent \textbf{Supplementary file.}
This paper has a supplementary file. Figure S.$*$ and Table S.$*$ indicate a figure and a table in the supplementary file, respectively.

\noindent \textbf{Code availability.}
The Python implementation of all preference-based quality indicators investigated in this work is available at \href{https://github.com/ryojitanabe/prefqi}{https://github.com/ryojitanabe/prefqi}.

%% file: preliminary.tex

\section{Preliminaries}
\label{sec:preliminaries}

\subsection{Multi-objective optimization}
\label{sec:def_MOPs}


%
Multi-objective optimization aims to find an $n$-dimensional solution $\mathbf{x} \in \Omega$ that simultaneously minimizes $m$ objective functions $\mathbf{F} = (f_1, \ldots, f_m)^{\top}: \Omega \rightarrow \mathbb{R}^M$, where $\Omega$ is the feasible set in the decision space $\mathbb{R}^n$.
 A solution $\mathbf{x}_1$ is said to \underline{Pareto dominate} $\mathbf{x}_2$ if and only if $f_i(\mathbf{x}_1)\leq f_i(\mathbf{x}_2)$ for all $i\in\{1,\ldots,m\}$ and $f_i(\mathbf{x}_1)<f_i(\mathbf{x}_2)$ for at least one index $i$. We denote $\mathbf{x}_1\prec\mathbf{x}_2$ when $\mathbf{x}_1$ dominates $\mathbf{x}_2$. In addition, $\mathbf{x}_1$ is said to \underline{weakly Pareto dominate} $\mathbf{x}_2$ if $f_i(\mathbf{x}_1)\leq f_i(\mathbf{x}_2)$ for all $i \in \{1,\ldots,m\}$. A solution $\mathbf{x}^\ast$ is a Pareto-optimal solution if $\mathbf{x}^\ast$ is not dominated by any solution in $\Omega$. The set of all Pareto-optimal solutions in $\Omega$ is called the \underline{Pareto-optimal set} (PS) $\mathcal{X}^\ast=\{\mathbf{x}^\ast\in\Omega\,|\, \nexists\mathbf{x}\in\Omega\: \mathrm{s.t.} \: \mathbf{x}\prec\mathbf{x}^\ast\}$.
The image of the PS in $\mathbb{R}^m$ is also called the PF $\mathcal{F} = \mathbf{F}(\mathcal{X}^\ast)$. The \underline{ideal point} $\mathbf{p}^{\mathrm{ideal}} \in \mathbb{R}^m$ consists of the minimum values of the PF for $m$ objective functions. The \underline{nadir point} $\mathbf{p}^{\mathrm{nadir}}\in\mathbb{R}^m$ consists of the maximum values of the PF for $m$ objective functions. 
For the sake of simplicity, we refer $\mathbf{F}(\mathbf{x})$ as a point $\mathbf{p}=(p_1,\ldots,p_m)^\top\in\mathbb{R}^m$ in the rest of this paper.

Apart from the DM, an analyst is a person or a computer system responsible for the process of finding a solution set~\cite{Miettinen98}.
For example, the analyst selects a suitable EMO algorithm for the problem.

\subsection{Quality indicators}
\label{sec:q_indicators}

A quality indicator is a metric $I: \mathbb{R}^m\rightarrow\mathbb{R}$, $I:\mathcal{P}\mapsto I(\mathcal{P})$ that quantitatively evaluates the quality of a point set $\mathcal{P} = \{\mathbf{p}^i\}_{i=1}^\mu$ of size $\mu$ in terms of at least one of the following four aspects~\cite{LiY19}: $i$) \underline{convergence}: the closeness of the points in $\mathcal{P}$ to the PF; $ii$) \underline{uniformity}: the distribution of the points in $\mathcal{P}$; $iii$) \underline{spread}: the range of the points in $\mathcal{P}$ along the PF; and $iv$) \underline{cardinality}: the number of non-dominated points in $\mathcal{P}$.
As discussed in~\cite{KnowlesTZ06}, a quality indicator $I$ is said to be \underline{Pareto-compliant} if $I(\mathcal{P}^1)<I(\mathcal{P}^2)$ for any pair of point sets $\mathcal{P}^1$ and $\mathcal{P}^2$ in $\mathbb{R}^m$, where $\exists\mathbf{p}\in\mathcal{P}^1$, $\forall\tilde{\mathbf{p}}\in\mathcal{P}^2$ we have $\mathbf{p}\prec\tilde{\mathbf{p}}$.
Note that we assume the quality indicator is to be minimized. Otherwise, we have $I(\mathcal{P}^1)>I(\mathcal{P}^2)$ instead.


%
%


A unary indicator maps a point set $\mathcal{P}$ to a scalar value $\mathbb{R}$.
In contrast, a $K$-ary indicator requires $K$ point sets $\mathcal{P}^1, \ldots, \mathcal{P}^K$ and 
 evaluates the quality of the $K$ point sets relatively.
Both unary and $K$-ary quality indicators have pros and cons~\cite{ZitzlerTLFF03,LiY19}.
For example, $K$-ary quality indicators generally do not require information about the PF.
This is attractive for real-world problems with unknown PFs.
However, $K$-ary quality indicators only provide information about the relative quality of the $K$ point sets.

Below, we describe two representative quality indicators widely used in the EMO community.

\subsubsection{Hypervolume (HV)~\cite{ZitzlerT98}}

It measures the volume of the region dominated by the points in $\mathcal{P}$ and bounded by the HV-reference point $\mathbf{y}\in\mathbb{R}^m$:
\begin{equation}
    \mathrm{HV}(\mathcal{P})=\Lambda\Biggl(\bigcup_{\mathbf{p}\in\mathcal{P}}\{\mathbf{q}\in\mathbb{R}^m \, | \, \mathbf{p}\prec\mathbf{q}\prec\mathbf{y}\} \Biggr),
    \label{eq:hv}
\end{equation}
where $\Lambda(\cdot)$ in \eqref{eq:hv} is the Lebesgue measure. $\mathrm{HV}(\mathcal{P})$ can evaluate the quality of $\mathcal{P}$ in terms of both convergence and diversity. HV is to be maximized.

\subsubsection{Inverted generational distance (IGD)~\cite{CoelloS04}}
\label{subsubsec:igd}

Let $\mathcal{S}$ be a set of IGD-reference points uniformly distributed on the PF, 
IGD measures the average distance between each IGD-reference point $\mathbf{s}\in\mathcal{S}$ and its nearest point $\mathbf{p}\in\mathcal{P}$:
\begin{equation}
    {\rm IGD} (\mathcal{P}) = \frac{1}{|\mathcal{S}|}\left(\sum_{\mathbf{s}\in\mathcal{S}}\min_{\mathbf{p}\in\mathcal{P}}\Bigl\{\mathrm{dist}(\mathbf{s}, \mathbf{p}) \Bigr\} \right),
    \label{eq:igd}
\end{equation}
where $\mathrm{dist}(\cdot,\cdot)$ returns the Euclidean distance between two inputs.
IGD in \eqref{eq:igd} is to be minimized.
In general, IGD-reference points in $\mathcal{S}$ are uniformly distributed on the PF.
Like HV, IGD can also measure the convergence and diversity of $\mathcal{P}$ while it prefers a uniform distribution of points~\cite{TanabeI20b}.

\begin{remark}
    The term \underline{reference point} has been used in various contexts in the EMO literature.
    To avoid confusion, we use the term \underline{HV-reference point} to indicate the reference point for HV.
    Similarly, we use the term \underline{IGD-reference point set} to indicate a reference point set for IGD.
\end{remark}

\begin{remark}
	Note that Pareto-compliant is an important, yet hardly met, characteristic of a quality indicator. 
    Since HV is Pareto-compliant, HV has been one of the most popular quality indicators.
    A previous study~\cite{Falcon-CardonaE22} showed that a new Pareto-compliant indicator can be designed by combining existing quality indicators and HV.
\end{remark}


\subsection{Achievement scalarizing function}
\label{sec:asf}

Wierzbicki \cite{Wierzbicki80} proposed the ASF $s: \mathbb{R}^m\rightarrow\mathbb{R},\mathbf{p}\mapsto s(\mathbf{p})$ in the context of MCDM.
Although a number of scalarizing functions have been proposed for preference-based multi-objective optimization \cite{MiettinenM02}, the ASF is one of the most popular scalarizing functions.
Most previous studies on PBEMO (e.g., \cite{ThieleMKL09,RuizSL15,LiDY18}) used the following version of the ASF:
\begin{align}
    \label{eqn:asf1}
    s(\mathbf{p}) = \max_{i\in\{1,\ldots,m\}} w_i (p_i-z_i),
\end{align}
where $\mathbf{z}\in\mathbb{R}^m$ is the reference point specified by the DM.
In \eqref{eqn:asf1}, $\mathbf{w}\in\mathbb{R}^m$ is the weight vector that represents the relative importance of each objective function, where $\sum^m_{i=1} w_i = 1$ and $w_i \geq 0$ for any $i$.
Like in most previous studies, we set $\mathbf{w}$ to $(1/m,\ldots,1/m)^\top$ throughout this paper.
The ASF is order-preserving in terms of the Pareto dominance relation \cite{Wierzbicki80}, i.e., $s(\mathbf{p}^1)<s(\mathbf{p}^2)$ if $\mathbf{p}^1\prec\mathbf{p}^2$.
A point with the minimum ASF value is also weakly Pareto optimal with respect to $\mathbf{z}$ and $\mathbf{w}$.

Only the Pareto-optimal point with respect to $\mathbf{z}$ and $\mathbf{w}$ can be obtained by minimizing the following augmented version of the ASF (AASF) \cite{MiettinenM02}:
\begin{equation}
    \label{eqn:aasf}
    s^{\mathrm{aug}}(\mathbf{p})=s(\mathbf{p})+\rho\sum_{i=1}^m(p_i-z_i),
\end{equation}
%
where $\rho$ is a small positive value, e.g., $\rho=10^{-6}$~\cite{DebK07}.

\subsection{PBEMO algorithms}
\label{sec:pemo_ref}

To be self-contained, we give a briefing of six representative PBEMO algorithms considered in our experiments: R-NSGA-II~\cite{DebSBC06}, r-NSGA-II~\cite{SaidBG10}, g-NSGA-II~\cite{MolinaSHCC09}, PBEA~\cite{ThieleMKL09}, R-MEAD2~\cite{MohammadiOLD14}, and MOEA/D-NUMS~\cite{LiCMY18}.
Their behavior was also investigated in~\cite{LiLDMY20}.
As their names suggest, R-NSGA-II, r-NSGA-II, and g-NSGA-II are extended versions of NSGA-II.
PBEA is a variant of IBEA while RMEAD2 and MOEA/D-NUMS are scalarizing function-based approaches.
%
Although R-NSGA-II, r-NSGA-II, and PBEA can handle multiple reference points, we only introduce the case when using a single reference point $\mathbf{z}$.
As in \cite{LiLDMY20}, we focus on preference-based multi-objective optimization using a single reference point as the first step.
Below, we use the terms ``point set'' and ``population'' synonymously.
%

\subsubsection{R-NSGA-II}
\label{sec:rnsga2}

As in NSGA-II, the primary criterion in environmental selection in R-NSGA-II is based on the non-domination level of each point $\mathbf{p}$.
While the secondary criterion in NSGA-II is based on the crowding distance, that of R-NSGA-II is based on the following weighted distance to $\mathbf{z}$:
\begin{equation}
    \label{eqn:wdist}
    d^{\mathrm{R}}(\mathbf{p})=\sqrt{\sum^{m}_{i=1} w_i\left(\frac{p_i-z_i}{p^{\mathrm{max}}_i-p^{\mathrm{min}}_i}\right)^2},
\end{equation}
where $p^{\mathrm{max}}_i$  and $p^{\mathrm{min}}_i$ are the maximum and minimum values of the $i$-th objective function $f_i$ in the population $\mathcal{P}$. 
The weight vector $\mathbf{w}$ in \eqref{eqn:wdist} plays a similar role in $\mathbf{w}$ in the ASF. 
When comparing individuals in the same non-domination level, ties are broken by their $d^{\mathrm{R}}$ values. 
Thus, non-dominated individuals close to $\mathbf{z}$ are likely to survive to the next iteration.

In addition, R-NSGA-II performs $\epsilon$-clearing to maintain the diversity in the population.
If the distance between two individuals in the objective space is less than $\epsilon$, a randomly selected one is removed from the population.

\subsubsection{r-NSGA-II}
\label{sec:rnsga2}

It is an extended version of NSGA-II by replacing the Pareto dominance relation with the r-dominance relation.
For two points $\mathbf{p}^1$ and $\mathbf{p}^2$ in $\mathcal{P}$, $\mathbf{p}^1$ is said to r-dominate $\mathbf{p}^2$ if one of the following two criteria is met: 1) $\mathbf{p}^1 \prec \mathbf{p}^2$; 2) $\mathbf{p}^1 \nprec \mathbf{p}^2$, $\mathbf{p}^1 \nsucc \mathbf{p}^2$, and $d^{\mathrm{r}}(\mathbf{p}^1, \mathbf{p}^2) < -\delta$.
Here, $d^{\mathrm{r}}(\mathbf{p}^1, \mathbf{p}^2)$ is defined as follows:
\begin{equation}
	\label{eqn:rdominance_d}
  	d^{\mathrm{r}} (\mathbf{p}^1, \mathbf{p}^2) = \frac{d^{\mathrm{R}}(\mathbf{p}^1) - d^{\mathrm{R}}(\mathbf{p}^2)}{\max_{\mathbf{p} \in \mathcal{P}} \{d^{\mathrm{R}}(\mathbf{p})\} - \min_{\mathbf{p} \in \mathcal{P}} \{d^{\mathrm{R}}(\mathbf{p})\}},
\end{equation}
%
where the definition of $d^{\mathrm{R}}$ in \eqref{eqn:rdominance_d} can be found in \eqref{eqn:wdist}.
The threshold $\delta \in [0, 1]$ determines the spread of individuals in the objective space.
When $\delta = 1$, the r-dominance relation is the same as the Pareto dominance relation.
When $\delta = 0$, the r-dominance relation between two non-dominated points $\mathbf{p}^1$ and $\mathbf{p}^2$ is determined by their $d^{\mathrm{R}}$ values.

\subsubsection{g-NSGA-II}
\label{sec:gnsga2}

It uses the g-dominance relation instead of the Pareto dominance relation.
Let $\mathcal{Q}$ be the set of all points in $\mathbb{R}^m$ that dominate the reference point $\mathbf{z}$ or are dominated by  $\mathbf{z}$, i.e., $\mathcal{Q}=\{\mathbf{p}\in\mathbb{R}^m \, | \, \mathbf{p} \prec \mathbf{z} \: \text{or} \: \mathbf{p} \succ \mathbf{z}\}$.
A point $\mathbf{p}^1$ is said to g-dominate $\mathbf{p}^2$ if one of the following three criteria is met: 1) $\mathbf{p}^1\in\mathcal{Q}$ and $\mathbf{p}^2\notin\mathcal{Q}$; 2) $\mathbf{p}^1, \mathbf{p}^2\in\mathcal{Q}$ and $\mathbf{p}^1\prec\mathbf{p}^2$; 3) $\mathbf{p}^1, \mathbf{p}^2 \notin\mathcal{Q}$ and $\mathbf{p}^1\prec\mathbf{p}^2$.

Unlike other PBEMO algorithms, g-NSGA-II does not have a control parameter that adjusts the size of a region $\mathcal{Q}$ to be approximated. 
However, g-NSGA-II can obtain only points in a very small region when $\mathbf{z}$ is close to the PF  \cite{SaidBG10}.
In contrast, g-NSGA-II is equivalent to NSGA-II when $\mathbf{z}$ is very far from the PF \cite{LiLDMY20}.
This is because $\mathcal{Q}$ covers the PF when $\mathbf{z}$ dominates the ideal point or is dominated by the nadir point.

\subsubsection{PBEA}
\label{sec:pbea}

It is a variant of IBEA using the binary additive $\epsilon$-indicator ($I_{\epsilon +}$)~\cite{ZitzlerTLFF03}.
For a point set $\mathcal{P}$, the $I_{\epsilon +}$ value of a point $\mathbf{p}\in\mathcal{P}$ to another point $\mathbf{q}\in\mathcal{P}\setminus\{\mathbf{p}\}$ is defined as:
\begin{equation}
	\label{eqn:i_epsilon}
	I_{\epsilon +}(\mathbf{p},\mathbf{q})=\max_{i\in\{1,\ldots,m\}}\{p^\prime_i-q^\prime_i\},
\end{equation}
where $\mathbf{p}^\prime$ and $\mathbf{q}^\prime$ in \eqref{eqn:i_epsilon} are normalized versions of $\mathbf{p}$ and $\mathbf{q}$ based on the maximum and minimum values of $\mathcal{P}$. The $I_{\epsilon +}$ value is the minimum objective value such that $\mathbf{p}^\prime$ dominates $\mathbf{q}^\prime$. PBEA uses the following preference-based indicator $I_{p}$, which takes into account the AASF value in \eqref{eqn:aasf}:
\begin{align}
	\label{eqn:i_pbea}
  	I_{p}(\mathbf{p},\mathbf{q})&=\frac{I_{\epsilon +}(\mathbf{p},\mathbf{q})}{s^\prime(\mathbf{p})},\\
  	\label{eqn:i_pbea2}  
	s^\prime(\mathbf{p})&=s^{\mathrm{aug}}(\mathbf{p})+\delta-\min_{\mathbf{u}\in\mathcal{P}} \{s^{\mathrm{aug}}(\mathbf{u})\},
\end{align}
where the previous study \cite{ThieleMKL09} used $s^{\mathrm{aug}}$ with $s$ in \eqref{eqn:asf1}.
In \eqref{eqn:i_pbea2}, $ s'(\mathbf{p})$ is the normalized AASF value of $\mathbf{p}$ by the minimum AASF value of $\mathcal{P}$.
In \eqref{eqn:i_pbea2}, $\delta$ controls the extent of the preferred region.
A large $I_{p}$ value indicates that the corresponding $\mathbf{p}$ is preferred.
As acknowledged in \cite{ThieleMKL09}, one drawback of PBEA is the difficulty in determining the $\delta$ value.

\subsubsection{R-MEAD2}
\label{sec:rmead2}

R-MEAD2 \cite{MohammadiOLD14} is a decomposition-based EMO algorithm using a set of $\mu$ weight vectors $\mathcal{W}=\{\mathbf{w}^i\}_{i=1}^\mu$.
Similar to MOEA/D \cite{ZhangL07}, R-MEAD2 aims to approximate $\mu$ Pareto optimal points by simultaneously minimizing $\mu$ scalar optimization problems with $\mathcal{W}$.
R-MEAD2 adaptively adjusts $\mathcal{W}$ so that the corresponding individuals move toward $\mathbf{z}$.
At the beginning of the search, R-MEAD2 initializes $\mathcal{W}$ randomly.
For each iteration, R-MEAD2 selects the weight vector $\mathbf{w}^{\mathrm{c}}$ from $\mathcal{W}$, where the corresponding point $\mathbf{p}^{\mathrm{c}}$ is closest to the reference point $\mathbf{z}$, i.e., $\mathbf{p}^{\mathrm{c}}=\underset{\mathbf{p}\in\mathcal{P}}{\argmin}\left\{\mathrm{dist}(\mathbf{p}, \mathbf{z})\right\}$.
Then, R-MEAD2 randomly reinitializes $\mathcal{W}$ in an $m$-dimensional hypersphere of radius $r$ centered at $\mathbf{w}^{\mathrm{c}}$.

\subsubsection{MOEA/D-NUMS}
\label{sec:nums}

It is featured by a nonuniform mapping scheme (NUMS) that shifts $\mu$ uniformly distributed weight vectors toward the reference point $\mathbf{z}$.
In particular, the distribution of the $\mu$ shifted weight vectors, denoted as $\mathcal{W}^\prime$, is expected to be biased toward $\mathbf{z}$.
In NUMS, a parameter $r$ controls the extent of $\mathcal{W}^\prime$.
In contrast to R-MEAD2, NUMS adjusts the weight vectors in an offline manner.
In theory, NUMS can be incorporated into any decomposition-based EMO algorithm by using $\mathcal{W}^\prime$ instead of the original $\mathcal{W}$, while MOEA/D-NUMS proposed in~\cite{LiCMY18} is built upon the vanilla MOEA/D.
In addition, MOEA/D-NUMS uses the AASF in \eqref{eqn:aasf} instead of a general scalarizing function. 

%% file: roi.tex

\section{Review of region of interests}
\label{sec:review_roi}

%
PBEMO algorithms are designed to search for a set of $\mu$ non-dominated points that approximate the ROI.
%
However, as pointed out in \cite{LiCMY18}, the ROI has been loosely defined in the EMO community.
According to the definition in~\cite{LiCMY18}, we define the ROI as a subset of the PF, denoted as $\mathcal{R}\subseteq\mathcal{F}$.
We assume that the DM is interested in not only the closest Pareto-optimal point $\mathbf{p}^{\mathrm{c}^\ast}$ to the reference point $\mathbf{z}$ but also a set of Pareto-optimal points around $\mathbf{p}^{\mathrm{c}^\ast}$. 
In some cases, the extent of $\mathcal{R}$ is defined by a parameter given by the analyst.

Below, \pref{sec:def_roi} describes three ROIs addressed in previous studies.
The reference point $\mathbf{z}$ is said to be feasible if it cannot dominate any Pareto-optimal point.
Otherwise, it is said to be infeasible if $\mathbf{z}$ can dominate at least one Pareto-optimal point.
Then, \pref{sec:discussion_roi} discusses the three ROIs.

\begin{figure}[t]
    \centering
    \subfloat[$\mathrm{ROI}^\mathrm{C}$ and $\mathrm{ROI}^\mathrm{A}$]{
    \includegraphics[width=0.21\textwidth]{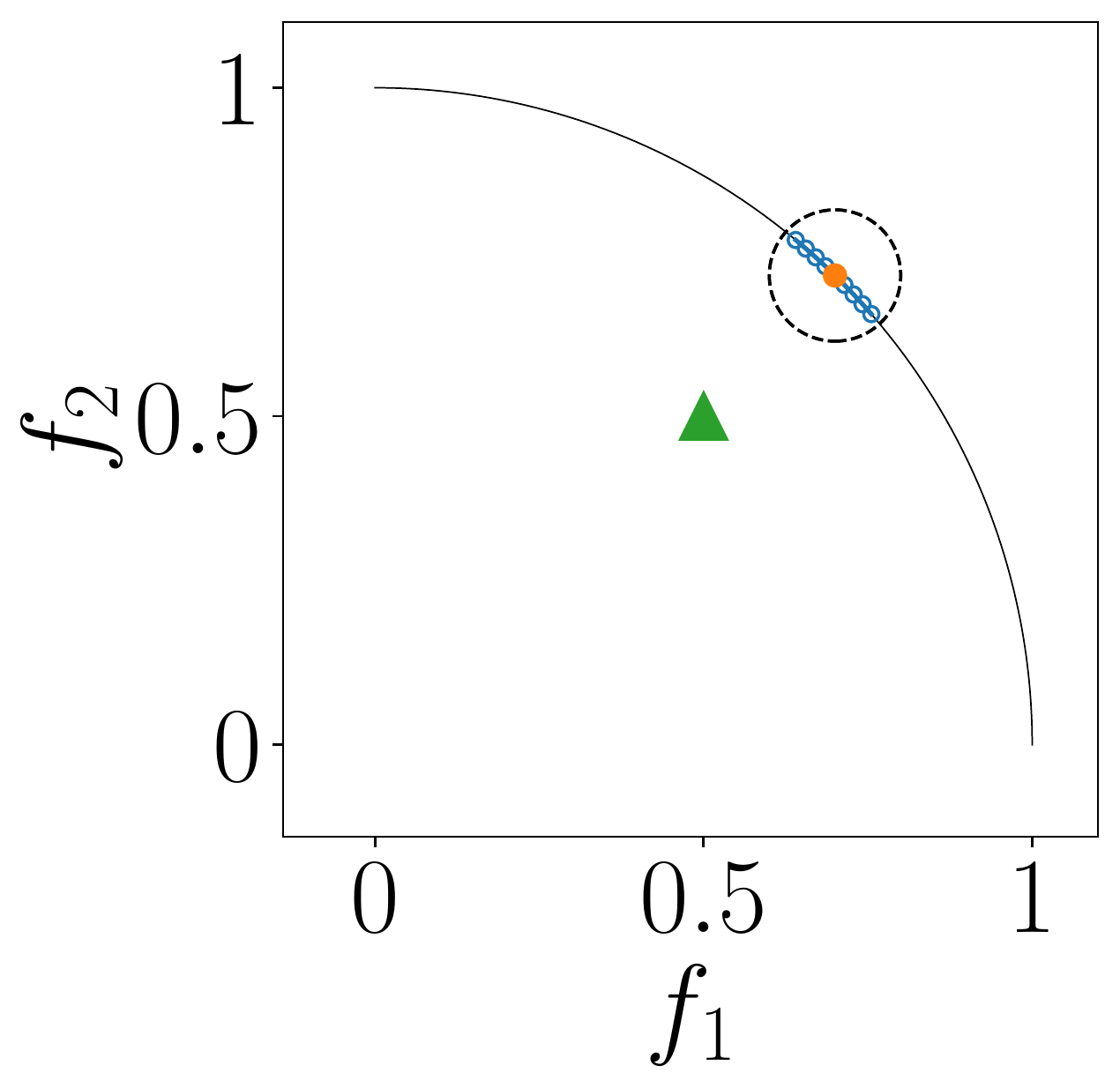}
    }
    \subfloat[$\mathrm{ROI}^\mathrm{P}$]{
    \includegraphics[width=0.21\textwidth]{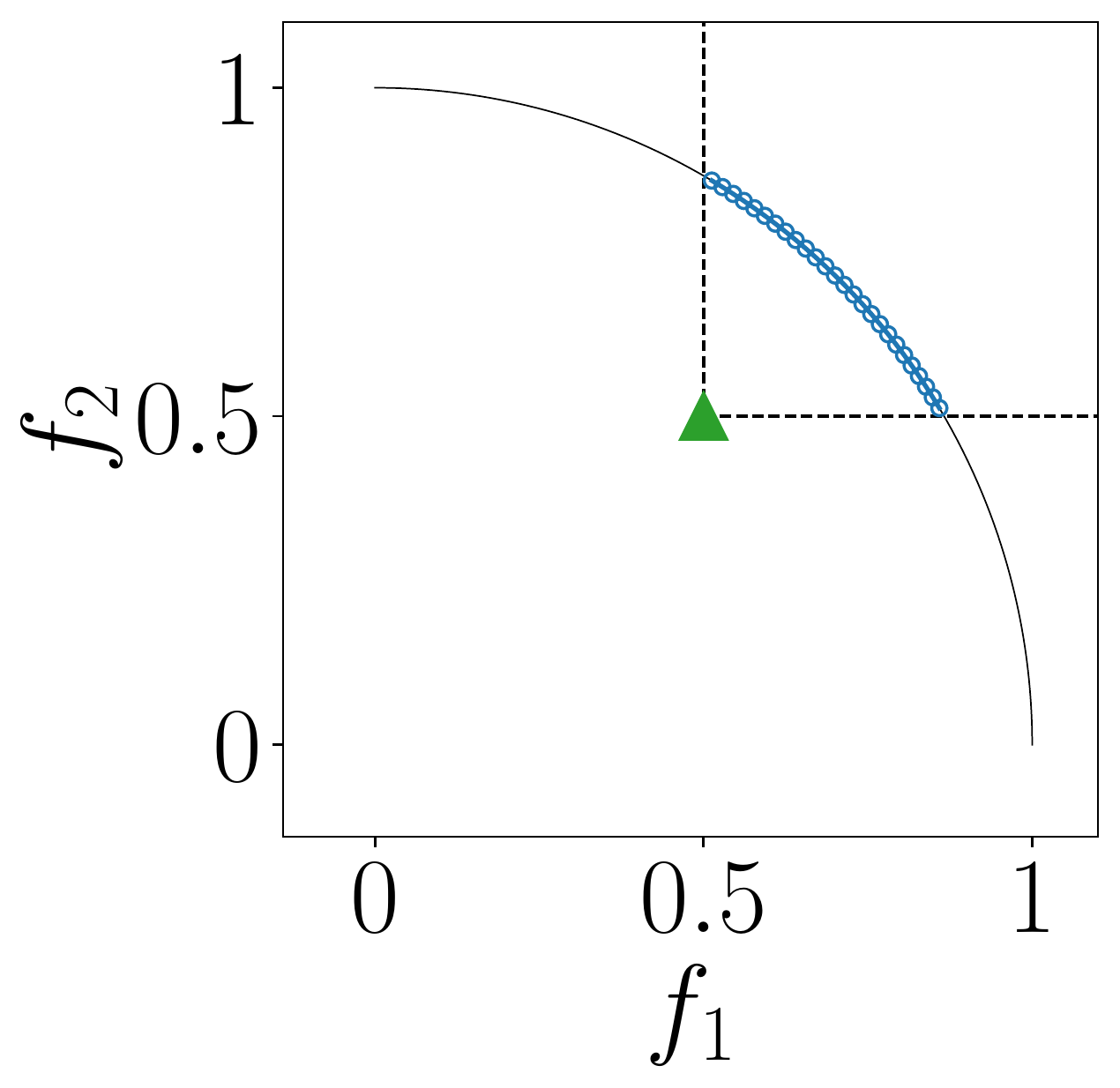}
    }
    \caption{Distributions of Pareto-optimal points \tabblue{$\bullet$} in the three ROIs on the DTLZ2 problem, where \tabgreen{$\blacktriangle$} is the reference point $\mathbf{z}^{0.5}=(0.5,0.5)^\top$, \taborange{$\blacksquare$} is $\mathbf{p}^{\mathrm{c}*}$ and $\mathbf{p}^{\mathrm{a}*}$, and $\zeta=0.1$. 
    }
    \label{fig:roi}
\end{figure}

\subsection{Definitions of three ROIs}
\label{sec:def_roi}

\subsubsection{ROI based on the closest point}

This might be the most intuitive ROI that consists of a set of Pareto-optimal points closest to $\mathbf{z}$ in terms of the Euclidean distance (e.g.,~\cite{MohammadiOL13} and~\cite{MohammadiOLD14}). Mathematically, it is defined as:
\begin{equation}
    \mathrm{ROI}^\mathrm{C}=\left\{\mathbf{p}^\ast\in\mathcal{F} \, | \, \mathrm{dist}(\mathbf{p}^\ast, \mathbf{p}^{\mathrm{c}^\ast})<\zeta\right\},
\end{equation}
where $\mathbf{p}^{\mathrm{c}^\ast}=\underset{\mathbf{p}^\ast\in\mathcal{F}}{\argmin}\left\{\mathrm{dist}(\mathbf{p}^\ast,\mathbf{z}) \right\}$ is the closest Pareto-optimal point to $\mathbf{z}$, and $\zeta$ is the radius of the $\mathrm{ROI}^\mathrm{C}$. As the  example shown in~\pref{fig:roi}(a), the $\mathrm{ROI}^\mathrm{C}$ is a set of points in a hypersphere of a radius $\zeta$ centered at $\mathbf{p}^{\mathrm{c}^\ast}$ while the extent of the $\mathrm{ROI}^\mathrm{C}$ depends on $\zeta$. We believe that R-NSGA-II, r-NSGA-II, and R-MEAD2 were designed for the $\mathrm{ROI}^\mathrm{C}$ implicitly.

\subsubsection{ROI based on the ASF}
\label{sec:def_roia}

As studied in~\cite{ThieleMKL09,LiCMY18} and~\cite{LiDY18}, this ROI consists of a set of the Pareto-optimal points closest to the one with the minimum ASF value: 
\begin{equation}
    \mathrm{ROI}^\mathrm{A}=\{\mathbf{p}^\ast\in\mathcal{F} \, | \, \mathrm{dist}(\mathbf{p}^\ast,\mathbf{p}^{\mathrm{a}^\ast})<\zeta\},
\end{equation}
where $\mathbf{p}^{\mathrm{a}^\ast}=\underset{\mathbf{p}^\ast\in\mathcal{F}}{\argmin}\left\{s(\mathbf{p}^*)\right\}$ is the Pareto-optimal point $\mathbf{p}^{\mathrm{a}\ast}$ having the minimum ASF value, and $s$ is the same as in~\eqref{eqn:asf1}. We believe that PBEA and MOEA/D-NUMS were designed for the $\mathrm{ROI}^\mathrm{A}$ implicitly.

\subsubsection{ROI based on the Pareto dominance relation}
\label{sec:def_roip}

This ROI is defined by an extension of the Pareto dominance relation with regard to the DM specified reference point $\mathbf{z}$ (e.g.,~\cite{RuizSL15,ZhuGDCX18,LuoSLC19,FilatovasKRF20}). When $\mathbf{z}$ is feasible, the $\mathrm{ROI}^\mathrm{P}$ is a set of Pareto-optimal points that dominate $\mathbf{z}$, i.e., $\mathrm{ROI}^\mathrm{P}=\{\mathbf{p}^\ast\in\mathcal{F} \,|\, \mathbf{p}^\ast\prec\mathbf{z}\}$. Otherwise, the $\mathrm{ROI}^\mathrm{P}$ is a set of Pareto-optimal points dominated by $\mathbf{z}$, i.e., $\mathrm{ROI}^\mathrm{P}=\{\mathbf{p}^\ast\in\mathcal{F}\,|\, \mathbf{p}^\ast\succ\mathbf{z}\}$ when $\mathbf{z}$ is infeasible.
We believe g-NSGA-II was designed for the $\mathrm{ROI}^\mathrm{P}$.
Note that $\mathrm{ROI}^\mathrm{P}$ is an empty set when $\mathbf{z}$ is on the PF.
For the same reason as in the g-dominance relation~\cite{LuoSLC19}, $\mathrm{ROI}^\mathrm{P}$ is also an empty set when $\mathbf{z}$ does not dominate any Pareto optimal point on a disconnected PF.


\subsection{Discussions}
\label{sec:discussion_roi}

To have an intuitive understanding of these three ROIs, \pref{fig:roi} shows distributions of Pareto-optimal points in the aforementioned ROIs on the $2$-objective DTLZ2 problem with a non-convex PF. 
As shown in Figs.~\ref{fig:roi}(a), the $\mathrm{ROI}^\mathrm{C}$ and $\mathrm{ROI}^\mathrm{A}$ are sets of the points in the hyper-spheres centered at $\mathbf{p}^{\mathrm{c}\ast}$ and $\mathbf{p}^{\mathrm{a}\ast}$, respectively. They are equivalent if the closest point to $\mathbf{z}$ and the point with the minimum ASF value are the same. For this reason, the $\mathrm{ROI}^\mathrm{C}$ and $\mathrm{ROI}^\mathrm{A}$ may have been considered as the same ROI in the literature.
However, \pref{sec:exp_inconsistency} shows that $\mathrm{ROI}^\mathrm{C}$ and $\mathrm{ROI}^\mathrm{A}$ are different depending on the position of $\mathbf{z}$. 
In contrast, as shown in~\pref{fig:roi}(b), the $\mathrm{ROI}^\mathrm{P}$ is a set of points dominated by $\mathbf{z}$ and its extent is larger than that of the $\mathrm{ROI}^\mathrm{C}$ and $\mathrm{ROI}^\mathrm{A}$. However, it is worth noting that the $\mathrm{ROI}^\mathrm{P}$ does not have any parameter to control its extent as done in the $\mathrm{ROI}^\mathrm{C}$ and $\mathrm{ROI}^\mathrm{A}$. Instead, the size of the $\mathrm{ROI}^\mathrm{P}$ depends on the position of $\mathbf{z}$. If it is too close to the PF, the size of the $\mathrm{ROI}^\mathrm{P}$ can be very small; otherwise it can be very large if $\mathbf{z}$ is too far away from the PF. In the extreme case, if $\mathbf{z}$ dominates the ideal point or is dominated by the nadir point, the $\mathrm{ROI}^\mathrm{P}$ is the same as the PF. Since a DM has little knowledge of the shape of the PF a priori, it is not recommended to use the $\mathrm{ROI}^\mathrm{P}$ in real-world black-box applications.

%% file: review.tex

\section{Review of quality indicators}
\label{sec:review_qi}

\begin{table*}[t!]
\setlength{\tabcolsep}{2pt} 
\centering
\caption{\small Properties of the $14$ quality indicators for PBEMO, including the number of point sets $K$, the type of target ROI, the convergence to the PF (C-PF), the convergence to the reference point $\mathbf{z}$ (C-$\mathbf{z}$), the diversity in the ROI (Div), the ability to handle point sets outside a preferred region (Out), no use of information about the PF (U-PF), control parameters (Param.), the extendability to the use of multiple reference points (Mul.), and the time complexity (Time. Comp.).}
\label{tab:qi}
 {\footnotesize
\scalebox{0.92}[1]{ 
  \begin{tabular}{lcccccccccc}
  \toprule
Indicators & $K$ & ROI & C-PF & C-$\mathbf{z}$ & Div & Out & U-PF & Param. & Mul. & Time. Comp.\\
  \midrule  
  Minimum achievement scalarizing function (MASF) \cite{ThieleMKL09} & unary & $\mathrm{ROI}^\mathrm{A}$ & \Checkmark & \Checkmark &  & \Checkmark & \Checkmark  & $\mathbf{w}$  & & $\mathcal{O}(\mu m)$\\
  Mean Euclidean distance (MED) \cite{TangLDWZF20} & unary & $\mathrm{ROI}^\mathrm{C}$ & &  \Checkmark &  & \Checkmark &  &  &  & $\mathcal{O}(\mu m)$\\
  IGD for $\mathrm{ROI}^\mathrm{C}$ (IGD-C) \cite{MohammadiOLD14} & unary & $\mathrm{ROI}^\mathrm{C}$ & \Checkmark & \Checkmark & \Checkmark  & \Checkmark &  & $r$, $\mathcal{S}$ & & $\max \{\mathcal{O}(\mu |\mathcal{S}'| m) , \mathcal{O}(|\mathcal{S}| m)\}$\\
  IGD for $\mathrm{ROI}^\mathrm{A}$ (IGD-A) & unary & $\mathrm{ROI}^\mathrm{A}$ & \Checkmark & \Checkmark & \Checkmark  & \Checkmark &  & $\mathbf{w}$, $r$, $\mathcal{S}$ & & $\max \{\mathcal{O}(\mu |\mathcal{S}'| m) , \mathcal{O}(|\mathcal{S}| m)\}$\\
  IGD for $\mathrm{ROI}^\mathrm{P}$ (IGD-P) \cite{LuoSLC19} & unary & $\mathrm{ROI}^\mathrm{P}$ & \Checkmark & \Checkmark & \Checkmark  & \Checkmark &  & $\mathcal{S}$ & & $\max \{\mathcal{O}(m |\mathcal{S}'| \mu) , \mathcal{O}( |\mathcal{S}| m)\}$\\
  HV of the ROI defined by $\mathbf{z}$ (HV$_{\mathbf{z}}$) \cite{RuizSL15} & unary & $\mathrm{ROI}^\mathrm{P}$ & \Checkmark & \Checkmark & \Checkmark &  &  \Checkmark & & \Checkmark & $\max\{\mathcal{O}(\mu^{m-1}) , \mathcal{O}( |\mathcal{S}| m)\}$\\
  Percentage of points in the ROI (PR) \cite{FilatovasLKZ17} & unary & $\mathrm{ROI}^\mathrm{P}$ & &  &  & & \Checkmark  & & & $\mathcal{O}(m \mu)$\\  
  PMOD \cite{HouYZZYR18} & unary & Unclear & & \Checkmark & \Checkmark & \Checkmark & \Checkmark  & $r$, $\alpha$ & & $\mathcal{O}(\mu^2 m)$\\
  IGD with a composite front (IGD-CF) \cite{MohammadiOL13} & $K$-ary & $\mathrm{ROI}^\mathrm{C}$ & \Checkmark & \Checkmark & \Checkmark & & \Checkmark  & $r$ & & $\max \{\mathcal{O}(m |\mathcal{P}^{\mathrm{CF}}| \mu), \mathcal{O}(K \mu m)\}$\\
  HV with a composite front (HV-CF) \cite{MohammadiOL13} & $K$-ary & $\mathrm{ROI}^\mathrm{C}$ & \Checkmark & \Checkmark & \Checkmark & & \Checkmark  & $r$, $\mathbf{y}$ & & $\max \{\mathcal{O}(\mu^{m-1}), \mathcal{O}(K \mu m)\}$\\
  Preference-based metric based on distances and angles (PMDA) \cite{YuZL15} & $K$-ary & Unclear & \Checkmark & \Checkmark &  & \Checkmark & \Checkmark  & $\alpha$, $\gamma$ & & $\mathcal{O}(K \mu^2 m)$\\
  R-metric version of IGD (R-IGD) \cite{LiDY18} & $K$-ary & $\mathrm{ROI}^\mathrm{A}$ & \Checkmark & \Checkmark & \Checkmark & \Checkmark &  & $r$, $\mathbf{z}^{\mathrm{w}}$, $\mathbf{w}$, $\mathcal{S}$ & \Checkmark & $\max{\mathcal{O}(\mu m), \mathcal{O}(K^2 \mu^2)}$\\
  R-metric version of HV (R-HV) \cite{LiDY18} & $K$-ary & $\mathrm{ROI}^\mathrm{A}$ & \Checkmark & \Checkmark & \Checkmark &  \Checkmark & \Checkmark & $r$, $\mathbf{z}^{\mathrm{w}}$, $\mathbf{w}$ & \Checkmark & $\max{\mathcal{O}(\mu^{m-1}), \mathcal{O}(K^2 \mu^2)}$\\
  Expanding hypercube metric (EH) \cite{BandaruS19} & $K$-ary & Unclear & \Checkmark & \Checkmark &  & \Checkmark  &  \Checkmark  &  & & $\mathcal{O}(K \mu^2 m)$\\
  \bottomrule
\end{tabular}
}
}
\end{table*}

This section reviews $14$ quality indicators proposed in the literature for assessing the performance of PBEMO algorithms. Their properties are summarized in~\pref{tab:qi}.
According to the definitions in~\pref{sec:def_roi}, we classify the target ROIs of the 14 quality indicators based on their preferred regions. In particular since the preferred regions of PMOD, PMDA, and EH do not belong to any of the three ROIs defined in~\pref{sec:def_roi}, their ROIs are labeled as unclear.
The target ROIs of R-IGD and R-HV are slightly different from the $\mathrm{ROI}^\mathrm{A}$, where they are based on a hypercube, instead of a hypersphere.
%
As shown in~\pref{tab:qi}, the previous studies assumed different ROIs.
This suggests that the ROI has not been standardized in the EMO community.

Although this paper uses a single reference point as mentioned in \pref{sec:pemo_ref}, some previous studies (e.g., \cite{DebSBC06,ThieleMKL09}) used multiple reference points.
As seen from \pref{tab:qi}, only HV$_{\mathbf{z}}$, R-IGD, and R-HV can handle multiple reference points.
Although the original HV$_{\mathbf{z}}$ can handle only a single reference point, its extended version was considered in \cite{ZhuGDCX18}.
Th extended HV$_{\mathbf{z}}$ calculates HV$_{\mathbf{z}}$ for each $\mathbf{z}$ and then calculates the sum of all HV$_{\mathbf{z}}$ values.
Other quality indicators may be able to handle multiple reference points in a similar manner, but investigating this idea is beyond the scope of this paper.

We referred to the time complexity of R-IGD, R-HV, and EH from the corresponding papers.
We calculated that of other quality indicators ourselves.
As seen from \pref{tab:qi}, MASF, MED, and PR have the lowest time complexity.
However, \pref{sec:results} demonstrates that none of them are very reliable.
Although IGD-based quality indicators (e.g., IGD-C) require computational cost for preprocessing the IGD-reference point set $\mathcal{S}$, the size of $\mathcal{S}$ can be extremely large as $m$ increases.
For example, the previous study \cite{MohammadiOLD14} set $|\mathcal{S}|$ to $10^m$ for $m \in \{4, 5, 6\}$ and $5^m$ for $m=7$.
This is because the number of IGD-reference points in the ROI can be drastically small as $m$ increases. 
%
As in the original HV indicator, HV-based quality indicators (e.g., HV$_{\mathbf{z}}$) require high computational cost for many-objective optimization.

In the following paragraphs, \pref{sec:discussions_qi} first discusses the desirable properties of a quality indicator for PBEMO.
Then, Sections~\ref{sec:masf} to \ref{sec:eh} delineate the underlying mechanisms of $14$ quality indicators, respectively. 
In particular, the technical details of some quality indicators (e.g., \cite{NguyenXB15,SiegmundND15,CinalliMSG16}) are missing.
In addition, the HV-based indicator developed in~\cite{WickramasingheCL10} and the spread-based indicator proposed in~\cite{FilatovasKS15} do not consider the preference information from the DM. 
%
It is also unclear how to extend HV-T~\cite{LiWTJE18} to evaluate the performance of PBEMO algorithms using a reference point.
Therefore, this paper does not intend to elaborate them.

\subsection{Desirable properties of quality indicators}
\label{sec:discussions_qi}

\begin{figure}[t!]
    \centering
    \subfloat[Distributions of $\mathcal{P}^1$ to $\mathcal{P}^{9}$]{  
    \includegraphics[width=0.22\textwidth]{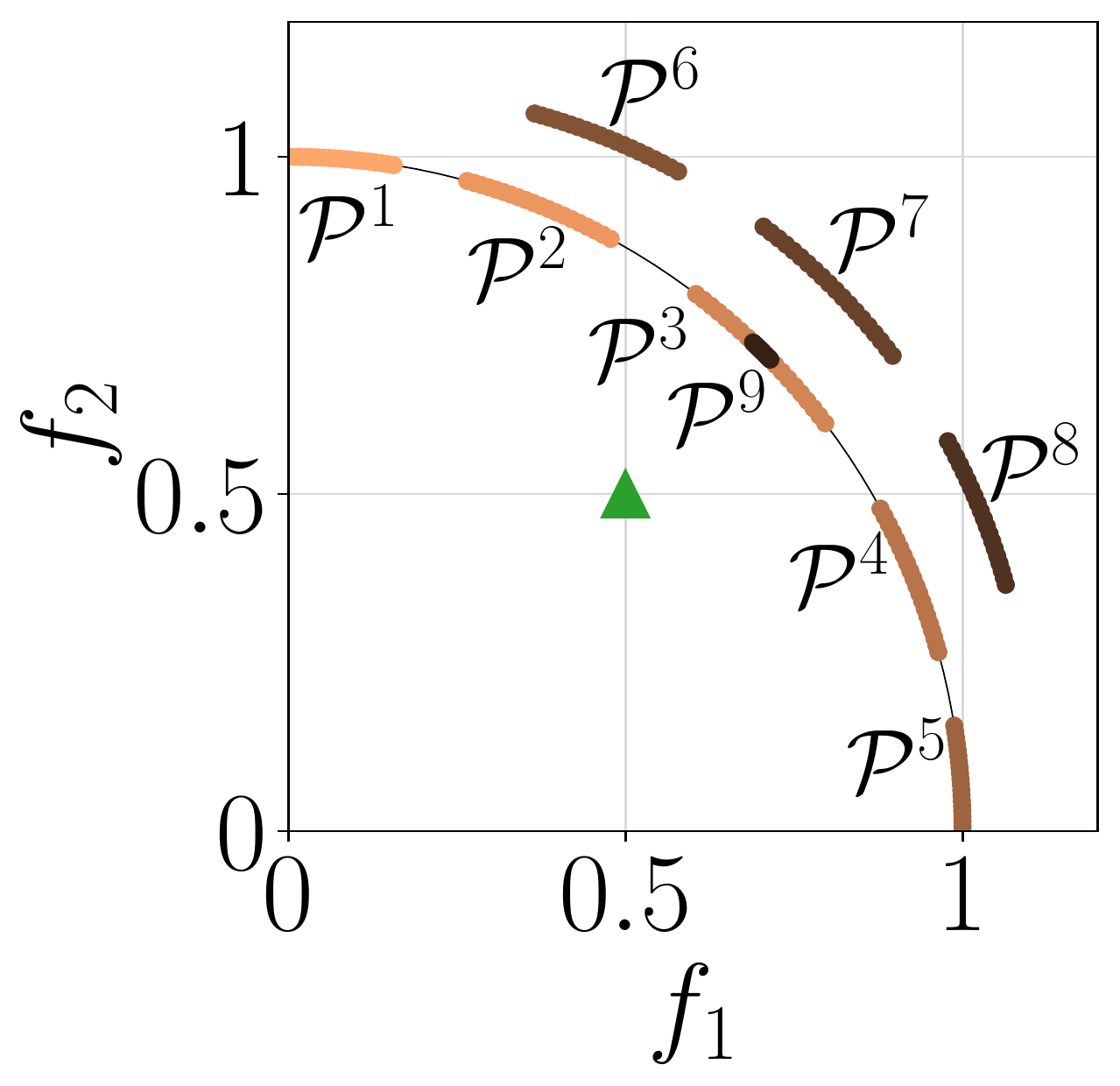}
    }
    \subfloat[Distribution of $\mathcal{P}^{10}$]{
        \includegraphics[width=0.21\textwidth]{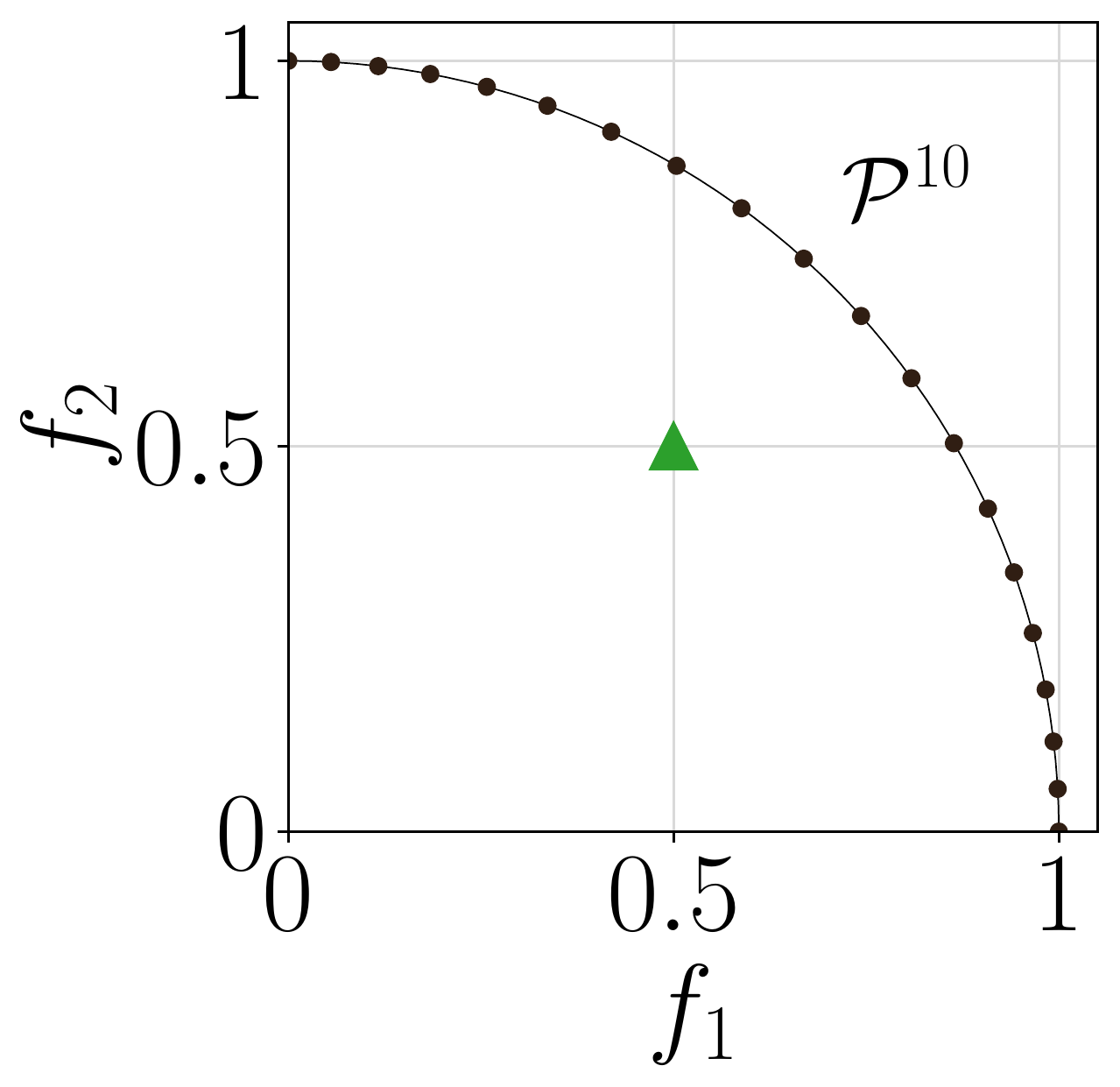}
    }
    \caption{Distributions of the $10$ point sets on the PF of the DTLZ2 problem when $m=2$, where \tabgreen{$\blacktriangle$} is the reference point.
    }
    \label{fig:toy}
\end{figure}

To facilitate our discussion, we generate $10$ synthetic point sets, each of which consists of $20$ uniformly distributed points, along the PF of the $2$-objective DTLZ2 problem as shown in~\pref{fig:toy}. More specifically, $\mathcal{P}^1$ to $\mathcal{P}^5$ are distributed on five different subregions of the PF. $\mathcal{P}^6,\mathcal{P}^7$ and $\mathcal{P}^8$ are the shifted versions of $\mathcal{P}^2,\mathcal{P}^3$ and $\mathcal{P}^4$ by adding $0.1$ to all elements, respectively. Thus, $\mathcal{P}^6, \mathcal{P}^7, \mathcal{P}^{8}$ are dominated by $\mathcal{P}^2, \mathcal{P}^3, \mathcal{P}^{4}$, respectively. $\mathcal{P}^9$ is on the PF, but the extent of $\mathcal{P}^9$ is worse than that of $\mathcal{P}^3$. Unlike the other point sets, the points in $\mathcal{P}^{10}$ are uniformly distributed on the whole PF. Given $\mathbf{z}=(0.5, 0.5)^\top$ as the DM specified reference point, $\mathcal{P}^3$ is the best point set with regard to the $\mathrm{ROI}^\mathrm{C}$, $\mathrm{ROI}^\mathrm{A}$, and $\mathrm{ROI}^\mathrm{P}$. In this paper, we argue that a desirable preference-based quality indicator is required to assess the four aspects including $i$) the convergence to the PF; $ii$) the convergence to $\mathbf{z}$; $iii$) the diversity of trade-off alternatives in a point set in the ROI; and $iv$) the ability to handle point sets outside the ROI.

Pour et al.~\cite{PourBAM22} discussed desirable properties of methods for evaluating the performance of interactive EMO algorithms using a reference point $\mathbf{z}$, where $\mathbf{z}$ is dynamically adjusted by the DM during the search. 
The desirable properties include nine general properties (GP1--GP9), two properties for the learning phase (LP1--LP2), and two properties for the decision phase (DP1--DP2).
In interactive EMO, the DM explores the objective space in the learning phase to determine her/his ROI.
Then, the DM exploits solutions in the ROI in the decision phase.
For example, LP1 is to measure how helpful solutions found by an interactive EMO algorithm are in the learning phase.
Some of GP1--GP9 can be applied to quality indicators for PBEMO.
In contrast, LP1--LP2 and DP1--DP2 are beyond the responsibility of quality indicators for PBEMO.
In fact, as discussed in \cite{PourBAM22}, none of R-HV, EH, HV-CF, PMOD, and PMDA is satisfied with LP1--LP2 and DP1--DP2.

Very recently, Pour et al.~\cite{PourBAEM23} proposed a preference-based hypervolume indicator (PHI), which is the first quality indicator for interactive EMO. 
PHI considers positive and negative HV contributions of points, where the latter is used to penalize points outside the ROI.
The previous study~\cite{PourBAEM23} demonstrated that PHI can evaluate the performance of two interactive EMO algorithms in the learning and decision phases.

\begin{remark}
The term \lq\lq convergence\rq\rq\ of a point set $\mathcal{P}$ has not been specified in the context of PBEMO. As shown in~\pref{tab:qi}, we distinguish the convergence to the PF and the convergence to $\mathbf{z}$. In~\pref{fig:toy}, $\mathcal{P}^1,\ldots,\mathcal{P}^5$, and $\mathcal{P}^9$ have a good convergence to the PF. In contrast, only $\mathcal{P}^3$ and $\mathcal{P}^9$ have a good convergence to $\mathbf{z}$.
\end{remark}

\begin{remark}
    The diversity of a point set in the ROI is also an important evaluation criterion. If the ROI contains both $\mathcal{P}^3$ and $\mathcal{P}^9$, a quality indicator should evaluate $\mathcal{P}^3$ as having higher diversity than $\mathcal{P}^9$.
\end{remark}

\begin{remark}
    Li et al.~\cite{LiDY18} pointed out that quality indicators should be able to distinguish point sets outside the ROI. In~\pref{fig:toy}, $\mathcal{P}^1$ and $\mathcal{P}^2$ are outside the ROI. However, $\mathcal{P}^2$ is closer to the ROI than $\mathcal{P}^1$. The same is true for the relation between $\mathcal{P}^4$ and $\mathcal{P}^5$. In this case, a quality indicator should evaluate $\mathcal{P}^2$ and $\mathcal{P}^4$ as having better quality than $\mathcal{P}^1$ and $\mathcal{P}^5$. Mohammadi et al. \cite{MohammadiOL13} pointed out the importance of not using information about the PF, which is generally unavailable in real-world problems. As shown in~\pref{tab:qi}, $9$ out of the $14$ indicators satisfy this criterion. Note that an approximation of the PF or the ROI found by PBEMO algorithms is available in practice. We believe that the remaining $5$ out of the $14$ indicators can address the issue by simply using the approximation.
\end{remark}

\subsection{MASF}
\label{sec:masf}


As done in some previous studies, e.g.,~\cite{ThieleMKL09,BrockhoffHK14} and~\cite{LiDY18}, the basic idea of this quality indicator is to use the minimum ASF (MASF) value of $\mathcal{P}$ to evaluate the closeness of $\mathcal{P}$ to $\mathbf{z}$:
\begin{equation}
    \mathrm{MASF}(\mathcal{P})=\min_{\mathbf{p}\in\mathcal{P}}\left\{s(\mathbf{p})\right\},
\end{equation}
where we use $s$ in \eqref{eqn:asf1}.
MASF can evaluate only the two types of convergence.
Since MASF does not consider the other $\mu-1$ points in $\mathcal{P}$, MASF cannot evaluate the diversity of $\mathcal{P}$.
As the example shown in~\pref{fig:toy}, MASF prefers $\mathcal{P}^9$ to $\mathcal{P}^3$.

\subsection{MED}
\label{sec:spp}

The mean Euclidean distance (MED) measures the average of Euclidean distance between each point in $\mathcal{P}$ to $\mathbf{z}$~\cite{TangLDWZF20}:
\begin{equation}
    \mathrm{MED}(\mathcal{P}) = \frac{1}{|\mathcal{P}|}\sum_{\mathbf{p} \in \mathcal{P}} \sqrt{\sum^{m}_{i=1} \left(\frac{p_i - z_i}{p_i^{\mathrm{nadir}} - p_i^{\mathrm{ideal}}}\right)^2}.
\end{equation}
MED can evaluate how close all points in $\mathcal{P}$ are to $\mathbf{z}$ and the PF when $\mathbf{z}$ is infeasible. MED cannot evaluate the convergence to the PF if $\mathbf{z}$ is feasible. This is because MED prefers the non-Pareto optimal points close to $\mathbf{z}$ than the Pareto optimal points. As the example shown in~\pref{fig:toy}, MED prefers the dominated $\mathcal{P}^7$ over the non-dominated $\mathcal{P}^3$ when $\mathbf{z}=(1.0,1.0)^\top$.

\subsection{IGD-based indicators}
\label{sec:igdroi}

Here, we introduce three quality indicators developed upon the IGD metric. In particular, since they are designed to deal with the $\mathrm{ROI}^\mathrm{C}$, $\mathrm{ROI}^\mathrm{A}$, and $\mathrm{ROI}^\mathrm{P}$ defined in~\pref{sec:review_roi}, respectively, they are thus denoted as IGD-C, IGD-A, and IGD-P accordingly in this paper. Note that both IGD-C and IGD-P were used in some previous studies~\cite{MohammadiOLD14,ZhuHJ16}, and~\cite{LuoSLC19}, respectively, whereas IGD-A is deliberately designed in this paper to facilitate our analysis.

In practice, the only difference between the original IGD and its three extensions is the choice of the IGD-reference point set $\mathcal{S}$.
In IGD, IGD-reference points in $\mathcal{S}$ are uniformly distributed on the whole PF.
In contrast, IGD-C, IGD-A, and IGD-P use a subset $\mathcal{S}' \subseteq \mathcal{S}$.
$\mathcal{S}'$ can also be a subset of each ROI. 
In the example in Fig. \ref{fig:roi}, $\mathcal{S}'$ of IGD-C, IGD-A, and IGD-P are in the $\mathrm{ROI}^\mathrm{C}$, $\mathrm{ROI}^\mathrm{A}$, and $\mathrm{ROI}^\mathrm{P}$, respectively.
Below, for each indicator, we describe how to select $\mathcal{S}'$ from $\mathcal{S}$.


\begin{itemize}
    \item For IGD-C, we first find the closest point $\mathbf{p}^{\mathrm{c}}$ to $\mathbf{z}$ from $\mathcal{S}$, i.e., $\mathbf{p}^{\mathrm{c}} = \mathrm{argmin}_{\mathbf{p} \in \mathcal{S}} \{\mathrm{dist}(\mathbf{p}, \mathbf{z}) \}$. Then, $\mathcal{S}'$ is a set of all points in the region of a hypersphere of radius $r$ centered at $\mathbf{p}^{\mathrm{c}}$, i.e., $\mathcal{S}' = \{\mathbf{p} \in \mathcal{S} \, | \, \mathrm{dist}(\mathbf{p}, \mathbf{p}^{\mathrm{c}}) < r \}$.

    \item The only difference between IGD-C and IGD-A is the choice of the center point. First, a point with the minimum ASF value  $\mathbf{p}^{\mathrm{a}}$ is selected from $\mathcal{S}$, i.e., $\mathbf{p}^{\mathrm{a}} = \mathrm{argmin}_{\mathbf{p} \in \mathcal{S}} \{s(\mathbf{p}) \}$. 
    Then, $\mathcal{S}'$ is a set of all points in the region of a hypersphere of radius $r$ centered at $\mathbf{p}^{\mathrm{a}}$, i.e., $\mathcal{S}' = \{\mathbf{p} \in \mathcal{S} \, | \, \mathrm{dist}(\mathbf{p}, \mathbf{p}^{\mathrm{a}}) < r \}$.

    \item In IGD-P, $\mathcal{S}'$ is selected from $\mathcal{S}$ based on the Pareto dominance relation as in the $\mathrm{ROI}^\mathrm{P}$. If $\mathbf{z}$ is feasible, $\mathcal{S}' = \{\mathbf{p} \in \mathcal{S} \,|\,  \mathbf{p} \prec \mathbf{z}\}$. Otherwise, $\mathcal{S}' = \{\mathbf{p} \in \mathcal{S} \,|\,  \mathbf{p} \succ \mathbf{z}\}$. Note that IGD-P does not require the radius $r$.
\end{itemize}

\subsection{HV$_{\mathbf{z}}$}
\label{sec:hvq}

\begin{figure}[t!]
  \centering
  \subfloat[Feasible $\mathbf{z} = (0.9, 0.9)^{\top}$]{  
    \includegraphics[width=0.21\textwidth]{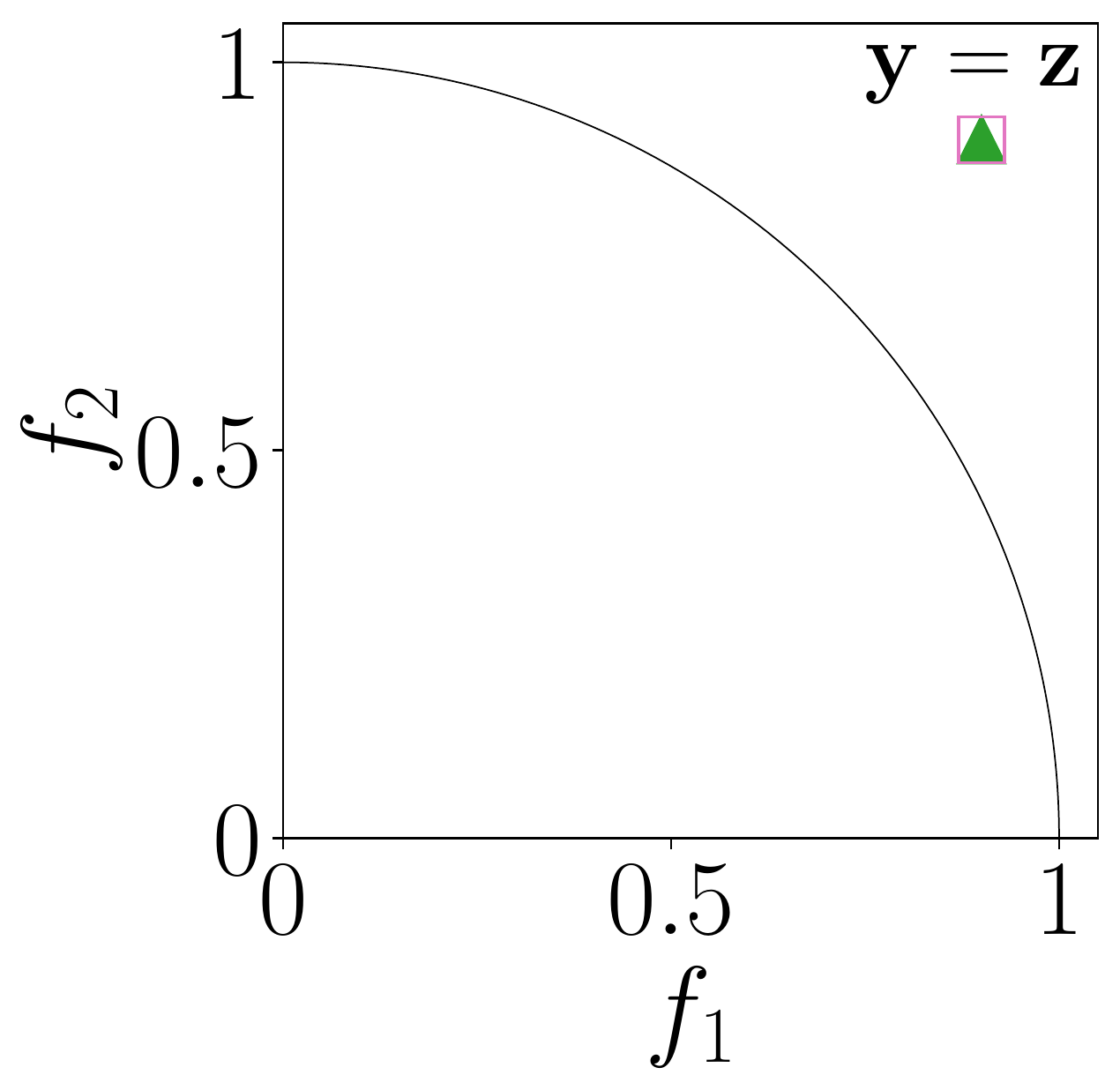}
  }
  \subfloat[Infeasible $\mathbf{z} = (0.5, 0.5)^{\top}$]{  
    \includegraphics[width=0.21\textwidth]{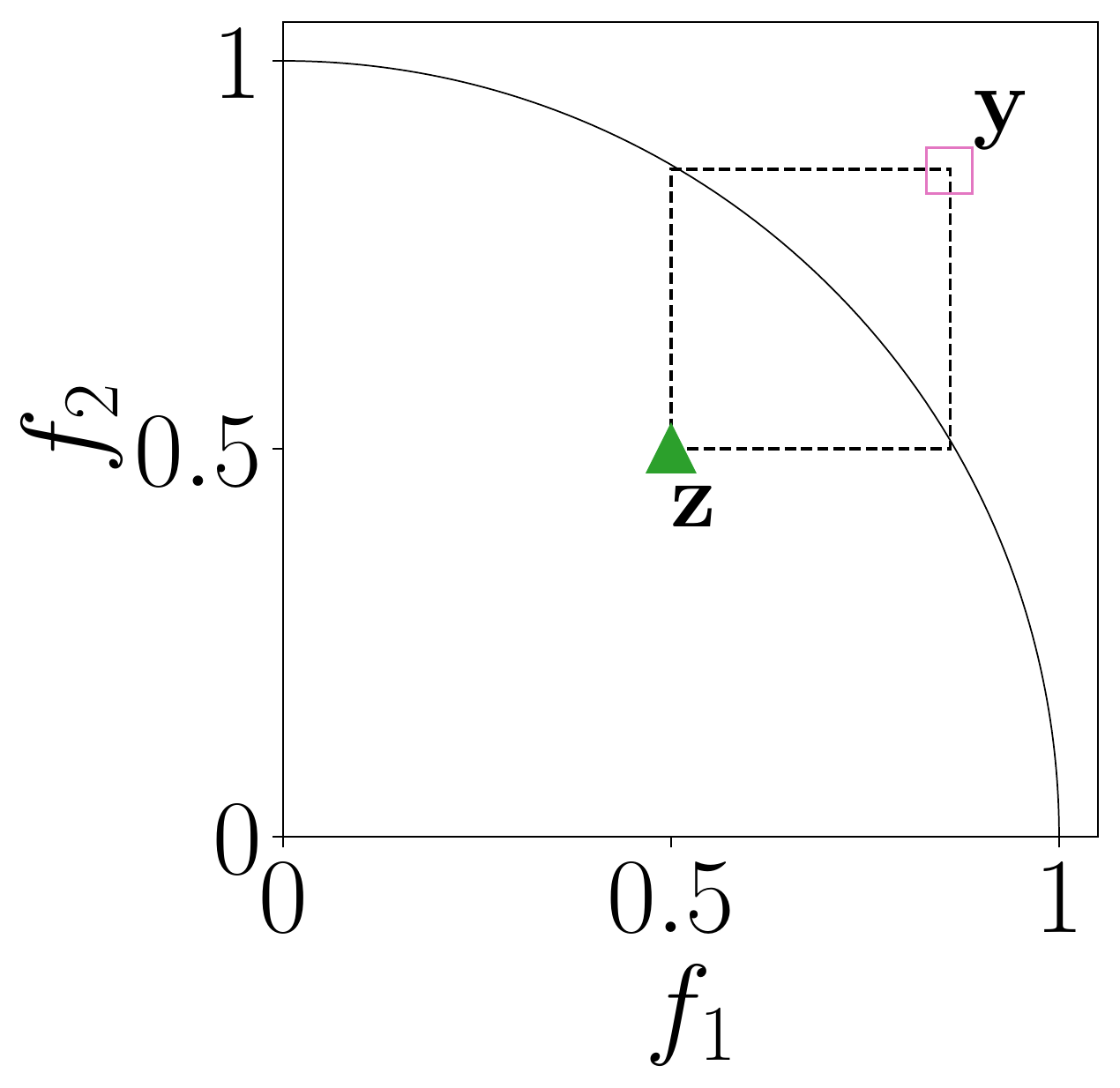}
  }
  \caption{Examples of the HV-reference point $\mathbf{y}$ (denoted as \tabpink{$\Box$}) in HV$_{\mathbf{z}}$, where where \tabgreen{$\blacktriangle$} is the reference point.}
   \label{fig:hvz}
\end{figure}

HV$_{\mathbf{z}}$ was originally named HV$_{\mathbf{q}}$ in~\cite{RuizSL15}, where $\mathbf{q}$ represents the reference point in \cite{RuizSL15}. Since this paper denotes the reference point as $\mathbf{z}$, we use the term ``HV$_{\mathbf{z}}$'' to make the consistency.
%
The only difference between HV and HV$_{\mathbf{z}}$ is the choice of the HV-reference point $\mathbf{y} \in \mathbb{R}^m$ as follows.
If $\mathbf{z}$ is feasible, $\mathbf{y} = \mathbf{z}$.
If $\mathbf{z}$ is infeasible, $y_i = \max_{\mathbf{p}^* \in \mathrm{ROI}^\mathrm{P}}\{p^*_i\}$ for each $i \in \{1, \dots, m\}$.
Fig. \ref{fig:hvz} shows the HV-reference point $\mathbf{y}$ in HV$_{\mathbf{z}}$ when setting $\mathbf{z}$ to $(0.9, 0.9)^{\top}$ and $(0.5, 0.5)^{\top}$, where the former is feasible while the latter is infeasible.

HV$_{\mathbf{z}}$ can evaluate the convergence and diversity of a point set. 
However, HV$_{\mathbf{z}}$  cannot handle the point sets outside the region dominated by the HV-reference point $\mathbf{y}$. 
In the example in Fig. \ref{fig:toy}, the HV$_{\mathbf{z}}$ values of $\mathcal{P}^1, \mathcal{P}^2, \mathcal{P}^4$, and $\mathcal{P}^5$ are 0.


\subsection{PR}
\label{sec:pr}

The percentage of points in the ROI (PR) evaluates the cardinality of $\mathcal{P}$~\cite{FilatovasLKZ17} lying in the DM specified ROI: 
\begin{equation}
    \mathrm{PR}(\mathcal{P})=\frac{|\{\mathbf{p}\in\mathcal{P}\cap\mathcal{R}\}|}{|\mathcal{P}|}\times 100\%,
\end{equation}
where $\mathcal{R}$ is the $\mathrm{ROI}^\mathrm{P}$ defined in~\cite{FilatovasLKZ17}, though it can be any type of ROI in principle. Note that PR is the only cardinality-based indicator considered in our study. A large PR means that many points in the corresponding $\mathcal{P}$ are in the $\mathrm{ROI}^\mathrm{P}$. Like HV$_{\mathbf{z}}$, it is clear that PR cannot distinguish point sets outside the ROI.

\subsection{PMOD}
\label{sec:pmod}

PMOD consists of two algorithmic steps~\cite{HouYZZYR18}. First, it maps each point $\mathbf{p}\in\mathcal{P}$ onto a hyperplane passing through $\mathbf{z}$ as: 
\begin{equation}
    \mathbf{p}' = \mathbf{p} + \left((\mathbf{z} - \mathbf{p}) \cdot \mathbf{\hat{z}} \right) \mathbf{\hat{z}},
\end{equation}
where $\mathbf{\hat{z}}$ is the unit vector of $\mathbf{z}$.
Then, PMOD aggregates three measurements including $i$) the distance between each mapped point $\mathbf{p}'$ and $\mathbf{z}$, $ii$) the distance between $\mathbf{p}$ and the origin $\mathbf{o}=(0,\ldots,0)^\top$, and $iii$) the unbiased standard deviation of all mapped points as:
\begin{align}
    \mathrm{PMOD}(\mathcal{P})&=\frac{1}{|\mathcal{P}|} \sum_{\mathcal{P}' \in \mathcal{P}'} \left(\mathrm{dist}(\mathbf{p}', \mathbf{z}) + \alpha \, \mathrm{dist}(\mathbf{p}, \mathbf{o})\right)\notag\\
&+ \mathrm{SD}\left(\{d_{\mathbf{p}'}\}_{\mathbf{p}' \in \mathcal{P}'}\right),
\label{eqn:pmod}
\end{align}
where $\mathcal{P}^\prime$ is a set of $|\mathcal{P}|$ mapped points.
In \eqref{eqn:pmod}, $\alpha$ is a penalty parameter for mapped points outside the preferred region of radius $r$ centered at $\mathbf{z}$, where $r$ is a parameter of PMOD. When $\mathbf{p}^\prime$ is inside the preferred region (i.e., $\mathrm{dist}(\mathbf{p}^\prime,\mathbf{z})\leq r$), $\alpha=1$. Otherwise, $\alpha>1$, e.g., $\alpha$ is set to $1.5$ in~\cite{HouYZZYR18}. $\mathrm{SD}$ returns the unbiased standard deviation of input values. For each $\mathbf{p}^\prime\in\mathcal{P}^\prime$, $d_{\mathbf{p}^\prime}$ in \eqref{eqn:pmod} is the minimum Manhattan distance between $\mathbf{p}'$ and another point $\mathbf{q}\in\mathcal{P}^\prime$, i.e., $d_{\mathbf{p}^\prime}=\underset{\mathbf{q}\in\mathcal{P}^\prime\setminus\{\mathbf{p}^\prime\}}{\min}\sum^m_{i=1}|p^\prime_i-q_i|$.

The smaller PMOD value is, the better quality $\mathcal{P}$ is in terms of $i$) the convergence to $\mathbf{z}$, $ii$) the convergence to \textit{the origin} (not the PF), and $iii$) the uniformity. Note that PMOD assumes the ideal point always does not dominate the origin. When the ideal point dominates the origin, PMOD prefers points far from the PF in view of above $ii$). Let us consider the shifted point sets in~\pref{fig:toy} again, if the offset is $-100$, PMOD is likely to prefer $\mathcal{P}^7$ to $\mathcal{P}^3$ because $\mathcal{P}^7$ is closer to the origin than $\mathcal{P}^3$.

\subsection{IGD-CF and HV-CF}
\label{sec:upcf}

A user-preference metric based on a composite front (UPCF) \cite{MohammadiOL13} is a framework for evaluating the quality of $\mathcal{P}$.
IGD-CF and HV-CF are the UPCF versions of IGD and HV, respectively.
%
Algorithm \ref{alg:upcf} shows how to calculate the IGD-CF and HV-CF values of $K$ points sets $\mathcal{P}^1,\ldots,\mathcal{P}^K$.
First, let $\mathcal{P}^{\mathrm{CF}}$ be a set of all non-dominated points in the $K$ point sets (line $1$), where $\mathcal{P}^{\mathrm{CF}}$ was called a composite front in \cite{MohammadiOL13}.
Then, the closest point $\mathbf{p}^{\mathrm{c}}$ to $\mathbf{z}$ is selected from $\mathcal{P}^{\mathrm{CF}}$ (line $2$).
A preferred region $\mathcal{R}^{\mathrm{pref}}$ is defined as the set of all points in the region of a hypersphere of radius $r$ centered at $\mathbf{p}^{\mathrm{c}}$ (line $3$).
Note that $\mathcal{R}^{\mathrm{pref}}$ can include dominated points.
If $\mathcal{P}^{\mathrm{CF}} = \mathcal{F}$, $\mathcal{R}^{\mathrm{pref}}$ is equivalent to the $\mathrm{ROI}^\mathrm{C}$ shown in Fig. \ref{fig:roi}(a).

For each $i\in\{1,\ldots,K\}$, the IGD-CF value of $\mathcal{P}^i$ is calculated based only on the points in $\mathcal{P}^i \cap \mathcal{R}^{\mathrm{pref}}$ (line $5$).
In other words, points outside $\mathcal{R}^{\mathrm{pref}}$ are removed from $\mathcal{P}^i$.
For example, in~\pref{fig:roi}(a), points outside the large dotted circle are removed from a point set.
IGD-CF uses $\mathcal{P}^{\mathrm{CF}}$ as an approximation of the IGD-reference point set $\mathcal{S}$.
The HV-CF value of $\mathcal{P}^i$ is calculated in a similar manner (line $6$), where the previous study \cite{MohammadiOL13} did not give a rule of thumb to set the HV-reference point $\mathbf{y}$.
When the trimmed $\mathcal{P}^i$ is empty, we set its IGD-CF value to $\infty$ and its HV-CF value to $0$ in this study.

Li et al.~\cite{LiDY18} pointed out that IGD-CF and HV-CF cannot distinguish point sets outside the preferred region.
This is because IGD-CF and HV-CF do not consider any point outside $\mathcal{R}^{\mathrm{pref}}$.
In the example in~\pref{fig:toy}, all point sets except for $\mathcal{P}^3$, $\mathcal{P}^9$, and $\mathcal{P}^{10}$ are equally bad, i.e., $\mathrm{IGD}$-$\mathrm{CF}(\mathcal{P}^i)=\infty$ and $\mathrm{HV}$-$\mathrm{CF}(\mathcal{P}^i)=0$ for $i\in\{1,2,4,5,6,7,8\}$.

\begin{algorithm}[t!]
\small
\SetSideCommentRight
\SetKwInOut{Input}{input}
\SetKwInOut{Output}{output}
$\mathcal{P}^{\mathrm{CF}} \leftarrow$ Select all non-dominated points from {\footnotesize $\mathcal{P}^1 \cup \cdots \cup \mathcal{P}^K$}\;
$\mathbf{p}^{\mathrm{c}} \leftarrow \argmin_{\mathbf{p} \in \mathcal{P}^{\mathrm{CF}}} \{\mathrm{dist}(\mathbf{p}, \mathbf{z})\}$\;
$\mathcal{R}^{\mathrm{pref}} \leftarrow \{\mathbf{p}  \in \mathbb{R}^m \, | \, \mathrm{dist}(\mathbf{p}, \mathbf{p}^{\mathrm{c}}) < r\}$\;
\For{$i \in \{1, \dots, K\}$}{
  $\mathrm{IGD}$-$\mathrm{CF}(\mathcal{P}^i) \leftarrow \mathrm{IGD}(\mathcal{P}^i \cap \mathcal{R}^{\mathrm{pref}})$ using $\mathcal{P}^{\mathrm{CF}}$ as $\mathcal{S}$\;
  $\mathrm{HV}$-$\mathrm{CF}(\mathcal{P}^i) \leftarrow \mathrm{HV}(\mathcal{P}^i \cap \mathcal{R}^{\mathrm{pref}})$\;
}
\caption{IGD-CF and HV-CF}
\label{alg:upcf}
\end{algorithm}

\subsection{PMDA}
\label{sec:pmda}

The preference-based metric based on specified distances and angles (PMDA)~\cite{YuZL15} is built upon the concept of light beams~\cite{JaszkiewiczS99}. It consists of three algorithmic steps.
\begin{enumerate}[Step 1:]
	\item It lets a set of points $\mathcal{Q}=
\{\mathbf{q}_i\}_{i=1}^{m+1}$ on a hyperplane passing through $\mathbf{z}$ while $m+1$ light beams pass from the origin $(0,\ldots,0)^\top$ to $\mathbf{q}_1,\ldots,\mathbf{q}_{m+1}$, respectively. For $i\in\{1,\ldots,m\}$, $\mathbf{q}_i$ is given as:
\begin{equation}
\label{eqn:pmda_lb}
	\mathbf{q}_i=\mathbf{z}+\alpha\, (\mathbf{e}_i-\mathbf{z}),
\end{equation}
where $\mathbf{e}_i$ is the standard-basis vector for the $i$-th objective function, e.g., $\mathbf{e}_1=(1, 0)^{\top}$ and $\mathbf{e}_2=(0, 1)^{\top}$ for $m=2$. In \eqref{eqn:pmda_lb}, $\alpha$ controls the spread of the light beams. The remaining  $\mathbf{q}_{m+1}$ in $\mathcal{Q}$ is set to $\mathbf{z}$.
	
	\item All the points in $\mathcal{Q}$ are further shifted as:
	\begin{equation}
		\mathcal{Q}^\prime=\beta\mathcal{Q},
	\end{equation} 
	where $\beta$ is the minimum objective value in $\mathcal{P}^\prime$ for all $m$ objectives, i.e., $\beta=\underset{\mathbf{p}\in\mathcal{P}^\prime}{\min}\{\underset{i\in\{1,\ldots,m\}}{\min}p_i\}$\footnote{\cite{YuZL15} defined that $\beta$ is the minimum objective value of $\mathcal{P}$, not $\mathcal{P}'$. Since $\beta$ can be different for different point sets in this case, this version of PMDA is not reliable.}. Here, $\mathcal{P}^\prime$ is a set of points in $ \cup_{i=1}^K\mathcal{P}^i$ that are in a preferred region defined by $m$ light beams, which pass through $\mathbf{q}_1, \ldots, \mathbf{q}_m$ but do not pass through $\mathbf{q}_{m+1}=\mathbf{z}$.

	\item PMDA measures the distance between each point in $\mathcal{P}$ and its closest point in $\mathcal{Q}^\prime$ as:
\begin{align}
  \label{eqn:pmda}
  \mathrm{PMDA}(\mathcal{P}) = \frac{1}{|\mathcal{P}|} \sum_{\mathbf{p}\in\mathcal{P}} \left(\min_{\mathbf{q}\in\mathcal{Q}^\prime}\{\mathrm{dist}(\mathbf{p},\mathbf{q})\} + \gamma \theta_{\mathbf{p}}\right),
\end{align}
where $\gamma$ is a penalty value and was set to $1/\pi$ in~\cite{YuZL15}. In~\eqref{eqn:pmda}, $\theta_{\mathbf{p}}$ is an angle between $\mathbf{p}$ and $\mathbf{z}$. If $\mathbf{p}$ is in the preferred region defined by the $m$ light beams passing through $\mathbf{q}_1, \ldots, \mathbf{q}_m$, $\theta_{\mathbf{p}} = 0$. Thus, points outside the preferred region are penalized.
\end{enumerate}

A small PMDA value indicates that points in $\mathcal{P}$ are close to the $m+1$ points in $\mathcal{Q}^\prime$ and the preferred region. Thus, PMDA does not evaluate the diversity of $\mathcal{P}$. Note that all elements of a point are implicitly assumed to be positive in~\cite{YuZL15}.
%





\begin{algorithm}[t]
\small
\SetSideCommentRight
\SetKwInOut{Input}{input}
\SetKwInOut{Output}{output}
$\mathcal{P}^{\mathrm{all}} \leftarrow$ Select all non-dominated points from {\footnotesize $\mathcal{P}^1 \cup \cdots \cup \mathcal{P}^K$}\;
\For{$i \in \{1, \ldots, K\}$}{
  $\mathcal{P}^i \leftarrow  \mathcal{P}^i \cap \mathcal{P}^{\mathrm{all}}$\;
  $\mathbf{p}^{\mathrm{a}} \leftarrow \argmin_{\mathbf{p} \in \mathcal{P}^i} \left\{s(\mathbf{p}) \right\}$\;
  $\mathcal{P}^i \leftarrow  \left\{ \mathbf{p} \in \mathcal{P}^i \, | \, |p_j  -  p^{\mathrm{a}}_j| \leq r \: \text{for} \: j \in \{1, \ldots, m\}  \right\}$\;
$k \leftarrow \argmax_{j \in \{1, \ldots, m\}}\left(\frac{p^{\mathrm{a}}_j - z_j}{z^{\mathrm{w}}_j - z_j}\right)$\;  
  $\mathbf{p}^{\mathrm{iso}} \leftarrow \mathbf{z} + \left(\frac{p^{\mathrm{a}}_k - z_k}{z^{\mathrm{w}}_k - z_k}\right) (\mathbf{z}^{\mathrm{w}} - \mathbf{z})$\;  
  \For{$\mathbf{p} \in \mathcal{P}^i$}{
    $\mathbf{p} \leftarrow  \mathbf{p}  + (\mathbf{p}^{\mathrm{iso}} - \mathbf{p}^{\mathrm{a}})$\;
  }
  $\mathrm{R\textrm{-}IGD}(\mathcal{P}^i) \leftarrow \mathrm{IGD}(\mathcal{P}^i)$ using a trimmed $\mathcal{S}$\; 
  $\mathrm{R\textrm{-}HV}(\mathcal{P}^i) \leftarrow \mathrm{HV}(\mathcal{P}^i)$ using $\mathbf{z}^{\mathrm{w}}$\;
}
\caption{R-IGD and R-HV}
\label{alg:r-metric}
\end{algorithm}

\begin{figure}[t]
  \centering
  \subfloat[The trimming operation]{  
    \includegraphics[width=0.21\textwidth]{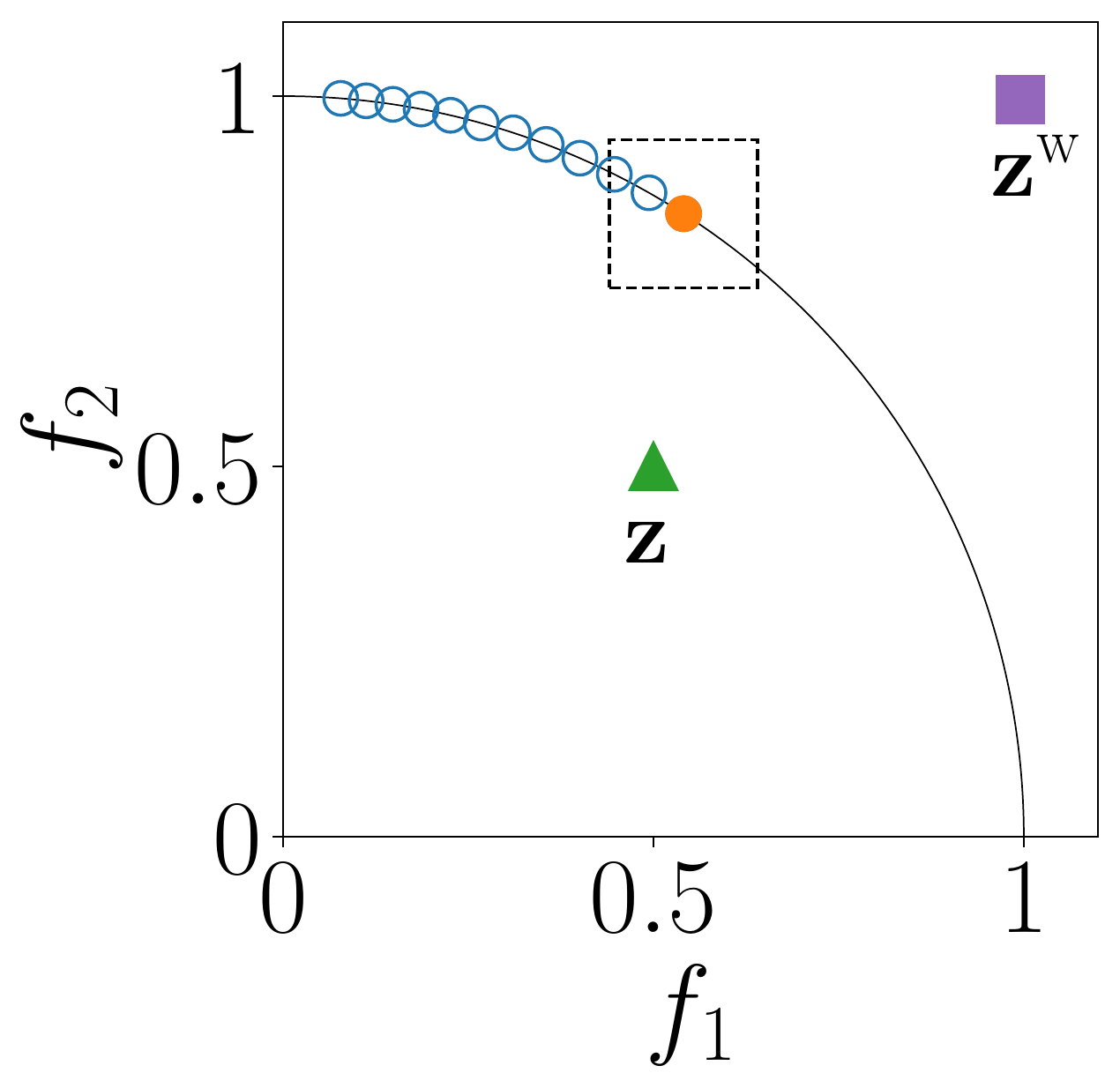}
  }
  \subfloat[The transfer operation]{
      \includegraphics[width=0.21\textwidth]{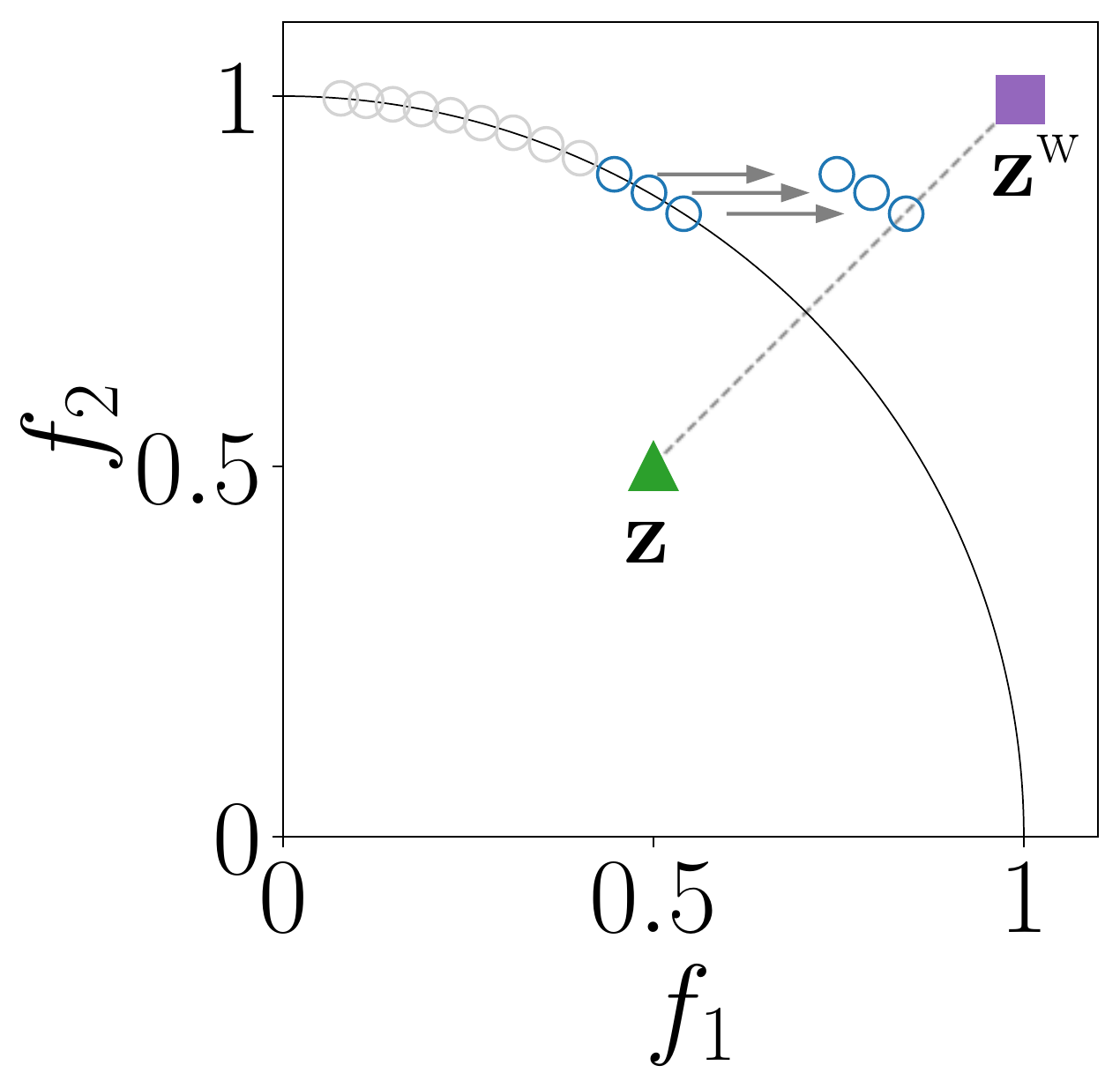}
  }
  \caption{Examples of the two operations in R-metric. In this example, $\mathbf{z}^{\mathrm{w}} = \mathbf{z} + 0.7 \times \mathbf{u}$.}
   \label{fig:rmetric}
\end{figure}

\subsection{R-IGD and R-HV}
\label{sec:rmetric}

R-metric \cite{LiDY18} is a framework that applies general quality indicators to the performance evaluation of $K$ PBEMO algorithms.
R-metric assumes that the DM prefers points along a line from $\mathbf{z}$ to the worst point $\mathbf{z}^{\mathrm{w}}$ defined by the DM.
As recommended in \cite{LiDY18}, we set $\mathbf{z}^{\mathrm{w}} = \mathbf{z} + 2 \times \mathbf{u}$, where $\mathbf{u}$ is a unit vector.
We set $\mathbf{u} = (1/\sqrt{m},\ldots,1/\sqrt{m})^{\top}$ in this study.
The previous study \cite{LiDY18} considered the R-metric versions of IGD and HV, denoted as R-IGD and R-HV.

\pref{alg:r-metric} gives the pseudo code for calculating R-IGD and R-HV.
A set of all non-dominated points $\mathcal{P}^{\mathrm{all}}$ are selected from the union of $K$ point sets (line $1$).
After that, the following steps are performed for each point set $\mathcal{P}^i$.
First, points dominated by any point in $\mathcal{P}^{\mathrm{all}}$ are removed from $\mathcal{P}^i$ (line $3$).
Then, the best point $\mathbf{p}^{\mathrm{a}}$ is selected from $\mathcal{P}^i$ in terms of the ASF (line $4$), where the previous study \cite{LiDY18} used a modified version of the ASF $s$ \cite{MaLQLJDWDHZW16}.
R-metric defines a preferred region based on a hypercube of size $2 \times r$ centered at $\mathbf{p}^{\mathrm{a}}$.
Points outside the preferred region are removed from $\mathcal{P}^i$  (line $5$).
For the example in \pref{fig:rmetric}(a), only the three points in the dotted box are considered for the R-metric calculation.
This trimming operation can penalize a point set that does not fit the preferred region.
Next, R-metric obtains a projection of $\mathbf{p}^{\mathrm{a}}$ on the line from $\mathbf{z}$ to $\mathbf{z}^{\mathrm{w}}$ by the ASF (lines $6$ and $7$).
This projection is called an iso-ASF point $\mathbf{p}^{\mathrm{iso}}$.
R-metric transfers all the points in $\mathcal{P}^i$ by the direction vector  from $\mathbf{p}^{\mathrm{iso}}$ to $\mathbf{p}^{\mathrm{a}}$ (lines $8$ and $9$). For the example shown in~\pref{fig:rmetric}(b), the three points are shifted horizontally.
This transfer operation redefines the convergence to the PF as the convergence to $\mathbf{z}$ along a line based on the DM's preference information.

Finally, the R-IGD and R-HV values of $\mathcal{P}^i$ are calculated  (lines $10$ and $11$).
More specifically, for R-IGD, the same trimming operation (lines $4$ and $5$) is first applied to the IGD-reference point set $\mathcal{S}$ in R-IGD.
Thus, all points in $\mathcal{S}$ are inside the preferred region.
Then, the IGD value of $\mathcal{P}^i$ is calculated using the trimmed $\mathcal{S}$.
For R-HV, $\mathbf{z}^{\mathrm{w}}$ is used as the HV-reference point $\mathbf{y}$.

\begin{algorithm}[t]
\small
\SetSideCommentRight
\SetKwInOut{Input}{input}
\SetKwInOut{Output}{output}
\For{$i \in \{1, \dots, K\}$}{
  $\mathcal{P}^i \leftarrow \{\mathbf{p} \in \mathcal{P} \, | \, \not\exists \mathbf{p}^{\mathrm{dup}} \in \mathcal{P} \: \text{s.t.} \: \mathbf{p}^{\mathrm{dup}}=\mathbf{p}\}$\;  
}
$\mathcal{P}^{\mathrm{all}} \leftarrow$ Select all non-dominated points from {\footnotesize $\mathcal{P}^1 \cup \cdots \cup \mathcal{P}^K$}\;
\For{$i \in \{1, \dots, K\}$}{
  $\mathcal{P}^i \leftarrow  \mathcal{P}^i \cap \mathcal{P}^{\mathrm{all}}$\;
}
$\mathbf{h}^{\mathrm{max}} \leftarrow \emptyset$\;
\For{$i \in \{1, \dots, K\}$}{
  $\mathbf{h} \leftarrow \emptyset$\;
  \For{$\mathbf{p} \in \mathcal{P}^i$}{
    $h \leftarrow \max_{j \in \{1, \dots, m\}} \{ |p_j - z_j| \}$\;
    $\mathbf{h} \leftarrow \mathbf{h} \cup \{h\}$\;
  }
  $\mathbf{h} \leftarrow$ Sort all elements in $\mathbf{h}$ in ascending order\;
  $h^{\mathrm{max}} \leftarrow \max_{h \in \mathbf{h}} \{h\}$\;
  $\mathbf{h}^{\mathrm{max}} \leftarrow \mathbf{h}^{\mathrm{max}} \cup \{h^{\mathrm{max}}\}$\;
  $a_{i} \leftarrow 0$\;
  \For{$l \in \{1, \dots, |\mathcal{P}^i|\}$}{
    $a_{i} \leftarrow a_{i} + \frac{l}{|\mathcal{P}^i|} \times (h_l - h_{l-1})$ \tcp*{$h_{0} = 0$}
  }
}
\For{$i \in \{1, \dots, K\}$}{
  \lIf{$\mathcal{P}^i = \emptyset$}{
    $\mathrm{EH}(\mathcal{P}^i) \leftarrow 0$
  }
  \lElse{
    $\mathrm{EH}(\mathcal{P}^i) \leftarrow a_{i} + \left(\max_{h \in \mathbf{h}^{\mathrm{max}}}\{h\} - h^{\mathrm{max}}_i\right)$
    }
}
\caption{EH}
\label{alg:eh}
\end{algorithm}

\subsection{EH}
\label{sec:eh}

The expanding hypercube metric (EH)~\cite{BandaruS19} is based on the size of a hypercube centered at $\mathbf{z}$ that contains each point and the fraction of points inside the hypercube.
While the former evaluates the convergence of a point set $\mathcal{P}$ to $\mathbf{z}$, the latter \textit{tries to} evaluate the diversity of $\mathcal{P}$.

The pseudo code of calculating the EH for $K$ point sets $\mathcal{P}^1,\ldots,\mathcal{P}^K$ is given in~\pref{alg:eh}. First, EH removes duplicated points for each point set (lines $1$ and $2$). In the meanwhile, it also removes dominated points from each point set (lines $3$ and $5$). Note that if a point set is empty after these removal operations, its EH value is set to $0$ (line $19$).

Then, the following steps are performed for each point set $\mathcal{P}^i$.
EH calculates the size of a hypercube centered at $\mathbf{z}$ that contains each point $\mathbf{p}$ in $\mathcal{P}^i$ (lines $10$ and $11$).
Thereafter, all elements in $\mathbf{h}$ are sorted (line $12$).
Note that $|\mathbf{h}| = |\mathcal{P}^i|$.
The maximum size $h^{\mathrm{max}}$ in $\mathbf{h}$ is maintained for an adjustment described later (lines $13$ and $14$).
EH calculates ``the area under the trade-off curve'' $a_i$ between the hypercube size and the fraction (lines $16$ and $17$).
While $l/|\mathcal{P}^i|$ is the fraction of points in the $l$-th hypercube, ``$(h_l - h_{l-1})$'' is the incremental size of the hypercube.
Finally, the EH value of each point set $\mathcal{P}^i$ is calculated by adjusting $a_i$ using $\mathbf{h}^{\mathrm{max}}$ (lines $18$ and $20$).

A large EH value means the corresponding $\mathcal{P}$ has good convergence to $\mathbf{z}$. 
Due to the operation for removing dominated points (line 3), EH also implicitly evaluates the convergence of $\mathcal{P}$ to the PF.
Since EH does not define a preferred region, EH fails to evaluate the diversity of $\mathcal{P}$ in some cases.
Let us consider a set of non-dominated unduplicated points that are close to $\mathbf{z}$ and distributed at intervals of $\Delta$.
EH is maximized when $\Delta$ is a positive value as close to zero as possible.
For the example shown in \pref{fig:toy}(a), EH prefers $\mathcal{P}^9$ to $\mathcal{P}^3$.

%% file: setup.tex

\section{Experimental Setup}
\label{sec:setting}

This section introduces the settings used in our experiments including the quality indicators, the benchmark problems, and preference-based point sets used in our analysis.

\subsection{Quality Indicators}
\label{sec:indicators}

In our experiments, we empirically analyze the performance and properties of $14$ quality indicators reviewed in \pref{sec:review_qi}. As a baseline, we also take the results of HV and IGD into account. In particular, the implementations of HV and R-IGD and R-HV are taken from \texttt{pygmo}~\cite{BiscaniI20} and \texttt{pymoo}~\cite{BlankD20}, respectively, while the other quality indicators are implemented by us in Python. The innate parameters of the $14$ quality indicators are set according to the recommendation in their original paper. For the IGD-based indicators, we uniformly generated $1\,000$ IGD-reference points on the PF of a problem. For those HV-based indicators, we set the HV-reference point $\mathbf{y}$ in HV and HV-CF to $(1.1,\ldots,1.1)^\top$. We set the radius $r$ of a preferred region to $0.1$ for all the quality indicators. We also set the radius $\zeta$ of the $\mathrm{ROI}^\mathrm{C}$ and $\mathrm{ROI}^\mathrm{A}$ to $0.1$.

\subsection{Benchmark Test Problems}
\label{sec:test_problems}

DTLZ1~\cite{DebTLZ05}, DTLZ2~\cite{DebTLZ05}, convDTLZ2~\cite{DebJ14} are chosen to constitute the benchmark test problems, which have linear, nonconvex, and convex PFs, respectively. To ensure the fairness of our experiments, the PF of the DTLZ1 problem is normalized to $[0,1]^m$. As a first attempt to investigate the properties of preference-based quality indicators, we mainly focus on the two-objective scenarios to facilitate the analysis and discussion about the impact of the distribution of points on the quality indicators.

\begin{remark}
We are aware of a previous study~\cite{BandaruS19} evaluated the performance of R-NSGA-II and g-NSGA-II on the DTLZ problems with $m\in\{3,5,8,10,15,20\}$ by EH and R-HV. The previous study discussed the influence of the distribution of $m$-dimensional points on EH and R-HV using the parallel coordinates plot. However, the parallel coordinates plot is likely to lead to a wrong conclusion~\cite{LiZY17}. In fact, the results in~\cite{BandaruS19} did not show the undesirable property of EH.
\end{remark}

\subsection{Experimental Settings}
\label{sec:experimental_setting}

We conduct two types of experiments.
\begin{itemize}
    \item One is an experiment using the $10$ synthetic point sets as shown in~\pref{fig:toy}. Fig.~\ref{supfig:toy} shows the distributions of the $10$ point sets on the PF of the DTLZ1 and convDTLZ2 problems. Fig.~\ref{supfig:toy} is similar to~\pref{fig:toy}.

    \item The other is an experiment using point sets found by the six PBEMO algorithms introduced in~\pref{sec:pemo_ref}. Note that comprehensive benchmarking of the PBEMO algorithms is beyond the scope of this paper. Instead, we focus on an analysis of the behavior of the PBEMO algorithms. This contributes to the understanding of \underline{RQ2}. Moreover, we also investigate how the choice of quality indicators influences the rankings of the PBEMO algorithms. This contributes to addressing \underline{RQ4}. In particular, the source code of the PBEMO algorithms are provided by Li~\cite{LiLDMY20} while the weight vectors used in MOEA/D-NUMS are generated by using the source code provided by Li~\cite{LiCMY18}. Each PBEMO algorithm is independently run $31$ times with different random seeds. The population size $\mu$ is set to $100$. The parameters associated with these PBEMO algorithms are set according to the recommendations in their original papers, except PBEA of which $\delta$ is set to $0.01$ in this study.
    Since a suitable $\delta$ value depends on the type of problems~\cite{ThieleMKL09}, we performed a rough-tuning of $\delta$ on our test problems.
\end{itemize}

%% file: results.tex
\begin{figure*}[t]
   \centering
\subfloat[100 points]{  
  \includegraphics[width=0.18\textwidth]{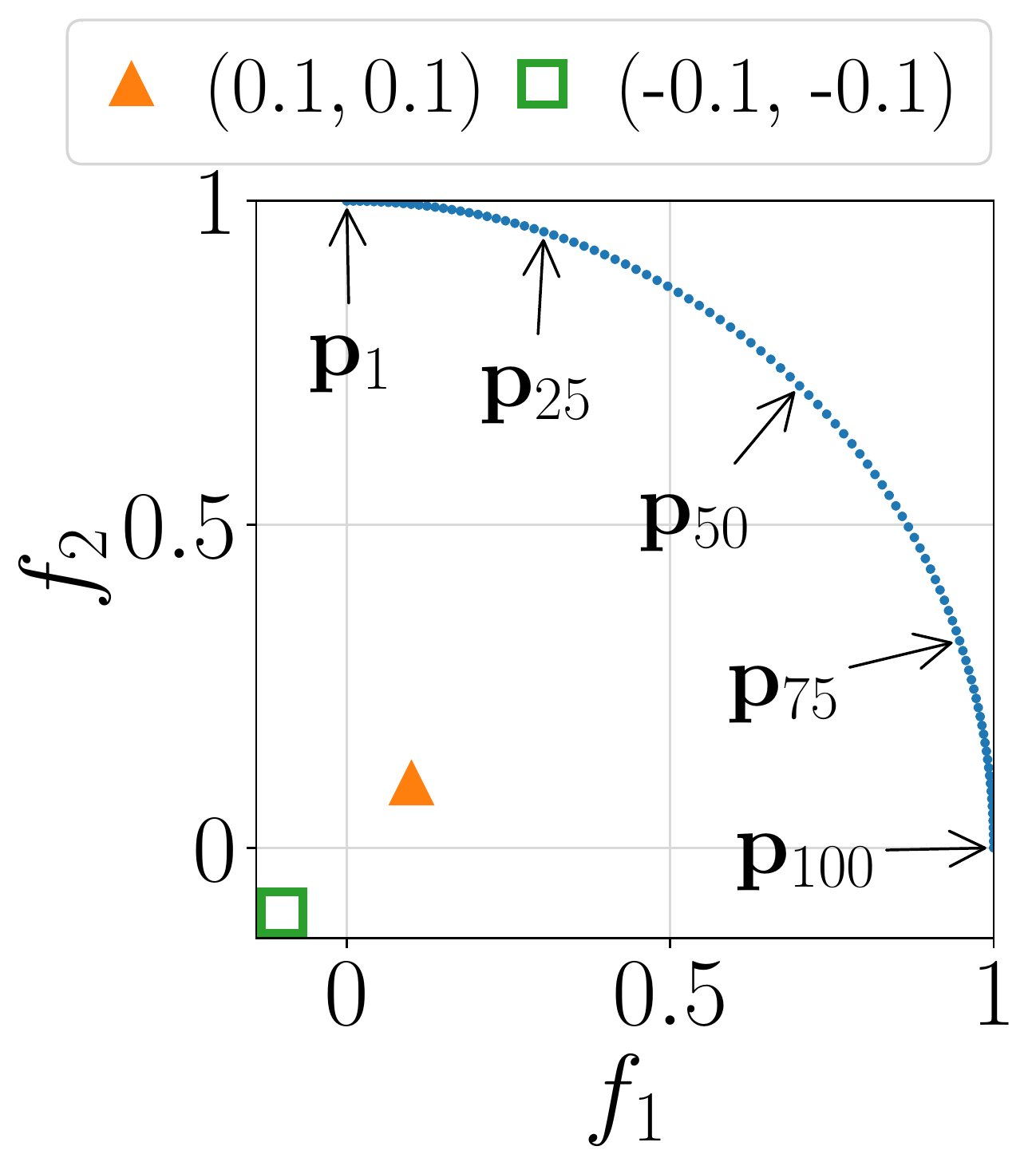}
}
\subfloat[Rankings  (distance)]{
  \includegraphics[width=0.22\textwidth]{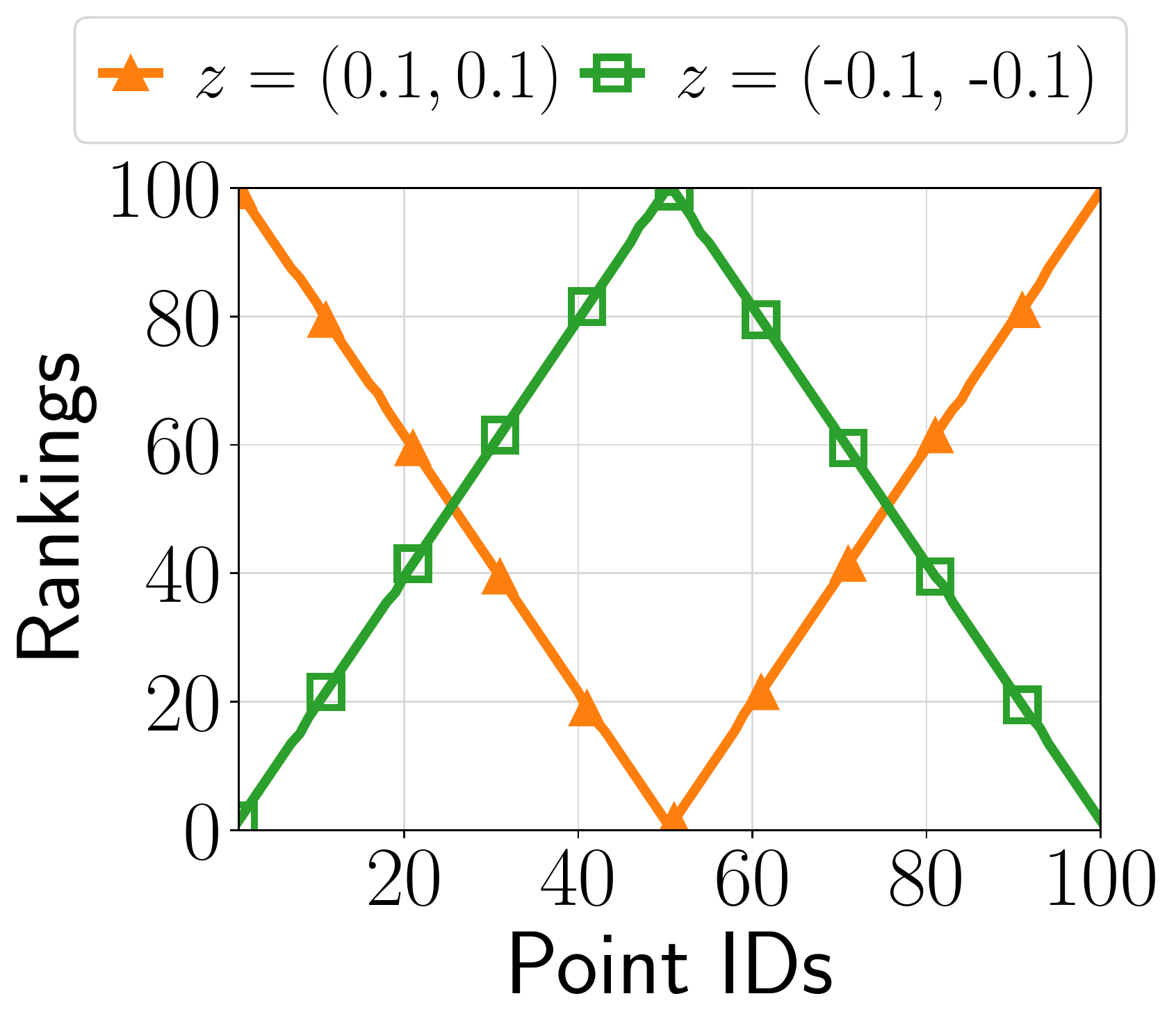}
}
\subfloat[Rankings (ASF)]{  
  \includegraphics[width=0.22\textwidth]{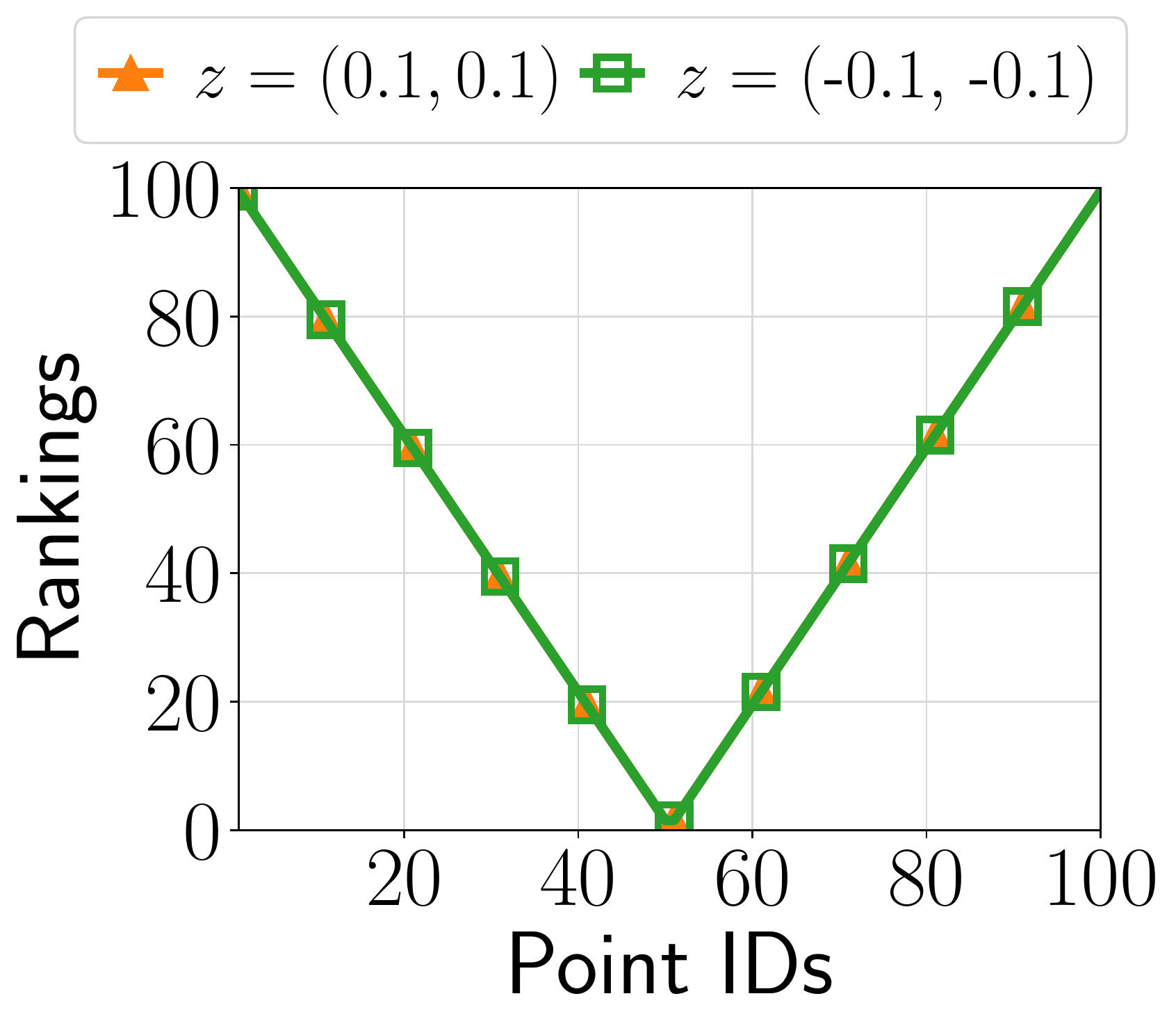}
}
\subfloat[Kendall $\tau$]{  
  \includegraphics[width=0.22\textwidth]{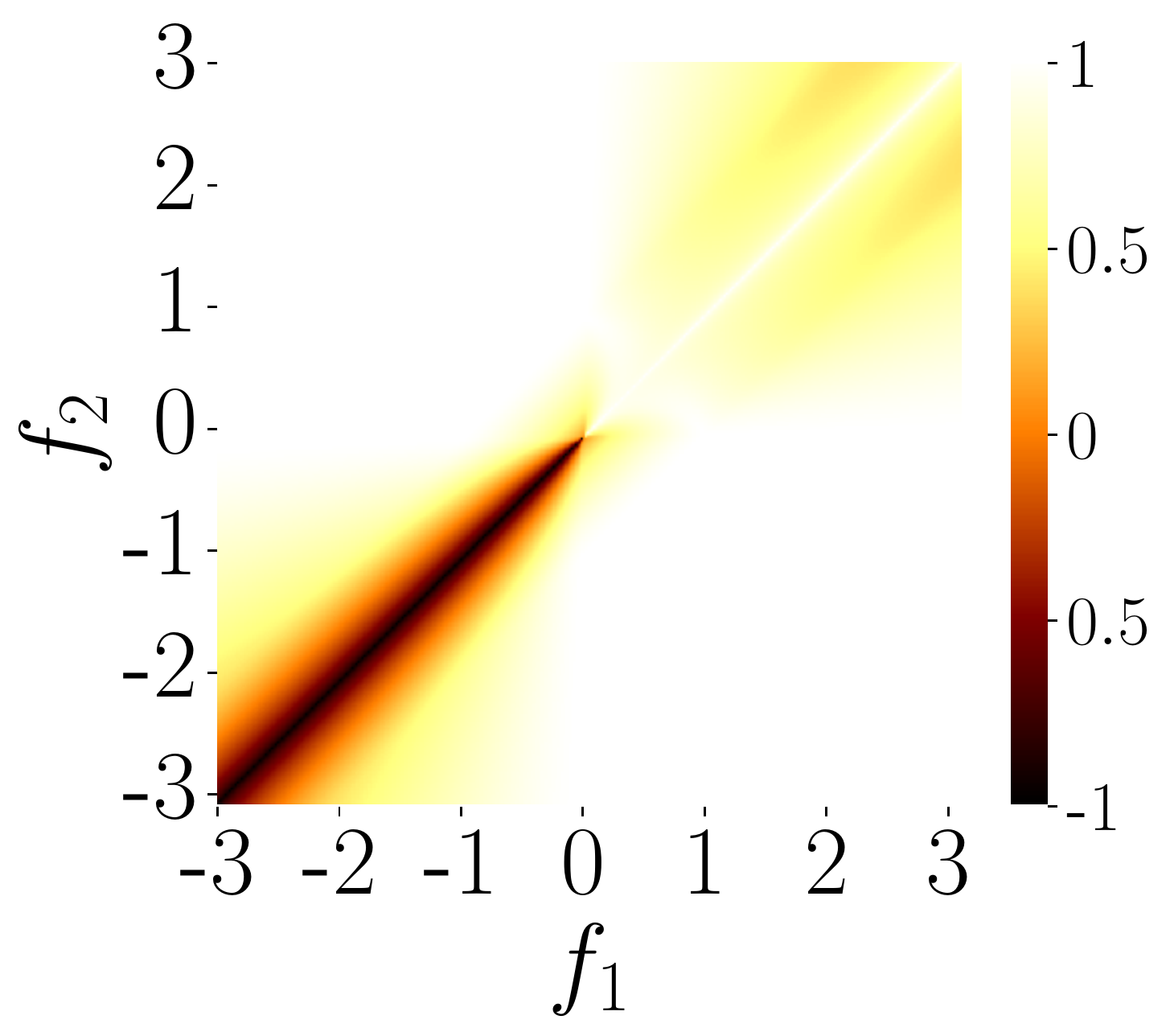}
}
\caption{(a) Distribution of 100 uniformly distributed points, (b) the rankings of the 100 points by the distance, (c) the ranking of the 100 points by the ASF, and (d) the Kendall $\tau$ values on the DTLZ2 problem, where \taborange{$\blacktriangle$} and \tabgreen{$\Box$} are the reference points.}
   \label{fig:100points}
\end{figure*}

\section{Results}
\label{sec:results}

This section is dedicated to addressing the four RQs raised in~\pref{sec:introduction}.
First, \pref{sec:dist_asf} analyzes the relation between the distance to the reference point $\mathbf{z}$ and the ASF value.
Then, \pref{sec:exp_inconsistency} investigates differences in the three ROIs and the behavior of EMO algorithms.
Thereafter, \pref{sec:exp_qi} examines the properties of the $14$ quality indicators using the synthetic point sets shown in Figs. \ref{fig:toy} and \ref{supfig:toy}.
\pref{sec:exp_rankings} analyzes the influence of quality indicators on the rankings of EMO algorithms.
Finally, \pref{sec:addtional_exp} considers the three-objective problems and a discontinuous ROI.


\subsection{Relation between the distance to $\mathbf{z}$ and the ASF value}
\label{sec:dist_asf}

\pref{fig:100points}(a) shows $100$ uniformly distributed points $\mathbf{p}_1,\ldots,\mathbf{p}_{100}$ on the PF of the DTLZ2 problem.
\pref{fig:100points}(a) also shows the two reference points $\mathbf{z}^{0.1} = (0.1, 0.1)^{\top}$ and $\mathbf{z}^{-0.1} = (-0.1, -0.1)^{\top}$.
While $\mathbf{z}^{0.1}$ is dominated by the ideal point, $\mathbf{z}^{-0.1}$ dominates the ideal point.

As shown in~\pref{fig:100points}(a), \textit{intuitively}, the $50$-th point $\mathbf{p}_{50}$ on the center of the PF is closest to both $\mathbf{z}^{0.1}$ and $\mathbf{z}^{-0.1}$.
However, this intuition is incorrect.
Figs. \ref{fig:100points}(b) and \ref{fig:100points}(c) show the rankings of the $100$ points by the Euclidean distance to $\mathbf{z}$ and the ASF in \eqref{eqn:asf1}, respectively.
A low ranking means that the corresponding point is close to $\mathbf{z}$ or obtains a small ASF value.
As seen from~\pref{fig:100points}(b), $\mathbf{p}_{50}$ is closest to $\mathbf{z}^{0.1}$.
In contrast, $\mathbf{p}_{50}$ is farthest from $\mathbf{z}^{-0.1}$.
The two extreme points ($\mathbf{p}_{1}$ and $\mathbf{p}_{100}$) are closest to $\mathbf{z}^{-0.1}$.
Thus, the closest points to $\mathbf{z}^{0.1}$ and $\mathbf{z}^{-0.1}$ are different.
As shown in~\pref{fig:100points}(c), the rankings by the ASF are consistent when using either one of $\mathbf{z}^{0.1}$ and $\mathbf{z}^{-0.1}$.
This is because both $\mathbf{z}^{0.1}$ and $\mathbf{z}^{-0.1}$ are in the same direction.

\pref{fig:100points}(d) shows the Kendall rank correlation $\tau$ value~\cite{Kendall38} of the distance to $\mathbf{z}$ and the ASF value, where $\tau\in[-1,1]$.
The $\tau$ value measures the similarity of the two rankings of the 100 points by the distance and the ASF.
A large positive $\tau$ value indicates that the two rankings are consistent.
In contrast, a small negative $\tau$ value indicates that the two rankings are inconsistent.
If $\tau = 0$, the two rankings are perfectly independent.
In \pref{fig:100points}(d), we uniformly generated $\mathbf{z}$ from $(-3, 3)^\top$ to $(3, -3)^\top$ at intervals of $0.01$. 
Then, we calculated the $\tau$ value for each $\mathbf{z}$.
The $\tau$ value quantifies the consistency of the two rankings, where one is based on the distance to the corresponding $\mathbf{z}$, and the other is based on the ASF value.
Positive and negative $\tau$ values indicate that the two rankings are consistent and inconsistent, respectively.
As seen from Fig. \ref{fig:100points}(d), the rankings by the distance to $\mathbf{z}$ and the ASF value are inconsistent when setting $\mathbf{z}$ close to the line passing through $(0,0)^{\top}$ and $(-3,-3)^{\top}$.
We can also see that the rankings are weakly inconsistent when setting $\mathbf{z}$ to other positions.


Note that the inconsistency between the distance to $\mathbf{z}$ and the ASF value depends on not only the position of $\mathbf{z}$, but also the shape of the PF.
Figs. \ref{supfig:100points_dtlz1} and \ref{supfig:100points_convdtlz2} show the results on the DTLZ1 and convDTLZ2 problems, respectively.
As shown in Fig. \ref{supfig:100points_convdtlz2}(a), we set $\mathbf{z}$ to $(2,2)^\top$ instead of $(-0.1,-0.1)^\top$ for the convDTLZ2 problem.
As shown in Figs. \ref{supfig:100points_dtlz1}(b) and (c), the rankings by the distance to $\mathbf{z}$ and the ASF value on the DTLZ1 problem are always consistent regardless of the position of $\mathbf{z}$.
In contrast, as seen from Figs. \ref{supfig:100points_convdtlz2}(b) and (c), the inconsistency of the rankings can be observed on the convDTLZ2 problem.
While Fig. \ref{supfig:100points_convdtlz2}(b) is similar to~\pref{fig:100points}(b), Fig. \ref{supfig:100points_convdtlz2}(d) is opposite from~\pref{fig:100points}(d).
Unlike~\pref{fig:100points}(d), Fig. \ref{supfig:100points_convdtlz2}(d) indicates that the inconsistency between the two rankings occurs when $\mathbf{z}$ is dominated by the nadir point $\mathbf{p}^{\mathrm{nadir}}$ on the convex PF.




\begin{tcolorbox}[sharpish corners, top=2pt, bottom=2pt, left=4pt, right=4pt, boxrule=0.0pt, colback=black!5!white,leftrule=0.75mm,]
    \textbf{\underline{Answers to RQ1:}} \textit{
        Our results show that the closest Pareto-optimal point to the reference point $\mathbf{z}$ does not always minimize the ASF. Although it has been believed that minimizing the ASF means moving closer to $\mathbf{z}$, this is not always correct. We observed that the inconsistency between the distance to $\mathbf{z}$ and the ASF value depends on the position of $\mathbf{z}$ and the shape of the PF. Roughly speaking, the inconsistency can be observed when $\mathbf{z}$ dominates the ideal point on a problem with a nonconvex PF, and $\mathbf{z}$ is dominated by the nadir point on a problem with the convex PF.
        }
\end{tcolorbox}




\begin{figure}[t]
   \centering
   \subfloat[$\mathrm{ROI}^\mathrm{C}$ ($\mathbf{z}^{0.1}$)]{  
    \includegraphics[width=0.145\textwidth]{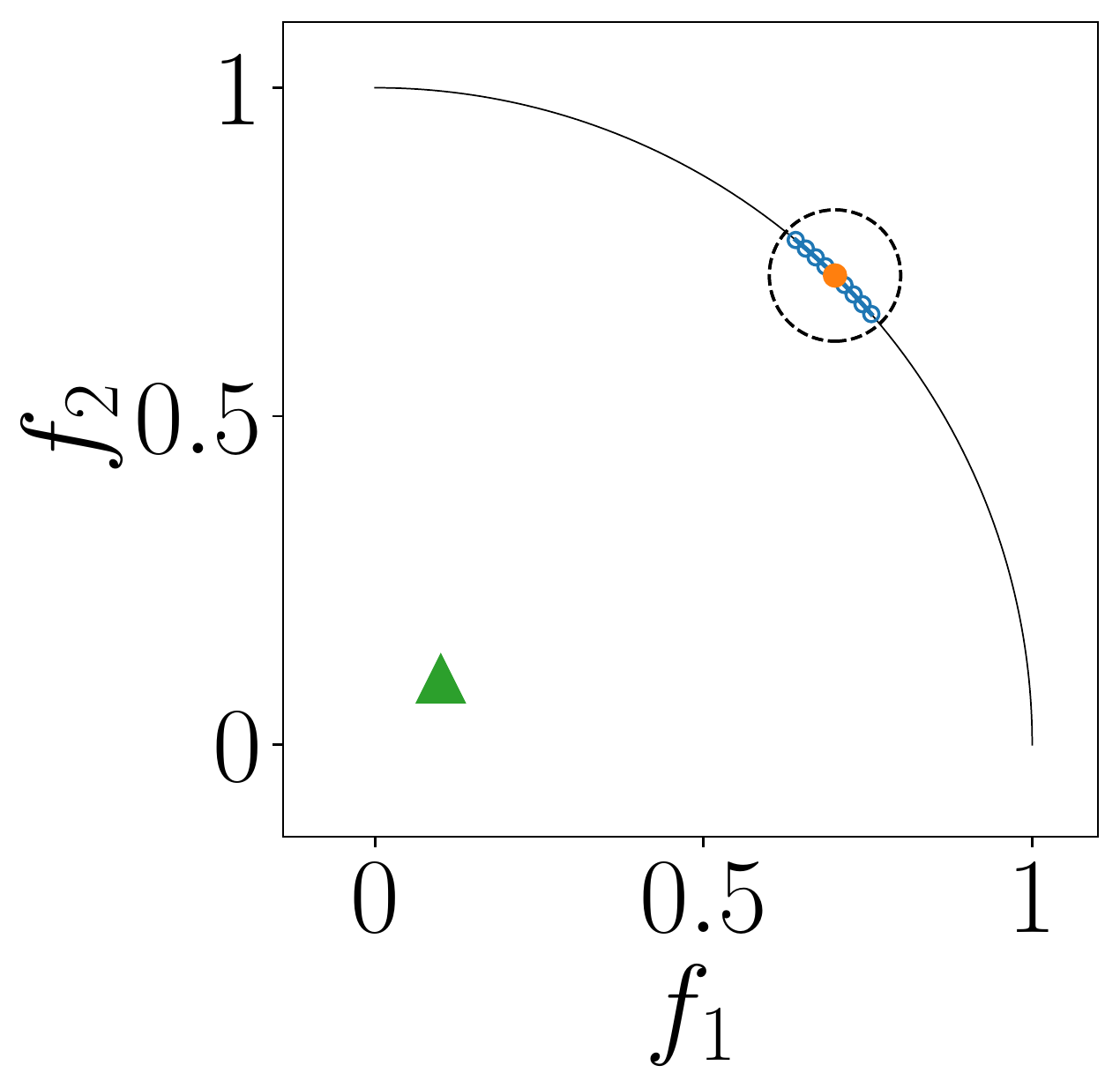}
    }
    \subfloat[$\mathrm{ROI}^\mathrm{A}$ ($\mathbf{z}^{0.1}$)]{  
    \includegraphics[width=0.145\textwidth]{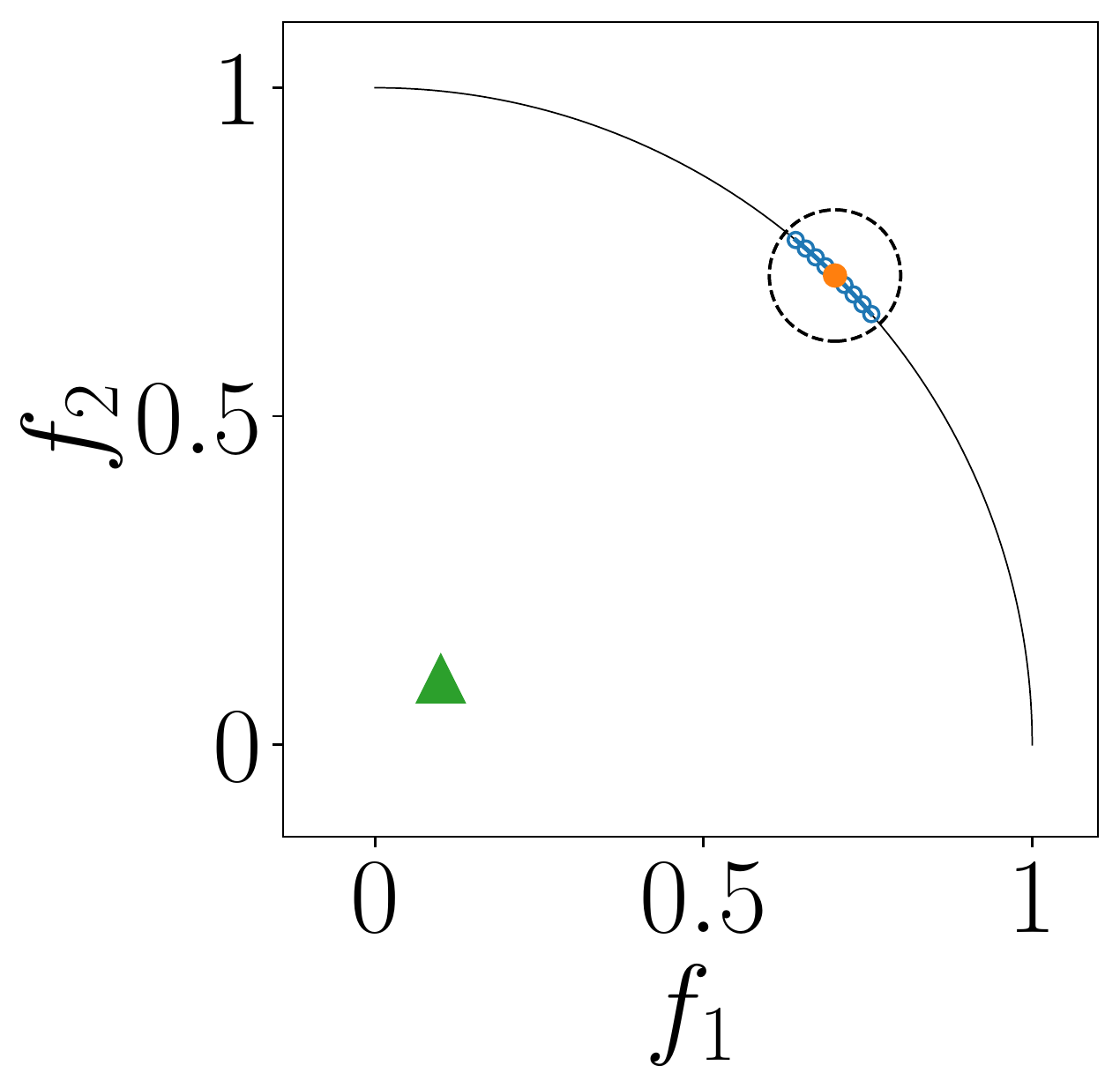}
    }
    \subfloat[$\mathrm{ROI}^\mathrm{P}$ ($\mathbf{z}^{0.1}$)]{  
    \includegraphics[width=0.145\textwidth]{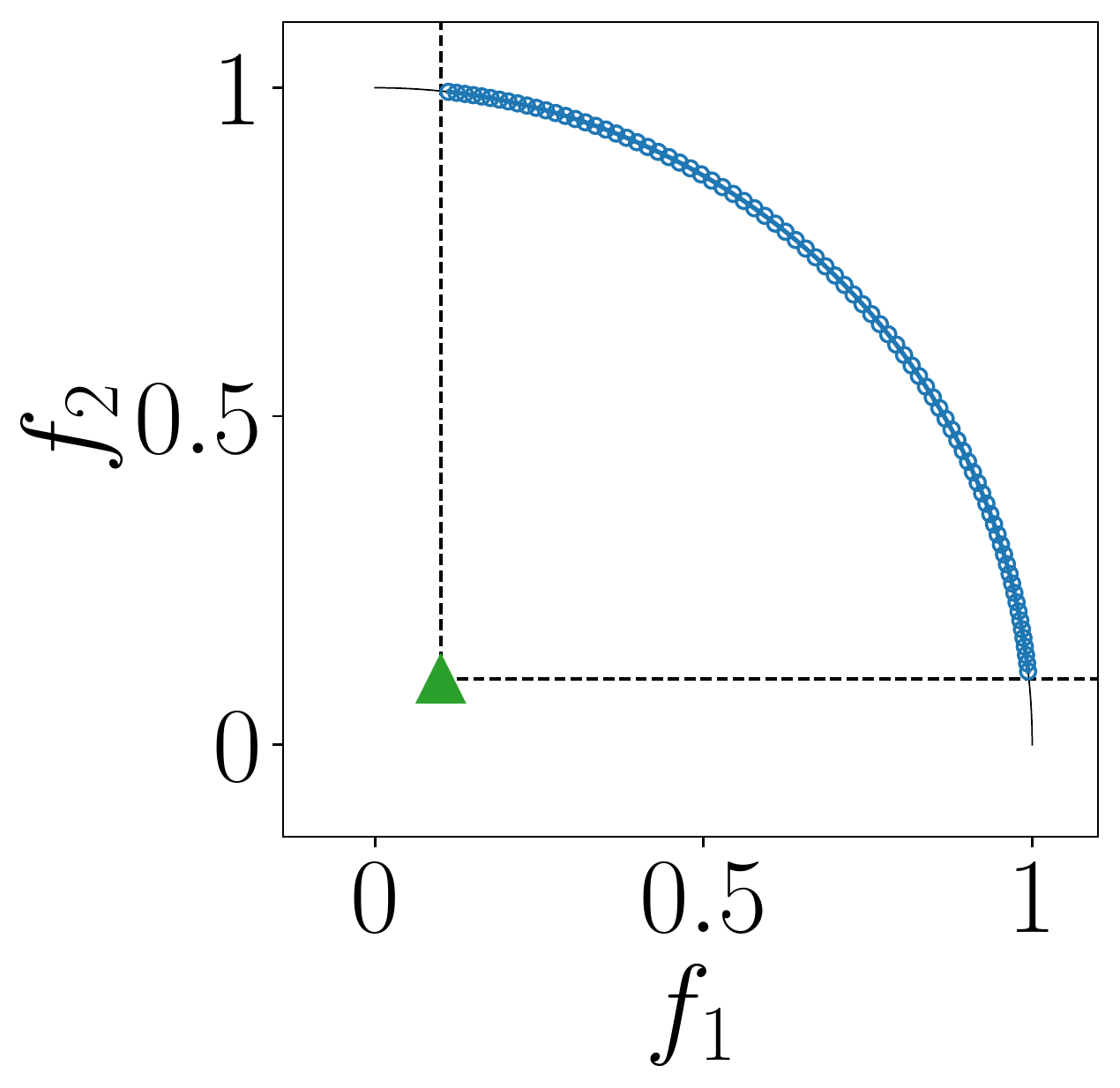}
    }
    \\
   \subfloat[$\mathrm{ROI}^\mathrm{C}$ ($\mathbf{z}^{-0.1}$)]{  
    \includegraphics[width=0.145\textwidth]{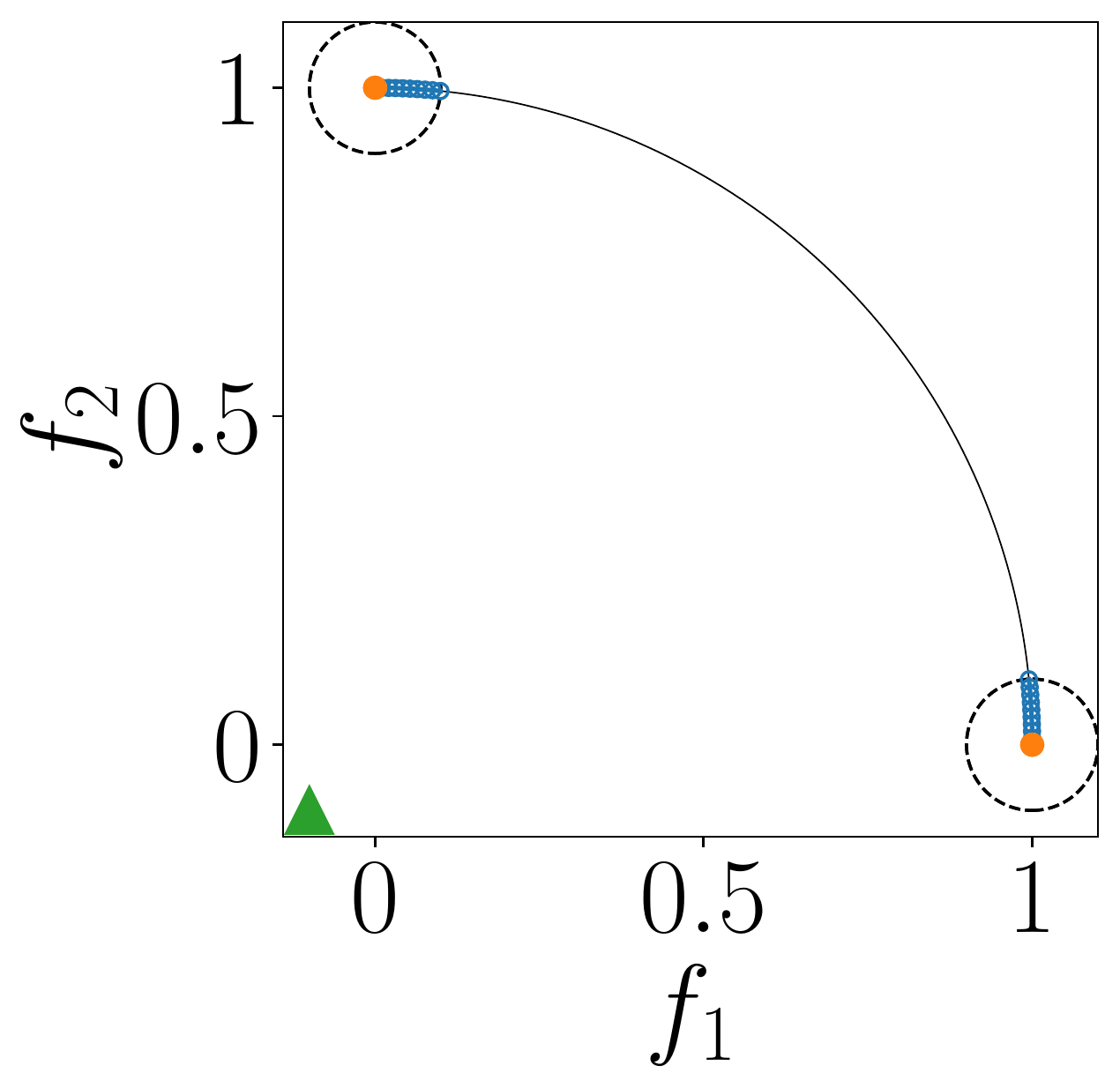}
    }
    \subfloat[$\mathrm{ROI}^\mathrm{A}$ ($\mathbf{z}^{-0.1}$)]{  
    \includegraphics[width=0.145\textwidth]{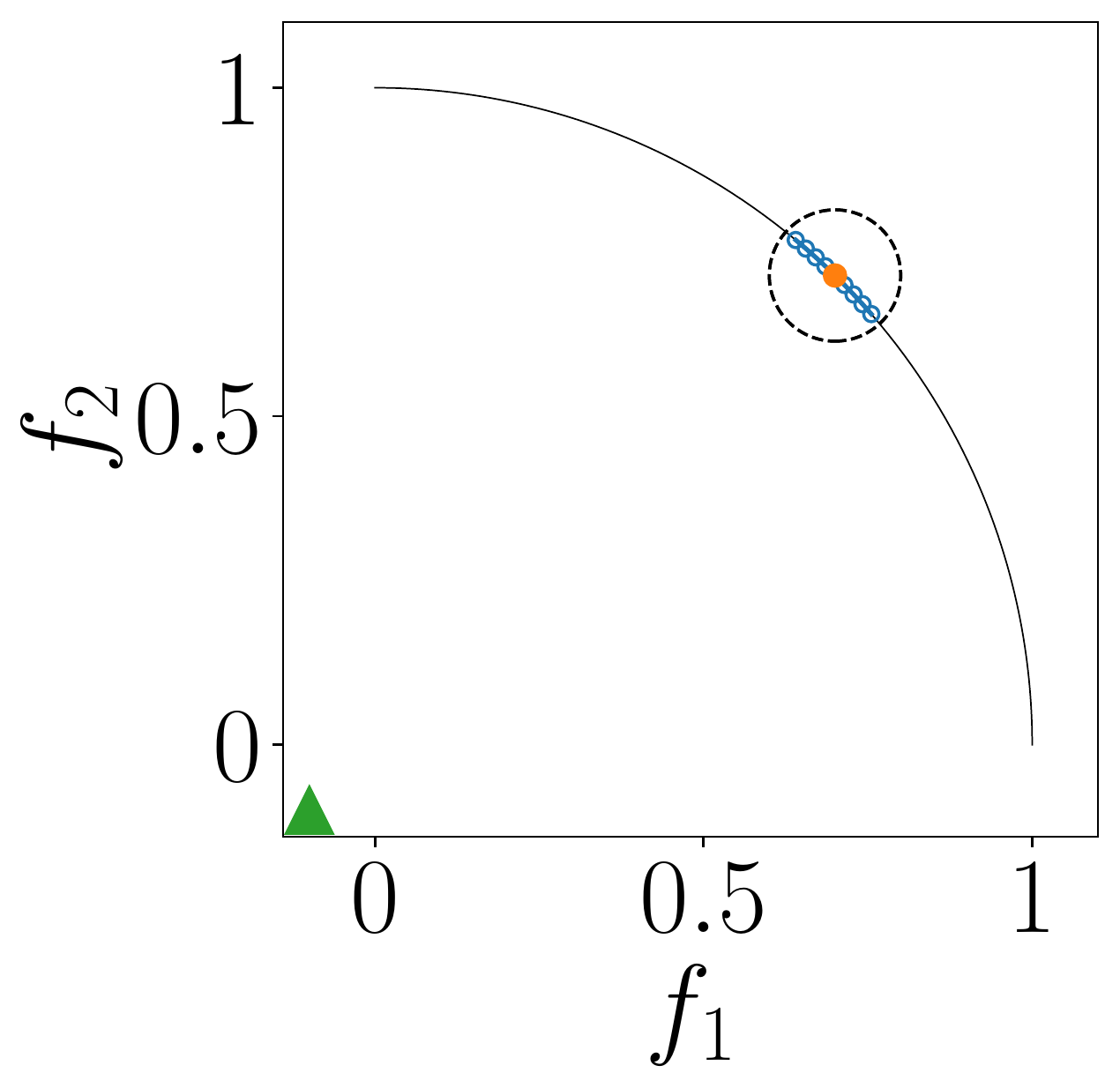}
    }
    \subfloat[$\mathrm{ROI}^\mathrm{P}$ ($\mathbf{z}^{-0.1}$)]{  
    \includegraphics[width=0.145\textwidth]{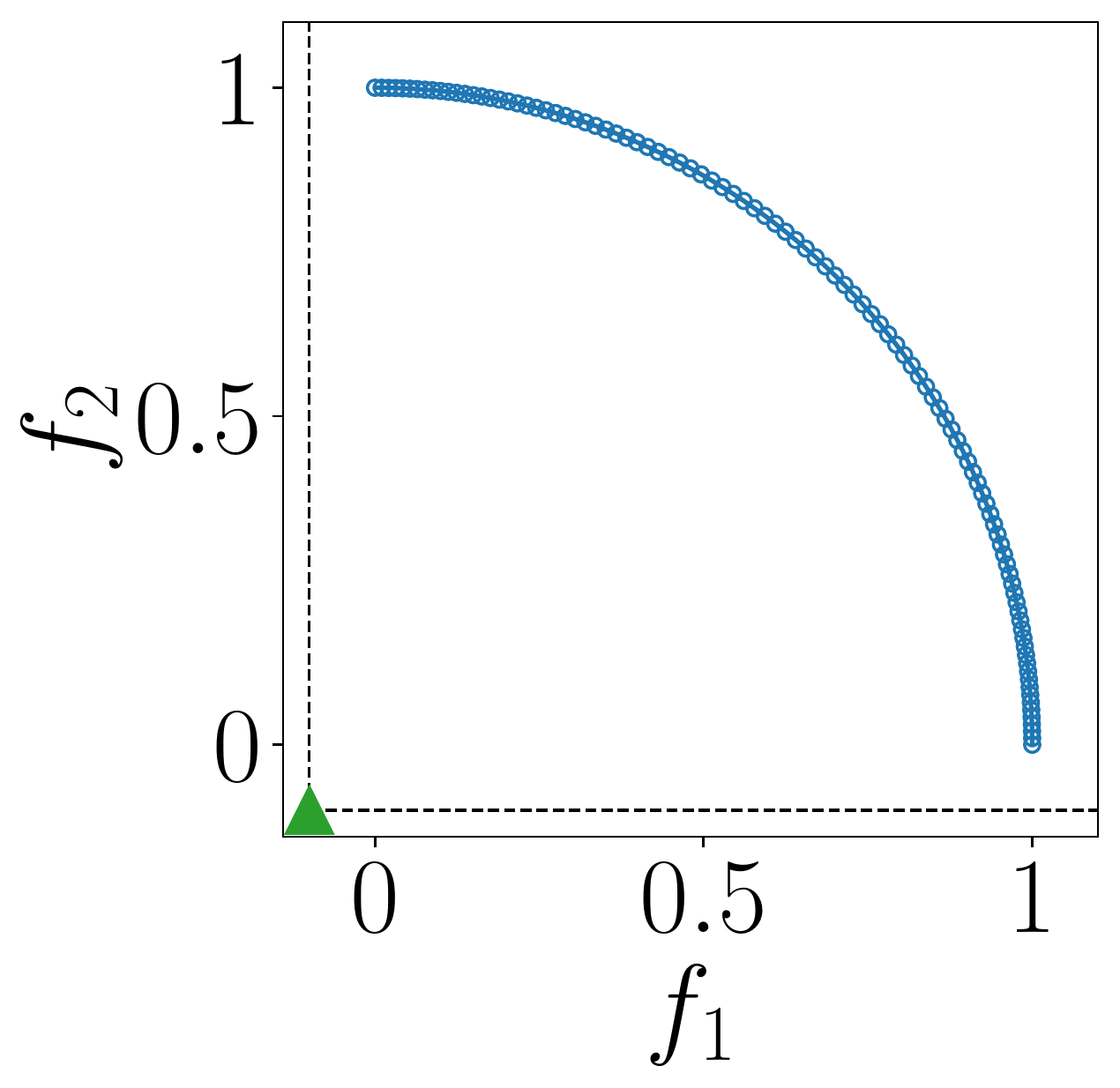}
    }
     \caption{Distributions of Pareto optimal points in the three ROIs on the DTLZ2 problem when using $\mathbf{z}^{0.1}$ and $\mathbf{z}^{-0.1}$, where \tabgreen{$\blacktriangle$} is the reference point $\mathbf{z}$, \taborange{$\blacksquare$} is $\mathbf{p}^{\mathrm{c}*}$ and $\mathbf{p}^{\mathrm{a}*}$.}
   \label{fig:roi_influence}
%
%
   \subfloat[R-NSGA-II]{  
     \includegraphics[width=0.145\textwidth]{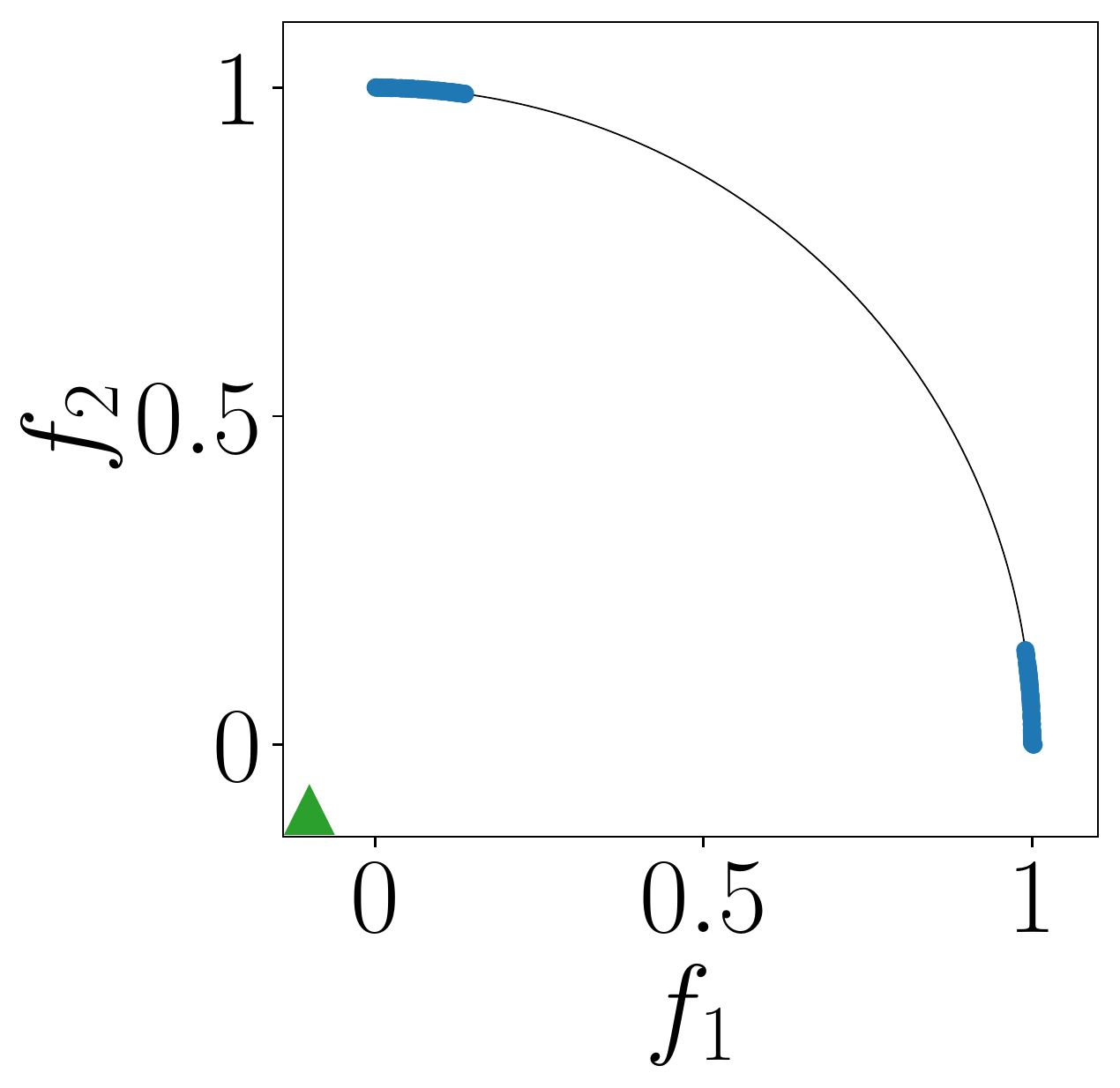}
     }
     \subfloat[MOEA/D-NUMS]{  
     \includegraphics[width=0.145\textwidth]{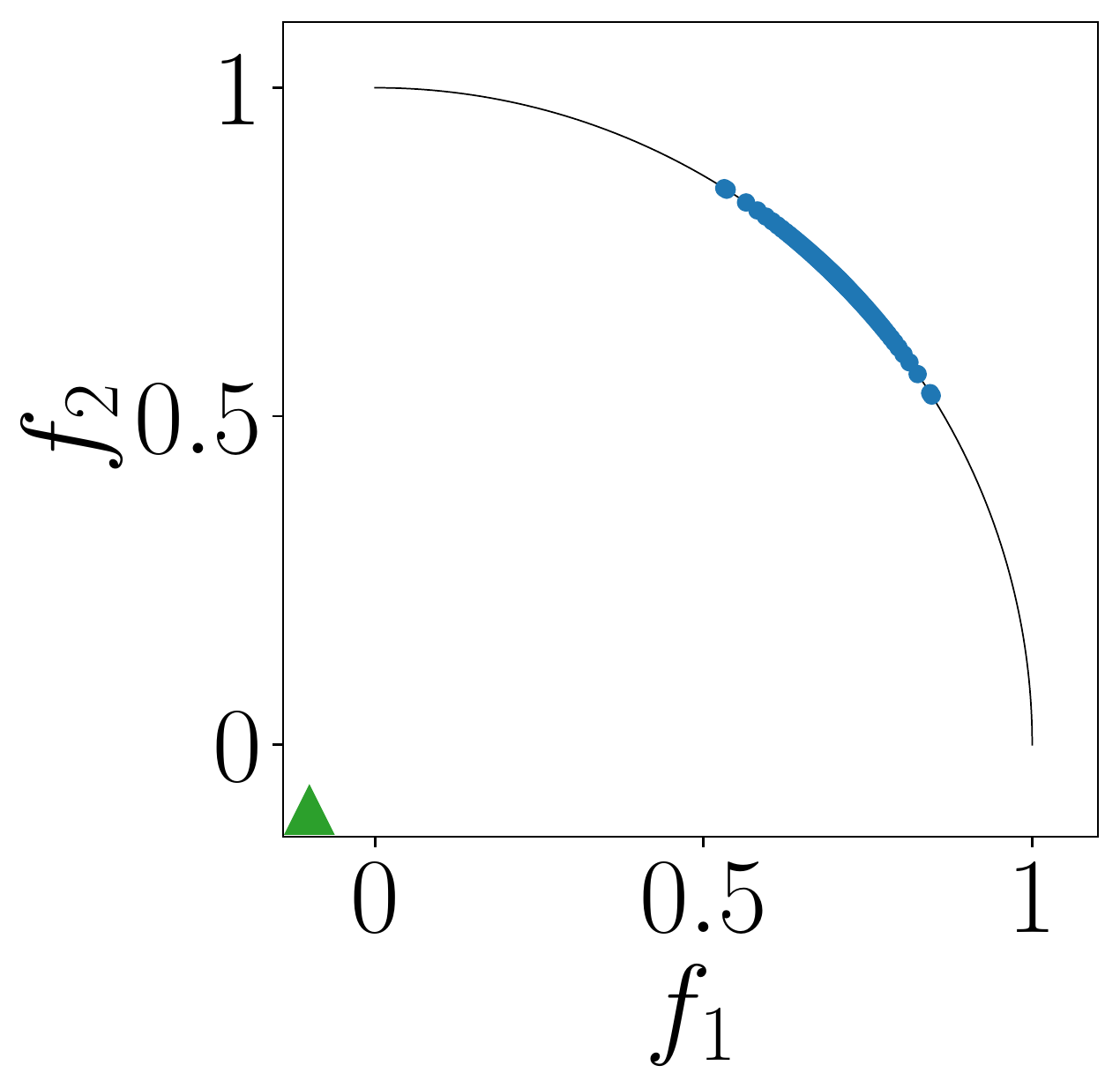}        
     }
   \subfloat[g-NSGA-II]{  
     \includegraphics[width=0.145\textwidth]{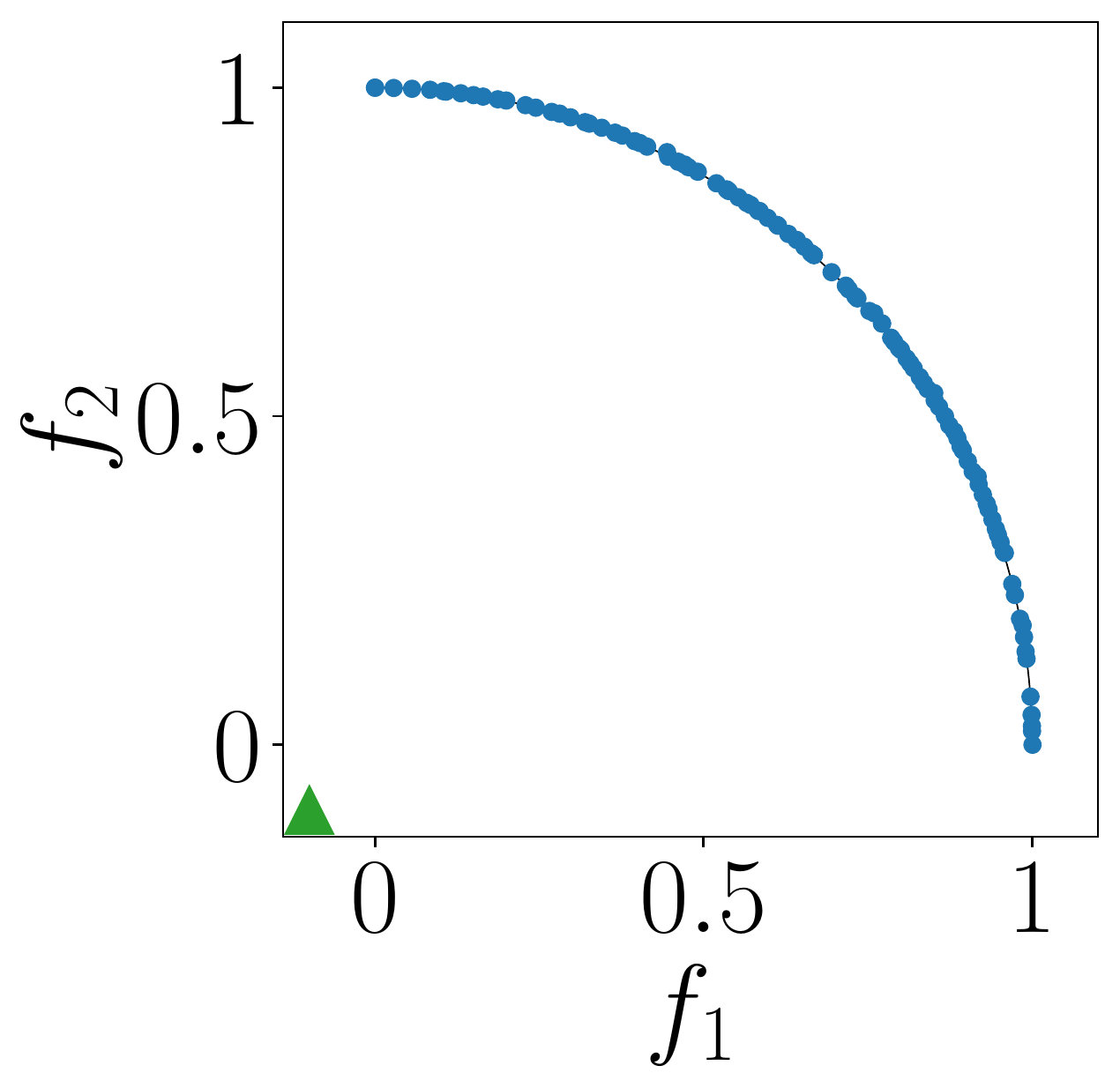}
     }
     \caption{Distributions of points found by three PBEMO algorithms on the DTLZ2 problem when using $\mathbf{z}^{-0.1}$, where \tabgreen{$\blacktriangle$} is the reference point $\mathbf{z}$.}
   \label{fig:emo_points_dtlz2_z-0.1}
\end{figure}

\subsection{Analysis of the three ROIs}
\label{sec:exp_inconsistency}




First, this section investigates the differences between the three ROIs.
Similar to~\pref{fig:roi}, \pref{fig:roi_influence} shows the distributions of Pareto optimal points in the three ROIs on the DTLZ2 problem when using $\mathbf{z}^{0.1}=(0.1,0.1)^\top$ and $\mathbf{z}^{-0.1}=(-0.1,-0.1)^\top$.
Figs.~\ref{supfig:roi_influence_dtlz1} and~\ref{supfig:roi_influence_convdtlz2} show the results on the DTLZ1 and convDTLZ2 problems, respectively.
Figs.~\ref{fig:roi_influence}(a) and (b) are exactly the same as Figs.~\ref{fig:roi}(a) and (b), respectively.
Thus, the $\mathrm{ROI}^\mathrm{C}$ and $\mathrm{ROI}^\mathrm{A}$ are the same even when using either one of $\mathbf{z}^{0.1}$ and $\mathbf{z}^{0.5}$.
In contrast, as shown in Figs.~\ref{fig:roi_influence}(d) and (e), the $\mathrm{ROI}^\mathrm{C}$ and $\mathrm{ROI}^\mathrm{A}$ are totally different when using $\mathbf{z}^{-0.1}$.
While the $\mathrm{ROI}^\mathrm{A}$ is on the center of the PF, the $\mathrm{ROI}^\mathrm{C}$ is on either one of the two extreme points $(1,0)^\top$ and $(0,1)^\top$.
Since the two extreme points $(1,0)^\top$ and $(0,1)^\top$ are equally close to $\mathbf{z}^{-0.1}$, \pref{fig:roi}(d) shows two possible ROIs.
This strange result is due to the inconsistency between the distance to $\mathbf{z}$ and the ASF value reported in~\pref{sec:dist_asf}.
As seen from Fig.~\ref{supfig:roi_influence_convdtlz2}(g), a similar result can be observed on the convDTLZ2 problem.
Results similar to those in~\pref{fig:roi_influence}(d) can be obtained by using $\mathbf{z}$ with a small Kendall $\tau$ value in~\pref{fig:100points}(d).


Fig. \ref{fig:roi_influence}(c) significantly differs from Fig. \ref{fig:roi}(c). 
The extent and cardinality of the $\mathrm{ROI}^\mathrm{P}$ in Fig. \ref{fig:roi_influence}(c) are  much larger than those in Fig. \ref{fig:roi}(c).
As shown in Fig. \ref{fig:roi_influence}(f), the $\mathrm{ROI}^\mathrm{P}$ and the PF are identical when the reference point dominates the ideal point.
The same is true when the reference point is dominated by the nadir point. 
In this case, preference-based multi-objective optimization is the same as general one.
The size of the $\mathrm{ROI}^\mathrm{P}$ increases as $\mathbf{z}$ moves away from the PF.
This undesirable property of the $\mathrm{ROI}^\mathrm{P}$ is similar to that of the g-dominance relation pointed out in~\cite{LiLDMY20}.
As seen from Figs.~\ref{supfig:roi_influence_dtlz1} and~\ref{supfig:roi_influence_convdtlz2}, this undesirable property of the $\mathrm{ROI}^\mathrm{P}$ can be observed on other problems.
Since the DM does not know any information about the PF in practice, it is difficult to set a reference point that is neither too close nor too far from the PF.


Next, we point out that the differences in the target ROIs caused the unexpected behavior of some PBEMO algorithms in~\cite{LiLDMY20}.
\pref{fig:emo_points_dtlz2_z-0.1} shows the final point sets found by R-NSGA-II, MOEA/D-NUMS, and g-NSGA-II on the DTLZ2 problem when using $\mathbf{z}^{-0.1}$.
The results in~\pref{fig:emo_points_dtlz2_z-0.1} are consistent with the results in~\cite{LiLDMY20}. 
Figs.~\ref{supfig:emo_points_dtlz1_z0.5}--\ref{supfig:emo_points_convdtlz2_z-0.1} show the results of the six PBEMO algorithms on the three problems.
As discussed in~\pref{sec:def_roi}, R-NSGA-II, MOEA/D-NUMS, and g-NSGA-II aim to approximate the $\mathrm{ROI}^\mathrm{C}$, $\mathrm{ROI}^\mathrm{A}$, and $\mathrm{ROI}^\mathrm{P}$, respectively.
As demonstrated here, the three ROIs can also be different.
For these reasons, the three EMO algorithms found different point sets, as shown in~\pref{fig:emo_points_dtlz2_z-0.1}.
Figs. \ref{supfig:rnsga2_dtlz2_z-0.1_multi_fevals}--\ref{supfig:gNSGA2_dtlz2_z-0.1_multi_fevals} show the distributions of points found by R-NSGA-
II, MOEA/D-NUMS, and g-NSGA-II at $1\,000$, $5\,000$, $10\,000$, $30\,000$, and $50\,000$ function evaluations, where the setting is the same as in \pref{fig:emo_points_dtlz2_z-0.1}.
As seen from Figs. \ref{supfig:rnsga2_dtlz2_z-0.1_multi_fevals} and \ref{supfig:gNSGA2_dtlz2_z-0.1_multi_fevals}, the point sets found by R-NSGA-II and g-NSGA-II are similar at $1\,000$ function evaluations.
However, point sets found by R-NSGA-II are biased to the $\mathrm{ROI}^\mathrm{C}$ after $5\,000$ function evaluations.
As shown in Fig. \ref{supfig:MOEADNUMS_dtlz2_z-0.1_multi_fevals}, MOEA/D-NUMS quickly finds a good approximation of the $\mathrm{ROI}^\mathrm{A}$.

The previous study~\cite{LiLDMY20} concluded that R-NSGA-II and g-NSGA-II failed to approximate the \lq\lq ROI\rq\rq\ when $\mathbf{z}$ is far from the PF.
However, this conclusion is not very correct.
Correctly speaking, as shown in Figs.~\ref{fig:emo_points_dtlz2_z-0.1}(a) and (c), R-NSGA-II and g-NSGA-II failed to approximate the \lq\lq $\mathrm{ROI}^\mathrm{A}$\rq\rq\ but succeeded in approximating the \lq\lq $\mathrm{ROI}^\mathrm{C}$\rq\rq\ and \lq\lq $\mathrm{ROI}^\mathrm{P}$\rq\rq, respectively.
%



%

\begin{tcolorbox}[sharpish corners, top=2pt, bottom=2pt, left=4pt, right=4pt, boxrule=0.0pt, colback=black!5!white,leftrule=0.75mm,]
    \textbf{\underline{Answers to RQ2:}} \textit{
        There are two takeaways generated from the analysis in this subsection. First, our results showed that the three ROIs can be significantly different depending on the position of the reference point $\mathbf{z}$ and the shape of the PF. We demonstrated that the $\mathrm{ROI}^\mathrm{A}$ is not always a subregion of the PF closest to $\mathbf{z}$ due to the inconsistency observed in~\pref{sec:dist_asf}. In addition, we found that the size of the $\mathrm{ROI}^\mathrm{P}$ significantly depends on the position of $\mathbf{z}$. Unless the DM knows the shape of the PF in advance, it would be better not to use the $\mathrm{ROI}^\mathrm{P}$. Second, we also demonstrated that the differences in the three ROIs could cause the unexpected behavior of PBEMO algorithms. For this reason, we argue the importance to clearly define a target ROI when benchmarking PBEMO algorithms and performing a practical decision-making.
        }
\end{tcolorbox}





 \arrayrulecolor{black}
 \setlength\arrayrulewidth{1.2pt}
 \setlength{\extrarowheight}{1.2pt}
 \newcolumntype{A}{>{\columncolor[rgb]{142, 153, 148}}p{4.8em}}

\newcolumntype{C}{>{\centering\arraybackslash}X}
\newcolumntype{L}{>{\raggedright\arraybackslash}X}
\newcolumntype{R}{>{\raggedleft\arraybackslash}X}


\definecolor{c10}{RGB}{62, 110, 92}
\definecolor{c9}{RGB}{73, 126, 105}
\definecolor{c8}{RGB}{80, 136, 113}
\definecolor{c7}{RGB}{90, 151, 125}
\definecolor{c6}{RGB}{100, 166, 138}
\definecolor{c5}{RGB}{110, 180, 149}
\definecolor{c4}{RGB}{120, 194, 160}
\definecolor{c3}{RGB}{130, 209, 173}
\definecolor{c2}{RGB}{140, 224, 185}
\definecolor{c1}{RGB}{153, 242, 200}

 \newcolumntype{A}{>{\columncolor[rgb]{0.9, 0.9, 0.9}\centering}p{1.8em}}

 \begin{table*}[t]
   \setlength{\tabcolsep}{2pt} 
\renewcommand{\arraystretch}{0.9}
  \centering
  \caption{\small Rankings of the 10 synthetic point sets on the DTLZ2 problem by the 16 quality indicators when using $\mathbf{z}^{0.5}$ and $\mathbf{z}^{-0.1}$.}
  \label{tab:rankings_psets}
        {\scriptsize
\subfloat[$\mathbf{z}^{0.5}=(0.5, 0.5)^{\top}$]{
\begin{tabularx}{70em}{|A|C|C|C|C|C|C|C|C|C|C|C|C|C|C|C|C|}
\hline
\rowcolor[rgb]{0.9, 0.9, 0.9} & MASF & MED & IGD-C & IGD-A & IGD-P & HV$_{\mathbf{z}}$ & PR & PMOD & IGD-CF & HV-CF & PMDA & R-IGD & R-HV & EH & HV & IGD \\
\hline
$\mathcal{P}^{1}$ & \cellcolor{c9}9 & \cellcolor{c10}10 & \cellcolor{c9}9 & \cellcolor{c9}9 & \cellcolor{c9}9 & \cellcolor{c5}5 & \cellcolor{c7}7 & \cellcolor{c7}7 & \cellcolor{c4}4 & \cellcolor{c4}4 & \cellcolor{c10}10 & \cellcolor{c6}6 & \cellcolor{c6}6 & \cellcolor{c6}6 & \cellcolor{c7}7 & \cellcolor{c9}9\\
$\mathcal{P}^{2}$ & \cellcolor{c5}5 & \cellcolor{c5}5 & \cellcolor{c5}5 & \cellcolor{c5}5 & \cellcolor{c5}5 & \cellcolor{c5}5 & \cellcolor{c7}7 & \cellcolor{c3}3 & \cellcolor{c4}4 & \cellcolor{c4}4 & \cellcolor{c5}5 & \cellcolor{c4}4 & \cellcolor{c4}4 & \cellcolor{c4}4 & \cellcolor{c3}3 & \cellcolor{c4}4\\
$\mathcal{P}^{3}$ & \cellcolor{c2}2 & \cellcolor{c2}2 & \cellcolor{c1}1 & \cellcolor{c1}1 & \cellcolor{c2}2 & \cellcolor{c1}1 & \cellcolor{c1}1 & \cellcolor{c1}1 & \cellcolor{c1}1 & \cellcolor{c1}1 & \cellcolor{c2}2 & \cellcolor{c1}1 & \cellcolor{c1}1 & \cellcolor{c2}2 & \cellcolor{c2}2 & \cellcolor{c2}2\\
$\mathcal{P}^{4}$ & \cellcolor{c5}5 & \cellcolor{c4}4 & \cellcolor{c6}6 & \cellcolor{c6}6 & \cellcolor{c5}5 & \cellcolor{c5}5 & \cellcolor{c7}7 & \cellcolor{c5}5 & \cellcolor{c4}4 & \cellcolor{c4}4 & \cellcolor{c4}4 & \cellcolor{c5}5 & \cellcolor{c4}4 & \cellcolor{c4}4 & \cellcolor{c3}3 & \cellcolor{c4}4\\
$\mathcal{P}^{5}$ & \cellcolor{c9}9 & \cellcolor{c9}9 & \cellcolor{c10}10 & \cellcolor{c10}10 & \cellcolor{c9}9 & \cellcolor{c5}5 & \cellcolor{c7}7 & \cellcolor{c9}9 & \cellcolor{c4}4 & \cellcolor{c4}4 & \cellcolor{c9}9 & \cellcolor{c7}7 & \cellcolor{c6}6 & \cellcolor{c6}6 & \cellcolor{c7}7 & \cellcolor{c9}9\\
$\mathcal{P}^{6}$ & \cellcolor{c7}7 & \cellcolor{c7}7 & \cellcolor{c7}7 & \cellcolor{c7}7 & \cellcolor{c7}7 & \cellcolor{c5}5 & \cellcolor{c4}4 & \cellcolor{c4}4 & \cellcolor{c4}4 & \cellcolor{c4}4 & \cellcolor{c8}8 & \cellcolor{c8}8 & \cellcolor{c8}8 & \cellcolor{c8}8 & \cellcolor{c9}9 & \cellcolor{c7}7\\
$\mathcal{P}^{7}$ & \cellcolor{c4}4 & \cellcolor{c3}3 & \cellcolor{c4}4 & \cellcolor{c4}4 & \cellcolor{c4}4 & \cellcolor{c4}4 & \cellcolor{c1}1 & \cellcolor{c2}2 & \cellcolor{c4}4 & \cellcolor{c4}4 & \cellcolor{c3}3 & \cellcolor{c8}8 & \cellcolor{c8}8 & \cellcolor{c8}8 & \cellcolor{c6}6 & \cellcolor{c3}3\\
$\mathcal{P}^{8}$ & \cellcolor{c7}7 & \cellcolor{c7}7 & \cellcolor{c8}8 & \cellcolor{c8}8 & \cellcolor{c7}7 & \cellcolor{c5}5 & \cellcolor{c4}4 & \cellcolor{c8}8 & \cellcolor{c4}4 & \cellcolor{c4}4 & \cellcolor{c7}7 & \cellcolor{c8}8 & \cellcolor{c8}8 & \cellcolor{c8}8 & \cellcolor{c10}10 & \cellcolor{c8}8\\
$\mathcal{P}^{9}$ & \cellcolor{c1}1 & \cellcolor{c1}1 & \cellcolor{c3}3 & \cellcolor{c3}3 & \cellcolor{c3}3 & \cellcolor{c3}3 & \cellcolor{c1}1 & \cellcolor{c6}6 & \cellcolor{c3}3 & \cellcolor{c3}3 & \cellcolor{c1}1 & \cellcolor{c2}2 & \cellcolor{c3}3 & \cellcolor{c1}1 & \cellcolor{c5}5 & \cellcolor{c6}6\\
$\mathcal{P}^{10}$ & \cellcolor{c3}3 & \cellcolor{c6}6 & \cellcolor{c2}2 & \cellcolor{c2}2 & \cellcolor{c1}1 & \cellcolor{c2}2 & \cellcolor{c6}6 & \cellcolor{c10}10 & \cellcolor{c2}2 & \cellcolor{c2}2 & \cellcolor{c6}6 & \cellcolor{c3}3 & \cellcolor{c2}2 & \cellcolor{c3}3 & \cellcolor{c1}1 & \cellcolor{c1}1\\
\hline
\end{tabularx}
}
\\
\subfloat[$\mathbf{z}^{-0.1}=(-0.1, -0.1)^{\top}$]{
\begin{tabularx}{70em}{|A|C|C|C|C|C|C|C|C|C|C|C|C|C|C|C|C|}
\hline
\rowcolor[rgb]{0.9, 0.9, 0.9} & MASF & MED & IGD-C & IGD-A & IGD-P & HV$_{\mathbf{z}}$ & PR & PMOD & IGD-CF & HV-CF & PMDA & R-IGD & R-HV & EH & HV & IGD \\
\hline
$\mathcal{P}^{1}$ & \cellcolor{c9}9 & \cellcolor{c2}2 & \cellcolor{c1}1 & \cellcolor{c9}9 & \cellcolor{c9}9 & \cellcolor{c8}8 & \cellcolor{c1}1 & \cellcolor{c7}7 & \cellcolor{c1}1 & \cellcolor{c1}1 & \cellcolor{c10}10 & \cellcolor{c6}6 & \cellcolor{c7}7 & \cellcolor{c6}6 & \cellcolor{c7}7 & \cellcolor{c9}9\\
$\mathcal{P}^{2}$ & \cellcolor{c5}5 & \cellcolor{c4}4 & \cellcolor{c3}3 & \cellcolor{c5}5 & \cellcolor{c4}4 & \cellcolor{c4}4 & \cellcolor{c1}1 & \cellcolor{c3}3 & \cellcolor{c3}3 & \cellcolor{c3}3 & \cellcolor{c5}5 & \cellcolor{c4}4 & \cellcolor{c4}4 & \cellcolor{c4}4 & \cellcolor{c3}3 & \cellcolor{c4}4\\
$\mathcal{P}^{3}$ & \cellcolor{c2}2 & \cellcolor{c6}6 & \cellcolor{c5}5 & \cellcolor{c1}1 & \cellcolor{c2}2 & \cellcolor{c2}2 & \cellcolor{c1}1 & \cellcolor{c1}1 & \cellcolor{c3}3 & \cellcolor{c3}3 & \cellcolor{c2}2 & \cellcolor{c1}1 & \cellcolor{c1}1 & \cellcolor{c2}2 & \cellcolor{c2}2 & \cellcolor{c2}2\\
$\mathcal{P}^{4}$ & \cellcolor{c5}5 & \cellcolor{c4}4 & \cellcolor{c8}8 & \cellcolor{c6}6 & \cellcolor{c4}4 & \cellcolor{c4}4 & \cellcolor{c1}1 & \cellcolor{c5}5 & \cellcolor{c3}3 & \cellcolor{c3}3 & \cellcolor{c4}4 & \cellcolor{c5}5 & \cellcolor{c4}4 & \cellcolor{c4}4 & \cellcolor{c3}3 & \cellcolor{c4}4\\
$\mathcal{P}^{5}$ & \cellcolor{c9}9 & \cellcolor{c1}1 & \cellcolor{c10}10 & \cellcolor{c10}10 & \cellcolor{c9}9 & \cellcolor{c7}7 & \cellcolor{c1}1 & \cellcolor{c9}9 & \cellcolor{c3}3 & \cellcolor{c3}3 & \cellcolor{c9}9 & \cellcolor{c7}7 & \cellcolor{c6}6 & \cellcolor{c6}6 & \cellcolor{c7}7 & \cellcolor{c9}9\\
$\mathcal{P}^{6}$ & \cellcolor{c7}7 & \cellcolor{c8}8 & \cellcolor{c4}4 & \cellcolor{c7}7 & \cellcolor{c7}7 & \cellcolor{c9}9 & \cellcolor{c1}1 & \cellcolor{c4}4 & \cellcolor{c3}3 & \cellcolor{c3}3 & \cellcolor{c8}8 & \cellcolor{c8}8 & \cellcolor{c8}8 & \cellcolor{c8}8 & \cellcolor{c9}9 & \cellcolor{c7}7\\
$\mathcal{P}^{7}$ & \cellcolor{c4}4 & \cellcolor{c10}10 & \cellcolor{c6}6 & \cellcolor{c4}4 & \cellcolor{c3}3 & \cellcolor{c6}6 & \cellcolor{c1}1 & \cellcolor{c2}2 & \cellcolor{c3}3 & \cellcolor{c3}3 & \cellcolor{c3}3 & \cellcolor{c8}8 & \cellcolor{c8}8 & \cellcolor{c8}8 & \cellcolor{c6}6 & \cellcolor{c3}3\\
$\mathcal{P}^{8}$ & \cellcolor{c7}7 & \cellcolor{c9}9 & \cellcolor{c9}9 & \cellcolor{c8}8 & \cellcolor{c8}8 & \cellcolor{c9}9 & \cellcolor{c1}1 & \cellcolor{c8}8 & \cellcolor{c3}3 & \cellcolor{c3}3 & \cellcolor{c7}7 & \cellcolor{c8}8 & \cellcolor{c8}8 & \cellcolor{c8}8 & \cellcolor{c10}10 & \cellcolor{c8}8\\
$\mathcal{P}^{9}$ & \cellcolor{c1}1 & \cellcolor{c7}7 & \cellcolor{c7}7 & \cellcolor{c3}3 & \cellcolor{c6}6 & \cellcolor{c3}3 & \cellcolor{c1}1 & \cellcolor{c6}6 & \cellcolor{c3}3 & \cellcolor{c3}3 & \cellcolor{c1}1 & \cellcolor{c2}2 & \cellcolor{c3}3 & \cellcolor{c1}1 & \cellcolor{c5}5 & \cellcolor{c6}6\\
$\mathcal{P}^{10}$ & \cellcolor{c3}3 & \cellcolor{c3}3 & \cellcolor{c2}2 & \cellcolor{c2}2 & \cellcolor{c1}1 & \cellcolor{c1}1 & \cellcolor{c1}1 & \cellcolor{c10}10 & \cellcolor{c2}2 & \cellcolor{c2}2 & \cellcolor{c6}6 & \cellcolor{c3}3 & \cellcolor{c2}2 & \cellcolor{c3}3 & \cellcolor{c1}1 & \cellcolor{c1}1\\
\hline
\end{tabularx}
}
}
\end{table*}

\subsection{Analysis of quality indicators}
\label{sec:exp_qi}



\pref{tab:rankings_psets} shows the rankings of the $10$ point sets $\mathcal{P}^1$ to $\mathcal{P}^{10}$ in~\pref{fig:toy} by each quality indicator when using $\mathbf{z}^{0.5} = (0.5, 0.5)^{\top}$ and $\mathbf{z}^{-0.1} = (-0.1, -0.1)^{\top}$.
For the sake of reference, we show the results of HV and IGD.
\pref{tab:rankings_psets} shows which point set is preferred by each quality indicator.
For example, $\mathcal{P}^9$ obtains the best MASF value in the 10 point sets.
Tables \ref{suptab:rankings_psets_dtlz1} and~\ref{suptab:rankings_psets_convdtlz2} show the results on the DTLZ1 and convDTLZ2 problems, respectively.
We do not intend to elaborate Tables~\ref{suptab:rankings_psets_dtlz1} and~\ref{suptab:rankings_psets_convdtlz2}, as they are similar to~\pref{tab:rankings_psets}.


\subsubsection{Results for $\mathbf{z}^{0.5}=(0.5, 0.5)^{\top}$}

First, we discuss the results shown in~\pref{tab:rankings_psets}(a).
$\mathcal{P}^3$ is the best in terms of $i$) the convergence to the PF, $ii$) convergence to the reference point $\mathbf{z}$, and $iii$) diversity.
Thus, quality indicators should give $\mathcal{P}^3$ the highest ranking.
$\mathcal{P}^3$ is ranked highest by 9 out of the 16 quality indicators.
However, four quality indicators (MASF, MED, PMDA, and EH) prefer $\mathcal{P}^9$ with the poorest diversity to $\mathcal{P}^3$.
This is because they do not take into account the diversity of points as shown in~\pref{tab:qi}.
Since HV and IGD do not handle the preference information, they prefer $\mathcal{P}^{10}$ that covers the whole PF.
Interestingly, IGD-P also prefers $\mathcal{P}^{10}$ the most.
This is because the IGD-reference points of IGD-P are relatively widely distributed around the center of PF.

Since PR evaluates only the cardinality, PR cannot distinguish the quality of $\mathcal{P}^{3}$, $\mathcal{P}^{7}$, and $\mathcal{P}^{9}$.
Since IGD is Pareto non-compliant, IGD prefers $\mathcal{P}^7$ to $\mathcal{P}^9$, where all the points in $\mathcal{P}^7$ are dominated by those in $\mathcal{P}^9$.
Similarly, PMOD gives $\mathcal{P}^{7}$ the second highest ranking.
Since PMOD does not take into account the convergence to the PF, PMOD can evaluate the quality of point sets inaccurately.
%

Since IGD-CF and HV-CF cannot distinguish point sets outside their preferred regions, most point sets obtain the same ranking.
Although this undesirable property was already pointed out in \cite{LiDY18}, this is the first time to demonstrate that.
The same is true for HV$_{\mathbf{z}}$ and PR.
Since R-IGD, R-HV, and EH remove dominated points from point sets, they cannot distinguish the three dominated point sets ($\mathcal{P}^{6}$, $ \mathcal{P}^{7}$, and $\mathcal{P}^{8}$).

\subsubsection{Results for $\mathbf{z}^{-0.1}=(-0.1, -0.1)^\top$}
Next, we discuss the results shown in~\pref{tab:rankings_psets}(b).
In this setting, $\mathcal{P}^1$ and $\mathcal{P}^5$ are the best in terms of all three criteria $i$), $ii$), and $iii$).
Thus, quality indicators should give $\mathcal{P}^1$ or $\mathcal{P}^5$ the highest ranking.

However, the rankings by four ASF-based quality indicators (MASF, IGD-A, R-IGD, and R-HV) are the same in Tables \ref{tab:rankings_psets}(a) and (b).
This is because the point with the minimum ASF value and the $\mathrm{ROI}^\mathrm{A}$ are the same regardless of whether $\mathbf{z}^{0.5}$ or $\mathbf{z}^{-0.1}$ is used, as demonstrated in Sections \ref{sec:dist_asf} and \ref{sec:exp_inconsistency}.
The same is true for PMOD, PMDA, and EH.
Thus, these quality indicators fail to evaluate the convergence of the point sets to the reference point. 


In contrast, the rankings by other quality indicators based on the $\mathrm{ROI}^\mathrm{C}$ and $\mathrm{ROI}^\mathrm{P}$ are different in Tables \ref{tab:rankings_psets}(a) and (b).
As demonstrated in Section \ref{sec:exp_inconsistency}, the $\mathrm{ROI}^\mathrm{C}$ is on either one of the two extreme points when using $\mathbf{z}^{-0.1}$.
For this reason, four quality indicators based on the $\mathrm{ROI}^\mathrm{C}$ (MED, IGD-C, IGD-CF, and HV-CF) prefer $\mathcal{P}^1$ or $\mathcal{P}^5$ to $\mathcal{P}^3$.
While IGD-C and IGD-A are perfectly consistent for the results of $\mathbf{z}^{0.5}$, they are inconsistent for the results of $\mathbf{z}^{-0.1}$.
This is due to the inconsistency revealed in Section \ref{sec:dist_asf}.

As discussed in Section \ref{sec:exp_inconsistency}, when $\mathbf{z}$ dominates the ideal point or is dominated by the nadir point, the $\mathrm{ROI}^\mathrm{P}$ is equivalent to the PF.
For this reason, three quality indicators based on the $\mathrm{ROI}^\mathrm{P}$ (IGD-P, HV$_{\mathbf{z}}$, and PR) cannot handle the DM's preference information.
Thus, like HV and IGD, IGD-P, HV$_{\mathbf{z}}$, and PR prefer $\mathcal{P}^{10}$ the most.
Since IGD-P and IGD use the same IGD-reference point set $\mathcal{S}$, their rankings are perfectly consistent.
Although the position of the HV-reference point $\mathbf{y}$ is different in HV$_{\mathbf{z}}$ and HV, their rankings are almost the same.
PR cannot distinguish all the 10 point sets.

\begin{tcolorbox}[sharpish corners, top=2pt, bottom=2pt, left=4pt, right=4pt, boxrule=0.0pt, colback=black!5!white,leftrule=0.75mm,]
    \textbf{\underline{Answers to RQ3:}} \textit{
        Our results indicated that most quality indicators have some undesirable properties, which have not been noticed even in their corresponding papers. We demonstrated that the quality indicators based on the $\mathrm{ROI}^\mathrm{A}$ cannot evaluate the convergence of a point set to the reference point accurately in some cases. We also demonstrated that the quality indicators based on the $\mathrm{ROI}^\mathrm{P}$ cannot take into account the DM's preference information. Our results imply that IGD-C may be the most reliable quality indicator when considering the practical $\mathrm{ROI}^\mathrm{C}$. However, IGD-C is Pareto non-compliant.
    }
\end{tcolorbox}


\newcolumntype{A}{>{\columncolor[rgb]{0.9, 0.9, 0.9}\centering}p{5em}}

\begin{table*}[t]
 \setlength{\tabcolsep}{2pt} 
\renewcommand{\arraystretch}{0.9}
  \centering
  \caption{\small 
  Rankings of the six PBEMO algorithms on the DTLZ2 problem by the 16 quality indicators when using $\mathbf{z}^{0.5}=(0.5, 0.5)^{\top}$. \lq\lq NUMS\rq\rq stands for MOEA/D-NUMS.}
  \label{tab:emo_ranks}
{\scriptsize
\begin{tabularx}{72em}{|A|C|C|C|C|C|C|C|C|C|C|C|C|C|C|C|C|}
\hline
\rowcolor[rgb]{0.9, 0.9, 0.9} & MASF & MED & IGD-C & IGD-A & IGD-P & HV$_{\mathbf{z}}$ & PR & PMOD & IGD-CF & HV-CF & PMDA & R-IGD & R-HV & EH & HV & IGD \\
\hline
R-NSGA-II & \cellcolor{c3}3 & \cellcolor{c2}2 & \cellcolor{c6}6 & \cellcolor{c6}6 & \cellcolor{c5}5 & \cellcolor{c6}6 & \cellcolor{c4}4 & \cellcolor{c4}4 & \cellcolor{c6}6 & \cellcolor{c6}6 & \cellcolor{c3}3 & \cellcolor{c5}5 & \cellcolor{c5}5 & \cellcolor{c1}1 & \cellcolor{c5}5 & \cellcolor{c5}5\\
r-NSGA-II & \cellcolor{c4}4 & \cellcolor{c3}3 & \cellcolor{c3}3 & \cellcolor{c3}3 & \cellcolor{c4}4 & \cellcolor{c4}4 & \cellcolor{c1}1 & \cellcolor{c3}3 & \cellcolor{c3}3 & \cellcolor{c4}4 & \cellcolor{c2}2 & \cellcolor{c3}3 & \cellcolor{c3}3 & \cellcolor{c3}3 & \cellcolor{c4}4 & \cellcolor{c4}4\\
g-NSGA-II & \cellcolor{c5}5 & \cellcolor{c5}5 & \cellcolor{c1}1 & \cellcolor{c1}1 & \cellcolor{c1}1 & \cellcolor{c1}1 & \cellcolor{c1}1 & \cellcolor{c1}1 & \cellcolor{c1}1 & \cellcolor{c1}1 & \cellcolor{c5}5 & \cellcolor{c2}2 & \cellcolor{c1}1 & \cellcolor{c6}6 & \cellcolor{c2}2 & \cellcolor{c2}2\\
PBEA & \cellcolor{c1}1 & \cellcolor{c6}6 & \cellcolor{c2}2 & \cellcolor{c2}2 & \cellcolor{c2}2 & \cellcolor{c2}2 & \cellcolor{c6}6 & \cellcolor{c5}5 & \cellcolor{c2}2 & \cellcolor{c2}2 & \cellcolor{c6}6 & \cellcolor{c1}1 & \cellcolor{c2}2 & \cellcolor{c5}5 & \cellcolor{c1}1 & \cellcolor{c1}1\\
R-MEAD2 & \cellcolor{c6}6 & \cellcolor{c4}4 & \cellcolor{c5}5 & \cellcolor{c5}5 & \cellcolor{c3}3 & \cellcolor{c3}3 & \cellcolor{c5}5 & \cellcolor{c6}6 & \cellcolor{c4}4 & \cellcolor{c3}3 & \cellcolor{c4}4 & \cellcolor{c6}6 & \cellcolor{c6}6 & \cellcolor{c4}4 & \cellcolor{c3}3 & \cellcolor{c3}3\\
NUMS & \cellcolor{c2}2 & \cellcolor{c1}1 & \cellcolor{c4}4 & \cellcolor{c4}4 & \cellcolor{c6}6 & \cellcolor{c5}5 & \cellcolor{c1}1 & \cellcolor{c2}2 & \cellcolor{c5}5 & \cellcolor{c5}5 & \cellcolor{c1}1 & \cellcolor{c4}4 & \cellcolor{c4}4 & \cellcolor{c2}2 & \cellcolor{c6}6 & \cellcolor{c6}6\\
\hline
\end{tabularx}
}
\end{table*}

\subsection{On the rankings of PBEMO algorithms by quality indicators}
\label{sec:exp_rankings}

\pref{tab:emo_ranks} shows the rankings of the six PBEMO algorithms on the DTLZ2 problem by the $16$ quality indicators, where $\mathbf{z}=(0.5,0.5)^\top$.
We calculated the rankings based on the average quality indicator values of the PBEMO algorithms over $31$ runs.
Tables \ref{suptab:emo_ranks_dtlz1}--\ref{suptab:emo_ranks_convdtlz2} show the rankings on the DTLZ1, DTLZ2, and convDTLZ2 problems when using various reference points.
Note that we are interested in the influence of quality indicators on the rankings of the PBEMO algorithms rather than the rankings themselves.


As shown in~\pref{tab:emo_ranks}, the rankings of the PBEMO algorithms are different depending on the choice of the quality indicator.
For example, R-NSGA-II performs the best in terms of EH but the worst in terms of five  quality indicators including IGD-C, IGD-A, HV$_{\mathbf{z}}$, IGD-CF, and HV-CF.
Likewise, g-NSGA-II is the worst performer in terms of EH but it is the best algorithm when considering the other nine quality indicators.
Since PBEA and MOEA/D-NUMS aim to minimize the ASF, they are the best and second-best performers in terms of MASF.
%
As shown in Table.~\ref{suptab:emo_ranks_convdtlz2}(b), R-MEAD2 performs the best on the convDTLZ2 problem in terms of EH.
In summary, our results suggest that any PBEMO algorithm can obtain the best ranking depending on the choice of the quality indicator.

These observations can be explained as the coupling relationship between innate mechanism of the PBEMO algorithms and the quality indicators.
As demonstrated in~\pref{sec:exp_inconsistency}, each PBEMO algorithm approximates its target ROI. 
As investigated in~\pref{sec:exp_qi}, each quality indicator prefers a point set that approximates its target ROI well.
Thus, when benchmarking PBEMO algorithms, it is important to clarify which type of ROI the DM wants to approximate and select a suitable quality indicator.
For example, if the DM wants to approximate the $\mathrm{ROI}^\mathrm{A}$, she/he should select either of IGD-A, R-IGD, and R-HV.
Otherwise, the DM can overestimate or underestimate the performance of PBEMO algorithms.




%

%

\begin{tcolorbox}[sharpish corners, top=2pt, bottom=2pt, left=4pt, right=4pt, boxrule=0.0pt, colback=black!5!white,leftrule=0.75mm,]
    \textbf{\underline{Answers to RQ4:}} \textit{
	Our results showed that the choice of the quality indicator significantly influences the performance rankings of EMO algorithms. For example, as seen from Table \ref{tab:emo_ranks}, PBEA performs the worst in terms of PMDA but the best in terms of R-IGD. This means that any PBEMO algorithm can possibly be ranked as the best (or the worst) depending on the choice of the quality indicator. We also discussed how to conduct meaningful benchmarking of PBEMO algorithms.        
    }
\end{tcolorbox}

\subsection{Further investigations}
\label{sec:addtional_exp}

\subsubsection{Results on three-objective problems}

Due to the reason described in \pref{sec:test_problems}, this paper focused on the two-objective problems.
Although designing a new visualization method for $m>3$ is beyond the scope of this paper, it is still possible to perform our analysis for $m=3$. 
Similar to \pref{fig:toy}, we generated 13 point sets $\mathcal{P}^1, \ldots, \mathcal{P}^{13}$ on the PFs of the three-objective DTLZ1, DTLZ2, and convDTLZ2 problems.
Fig. \ref{supfig:toy_3d} shows distributions of the 13 point sets, where the size of each point set is 21.
$\mathcal{P}^6$ is on the center of the PF. $\mathcal{P}^1$, $\mathcal{P}^4$, and $\mathcal{P}^{10}$ are on the three extreme regions, respectively. $\mathcal{P}^2$, $\mathcal{P}^3$, $\mathcal{P}^5$, $\mathcal{P}^7$--$\mathcal{P}^9$, are on the edge of the PF. $\mathcal{P}^{11}$ is a shifted version of $\mathcal{P}^6$ by adding 0.1 to all elements. Similar to $\mathcal{P}^{9}$ in \pref{fig:toy}, $\mathcal{P}^{12}$ is worse than $\mathcal{P}^6$ in terms of the diversity. The 21 points in $\mathcal{P}^{13}$ are uniformly distributed on the PF.

Tables \ref{suptab:rankings_psets_dtlz1_3d}--\ref{suptab:rankings_psets_convdtlz2_3d} show the results on the DTLZ1, DTLZ2, and convDTLZ2 problems, respectively.
Here, we used the reference points $\mathbf{z}^{0.5}=(0.5, 0.5, 0.5)^{\top}$ and $\mathbf{z}^{-0.1}=(-0.1, -0.1, -0.1)^{\top}$ for the DTLZ1 and DTLZ2 problems, and $\mathbf{z}^{0.2}=(0.2, 0.2, 0.2)^{\top}$ and $\mathbf{z}^{2}=(2, 2, 2)^{\top}$ for the convDTLZ2 problem.
For the DTLZ2 and convDTLZ2 problems, when using $\mathbf{z}^{0.5}$ and $\mathbf{z}^{0.2}$, $\mathcal{P}^6$ is the best point set.
When using $\mathbf{z}^{-0.1}$ and $\mathbf{z}^{2}$, either of $\mathcal{P}^1$, $\mathcal{P}^4$, and $\mathcal{P}^{10}$ is the best point set.

As seen from Tables \ref{suptab:rankings_psets_dtlz1_3d}--\ref{suptab:rankings_psets_convdtlz2_3d}, the results for $m=3$ are almost similar to the results for $m=2$ described in \pref{sec:exp_qi}.
For example, for the results on the three-objective DTLZ2 problem shown in Table \ref{suptab:rankings_psets_dtlz2_3d}, MASF, MED, PMDA, and EH prefer $\mathcal{P}^{12}$ with the poorest diversity to $\mathcal{P}^{6}$.
The rankings by the four ASF-based quality indicators (MASF, IGD-A, R-IGD, and R-HV) are the same for $\mathbf{z}^{0.5}$ and $\mathbf{z}^{-0.1}$, i.e., they fail to evaluate the convergence of the point set to the reference point.
In summary, the results suggest that our findings for $m=2$ can be applied to $m=3$.

\subsubsection{Results on a discontinuous ROI}

Here, we investigate the effectiveness of the 14 quality indicators on a discontinuous ROI.
As in \cite{LiDY18}, we used the PF of the ZDT3 problem~\cite{ZitzlerDT00}, where the PF consists of five disconnected subsets.
Fig. \ref{supfig:toy_zdt3} shows distributions of seven point sets $\mathcal{P}^1, \ldots, \mathcal{P}^7$, where the size of each point set is $\mu=20$.
$\mathcal{P}^1, \ldots, \mathcal{P}^5$ are on the five subsets of the PF, respectively.
Here, the ROI contains some points in $\mathcal{P}^2$ and all the 20 points in $\mathcal{P}^3$.
Only in this experiment, the radius of the ROI $r$ is set to $0.25$ so that the ROI is disconnected.
Since all the 20 points in $\mathcal{P}^6$ are on the ROI, $\mathcal{P}^6$ is the best approximation.
The 20 points in $\mathcal{P}^7$ are uniformly distributed on the whole PF.

Table~\ref{suptab:rankings_psets_zdt3} shows the rankings of the seven point sets by each quality indicator.
As seen from Table~\ref{suptab:rankings_psets_zdt3}, as expected, most quality indicators prefer $\mathcal{P}^6$.
However, MED, PR, PMDA, and EH prefer $\mathcal{P}^3$ to $\mathcal{P}^6$.
As described in \pref{tab:qi}, they do not evaluate the diversity of a point set in the ROI.
Thus, the ability to evaluate the diversity may play an important role in handling a discontinuous ROI.


%% file: conclusion.tex
\section{Conclusion}
\label{sec:conclusion}

In this paper, we first reviewed the $3$ ROIs and $14$ existing quality indicators for PBEMO algorithms using the reference point.
Different from the descriptions in their corresponding papers, we classified the properties of the quality indicators from the perspective of their working principle.
As a result, we found that some quality indicators have undesirable properties.
For example, PMDA and EH cannot evaluate the diversity of a point set.
We also discussed the target ROI of each quality indicator.

Next, we empirically analyzed the performance and properties of those $14$ quality indicators to address 4 RQs (\underline{RQ1} to \underline{RQ4}).
Our findings are helpful for benchmarking PBEMO algorithms and decision-making in real-world problems.
In any case, we argue the importance of determining a target ROI first of all.
Our results suggested the use of the $\mathrm{ROI}^\mathrm{C}$.
%
%
Afterward, an analyst and the DM should select a PBEMO algorithm and quality indicator based on their target ROI.
For example, R-NSGA-II aims to approximate the $\mathrm{ROI}^\mathrm{C}$.
In contrast, HV$_{\mathbf{z}}$ is to evaluate how a point set approximates the $\mathrm{ROI}^\mathrm{P}$.
Thus, HV$_{\mathbf{z}}$ is not suitable for evaluating the performance of R-NSGA-II.

%
%




As demonstrated in the three IGD variants (IGD-C, IGD-A, and IGD-P), we believe that the target ROI of some quality indicators can be changed easily.
For example, the target ROI of R-IGD can be changed from the $\mathrm{ROI}^\mathrm{A}$ to the $\mathrm{ROI}^\mathrm{C}$ by revising the line $4$ in~\pref{alg:r-metric}, i.e., $\mathbf{p}^{\mathrm{c}} = \mathrm{argmin}_{\mathbf{p} \in \mathcal{P}_i} \{\mathrm{dist}(\mathbf{p}, \mathbf{z}) \}$.
An investigation of this concept is an avenue for future work.
Note that the analysis conducted in this paper focused on quality indicators for PBEMO algorithms using the reference point. It is questionable and important to extend our analysis for other preference-based optimization (e.g., a value function) in future research.
It is promising to design a general framework that extends the 14 quality indicators to other preference elicitation methods.
%
There is room for discussion about a systematic benchmarking methodology for PBEMO.




%% file: supplement.tex
\renewcommand{\thesection}{S.\arabic{section}}
\renewcommand{\thetable}{S.\arabic{table}}
\renewcommand{\thefigure}{S.\arabic{figure}}
\renewcommand\thealgocf{S.\arabic{algocf}} 
\renewcommand{\theequation}{S.\arabic{equation}}

\makeatletter
\renewcommand{\@biblabel}[1]{[S.#1]}
\renewcommand{\@cite}[1]{[S.#1]}
\makeatother


\begin{figure*}[t]
  \centering
  \subfloat[DTLZ1 ($\mathcal{P}^1, \dots, \mathcal{P}^{9}$)]{  
    \includegraphics[width=0.31\textwidth]{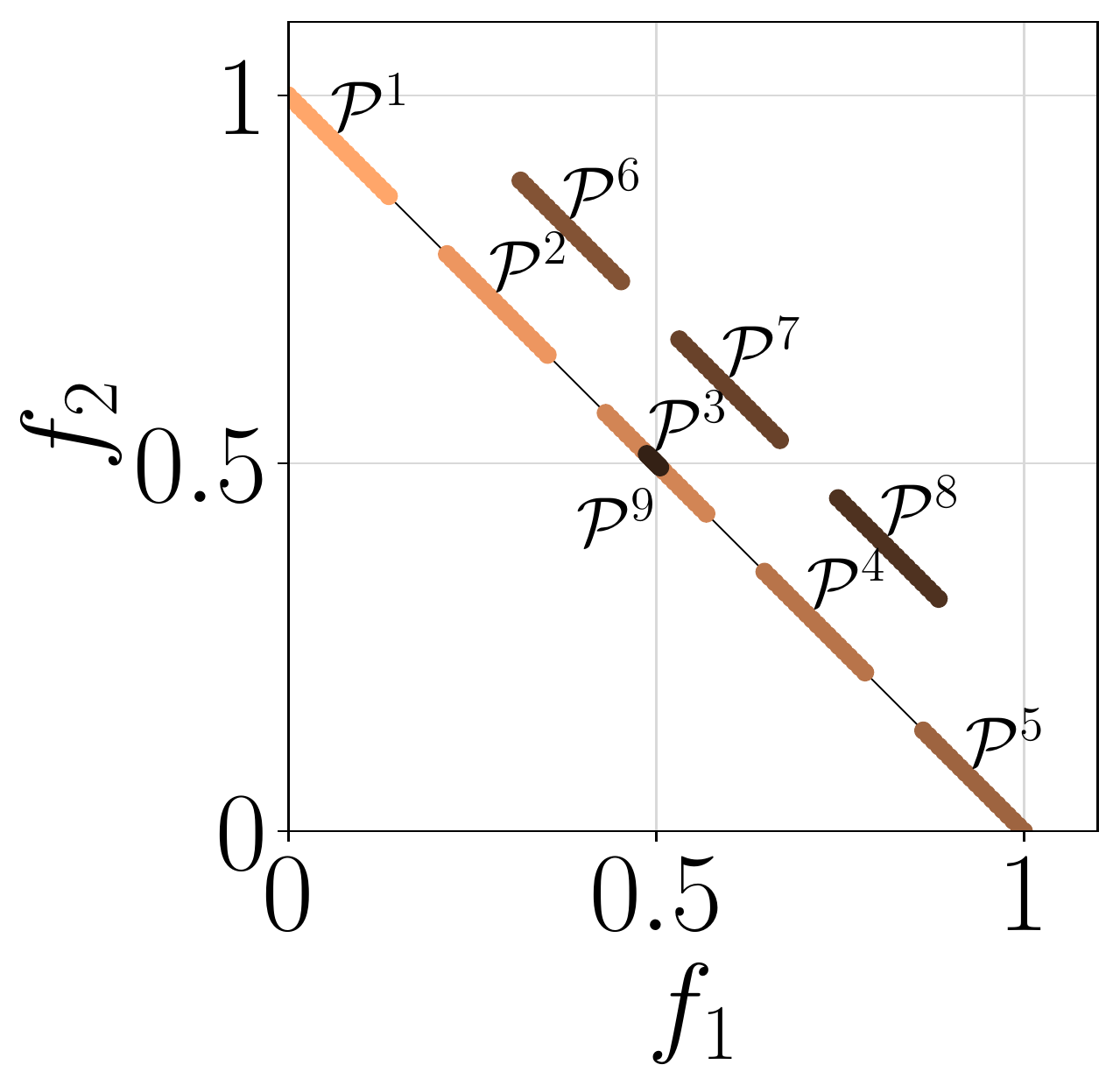}
  }
  \subfloat[DTLZ1 ($\mathcal{P}^{10}$)]{
    \includegraphics[width=0.3\textwidth]{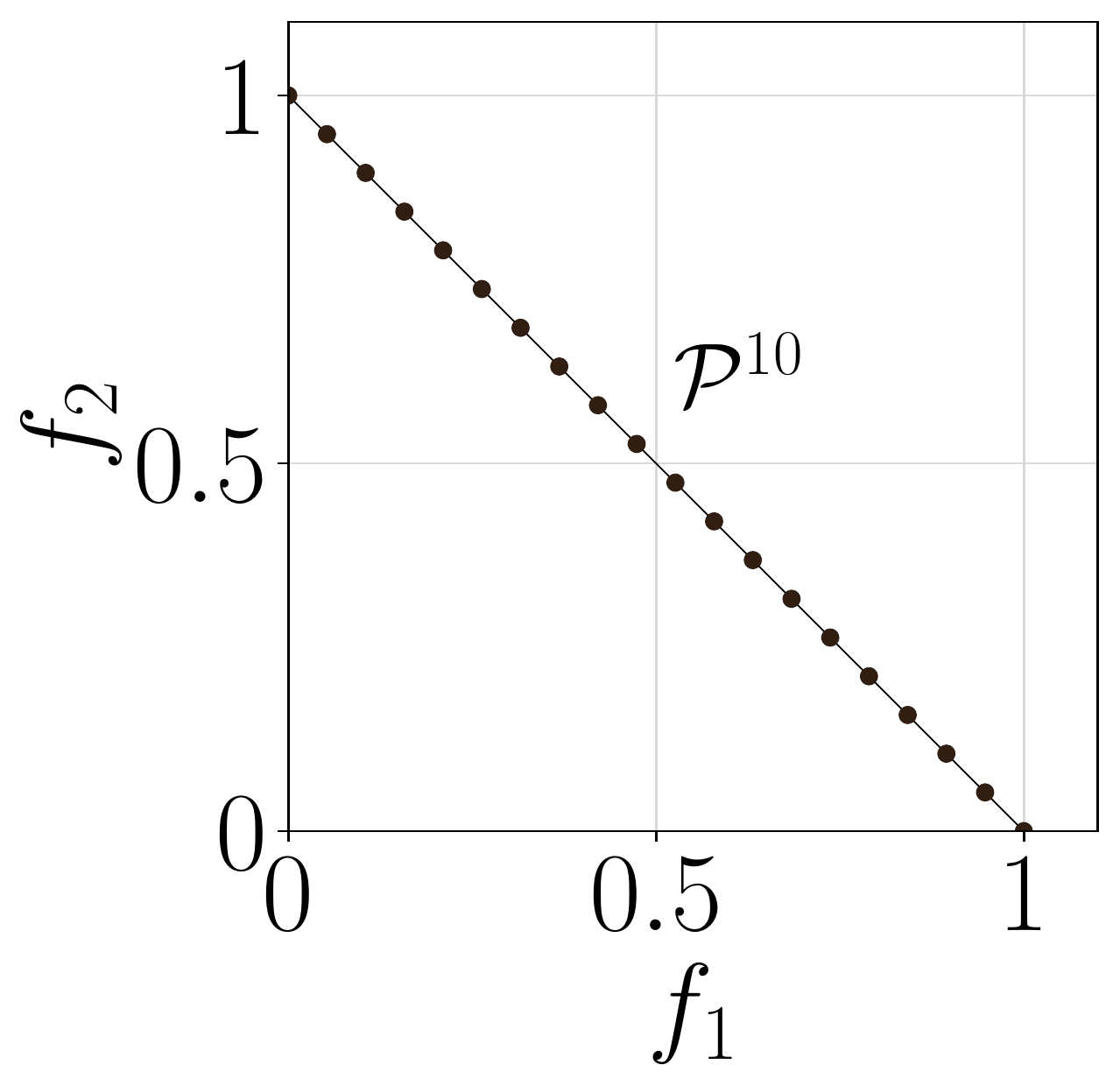}
  }
  \\
  \subfloat[convDTLZ2 ($\mathcal{P}^1, \dots, \mathcal{P}^{9}$)]{  
    \includegraphics[width=0.31\textwidth]{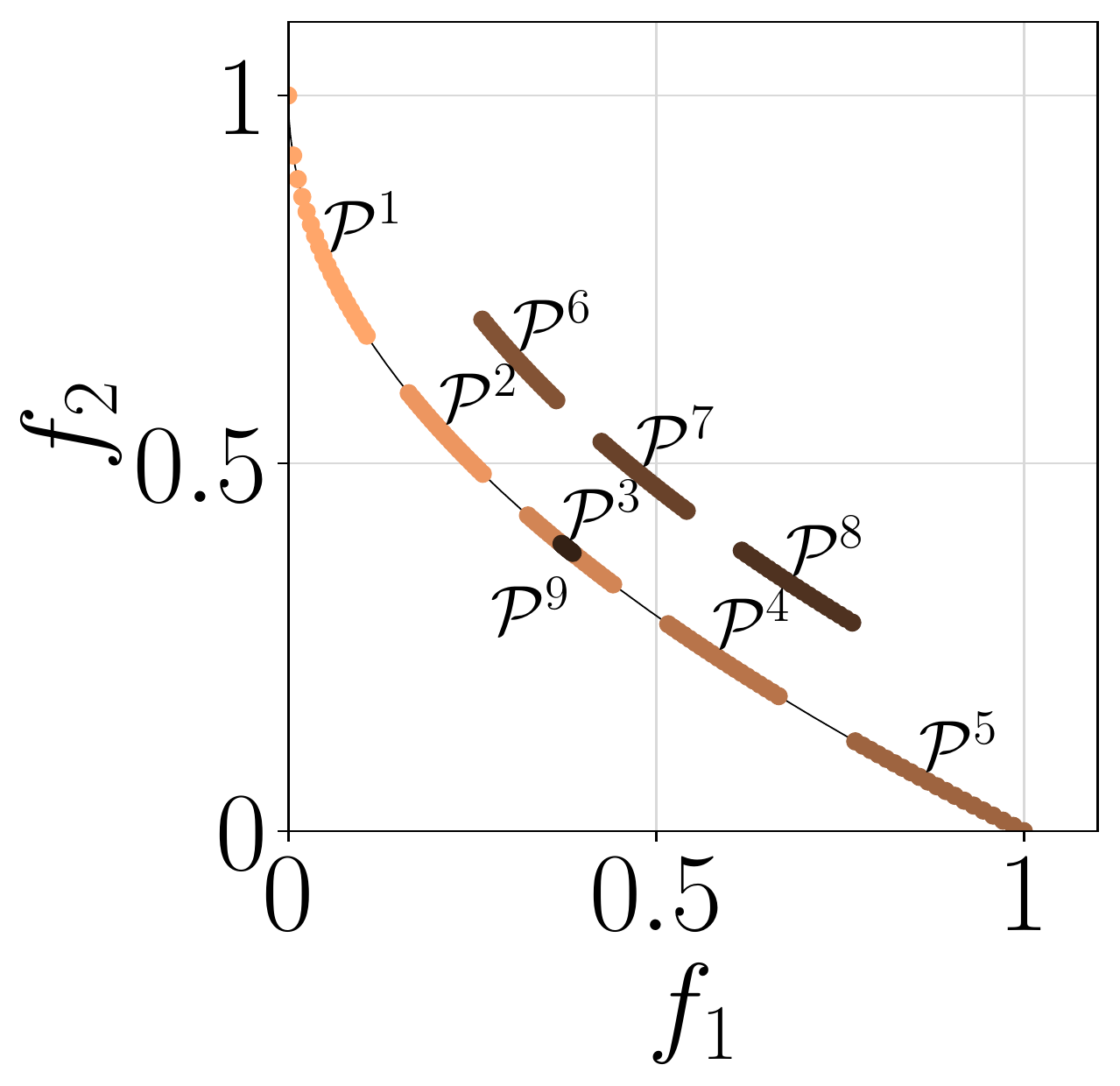}
  }
  \subfloat[convDTLZ2 ($\mathcal{P}^{10}$)]{
    \includegraphics[width=0.3\textwidth]{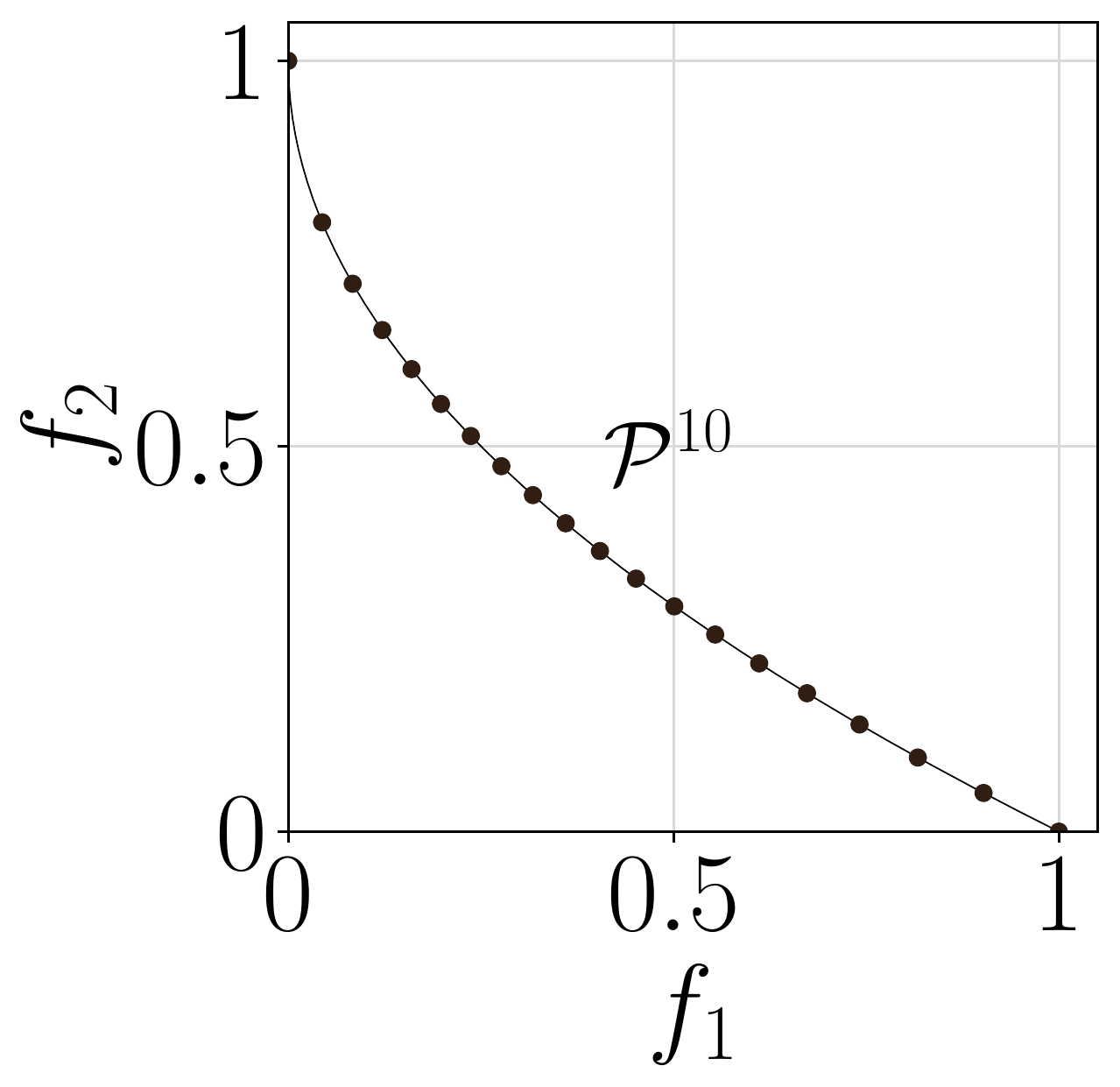}
  }
  \caption{Distributions of the 10 point sets on the Pareto fronts of the DTLZ1 and convDTLZ2 problems.}
   \label{supfig:toy}
\end{figure*}

\begin{figure*}[t]
   \centering
\subfloat[100 points]{  
  \includegraphics[width=0.18\textwidth]{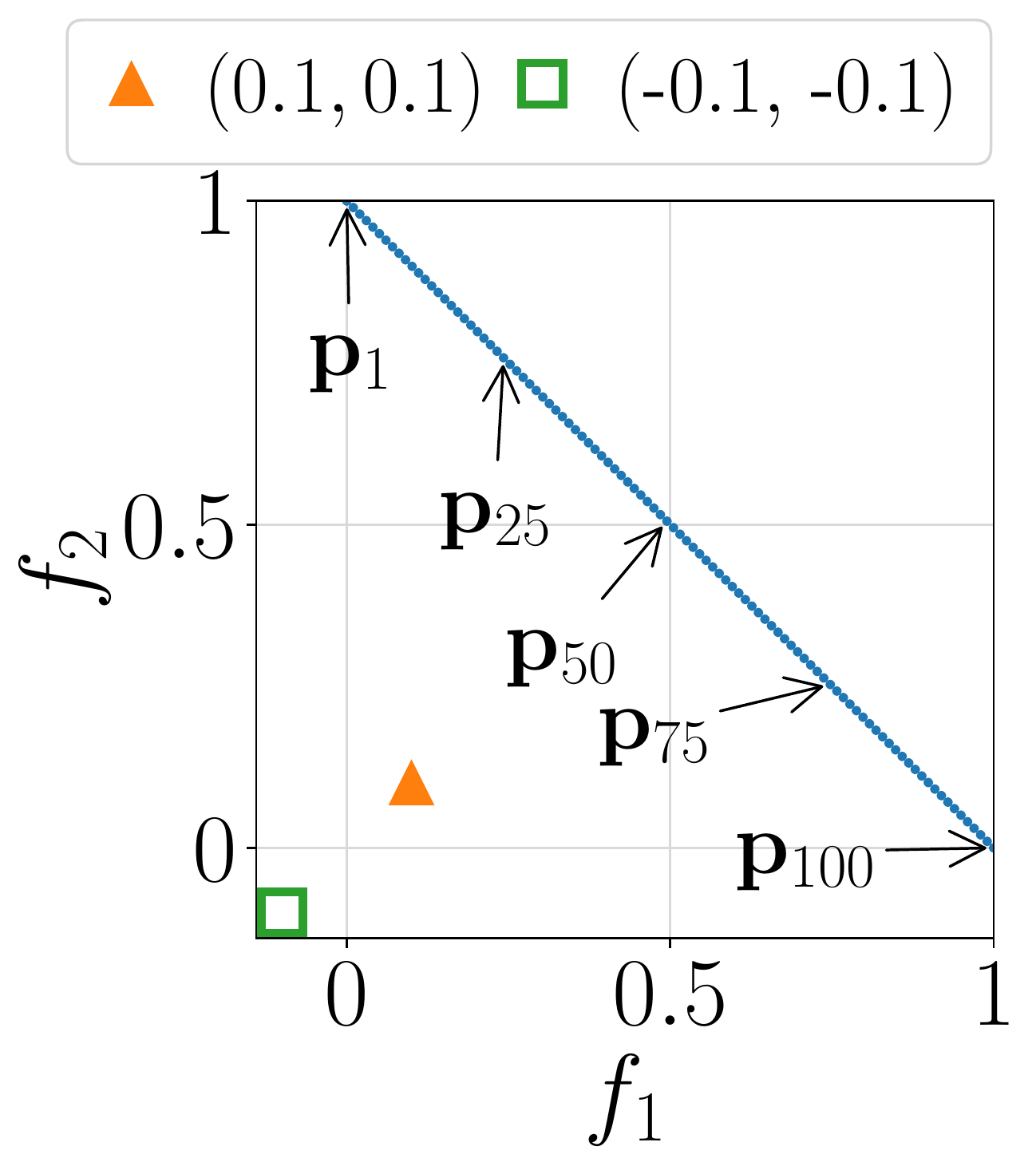}
}
\subfloat[Rankings  (distance)]{  
  \includegraphics[width=0.22\textwidth]{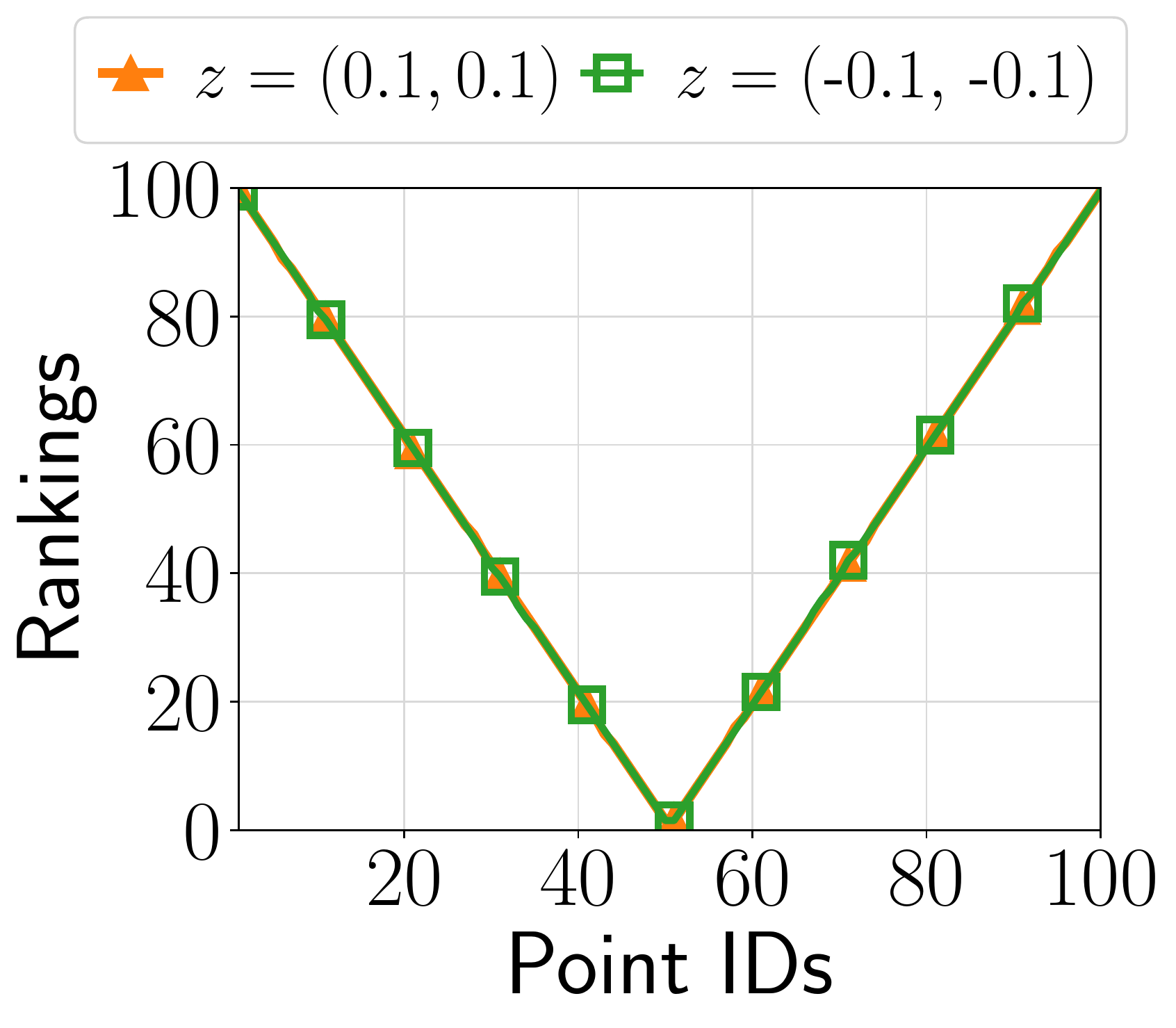}
}
\subfloat[Rankings (ASF)]{  
  \includegraphics[width=0.22\textwidth]{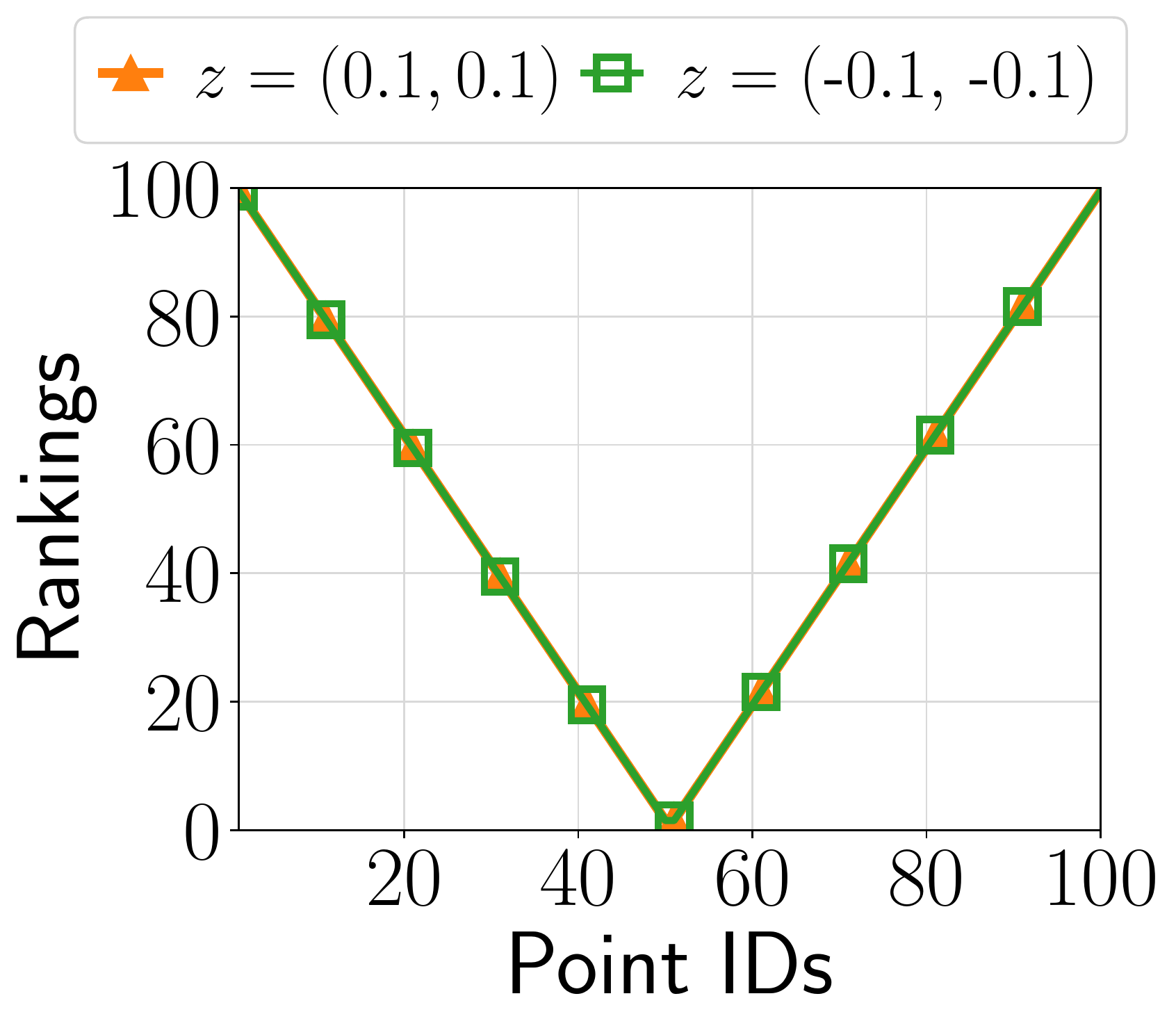}
}
\subfloat[Kendall $\tau$]{  
  \includegraphics[width=0.22\textwidth]{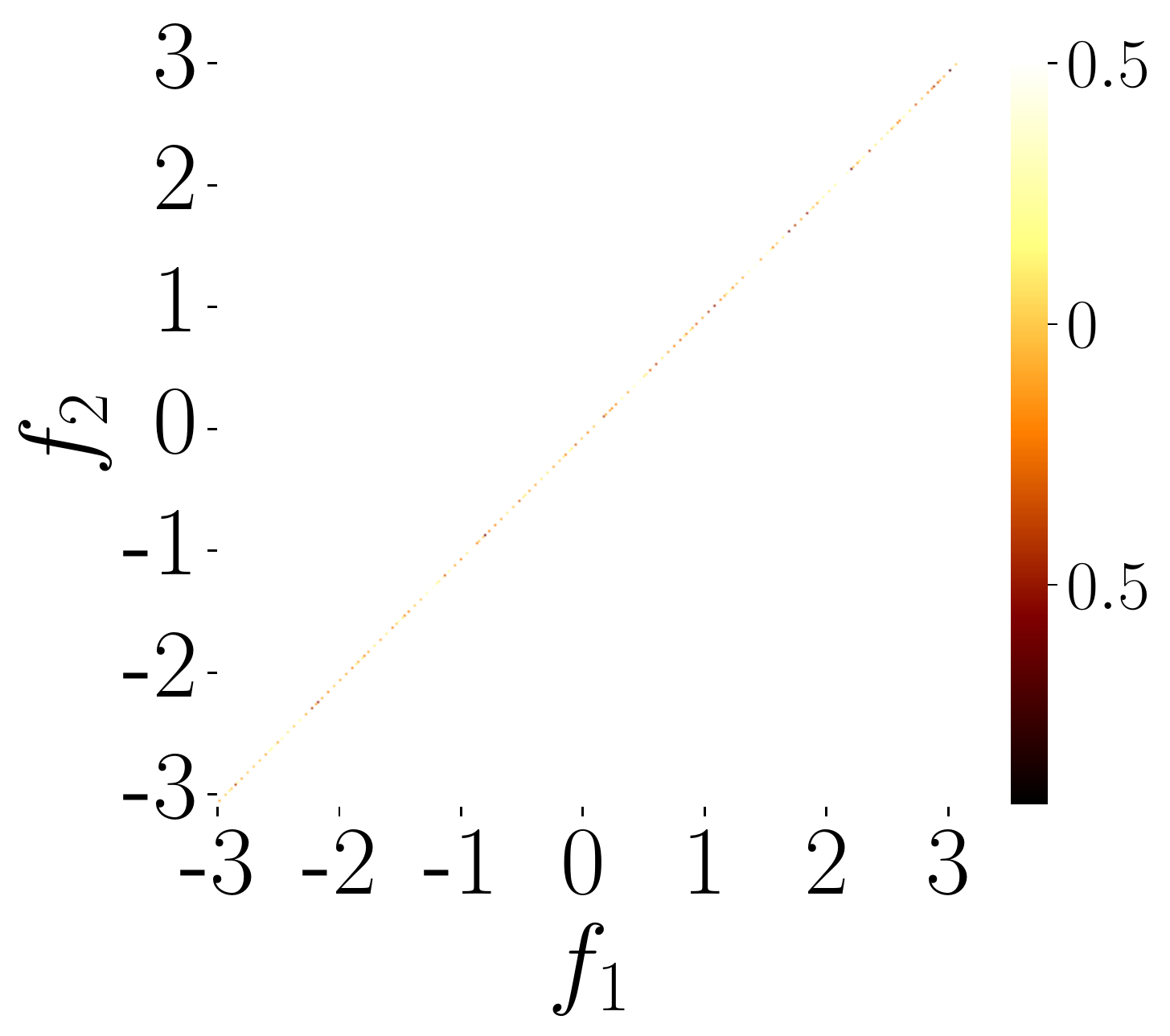}
}
\caption{(a) Distribution of 100 uniformly distributed points, (b) the rankings of the 100 points by the distance, (c) the ranking of the 100 points by the ASF , and (d) the Kendall $\tau$ values on the DTLZ1 problem. We normalized the PF of the DTLZ1 problem in the range $[0,1]^M$.}
   \label{supfig:100points_dtlz1}
%
   \centering
\subfloat[100 points]{  
  \includegraphics[width=0.18\textwidth]{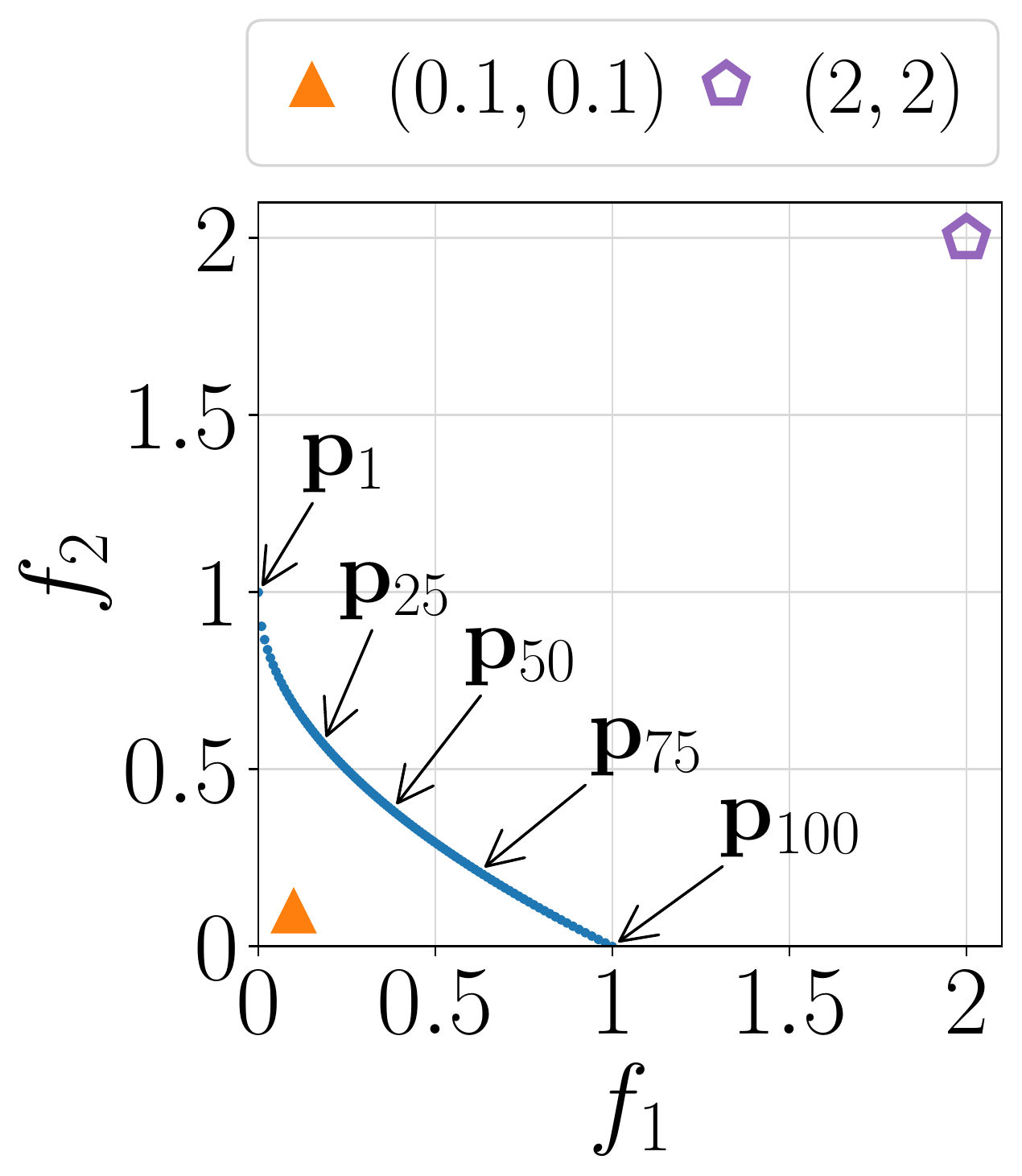}
}
\subfloat[Rankings  (distance)]{  
  \includegraphics[width=0.22\textwidth]{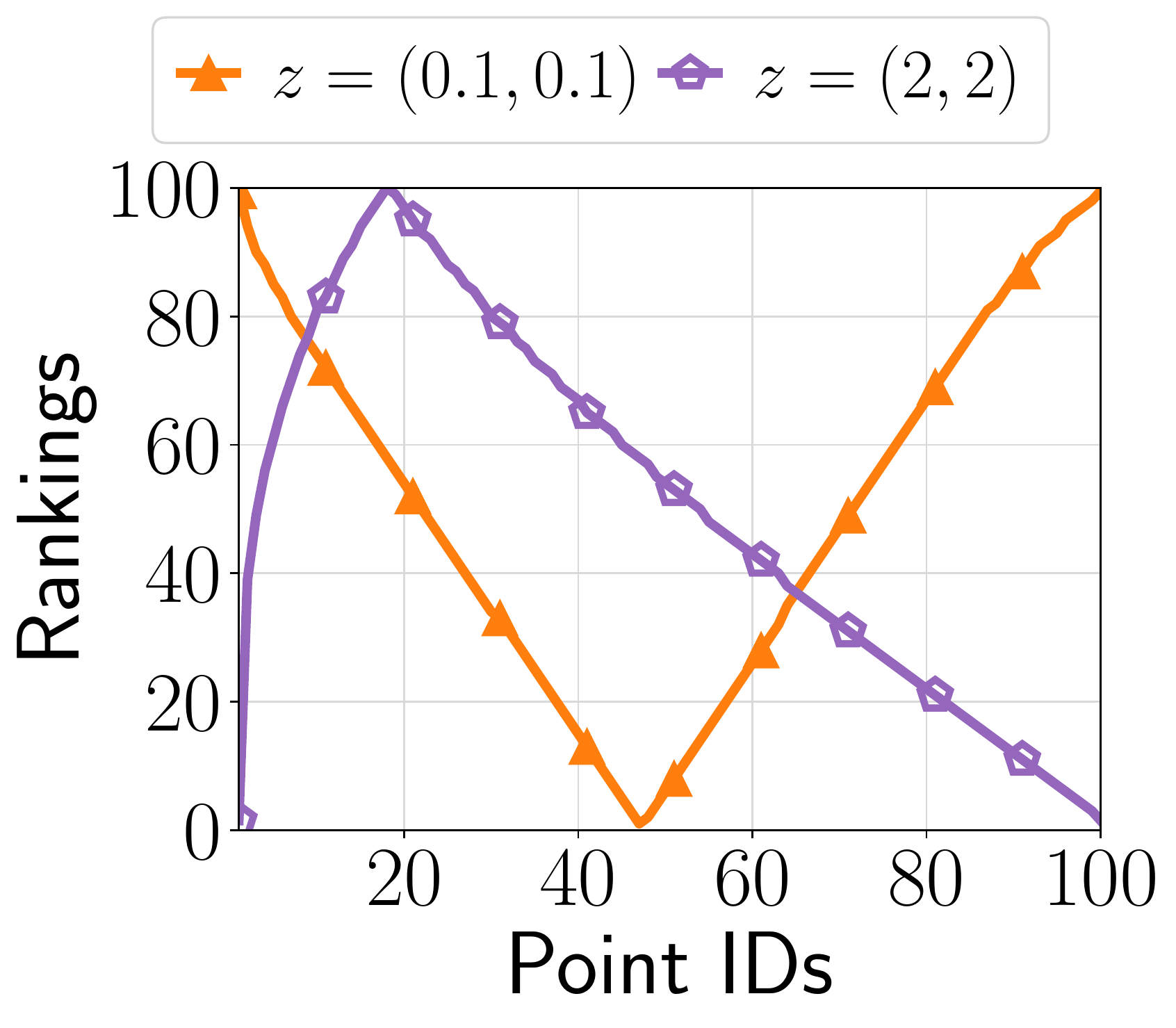}
}
\subfloat[Rankings (ASF)]{  
  \includegraphics[width=0.22\textwidth]{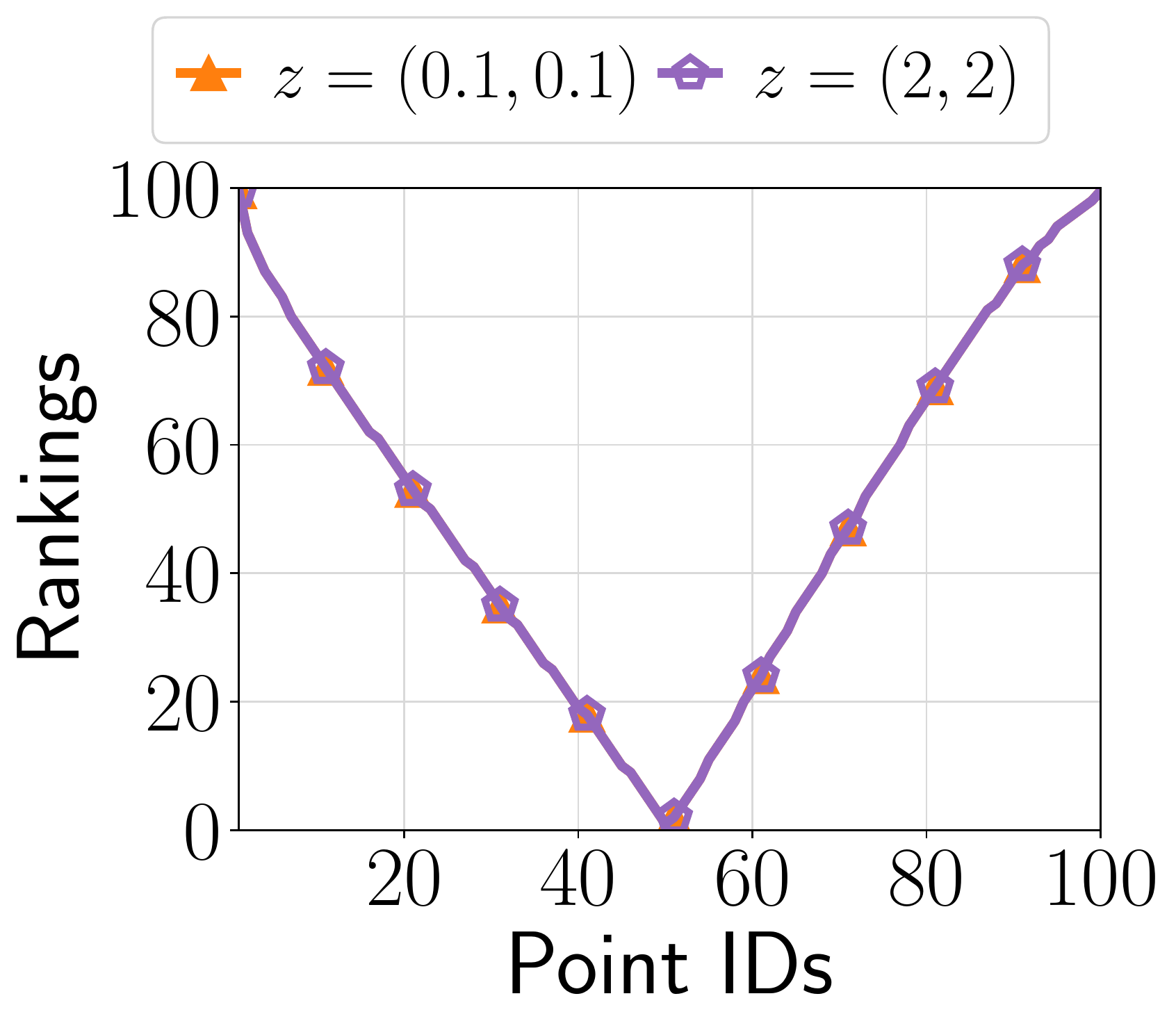}
}
\subfloat[Kendall $\tau$]{  
  \includegraphics[width=0.22\textwidth]{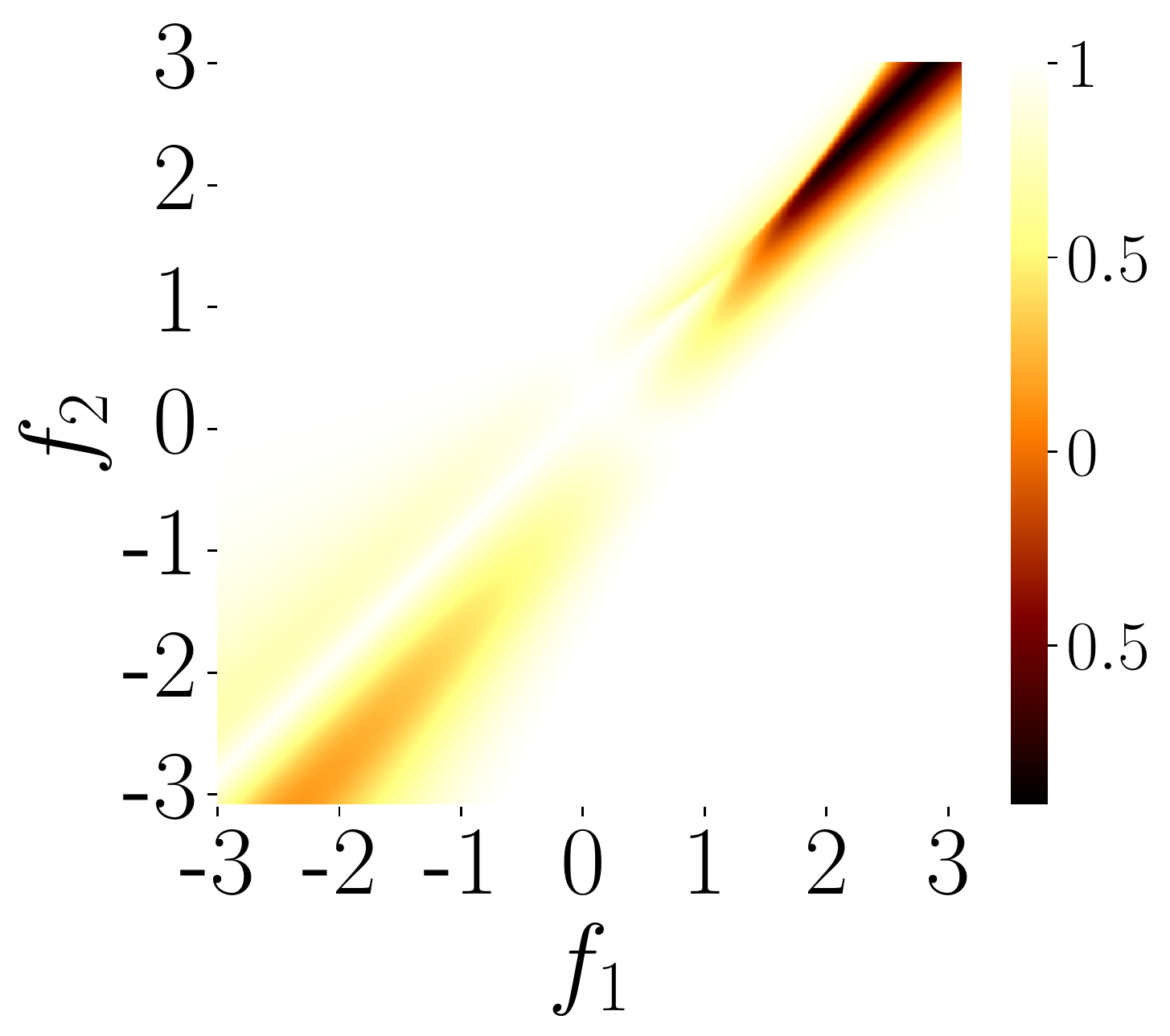}
}
\caption{(a) Distribution of 100 uniformly distributed points, (b) the rankings of the 100 points by the distance, (c) the ranking of the 100 points by the ASF , and (d) the Kendall $\tau$ values on the convDTLZ2 problem. Since the shape of the PF of the convDTLZ2 problem is asymmetric, the rankings in Fig. \ref{supfig:100points_convdtlz2}(b) is asymmetric.}
   \label{supfig:100points_convdtlz2}
\end{figure*}


\begin{figure*}[t]
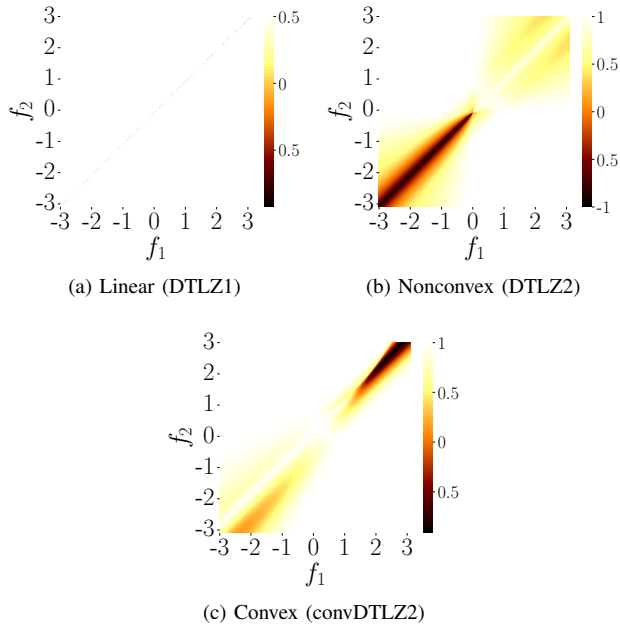

   \centering
\subfloat[Linear (DTLZ1)]{  
  \includegraphics[width=0.22\textwidth]{figs/kendall_heatmap/DTLZ1_d2_n100.pdf}
}
\subfloat[Nonconvex (DTLZ2)]{  
  \includegraphics[width=0.22\textwidth]{figs/kendall_heatmap/DTLZ2_d2_n100.pdf}
}
\\
\subfloat[Convex (convDTLZ2)]{  
  \includegraphics[width=0.22\textwidth]{figs/kendall_heatmap/convDTLZ2_d2_n100.pdf}
}
\caption{Kendall $\tau$ values on the four PFs. A darker area means that the distance to the reference point $\mathbf{z}$ and the ASF value are strongly inconsisitent with each other.}
\label{supfig:kendall_pf}
\end{figure*}

\begin{figure*}[t]
   \centering
    \subfloat[ROI-C ($\mathbf{z}^{0.5}$)]{  
    \includegraphics[width=0.145\textwidth]{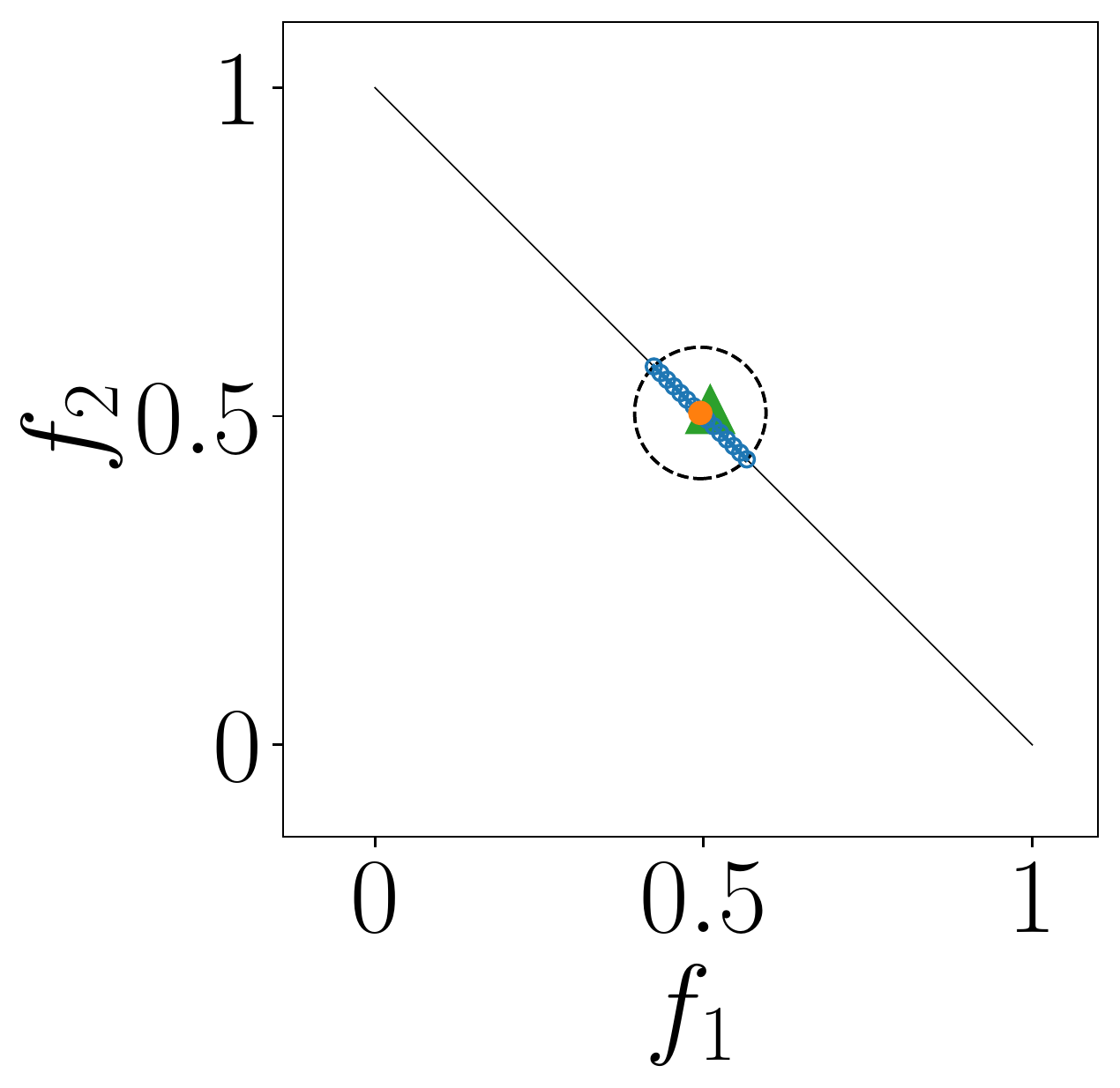}
    }
    \subfloat[ROI-A ($\mathbf{z}^{0.5}$)]{  
    \includegraphics[width=0.145\textwidth]{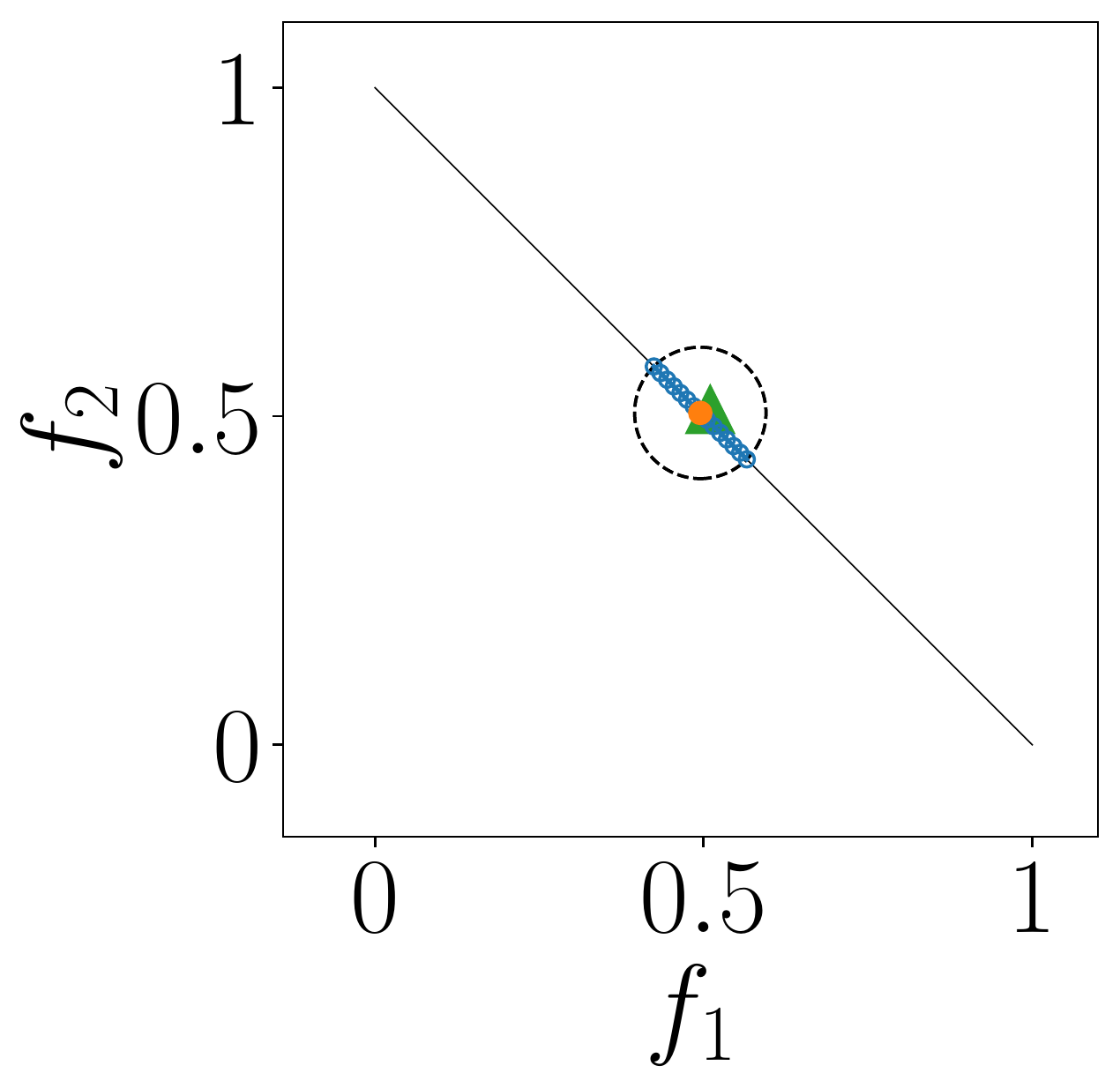}
    }
    \subfloat[ROI-P ($\mathbf{z}^{0.5}$)]{  
    \includegraphics[width=0.145\textwidth]{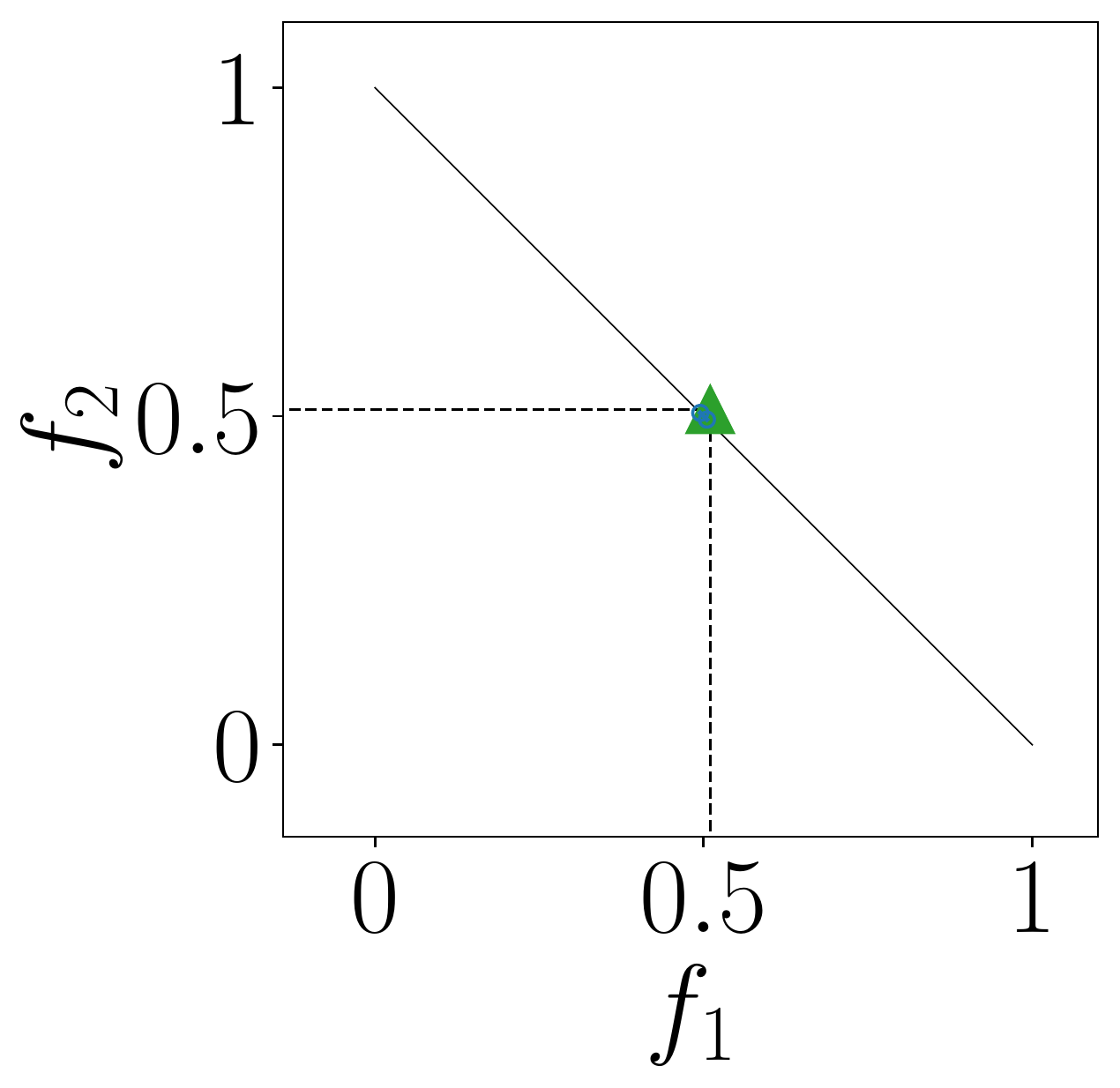}
    }
    \\
    \subfloat[ROI-C ($\mathbf{z}^{0.1}$)]{  
    \includegraphics[width=0.145\textwidth]{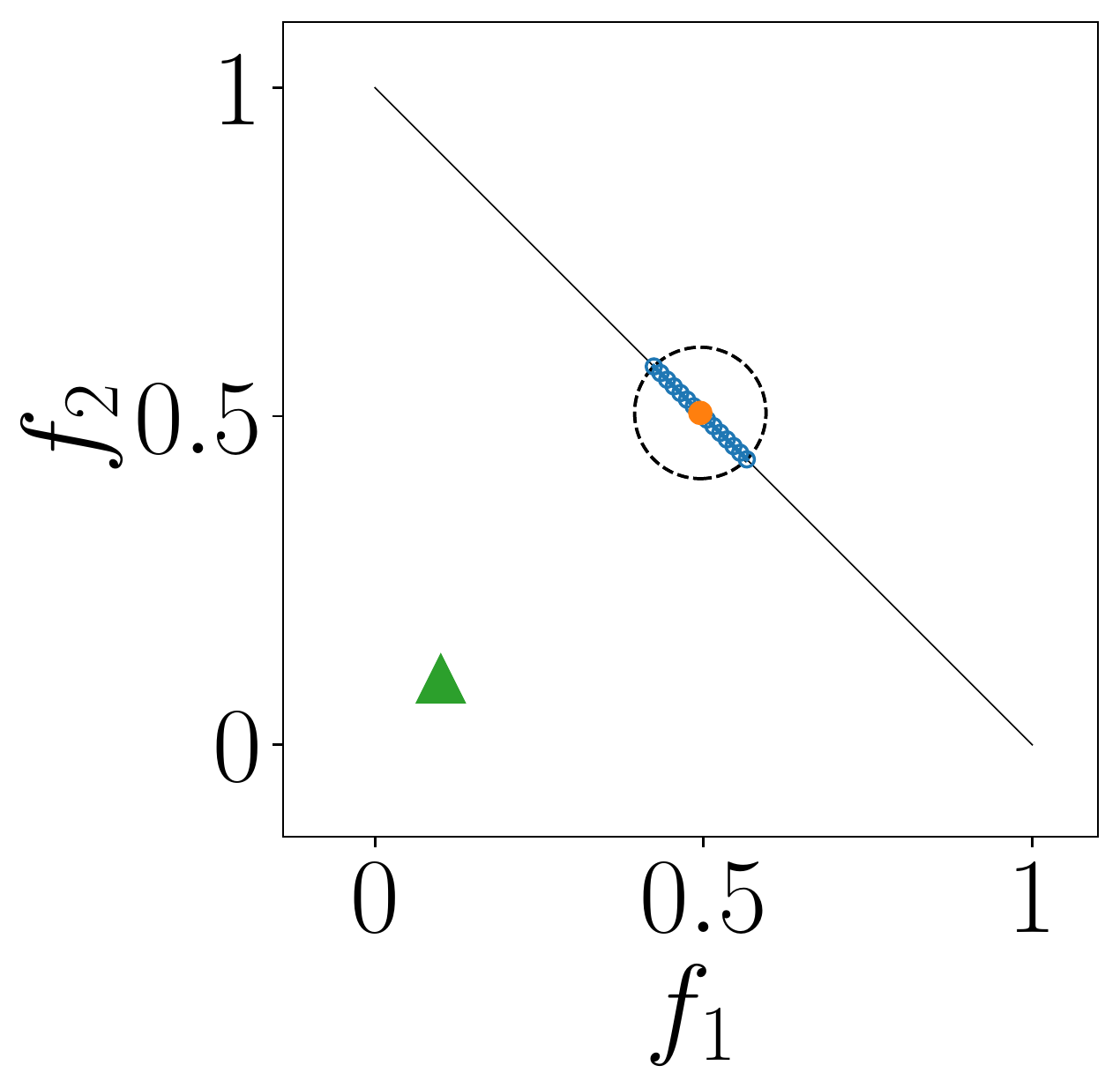}
    }
    \subfloat[ROI-A ($\mathbf{z}^{0.1}$)]{  
    \includegraphics[width=0.145\textwidth]{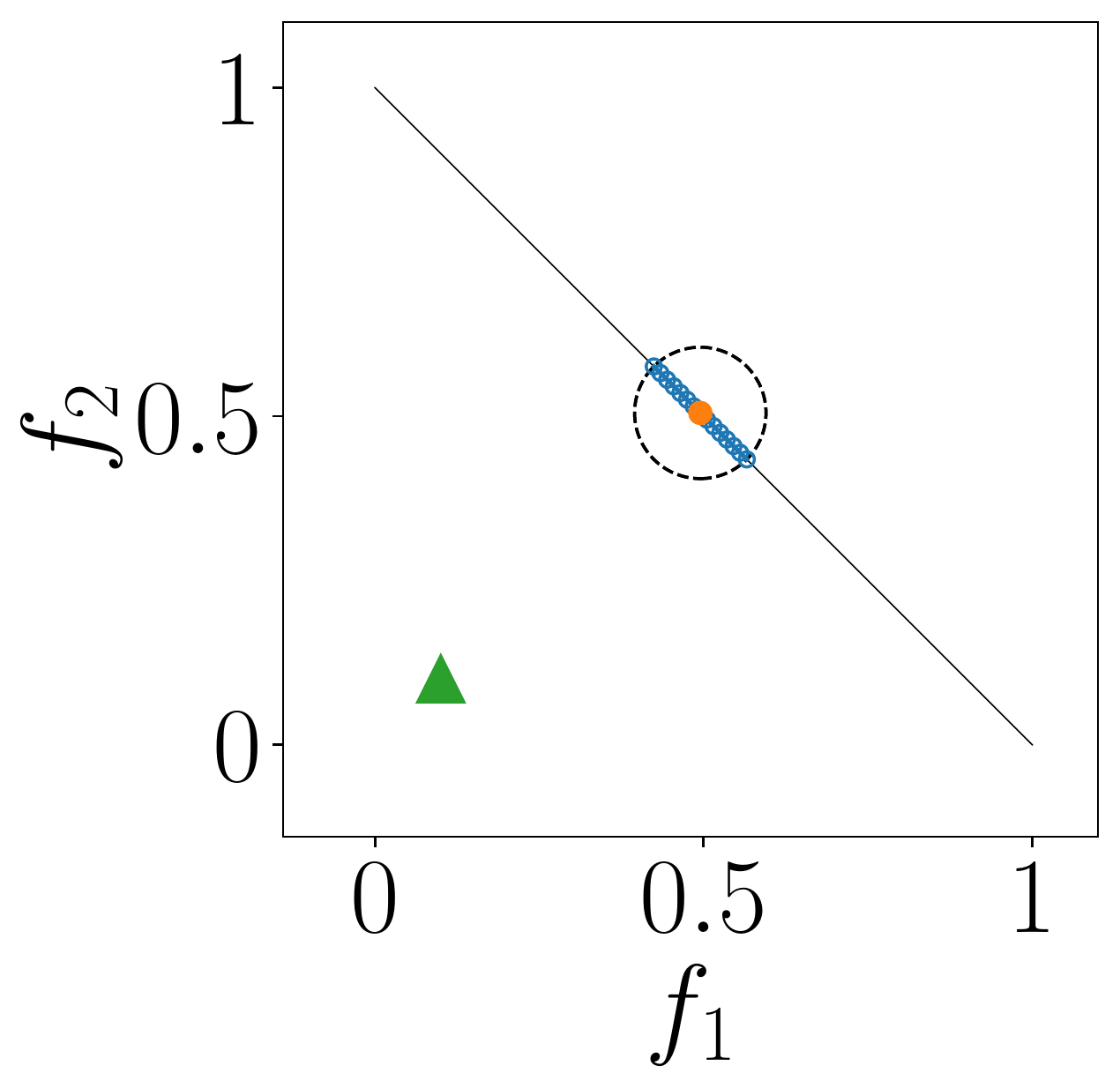}
    }
    \subfloat[ROI-P ($\mathbf{z}^{0.1}$)]{  
    \includegraphics[width=0.145\textwidth]{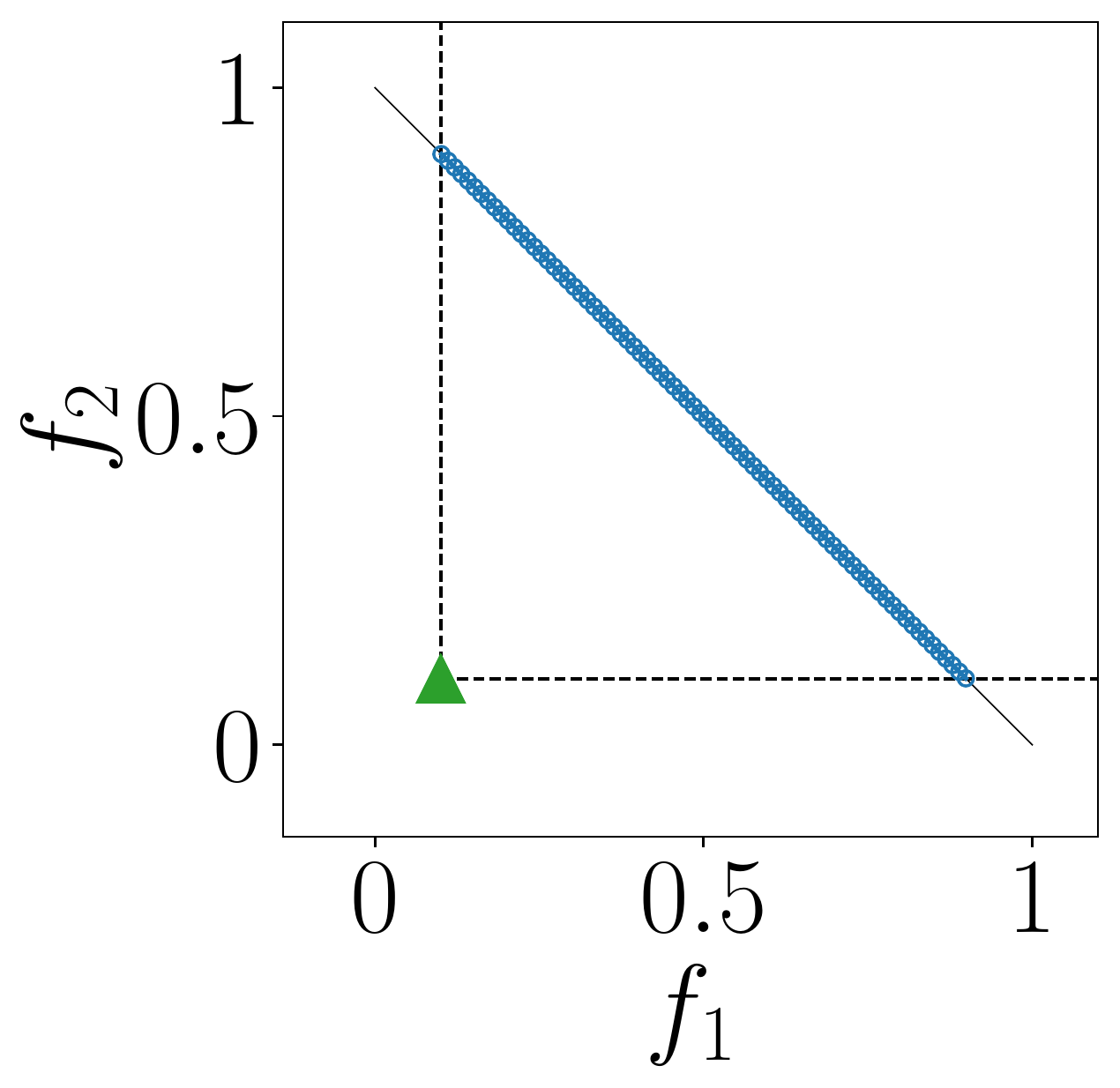}
    }
    \\
    \subfloat[ROI-C ($\mathbf{z}^{2}$)]{  
    \includegraphics[width=0.145\textwidth]{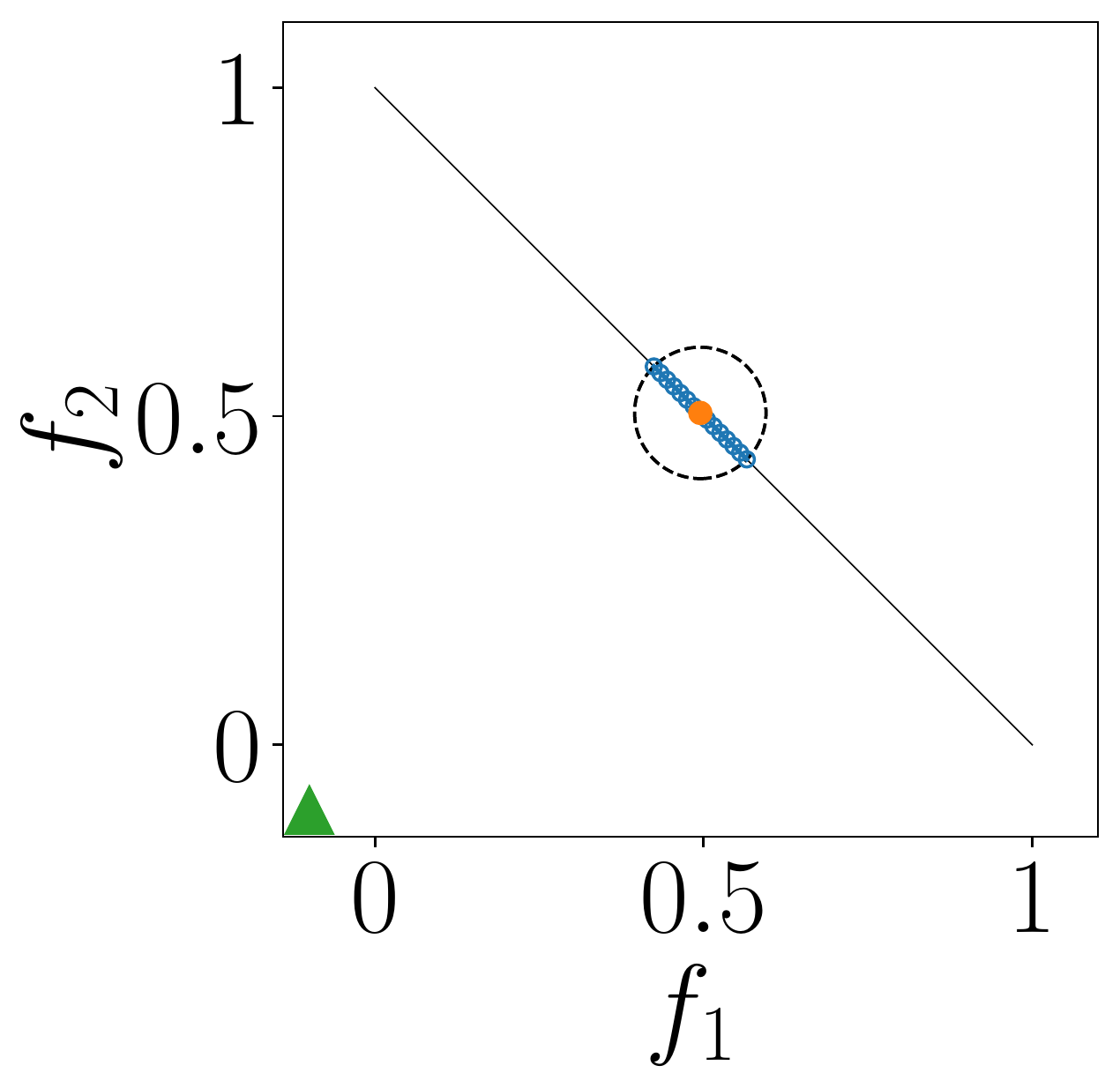}
    }
    \subfloat[ROI-A ($\mathbf{z}^{2}$)]{  
    \includegraphics[width=0.145\textwidth]{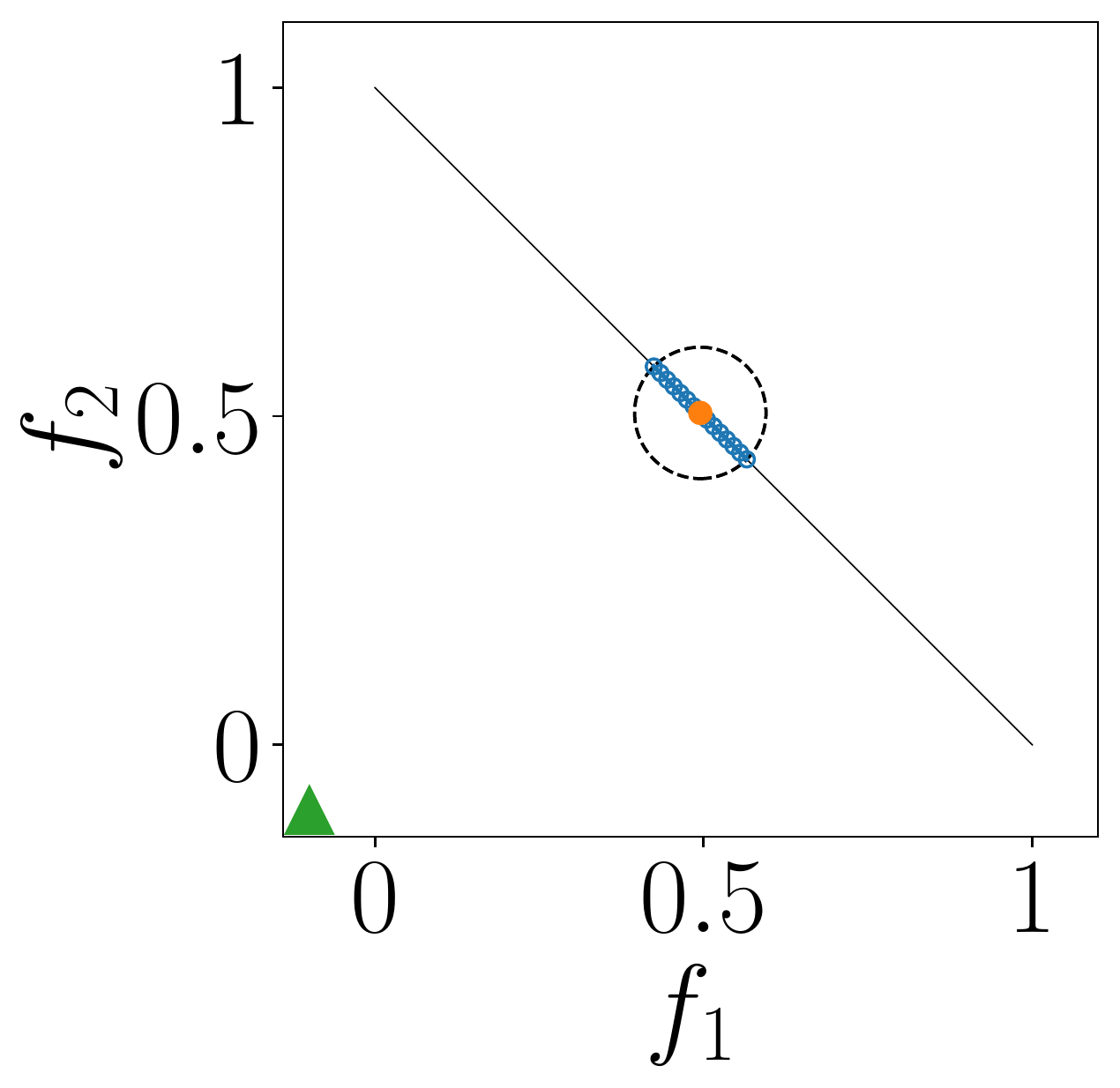}
    }
    \subfloat[ROI-P ($\mathbf{z}^{2}$)]{  
    \includegraphics[width=0.145\textwidth]{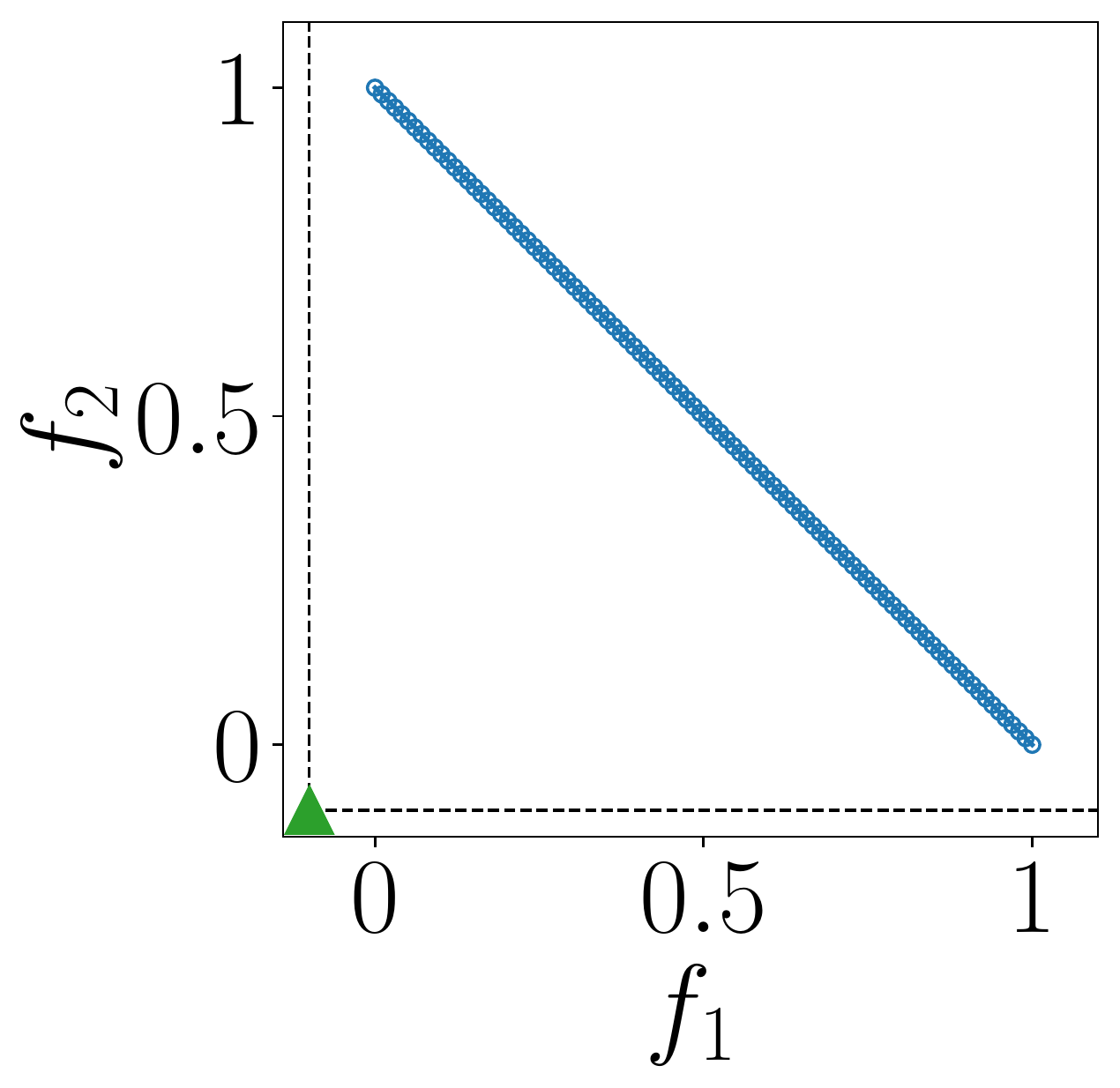}
    }
     \caption{Distributions of Pareto optimal points in the three ROIs on the DTLZ1 problem when using $\mathbf{z}^{0.51} = (0.51, 0.51)^{\top}$, $\mathbf{z}^{0.1}$, and $\mathbf{z}^{-0.1}$. When the reference point is on the Pareto front, the ROI-P contains no point. For this reason, we used $\mathbf{z}^{0.51}$ instead of $\mathbf{z}^{0.5}$.}
   \label{supfig:roi_influence_dtlz1}
\end{figure*}

\begin{figure*}[t]
   \centering
    \subfloat[ROI-C ($\mathbf{z}^{0.5}$)]{  
    \includegraphics[width=0.145\textwidth]{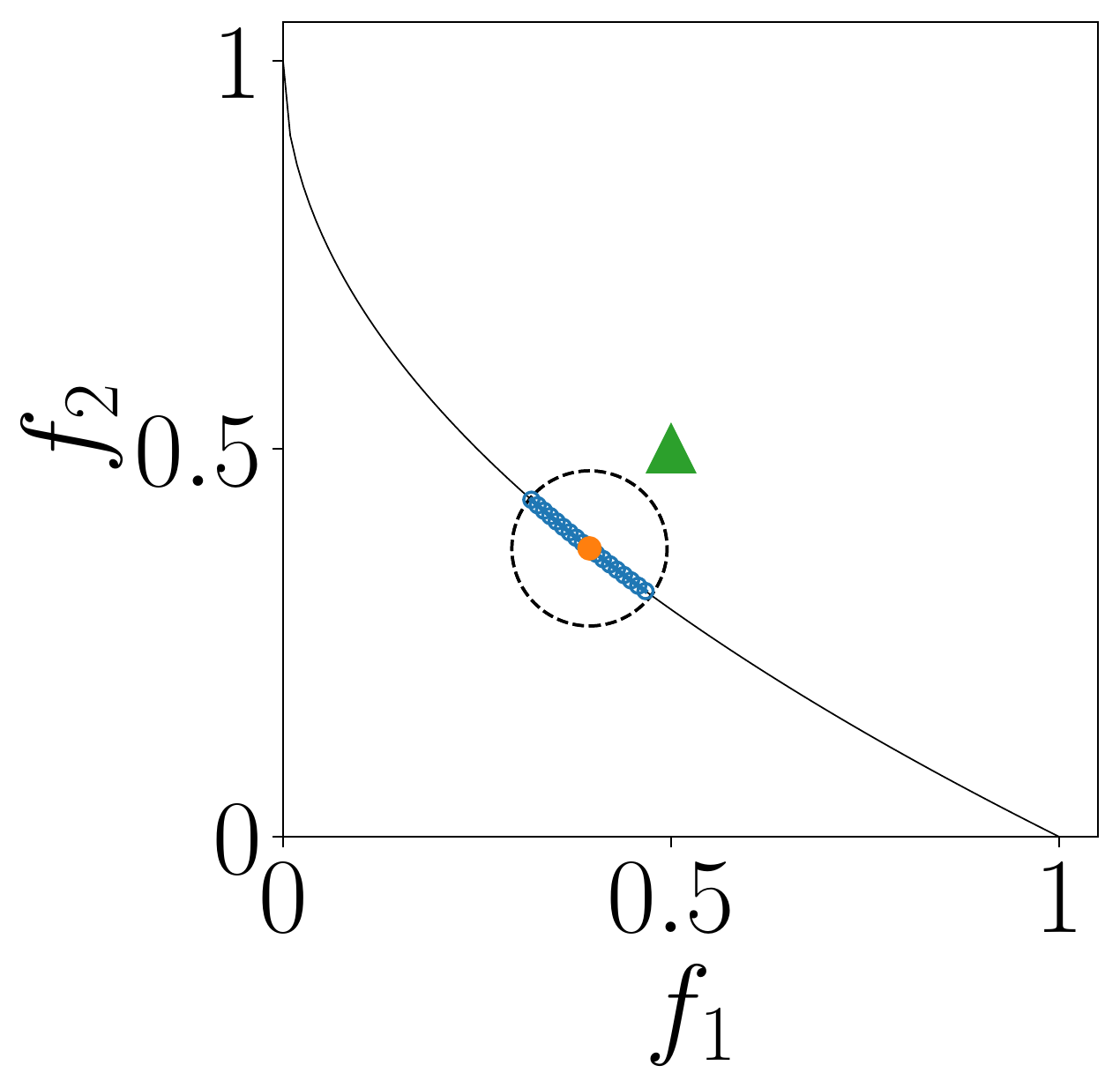}
    }
    \subfloat[ROI-A ($\mathbf{z}^{0.5}$)]{  
    \includegraphics[width=0.145\textwidth]{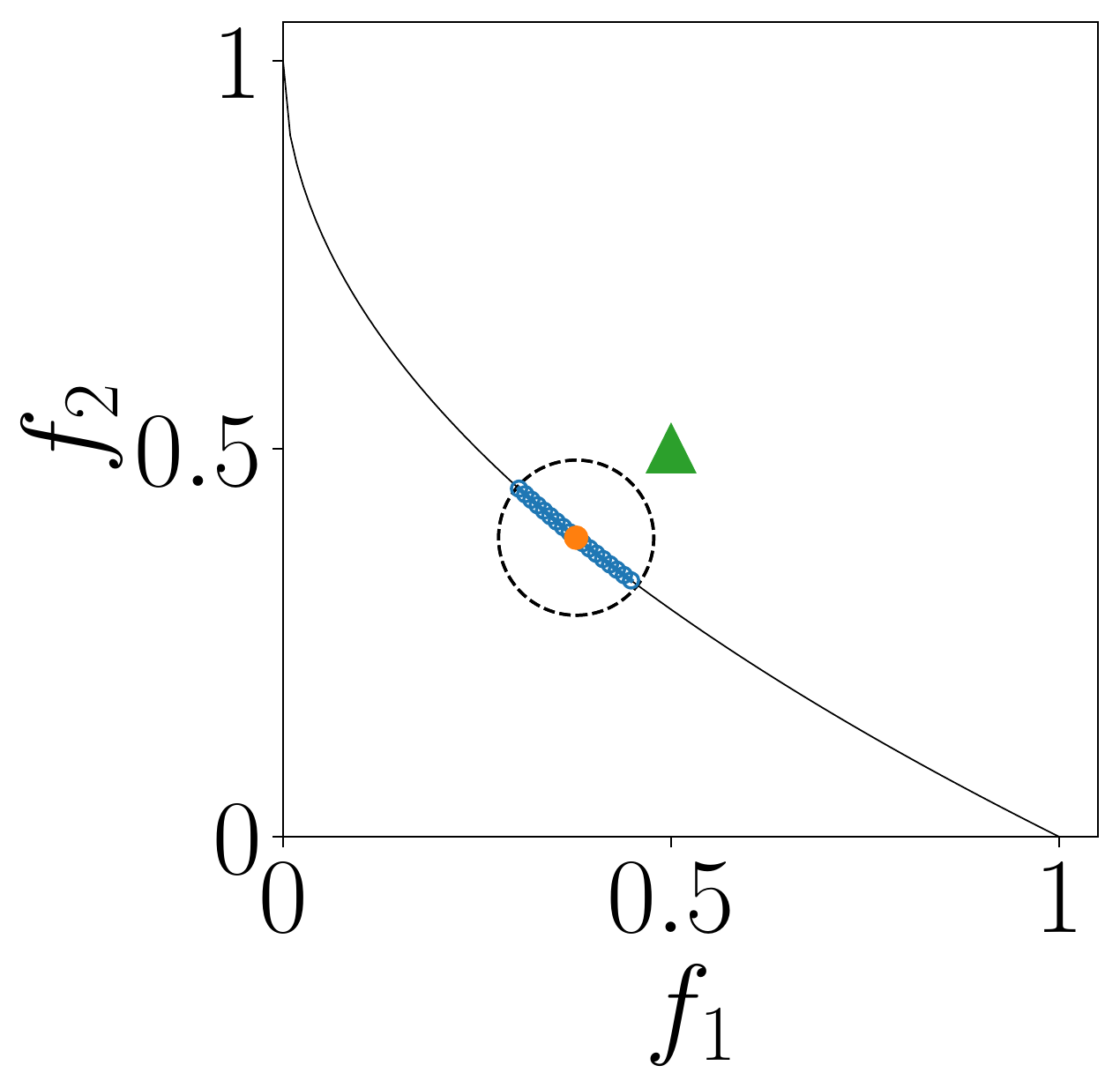}
    }
    \subfloat[ROI-P ($\mathbf{z}^{0.5}$)]{  
    \includegraphics[width=0.145\textwidth]{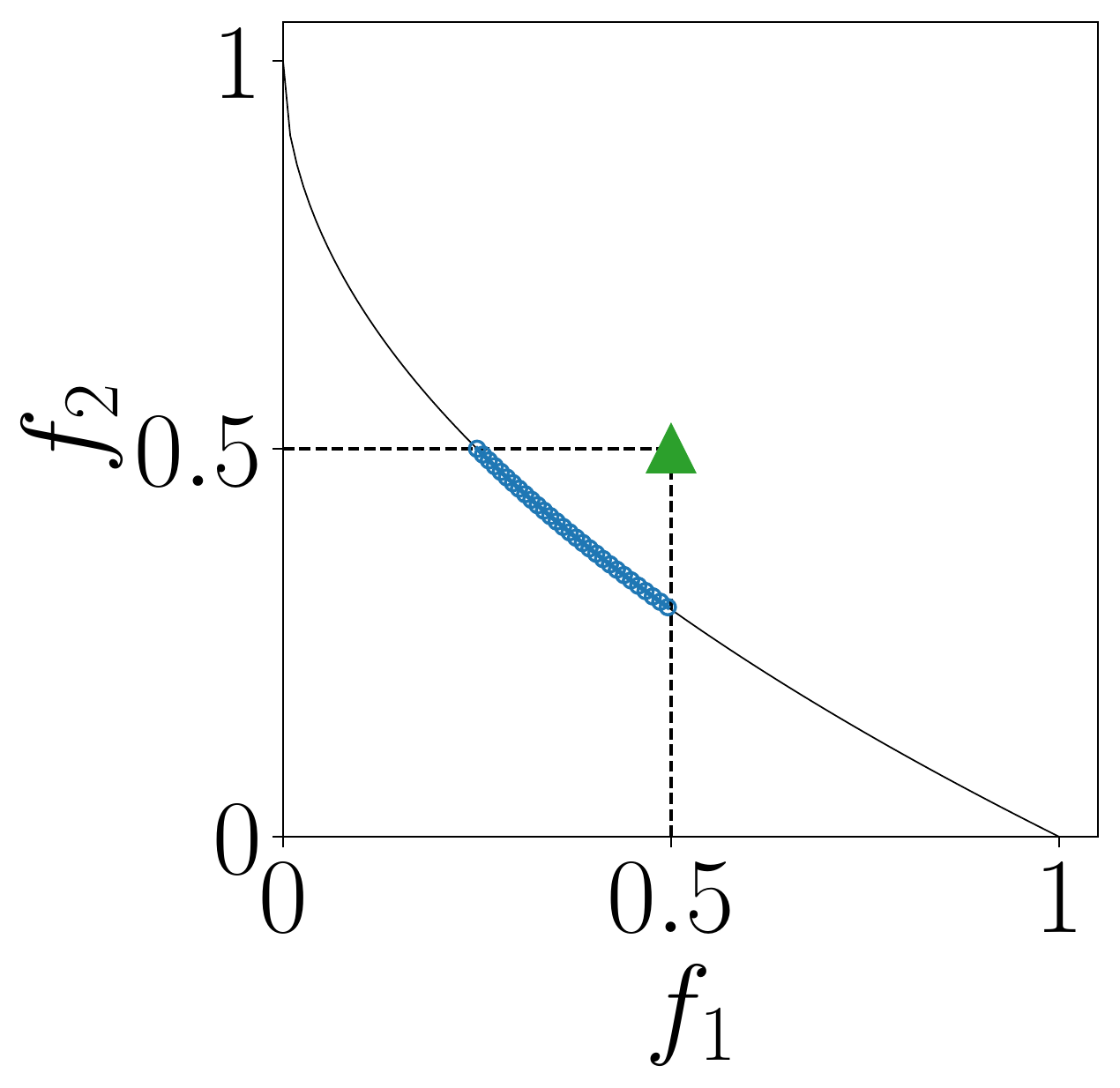}
    }
    \\
    \subfloat[ROI-C ($\mathbf{z}^{0.1}$)]{  
    \includegraphics[width=0.145\textwidth]{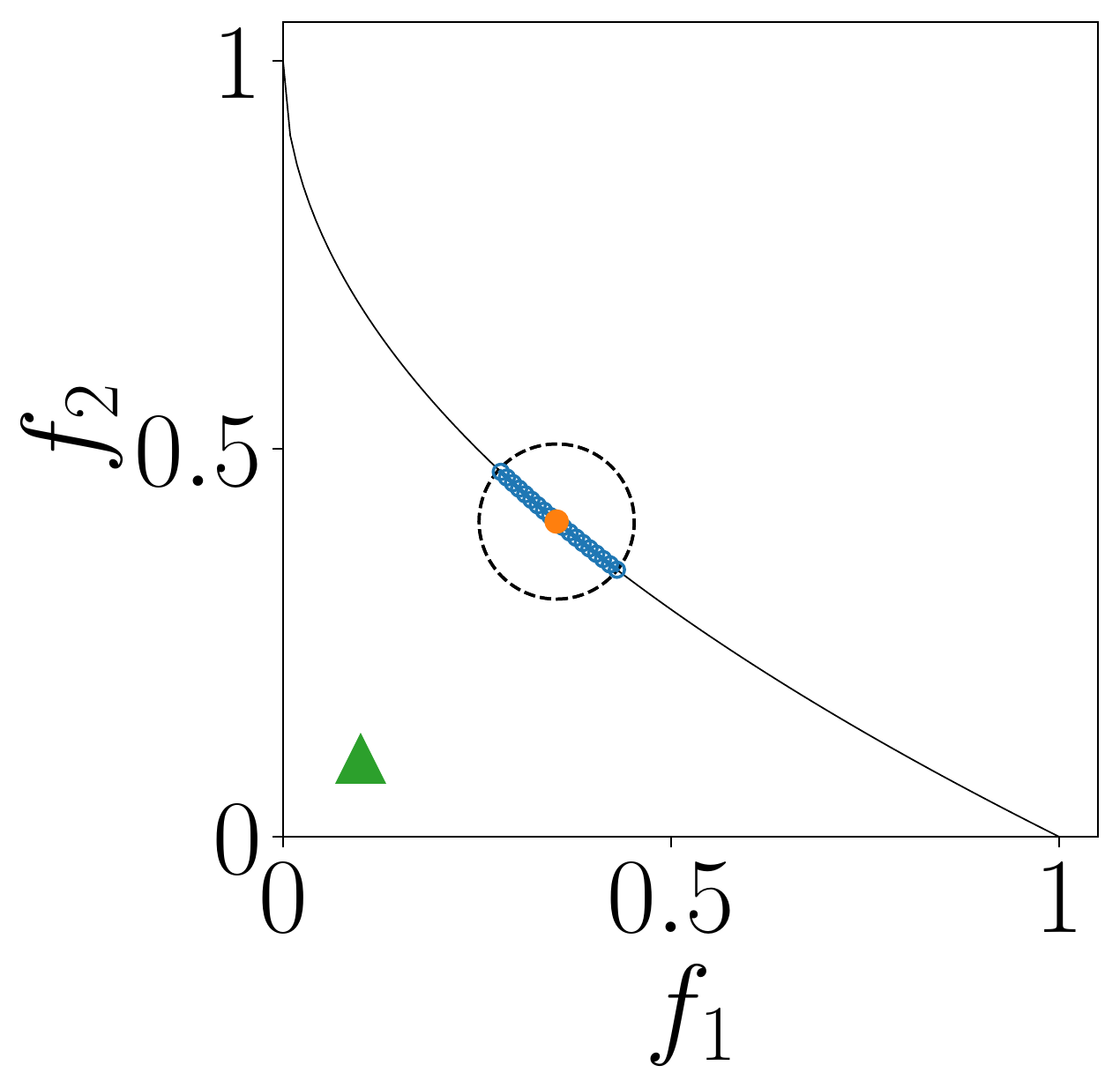}
    }
    \subfloat[ROI-A ($\mathbf{z}^{0.1}$)]{  
    \includegraphics[width=0.145\textwidth]{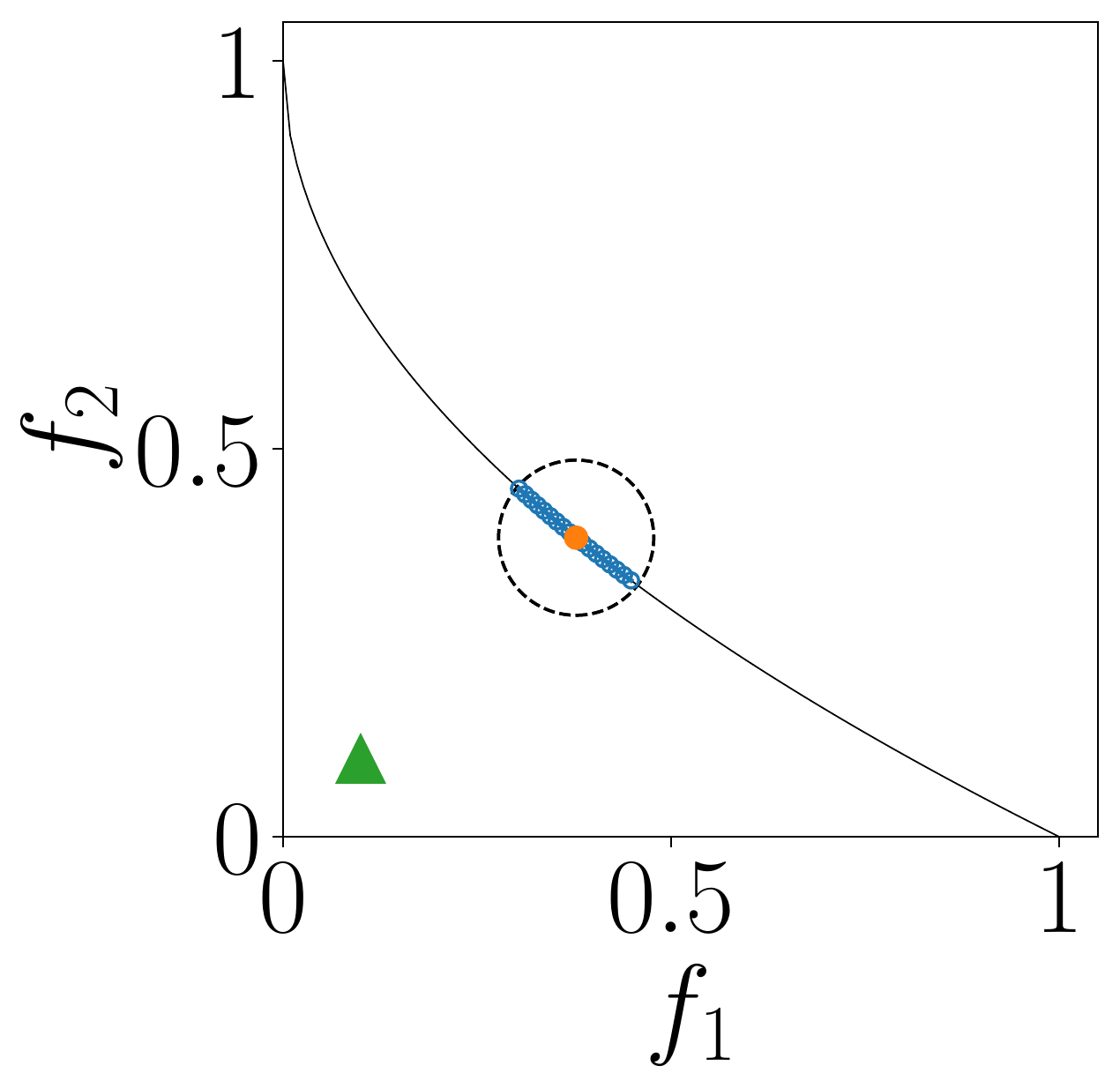}
    }
    \subfloat[ROI-P ($\mathbf{z}^{0.1}$)]{  
    \includegraphics[width=0.145\textwidth]{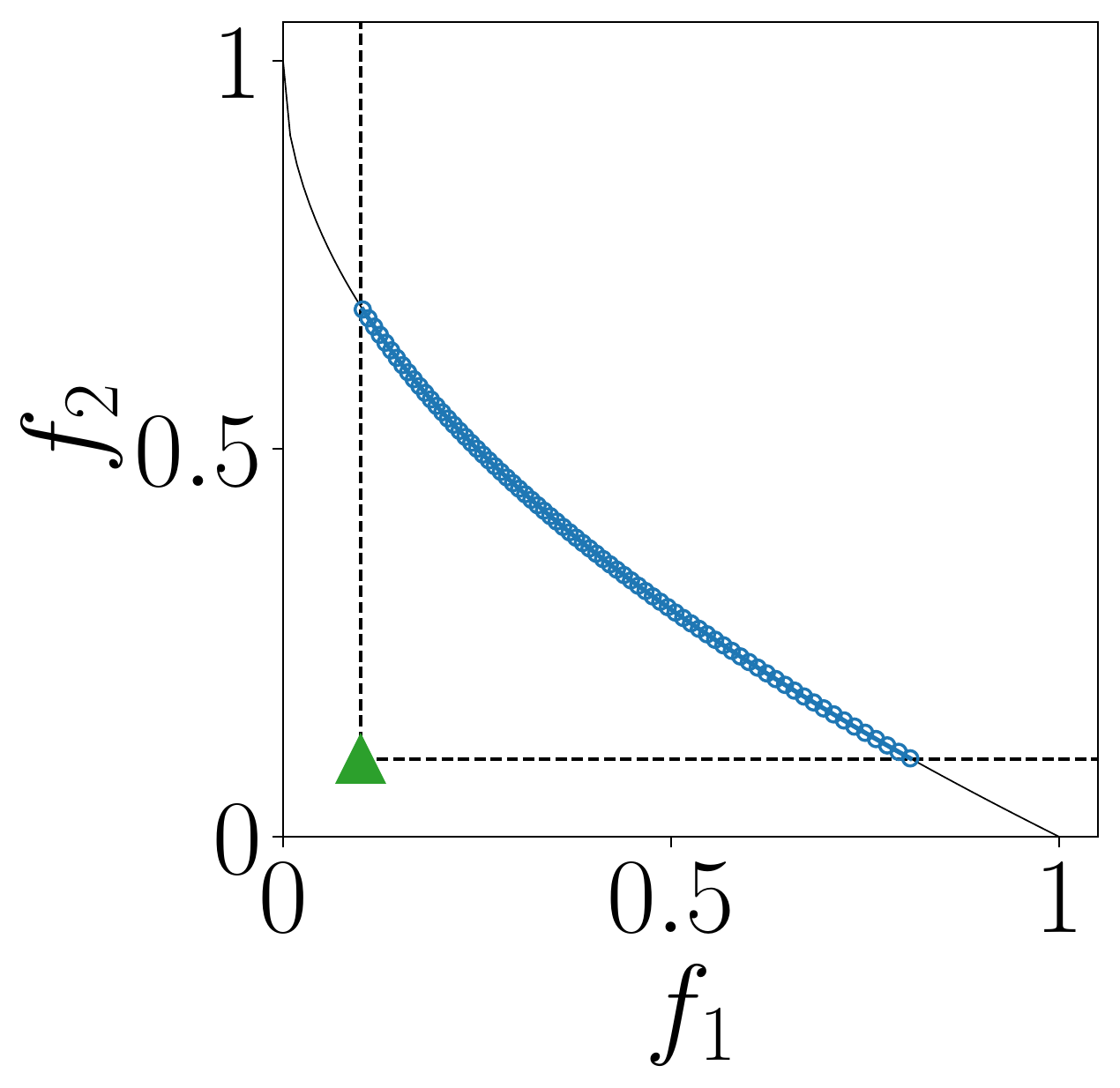}
    }
    \\
    \subfloat[ROI-C ($\mathbf{z}^{2}$)]{  
    \includegraphics[width=0.145\textwidth]{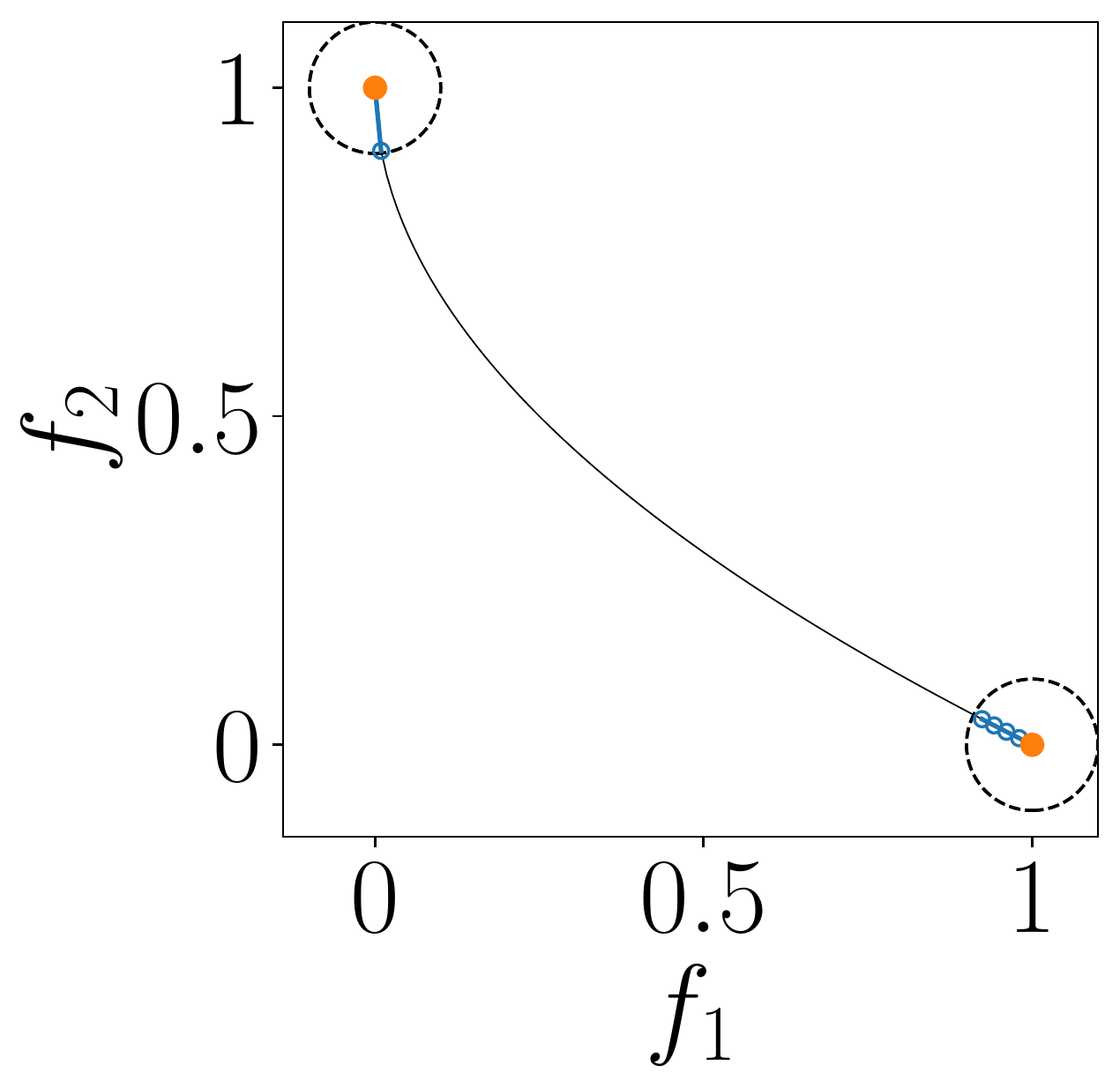}
    }
    \subfloat[ROI-A ($\mathbf{z}^{2}$)]{  
    \includegraphics[width=0.145\textwidth]{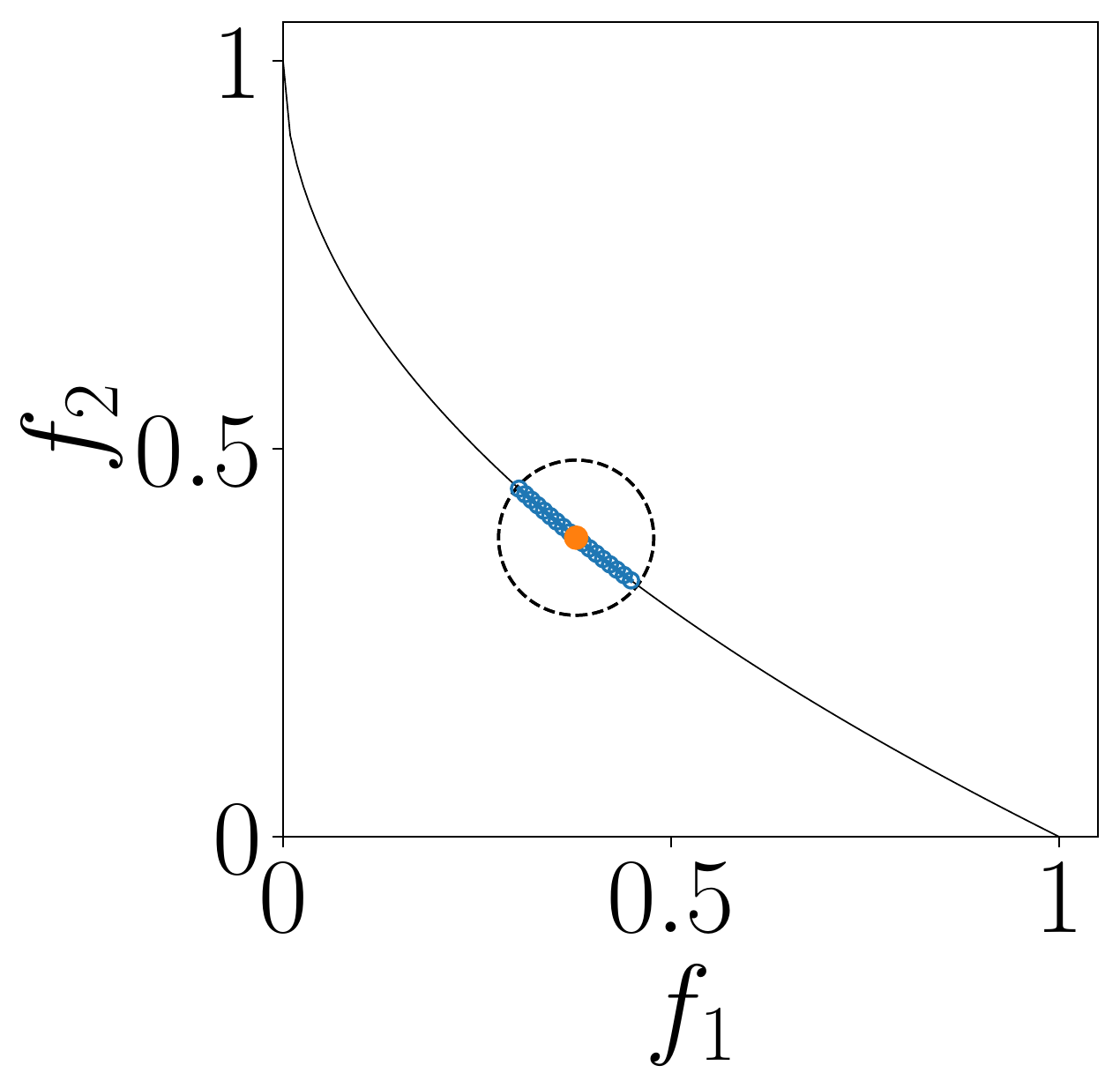}
    }
    \subfloat[ROI-P ($\mathbf{z}^{2}$)]{  
    \includegraphics[width=0.145\textwidth]{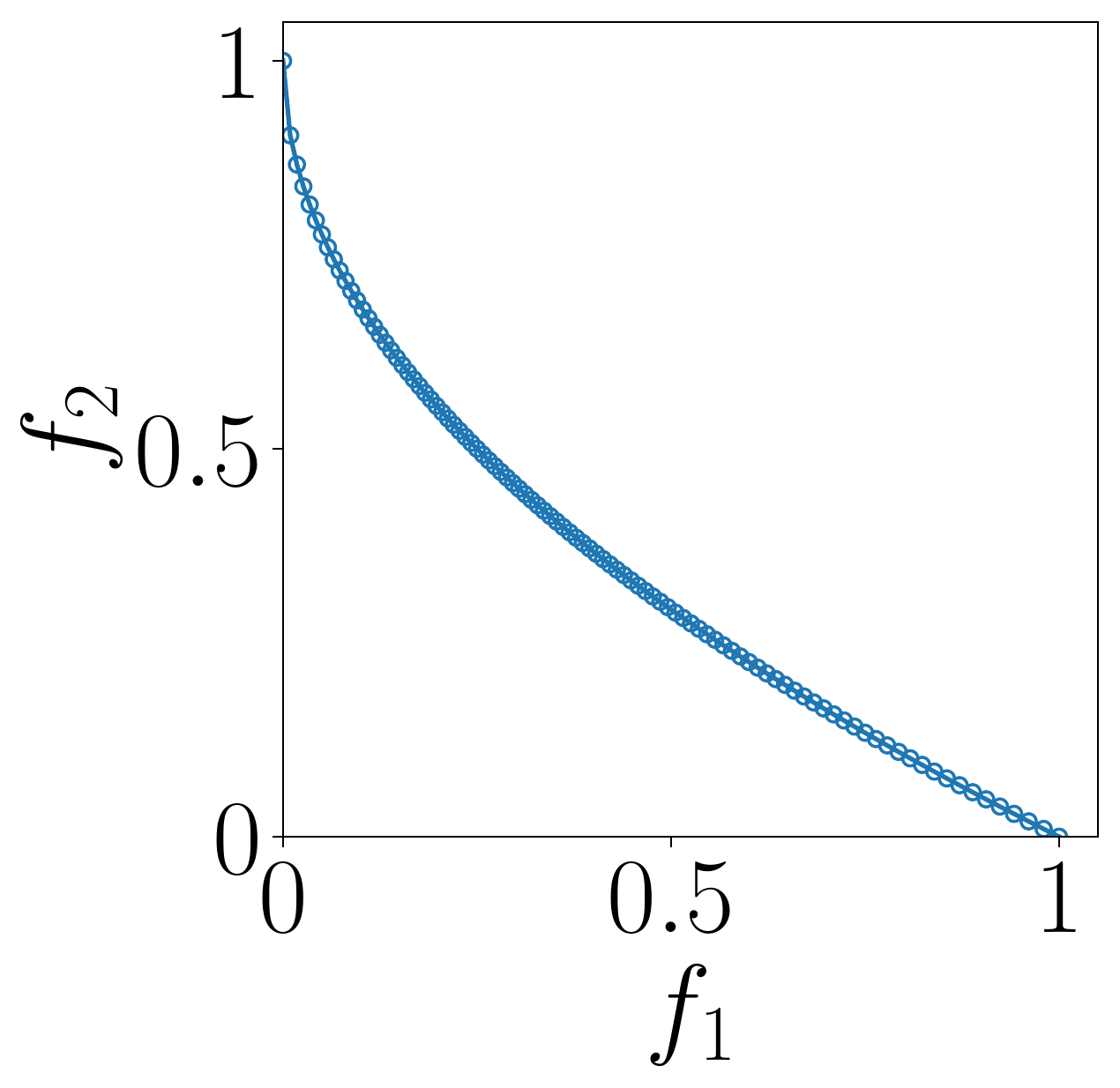}
    }
     \caption{Distributions of Pareto optimal points in the three ROIs on the convDTLZ2 problem when using $\mathbf{z}^{0.5}$, $\mathbf{z}^{0.1}$, and $\mathbf{z}^{2}$.}
   \label{supfig:roi_influence_convdtlz2}
\end{figure*}


\begin{figure*}[t]
   \centering
   \subfloat[R-NSGA-II]{  
     \includegraphics[width=0.145\textwidth]{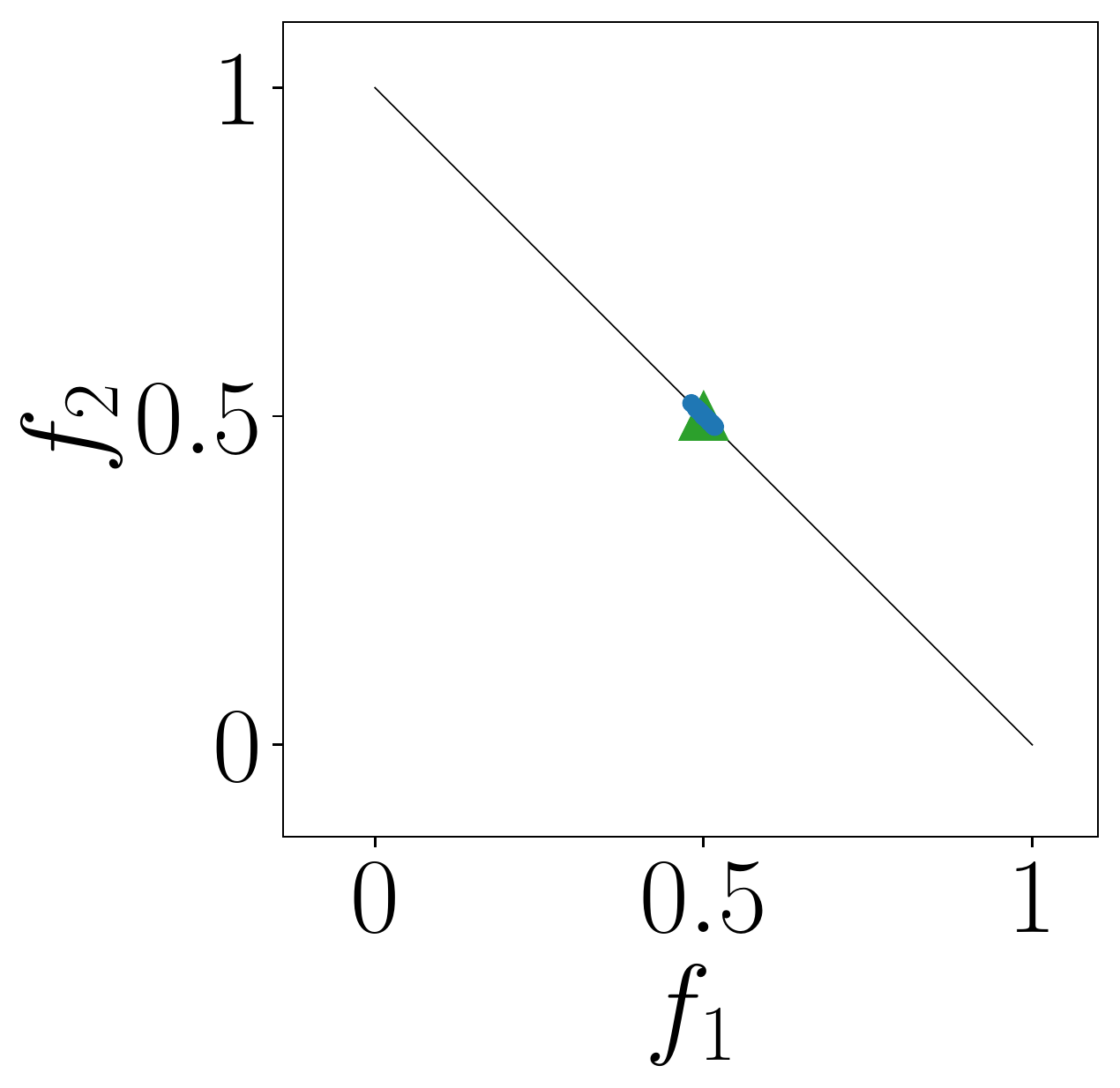}
     }
   \subfloat[r-NSGA-II]{  
     \includegraphics[width=0.145\textwidth]{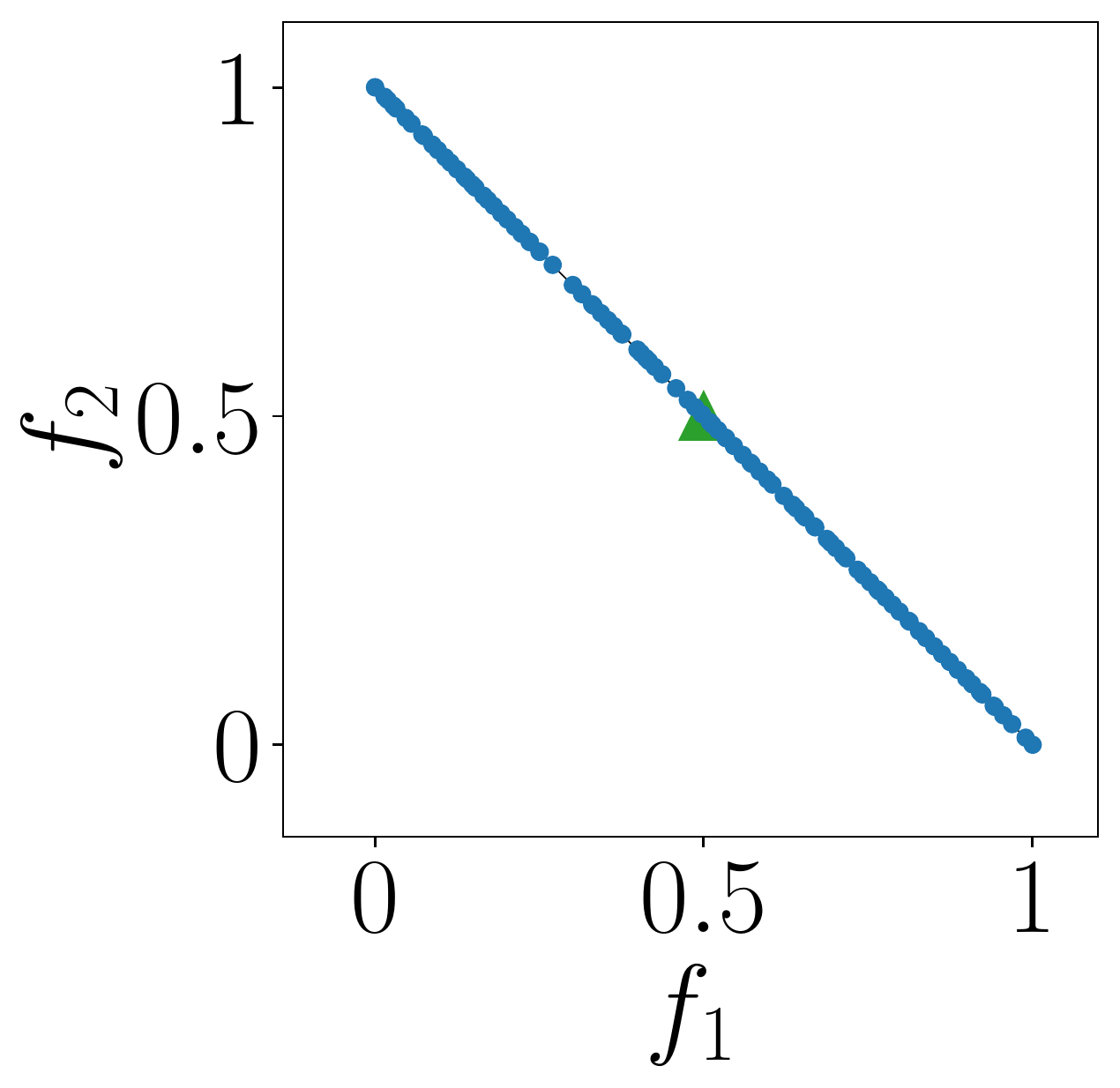}      
     }                                                  
   \subfloat[g-NSGA-II]{  
     \includegraphics[width=0.145\textwidth]{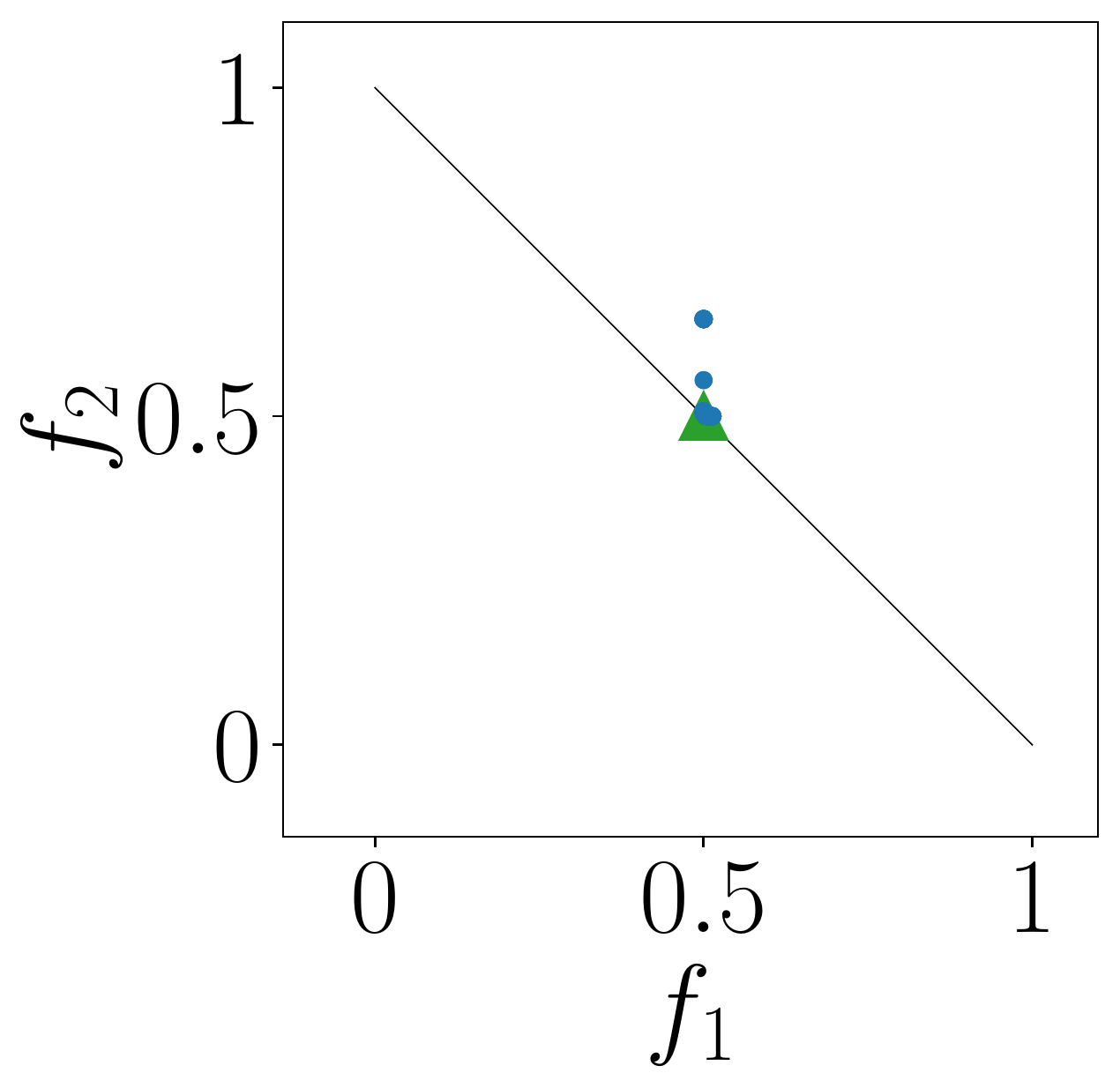}
     }
   \subfloat[PBEA]{  
     \includegraphics[width=0.145\textwidth]{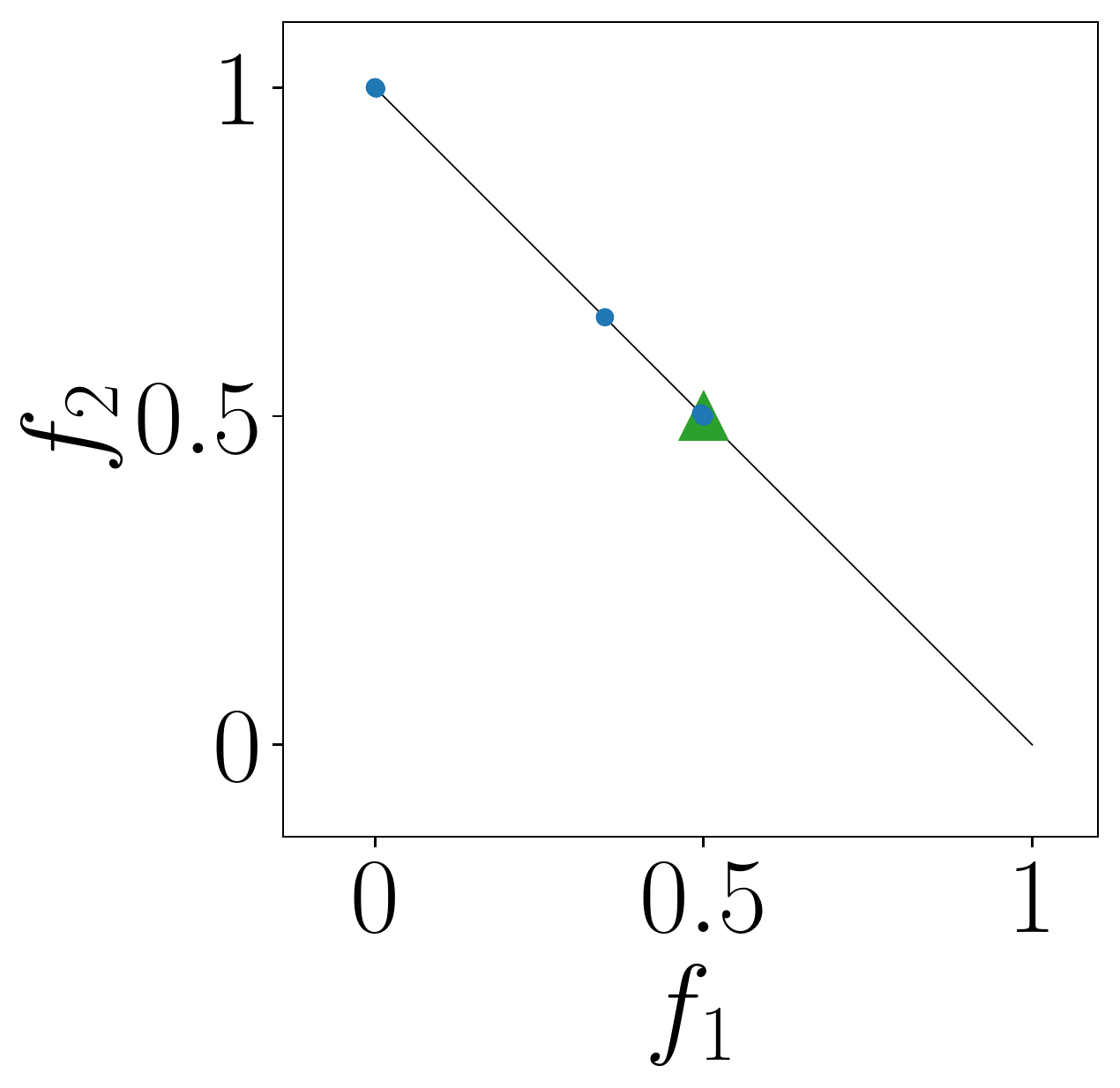}      
     }     
\subfloat[R-MEAD2]{  
     \includegraphics[width=0.145\textwidth]{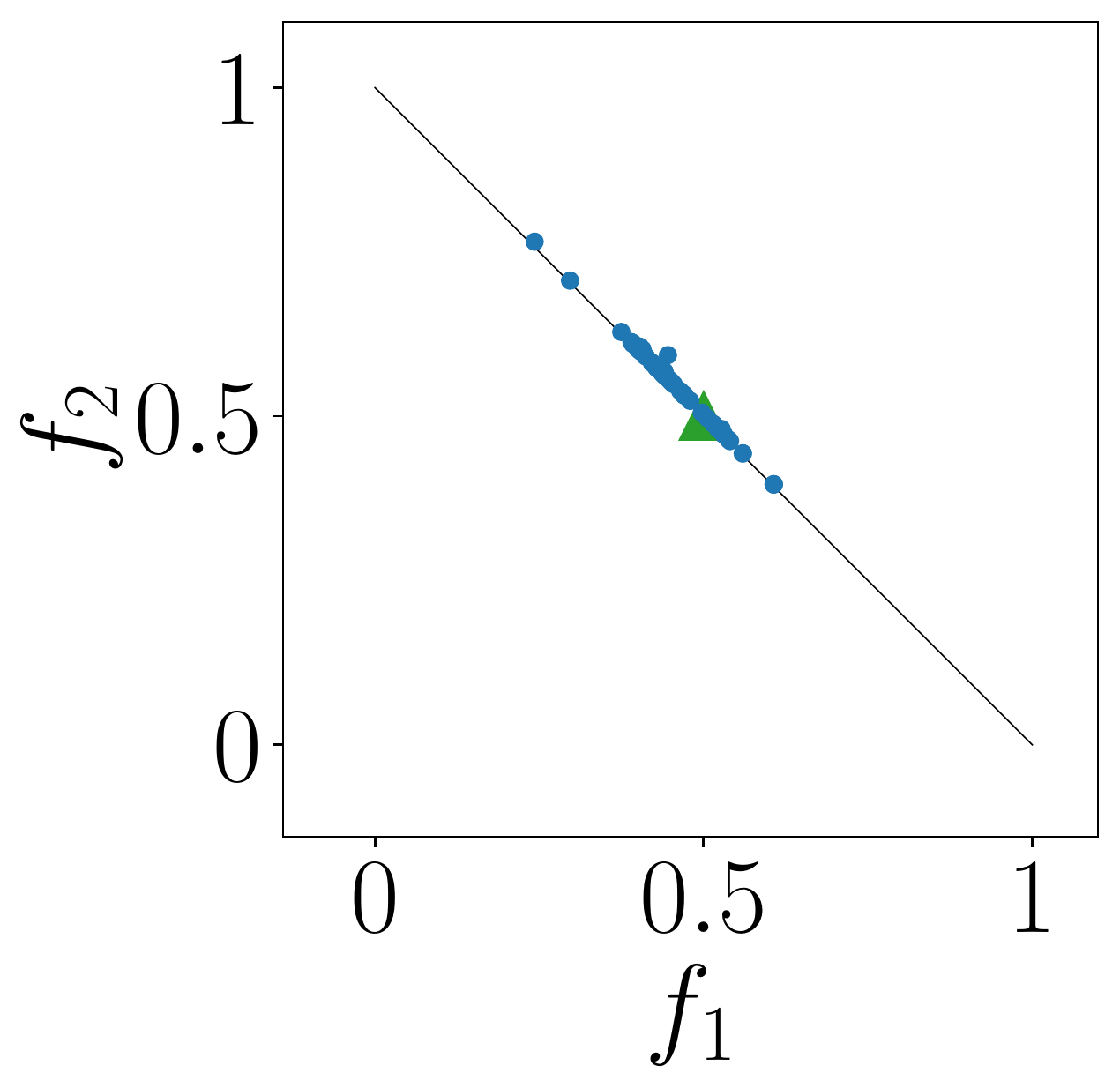}      
     }         
     \subfloat[MOEA/D-NUMS]{  
     \includegraphics[width=0.145\textwidth]{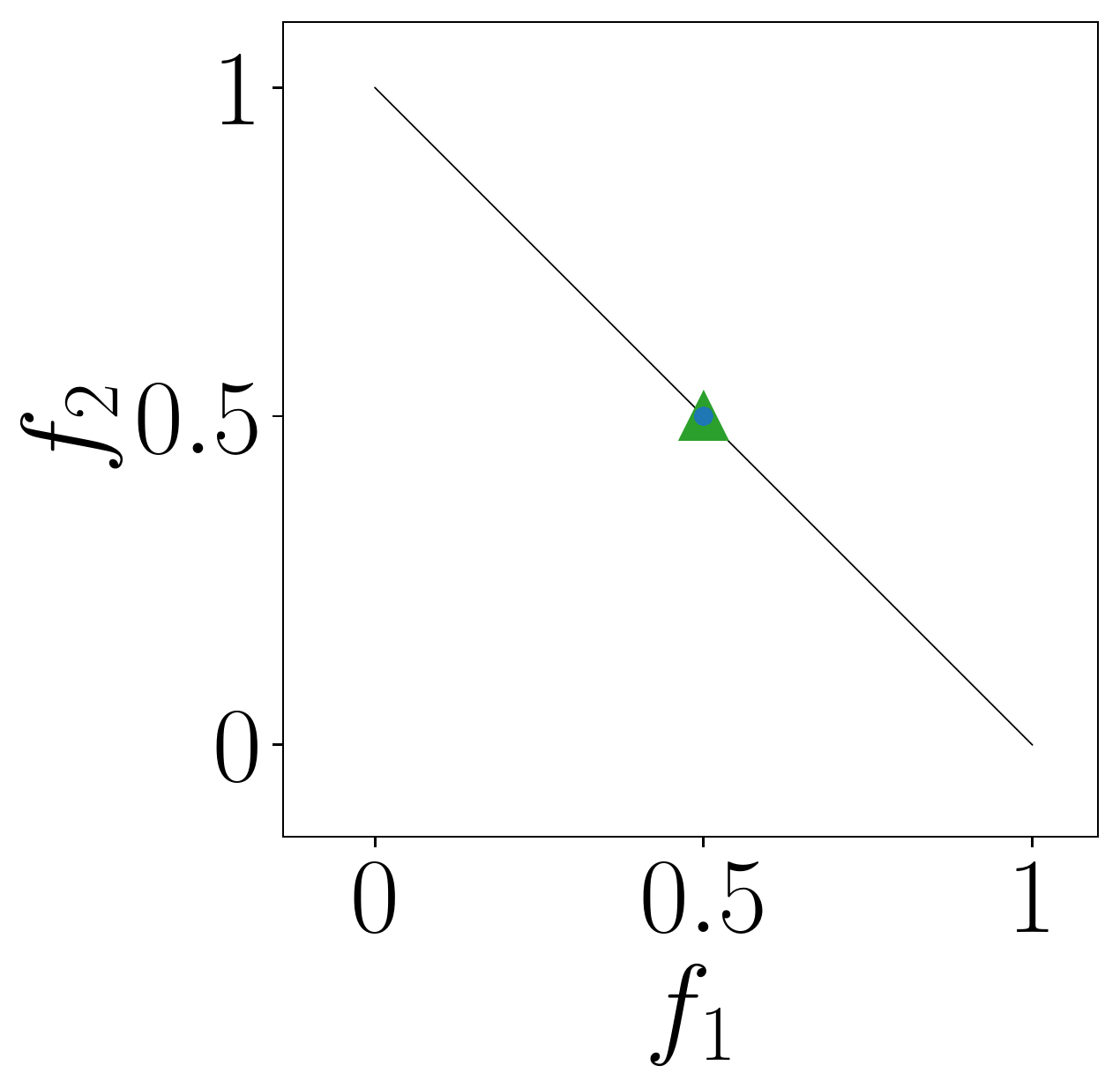}        
     }                               
     \caption{Distributions of points found by the six EMO algorithms on the DTLZ1 problem when using $\mathbf{z}^{0.5} =(0.5, 0.5)^{\top}$.}
   \label{supfig:emo_points_dtlz1_z0.5}
   \subfloat[R-NSGA-II]{  
     \includegraphics[width=0.145\textwidth]{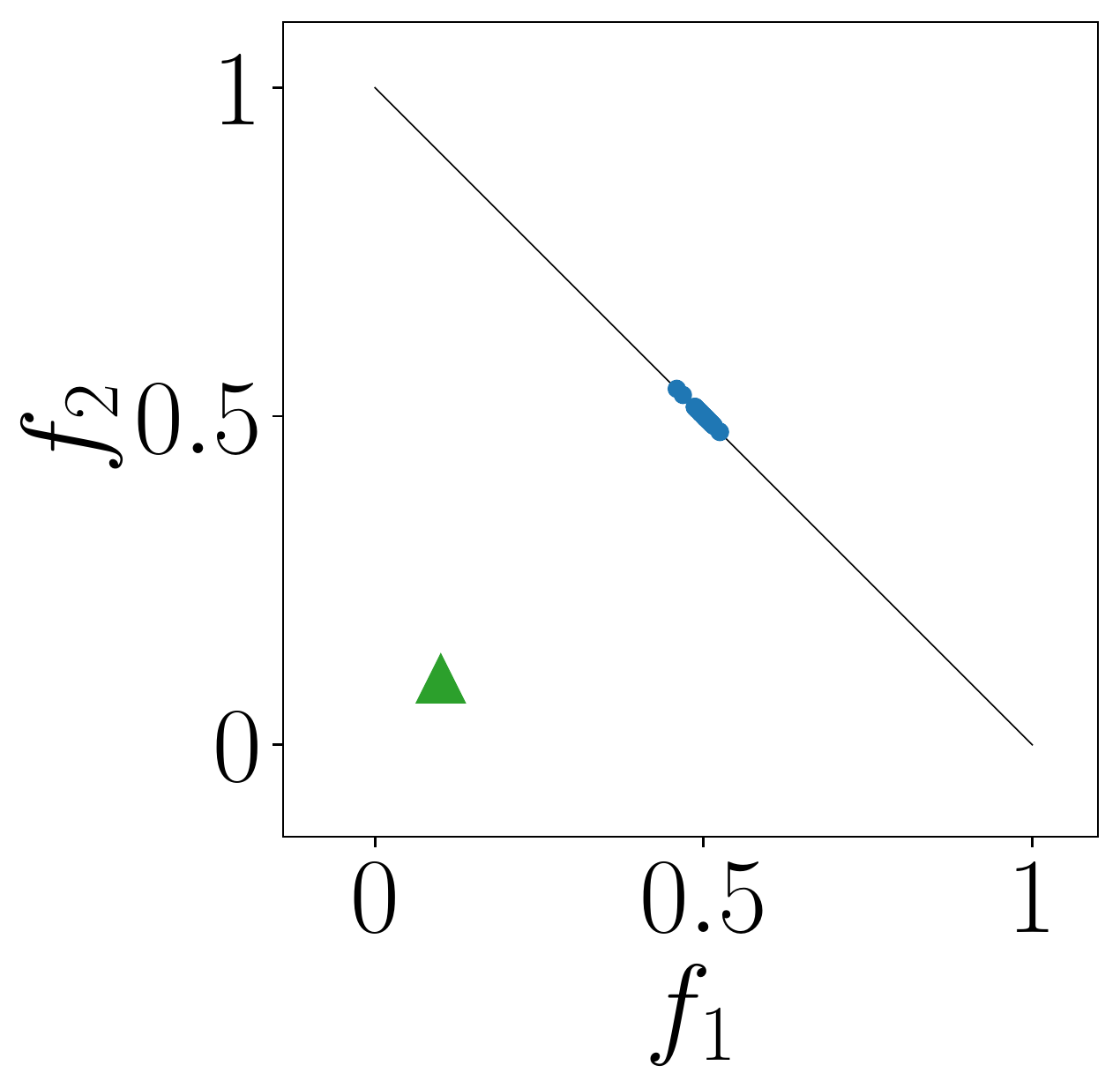}
     }
   \subfloat[r-NSGA-II]{  
     \includegraphics[width=0.145\textwidth]{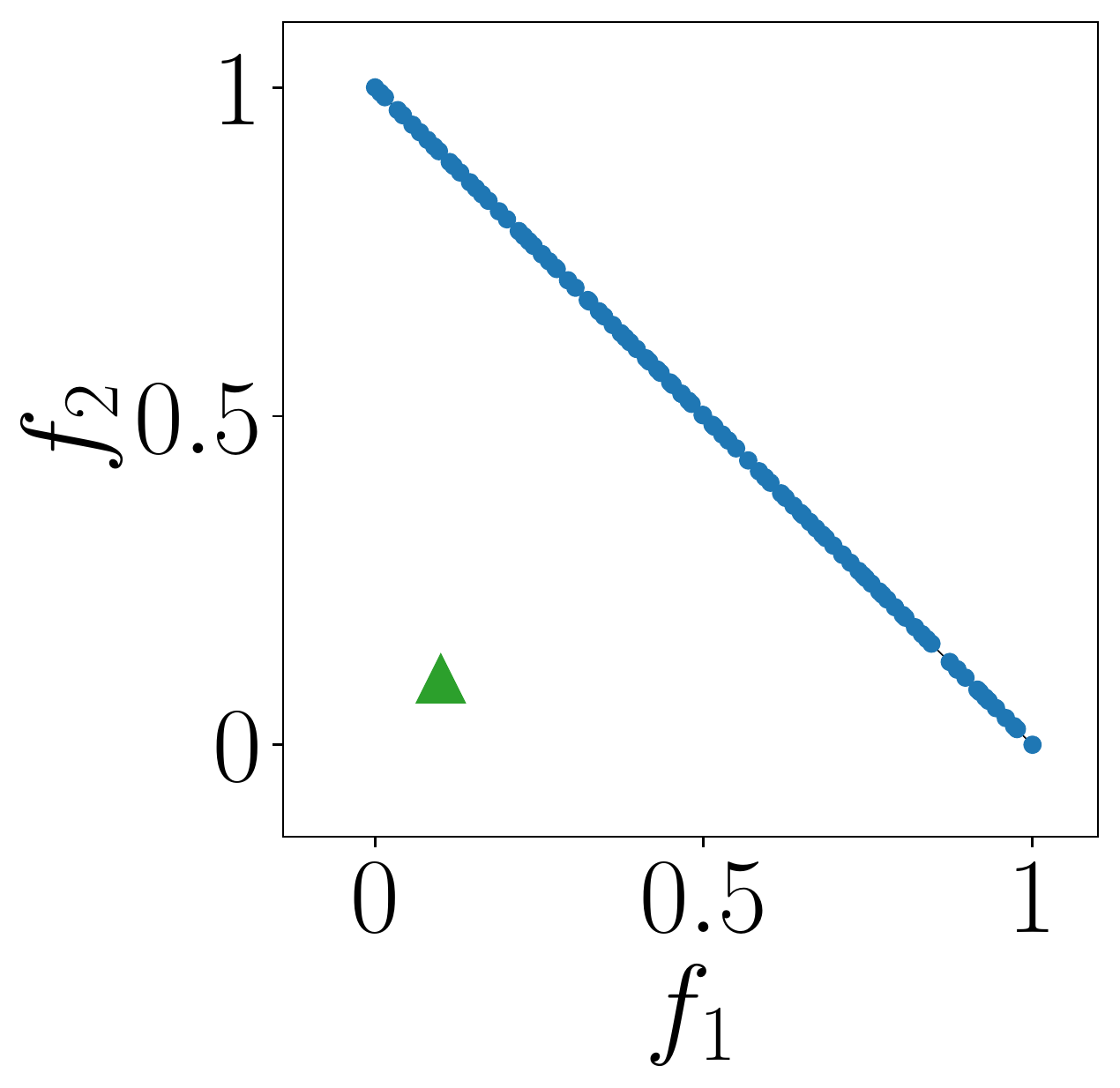}      
     }                                                                                      
   \subfloat[g-NSGA-II]{  
     \includegraphics[width=0.145\textwidth]{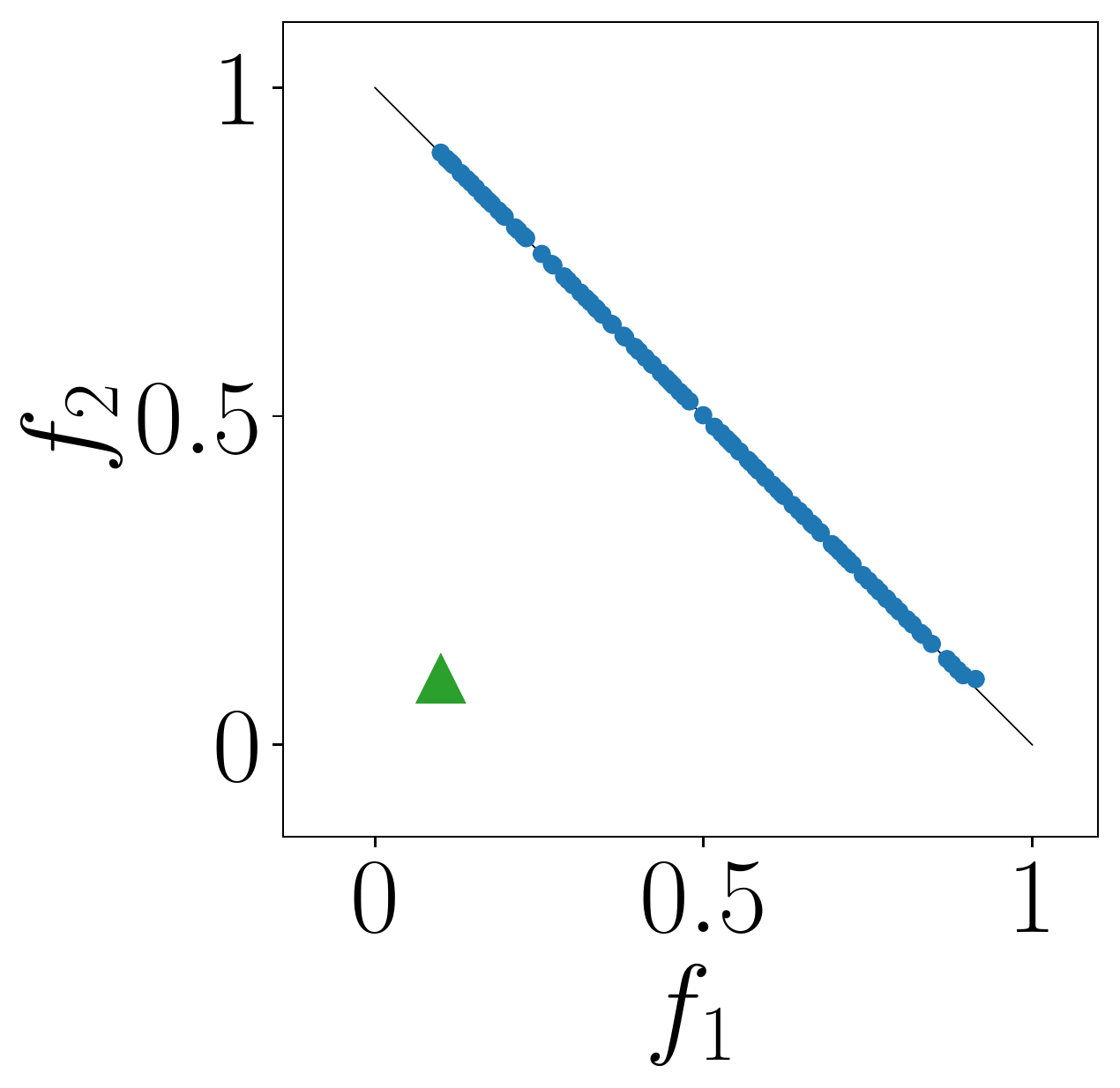}
     }
   \subfloat[PBEA]{  
     \includegraphics[width=0.145\textwidth]{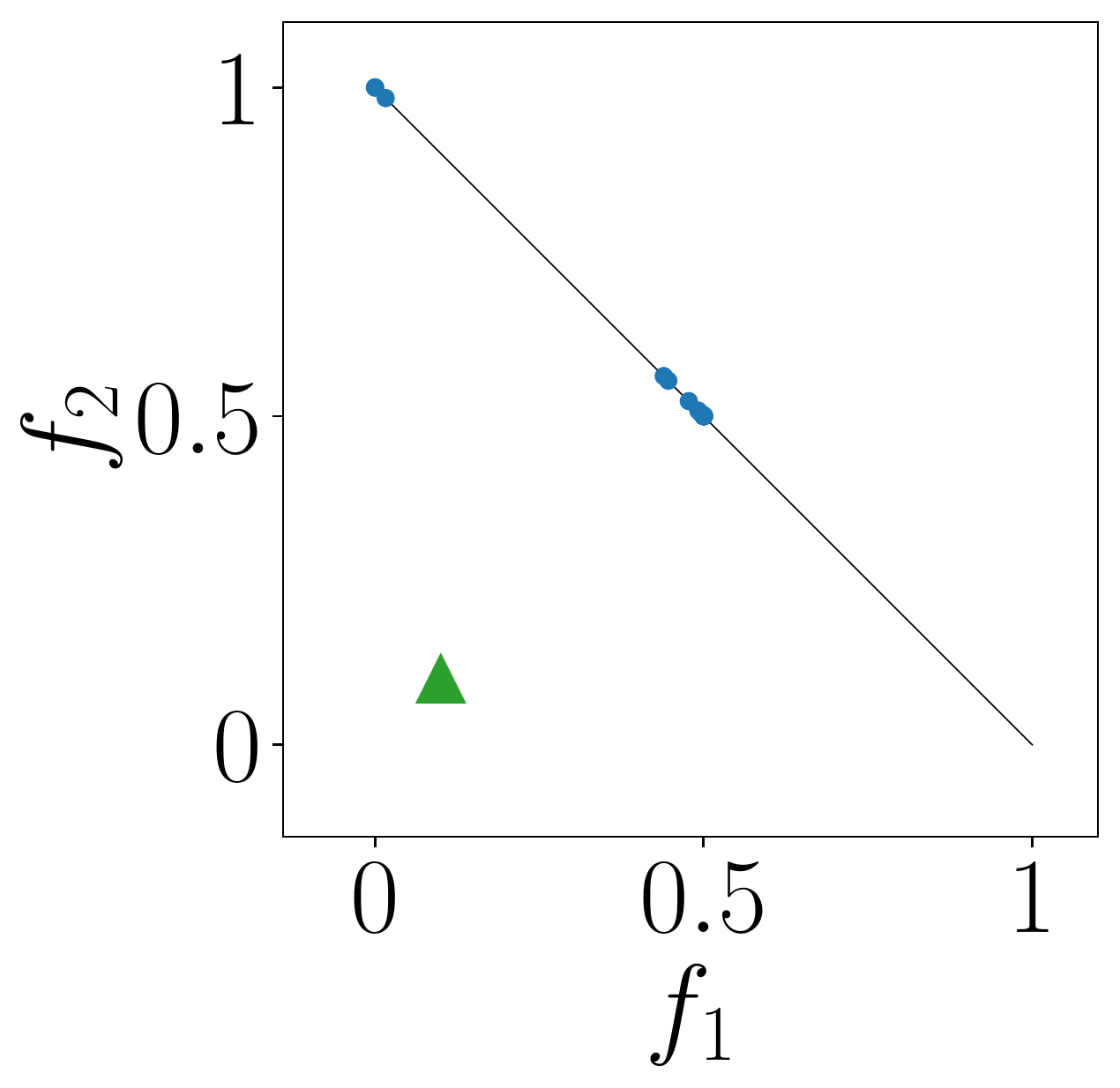}      
     }     
\subfloat[R-MEAD2]{  
     \includegraphics[width=0.145\textwidth]{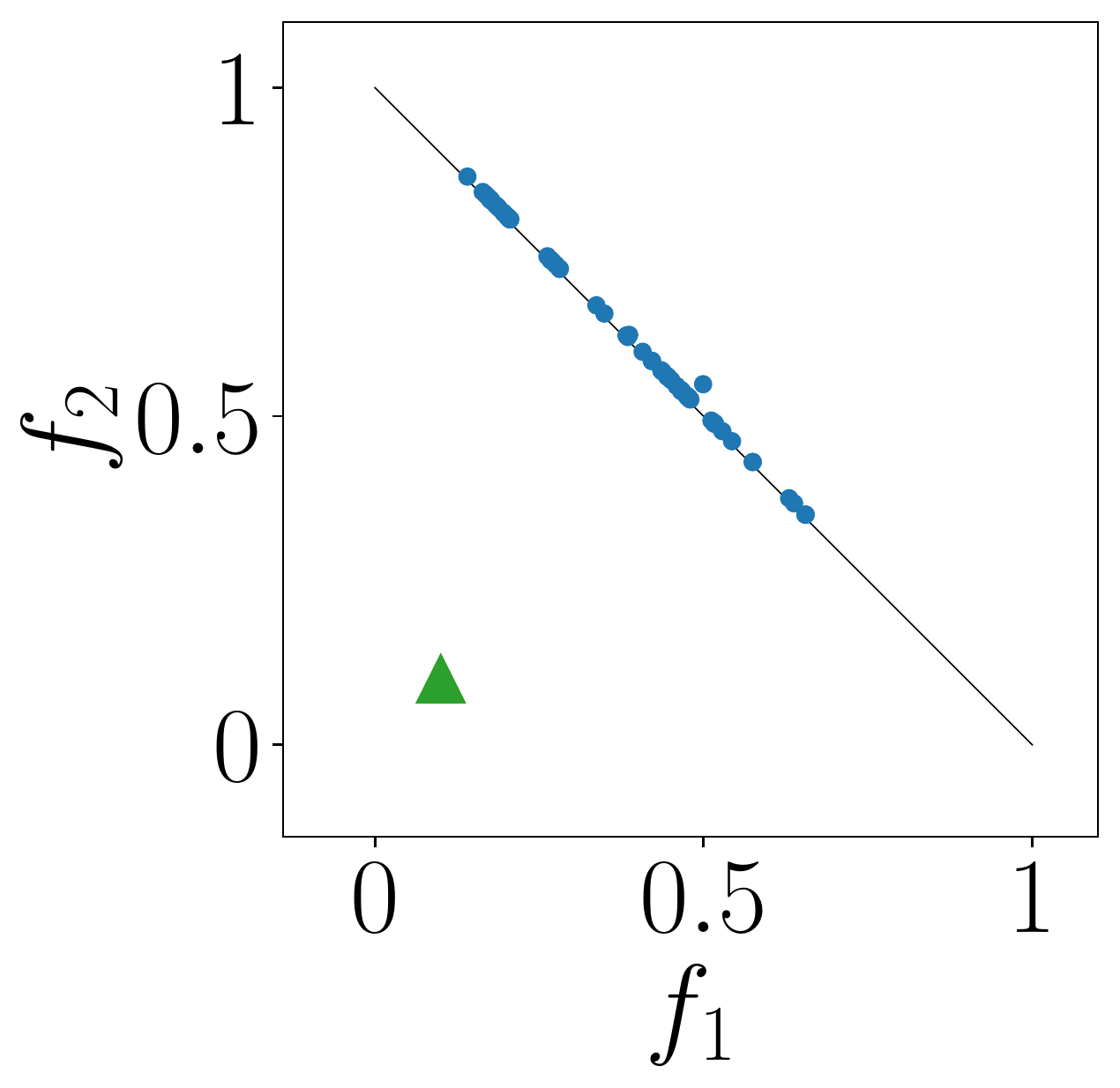}      
     }         
     \subfloat[MOEA/D-NUMS]{  
     \includegraphics[width=0.145\textwidth]{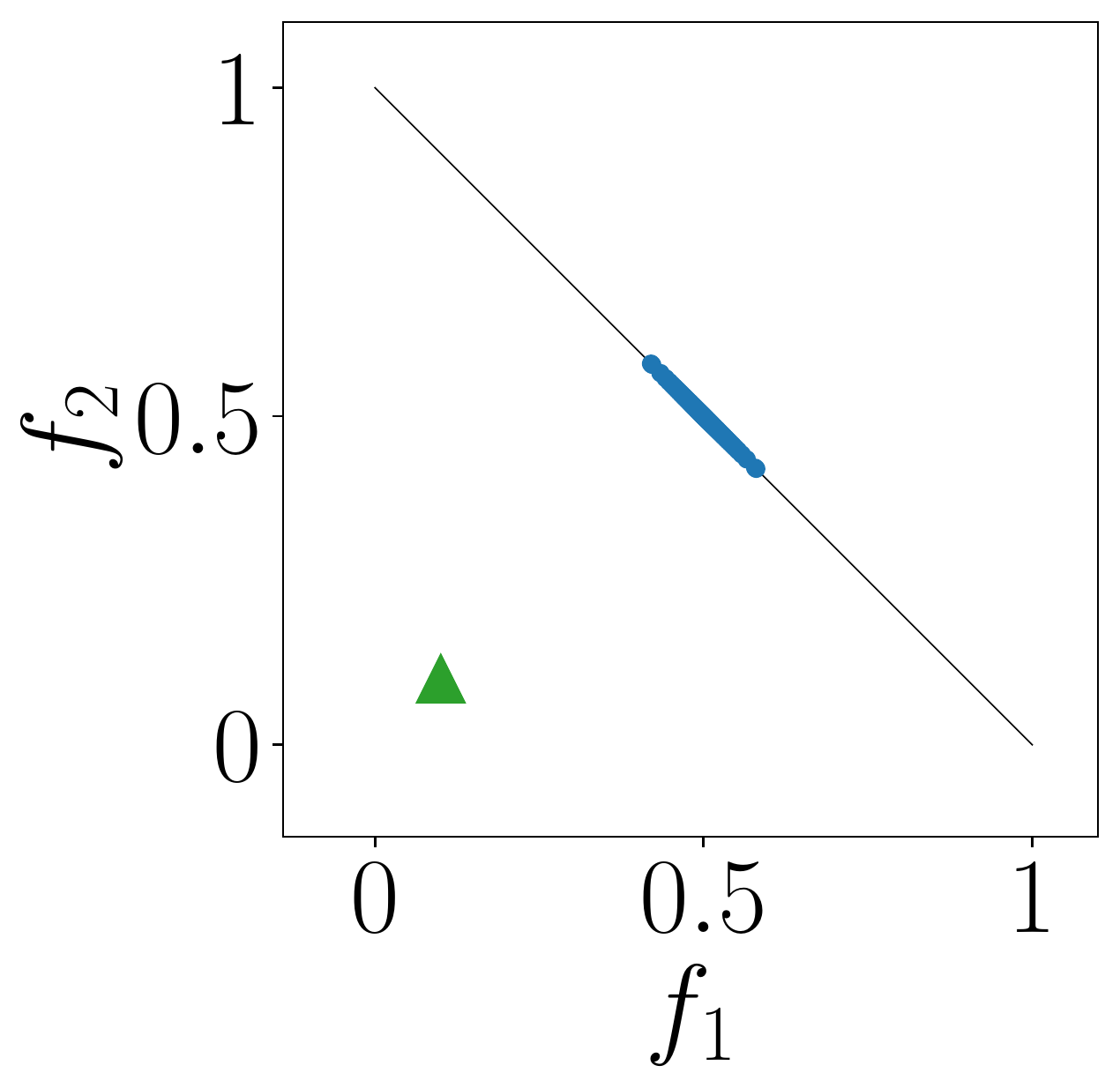}        
     }                                  
     \caption{Distributions of points found by the six EMO algorithms on the DTLZ1 problem when using $\mathbf{z}^{0.1} =(0.1, 0.1)^{\top}$.}
   \label{supfig:emo_points_dtlz1_z0.1}
   \subfloat[R-NSGA-II]{  
     \includegraphics[width=0.145\textwidth]{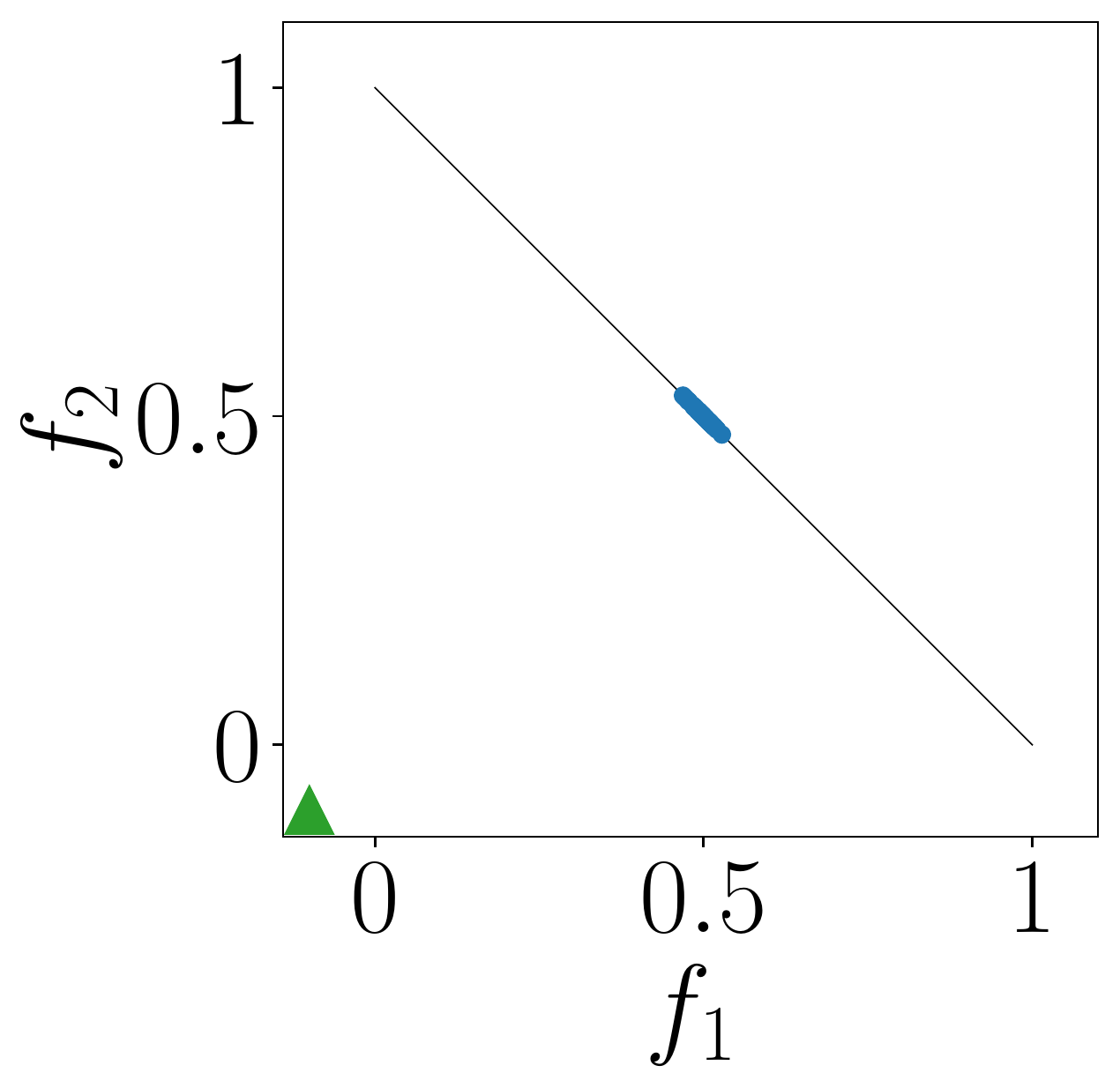}
     }
   \subfloat[r-NSGA-II]{  
     \includegraphics[width=0.145\textwidth]{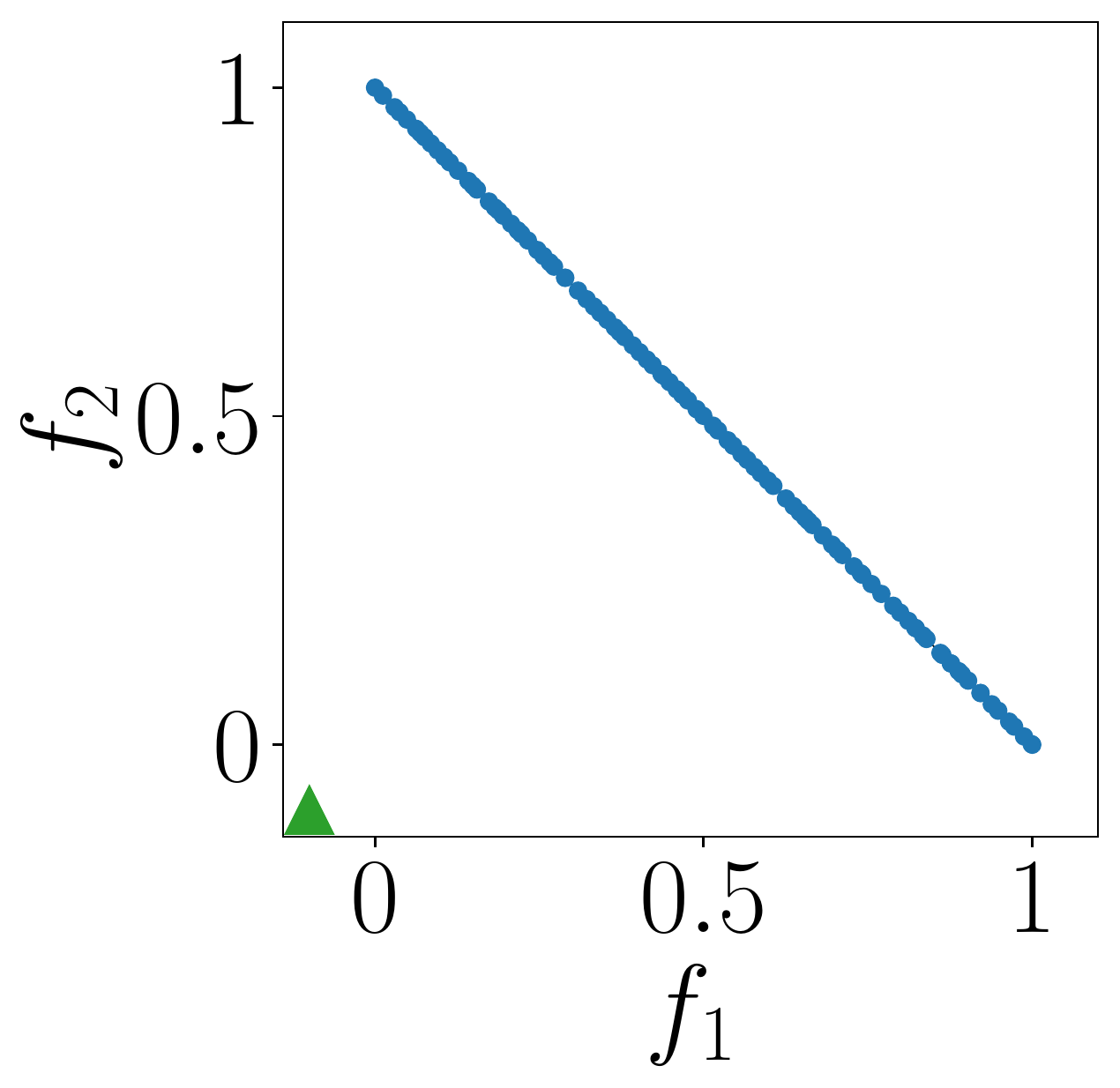}      
     }                                                                                      
   \subfloat[g-NSGA-II]{  
     \includegraphics[width=0.145\textwidth]{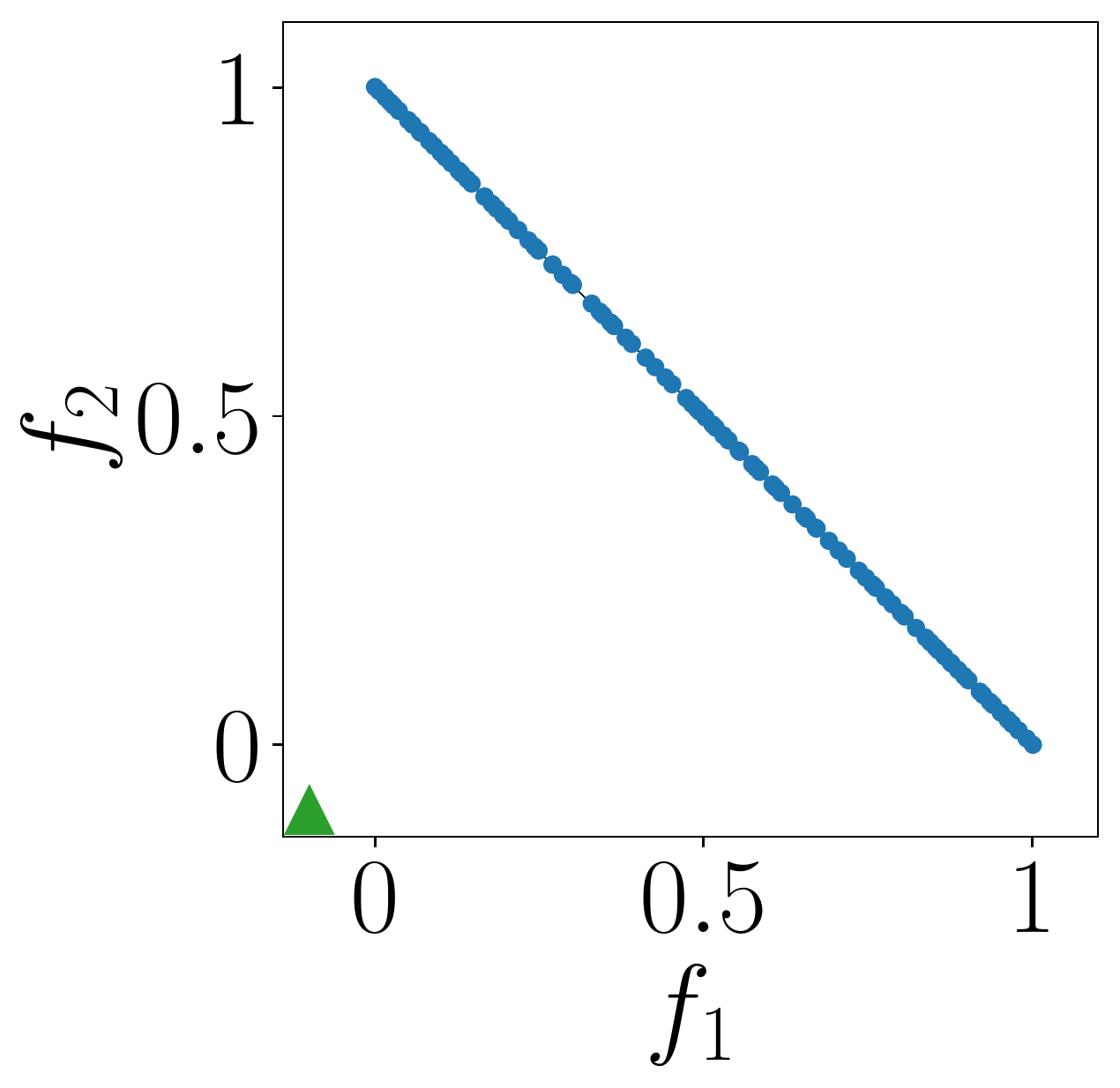}
     }
   \subfloat[PBEA]{  
     \includegraphics[width=0.145\textwidth]{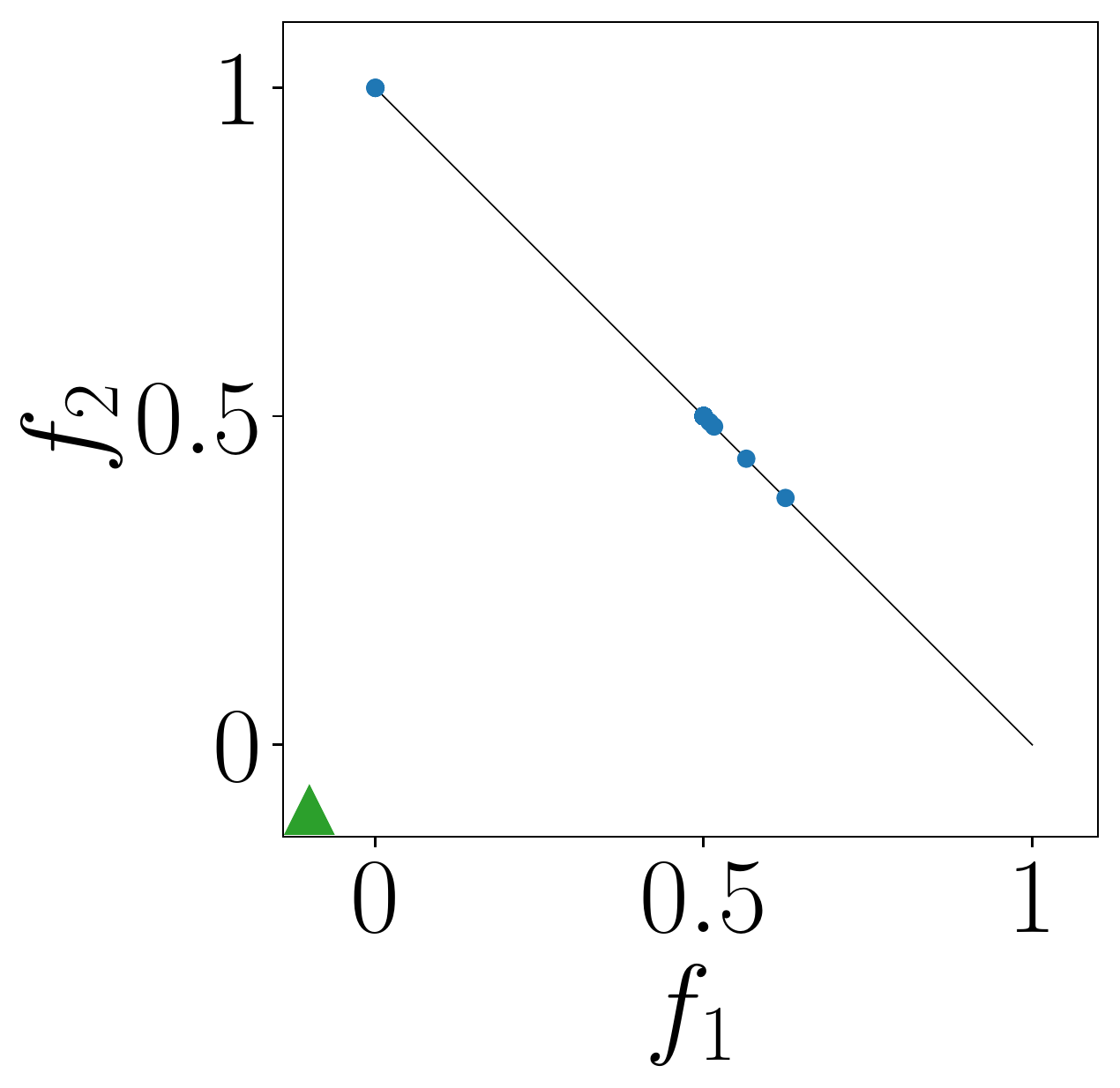}      
     }     
\subfloat[R-MEAD2]{  
     \includegraphics[width=0.145\textwidth]{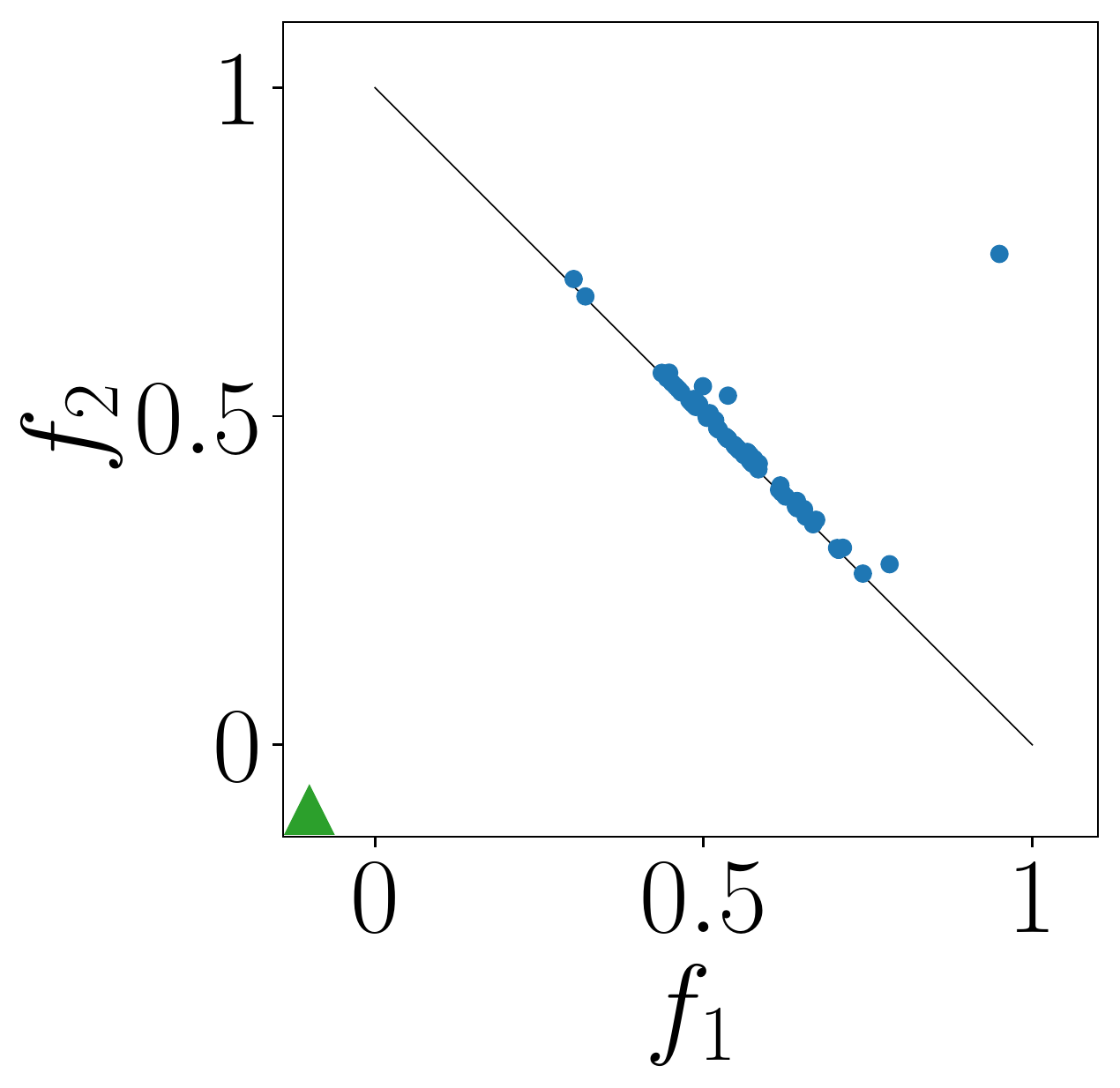}      
     }         
     \subfloat[MOEA/D-NUMS]{  
     \includegraphics[width=0.145\textwidth]{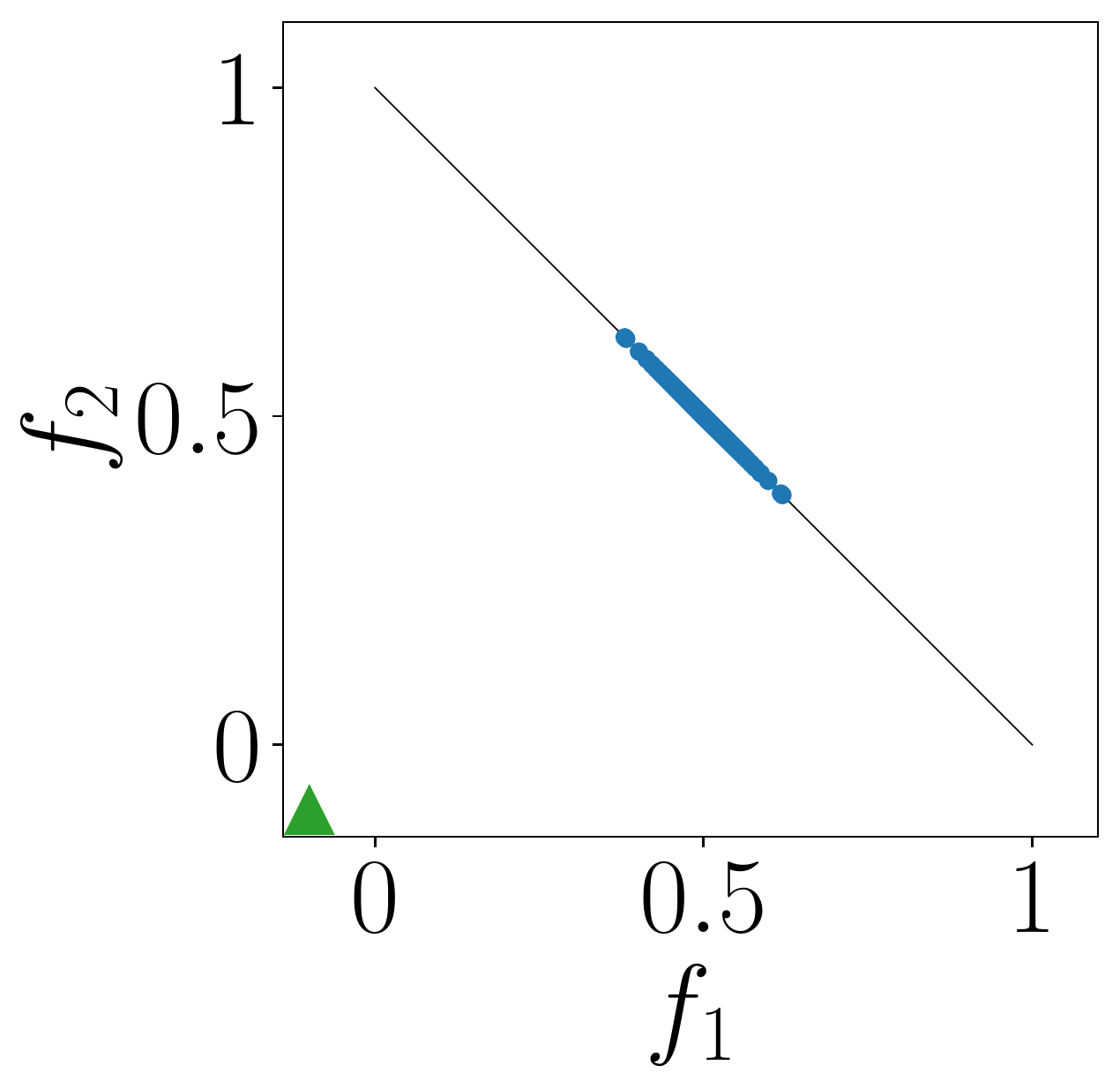}        
     }                               
     \caption{Distributions of points found by the six EMO algorithms on the DTLZ1 problem when using $\mathbf{z}^{-0.1} =(-0.1, -0.1)^{\top}$.}
   \label{supfig:emo_points_dtlz1_z-0.1}
\end{figure*}

\begin{figure*}[t]
   \centering
   \subfloat[R-NSGA-II]{  
     \includegraphics[width=0.145\textwidth]{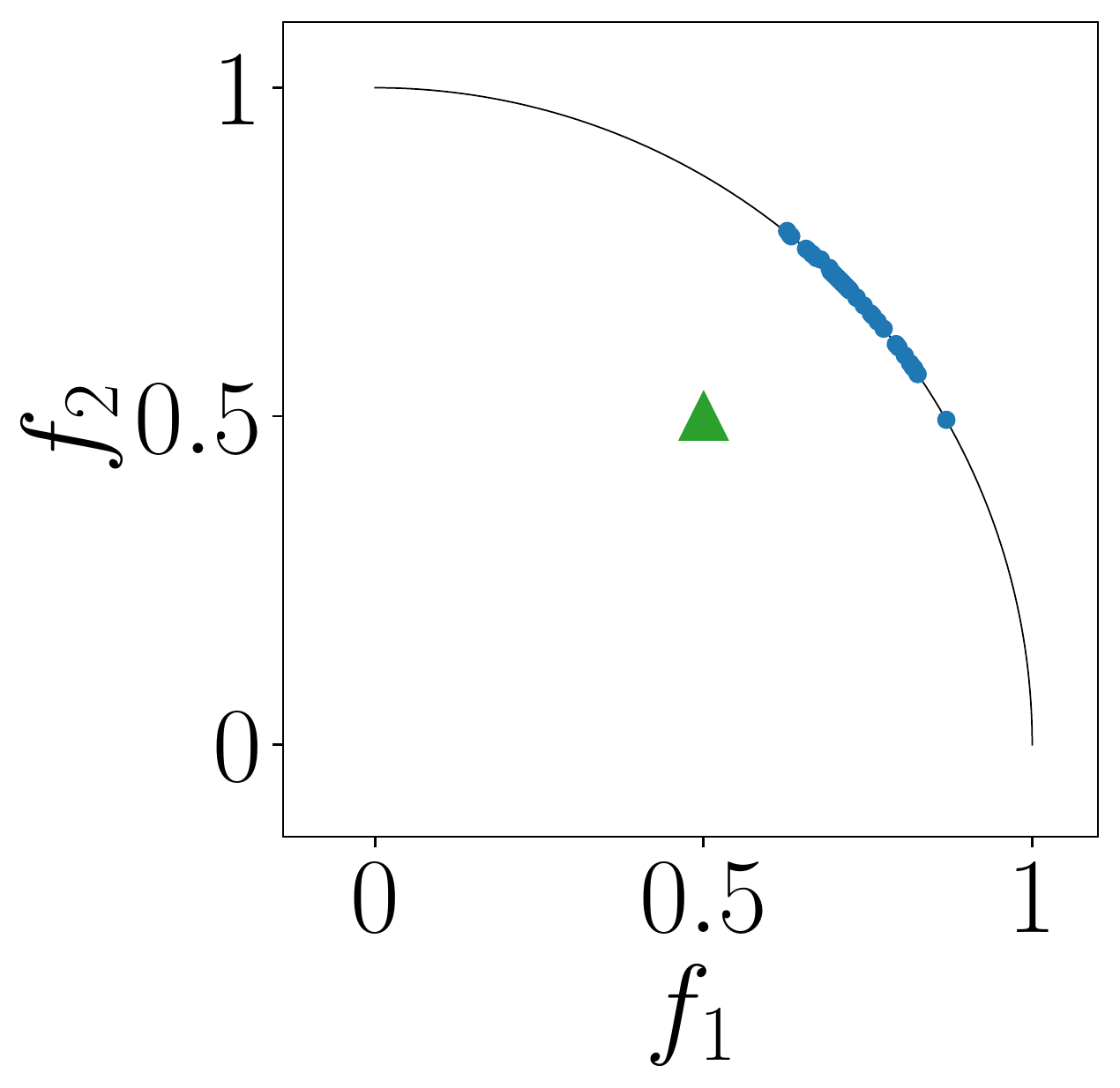}
     }
   \subfloat[r-NSGA-II]{  
     \includegraphics[width=0.145\textwidth]{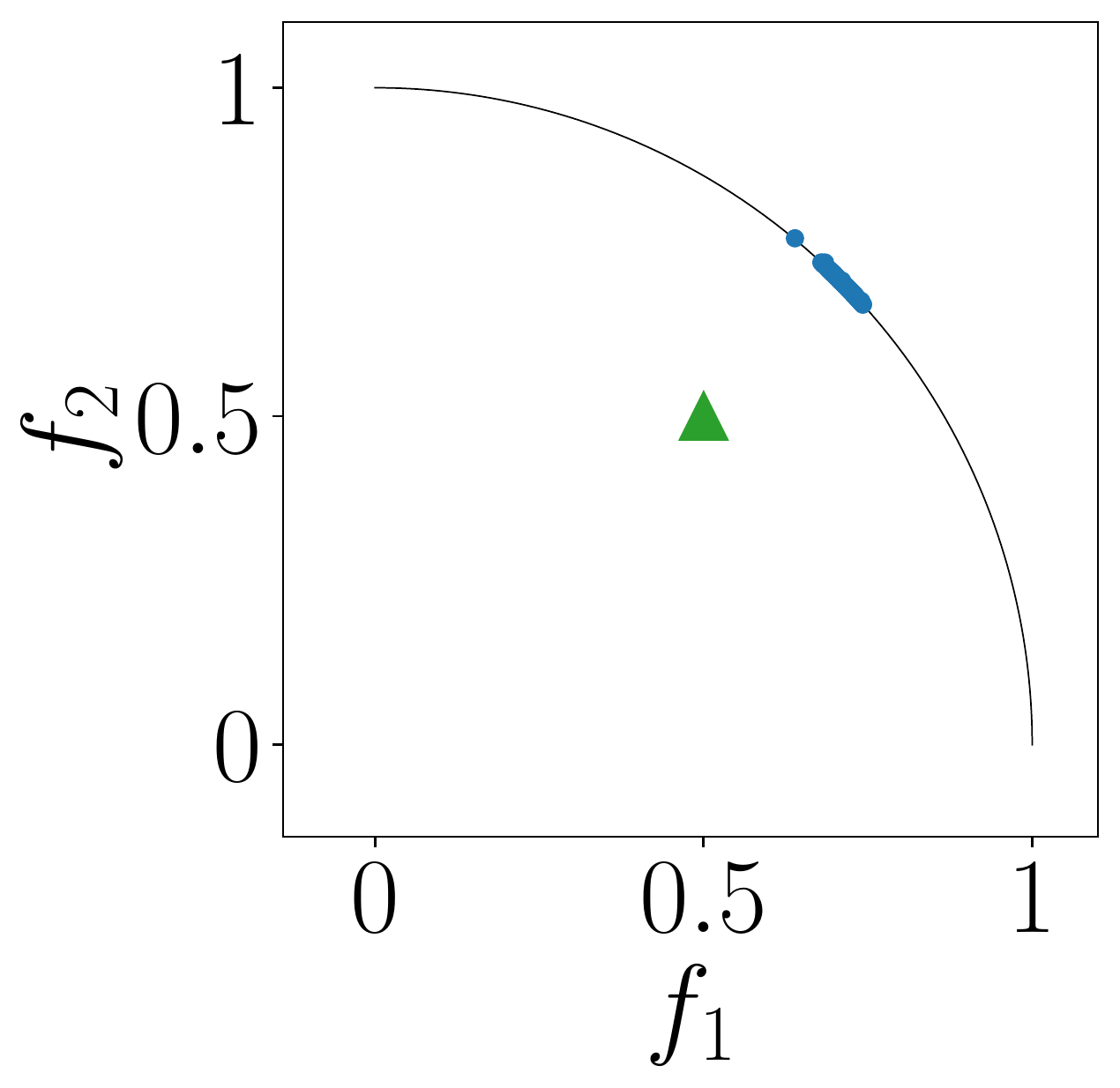}      
     }                                                                                      
   \subfloat[g-NSGA-II]{  
     \includegraphics[width=0.145\textwidth]{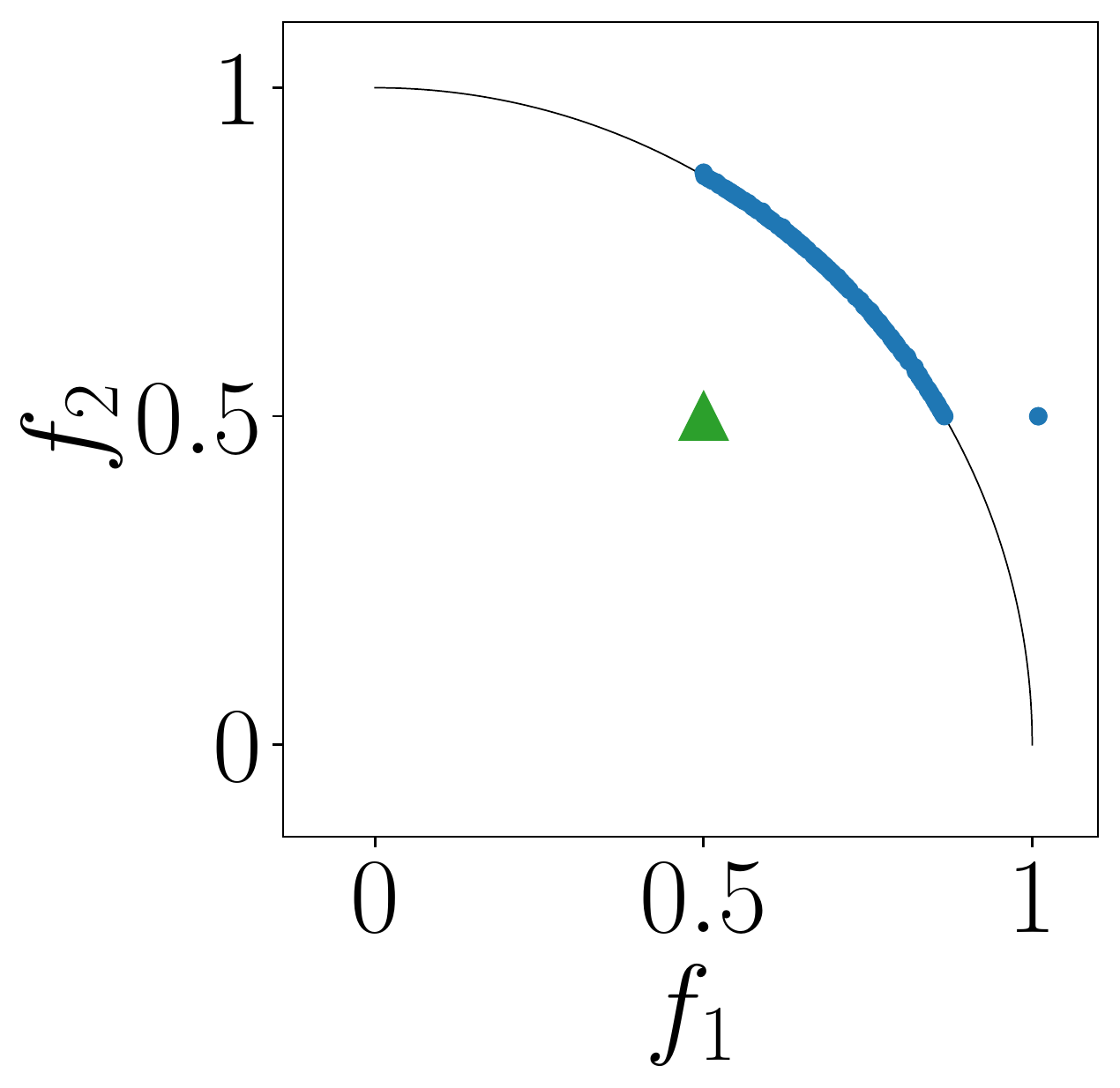}
     }
   \subfloat[PBEA]{  
     \includegraphics[width=0.145\textwidth]{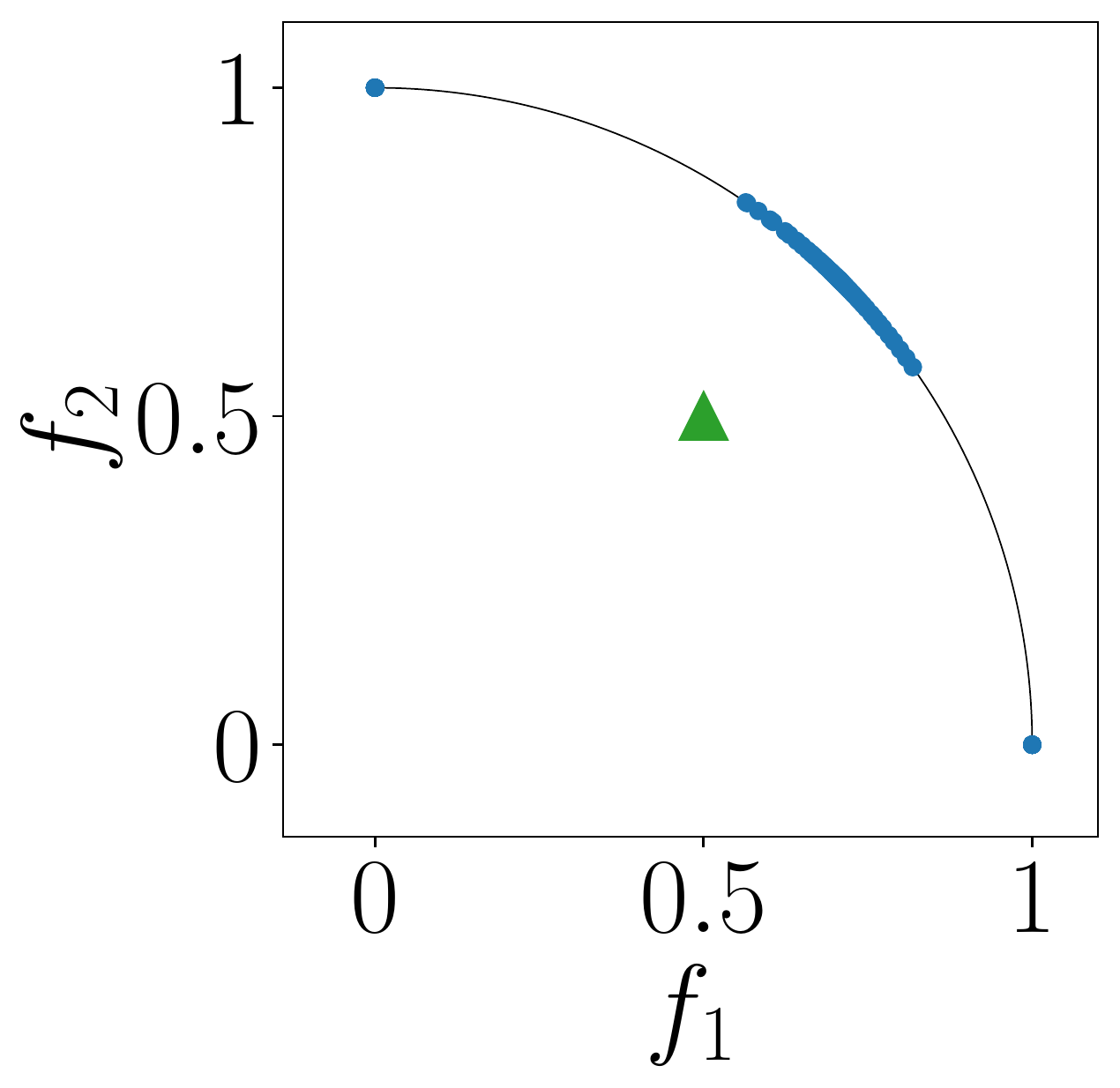}      
     }     
\subfloat[R-MEAD2]{  
     \includegraphics[width=0.145\textwidth]{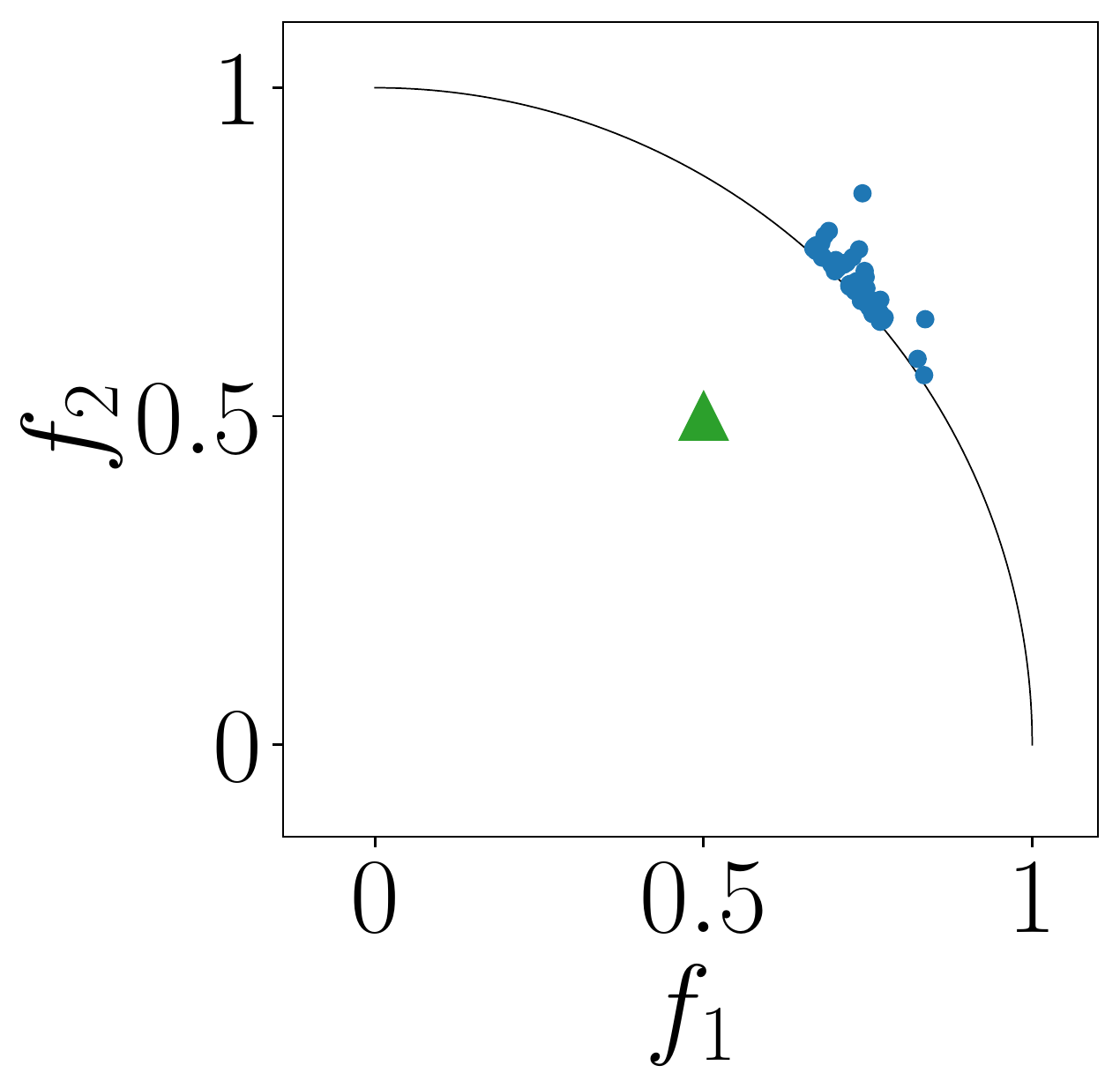}      
     }         
     \subfloat[MOEA/D-NUMS]{  
     \includegraphics[width=0.145\textwidth]{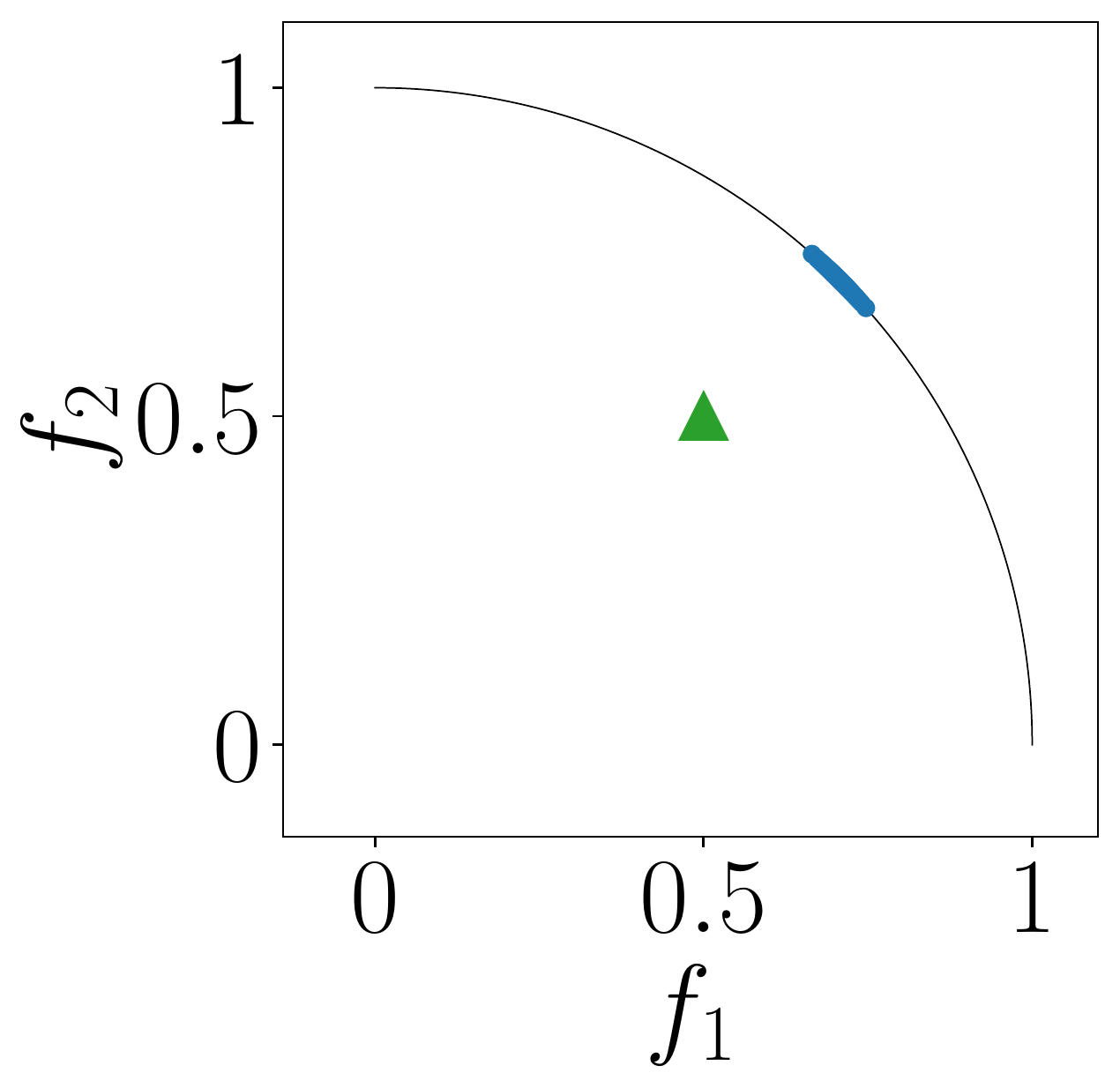}        
     }                               
     \caption{Distributions of points found by the six EMO algorithms on the DTLZ2 problem when using $\mathbf{z}^{0.5} =(0.5, 0.5)^{\top}$.}
   \label{supfig:emo_points_dtlz2_z0.5}
   \subfloat[R-NSGA-II]{  
     \includegraphics[width=0.145\textwidth]{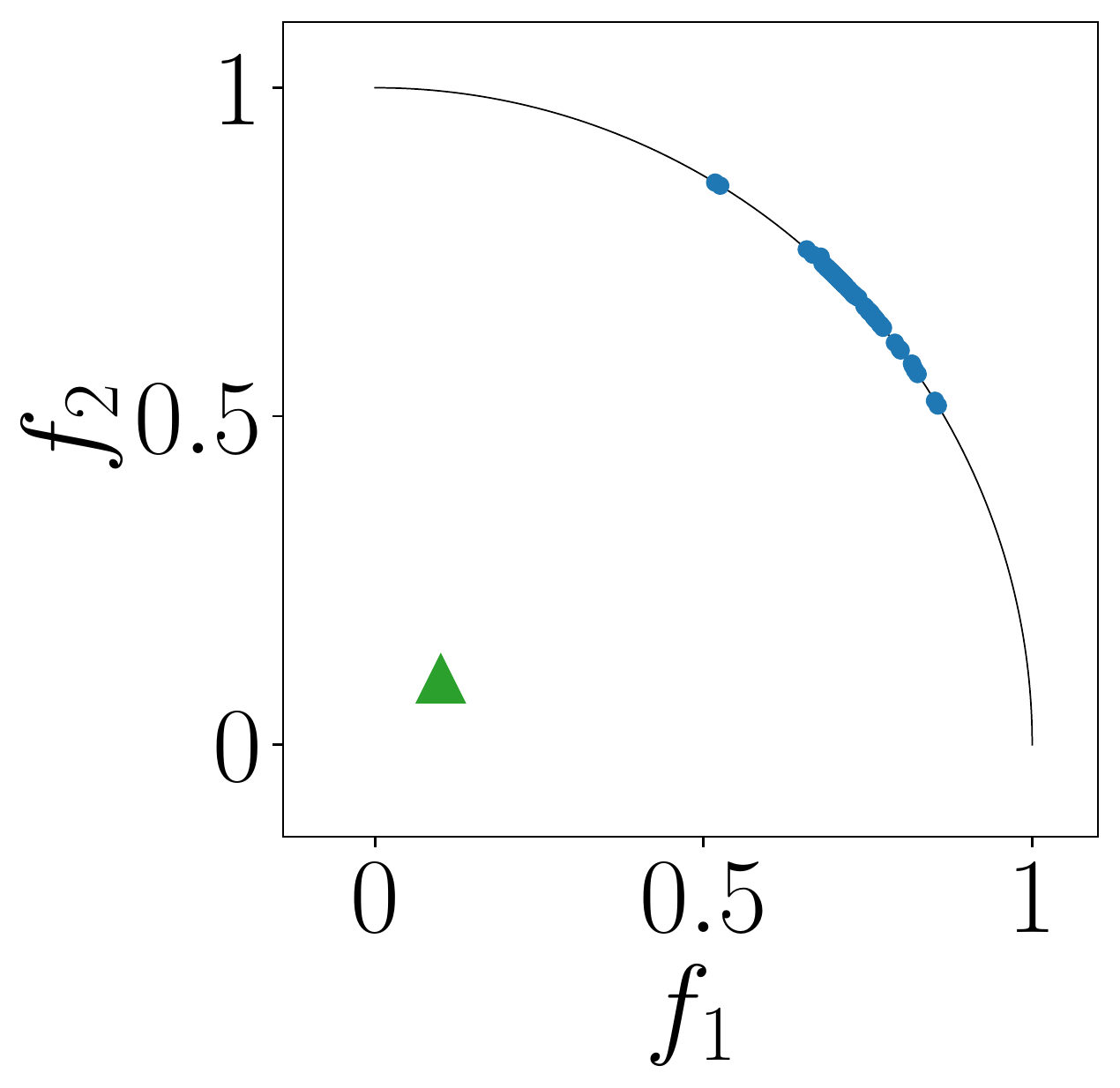}
     }
   \subfloat[r-NSGA-II]{  
     \includegraphics[width=0.145\textwidth]{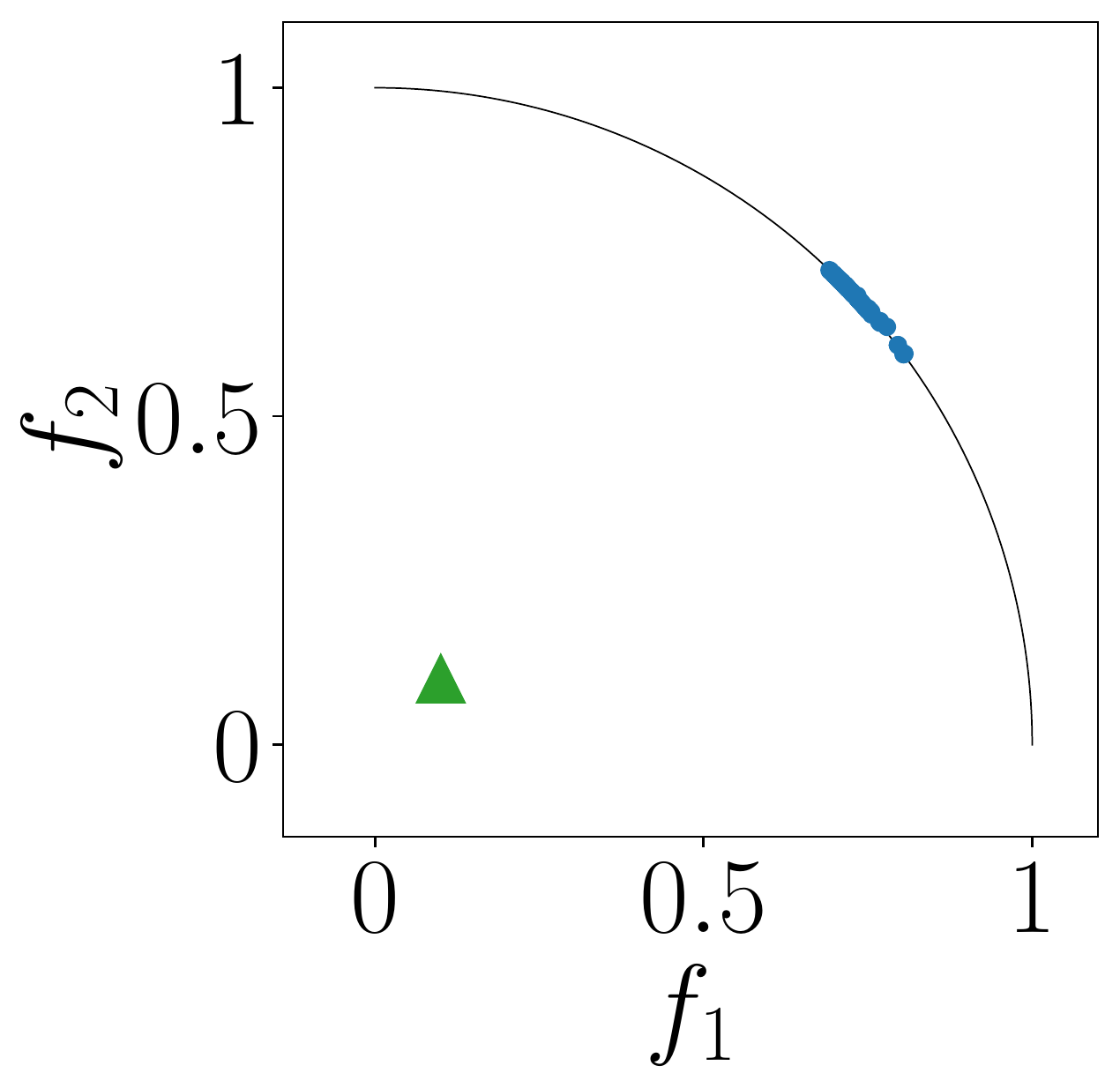}      
     }                                                                                      
   \subfloat[g-NSGA-II]{  
     \includegraphics[width=0.145\textwidth]{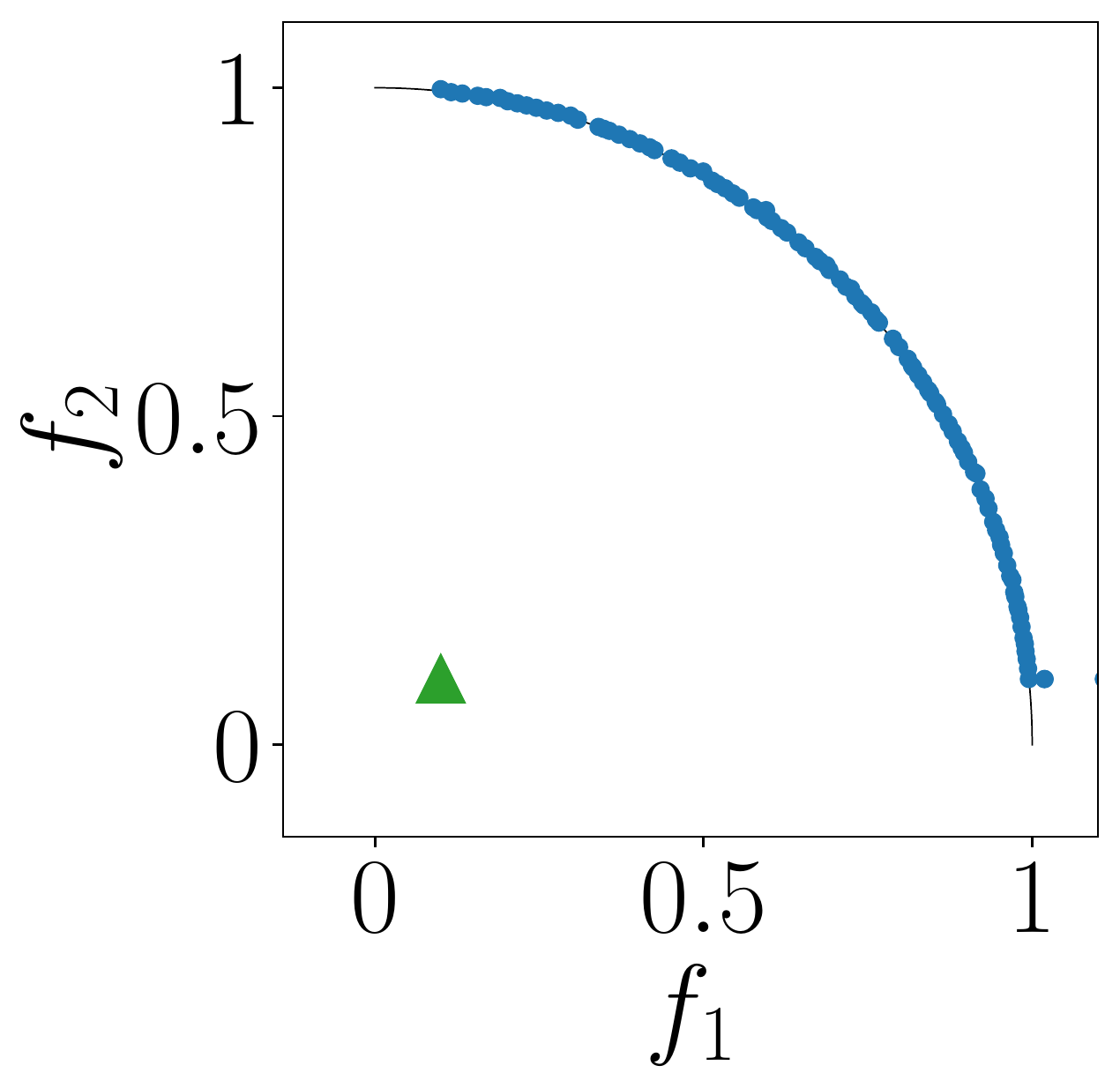}
     }
   \subfloat[PBEA]{  
     \includegraphics[width=0.145\textwidth]{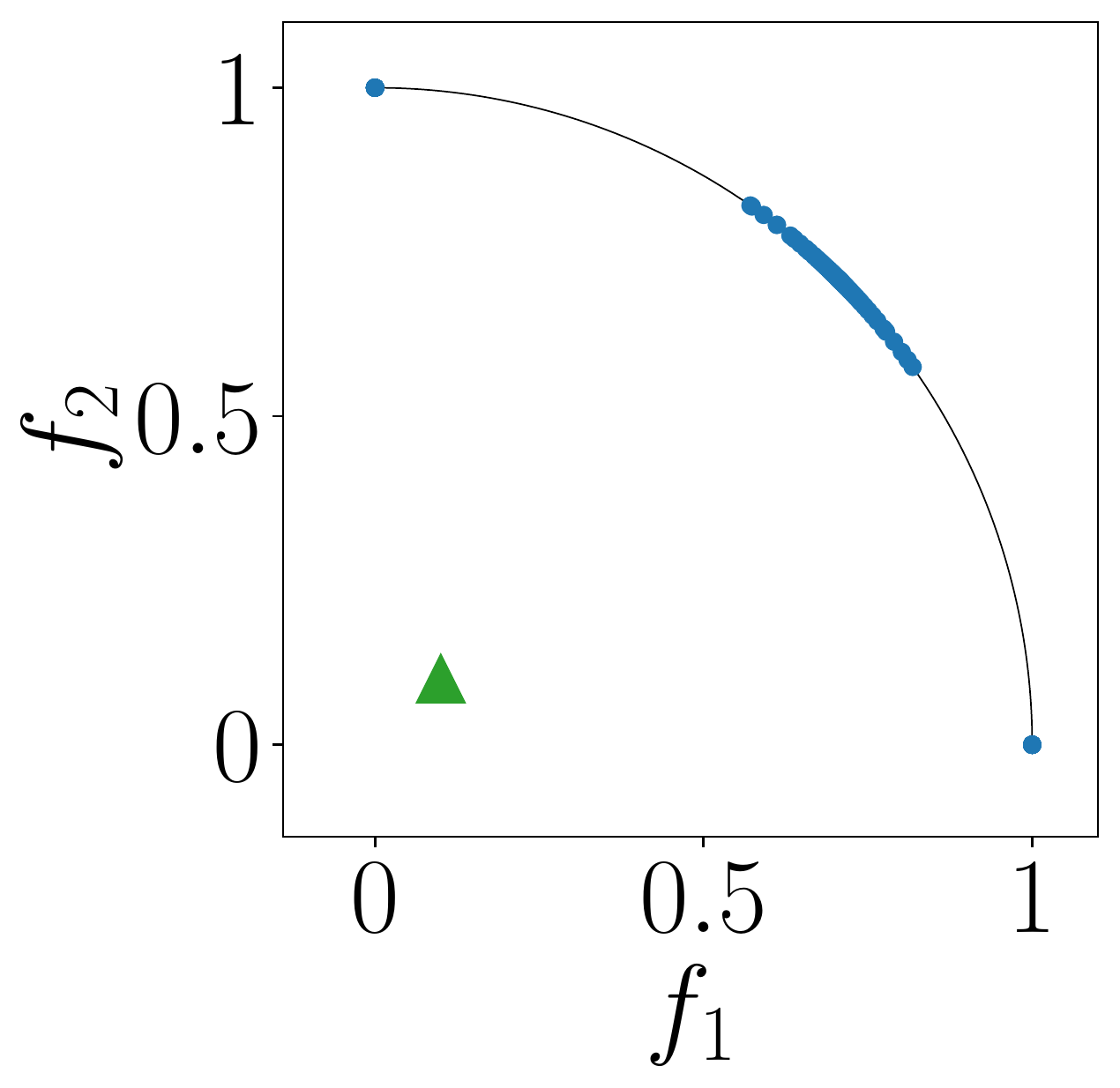}      
     }     
\subfloat[R-MEAD2]{  
     \includegraphics[width=0.145\textwidth]{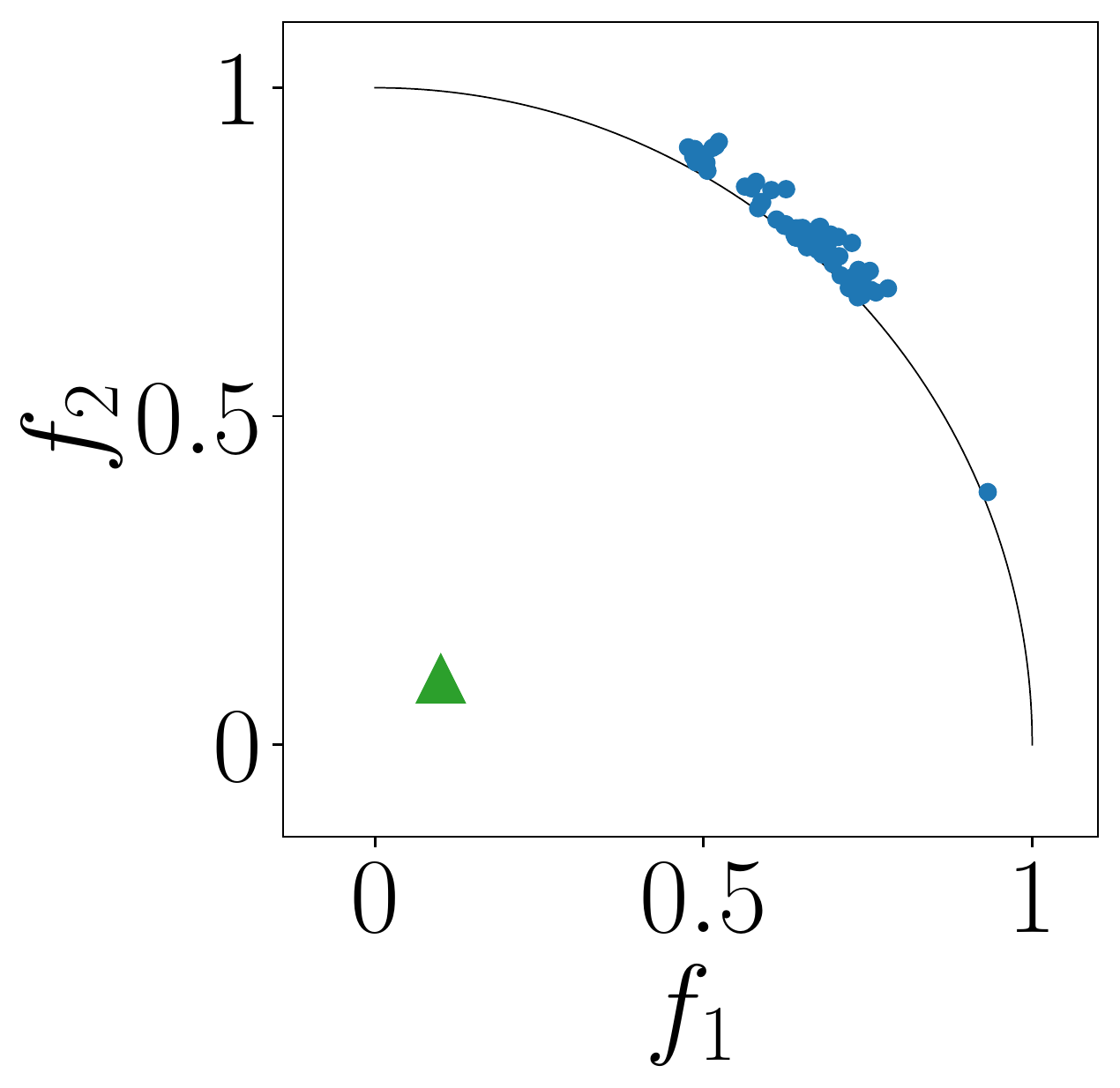}      
     }         
     \subfloat[MOEA/D-NUMS]{  
     \includegraphics[width=0.145\textwidth]{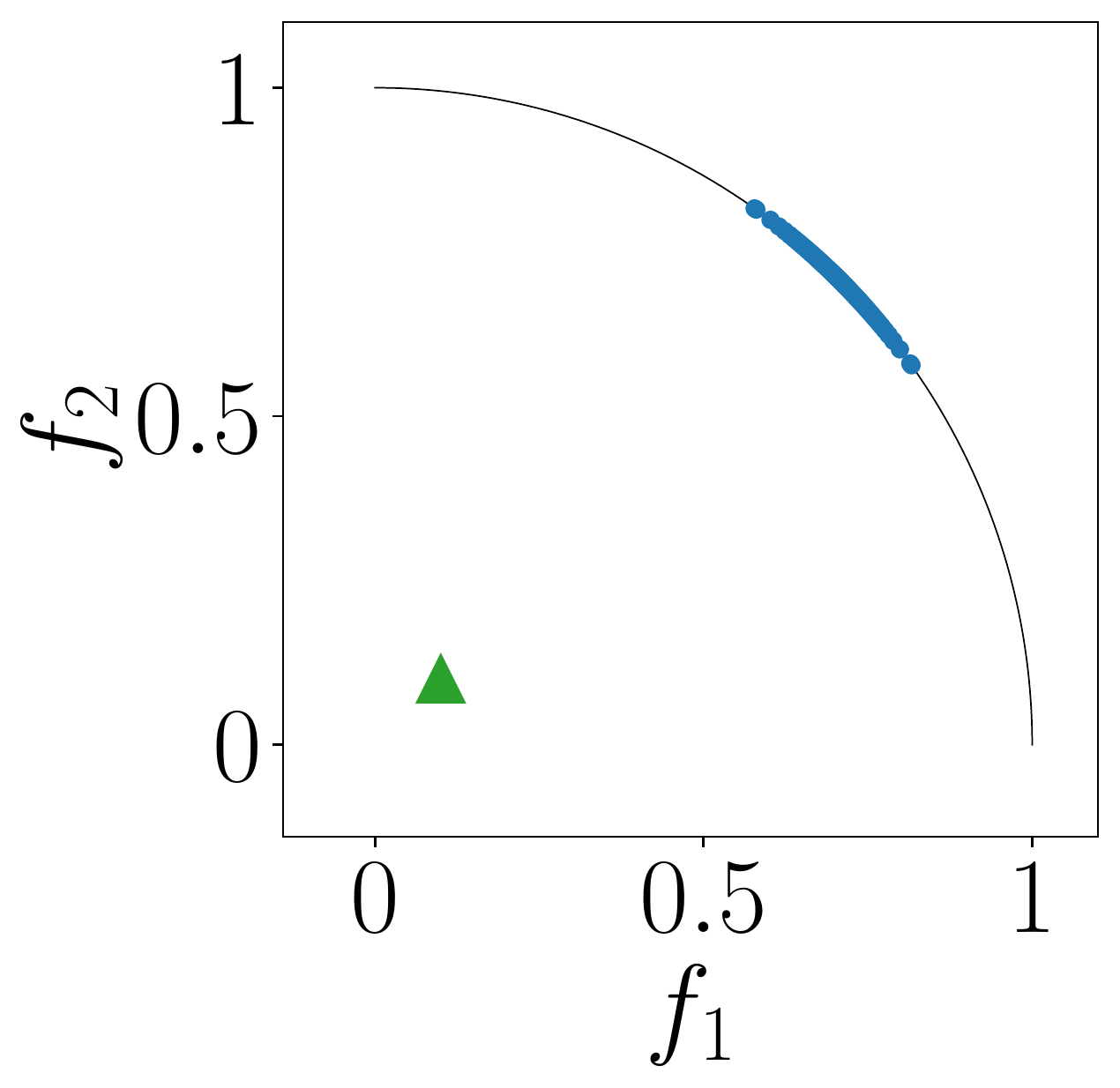}        
     }                               
     \caption{Distributions of points found by the six EMO algorithms on the DTLZ2 problem when using $\mathbf{z}^{0.1} =(0.1, 0.1)^{\top}$.}
   \label{supfig:emo_points_dtlz2_z0.1}
   \subfloat[R-NSGA-II]{  
     \includegraphics[width=0.145\textwidth]{figs/emo_points/RNSGA2_mu100_DTLZ2_m2_z-0.1_-0.1.pdf}
     }
   \subfloat[r-NSGA-II]{  
     \includegraphics[width=0.145\textwidth]{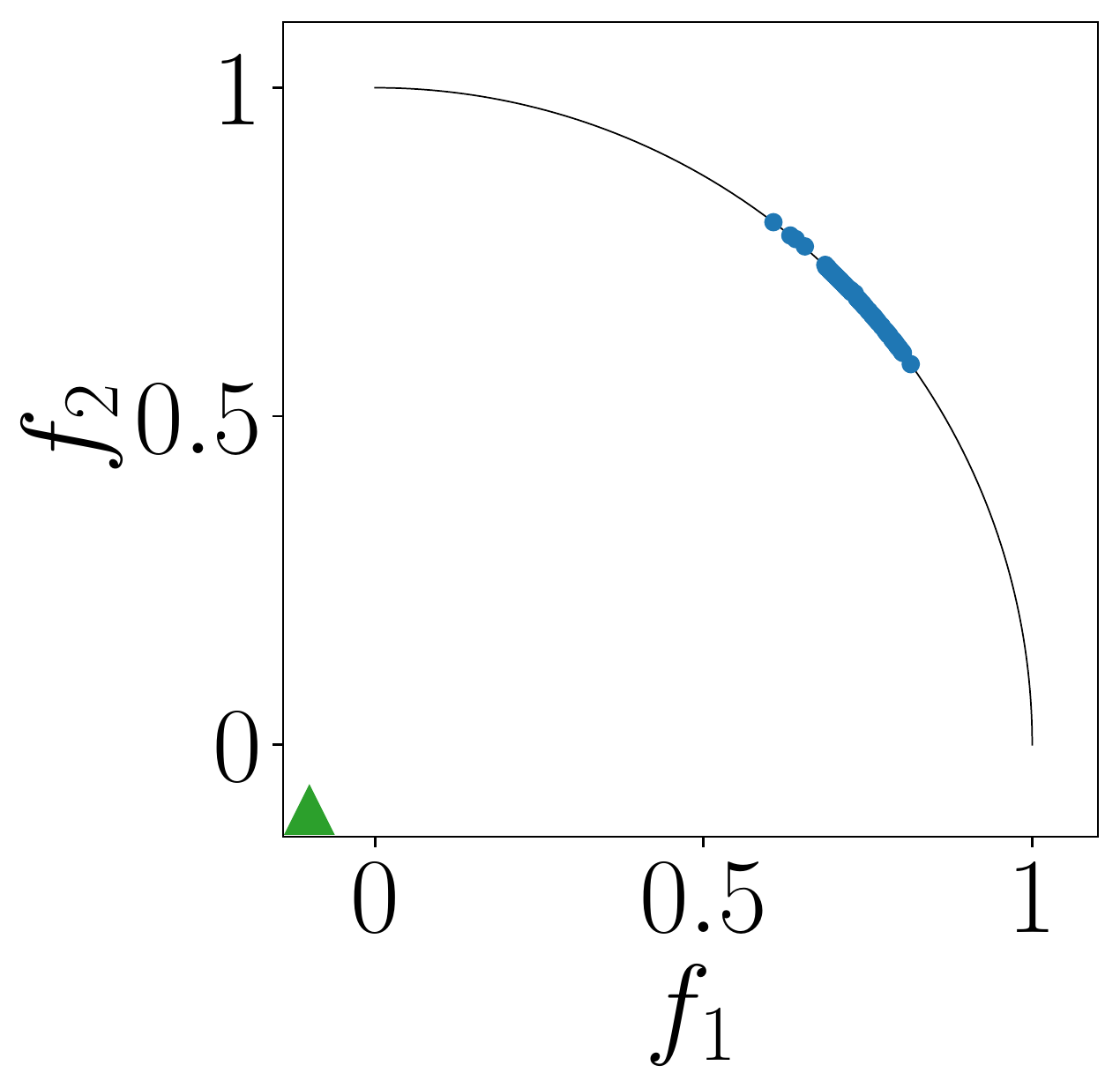}      
     }                                                                                      
   \subfloat[g-NSGA-II]{  
     \includegraphics[width=0.145\textwidth]{figs/emo_points/gNSGA2_mu100_DTLZ2_m2_z-0.1_-0.1.pdf}
     }
   \subfloat[PBEA]{  
     \includegraphics[width=0.145\textwidth]{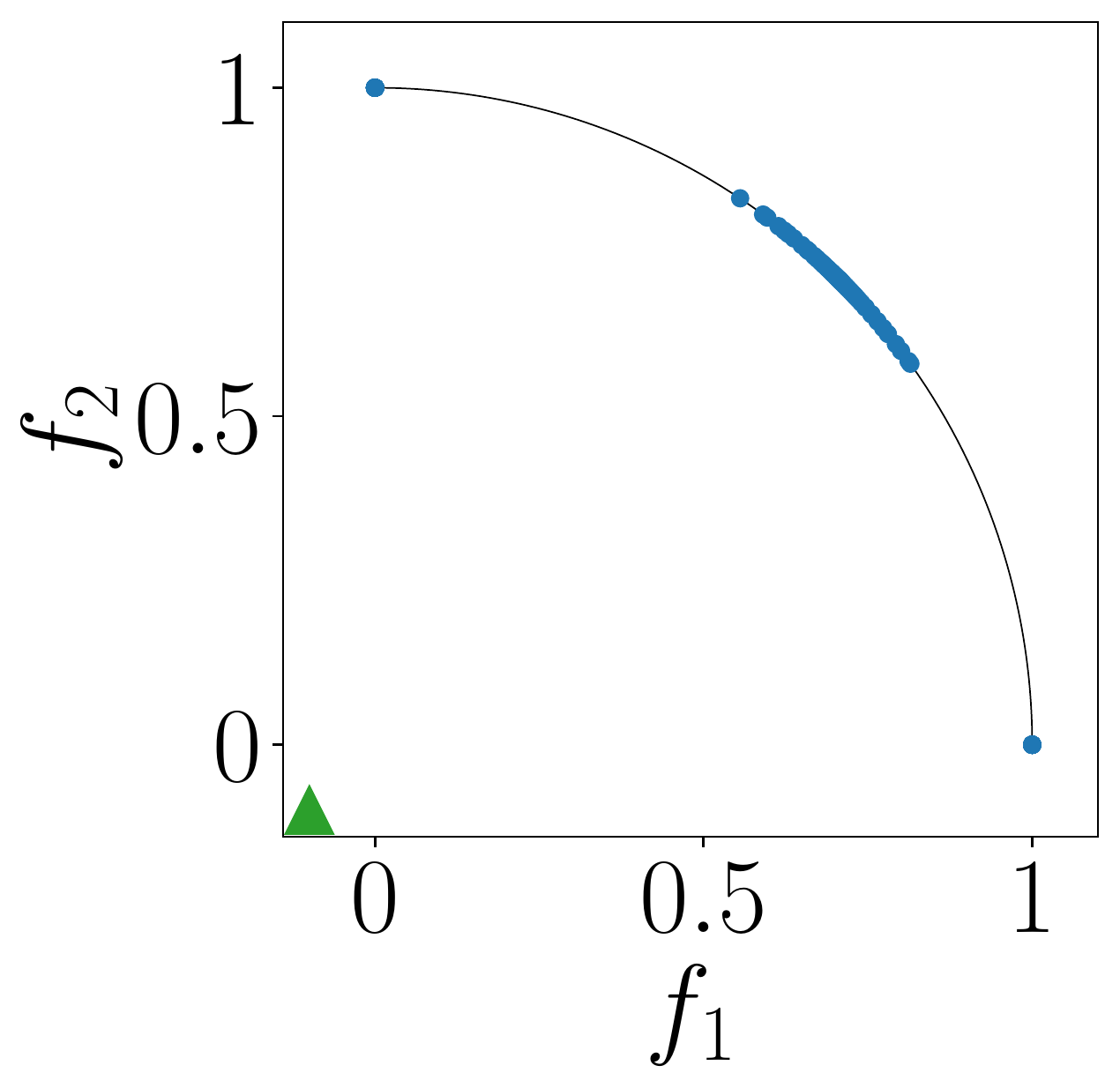}      
     }     
\subfloat[R-MEAD2]{  
     \includegraphics[width=0.145\textwidth]{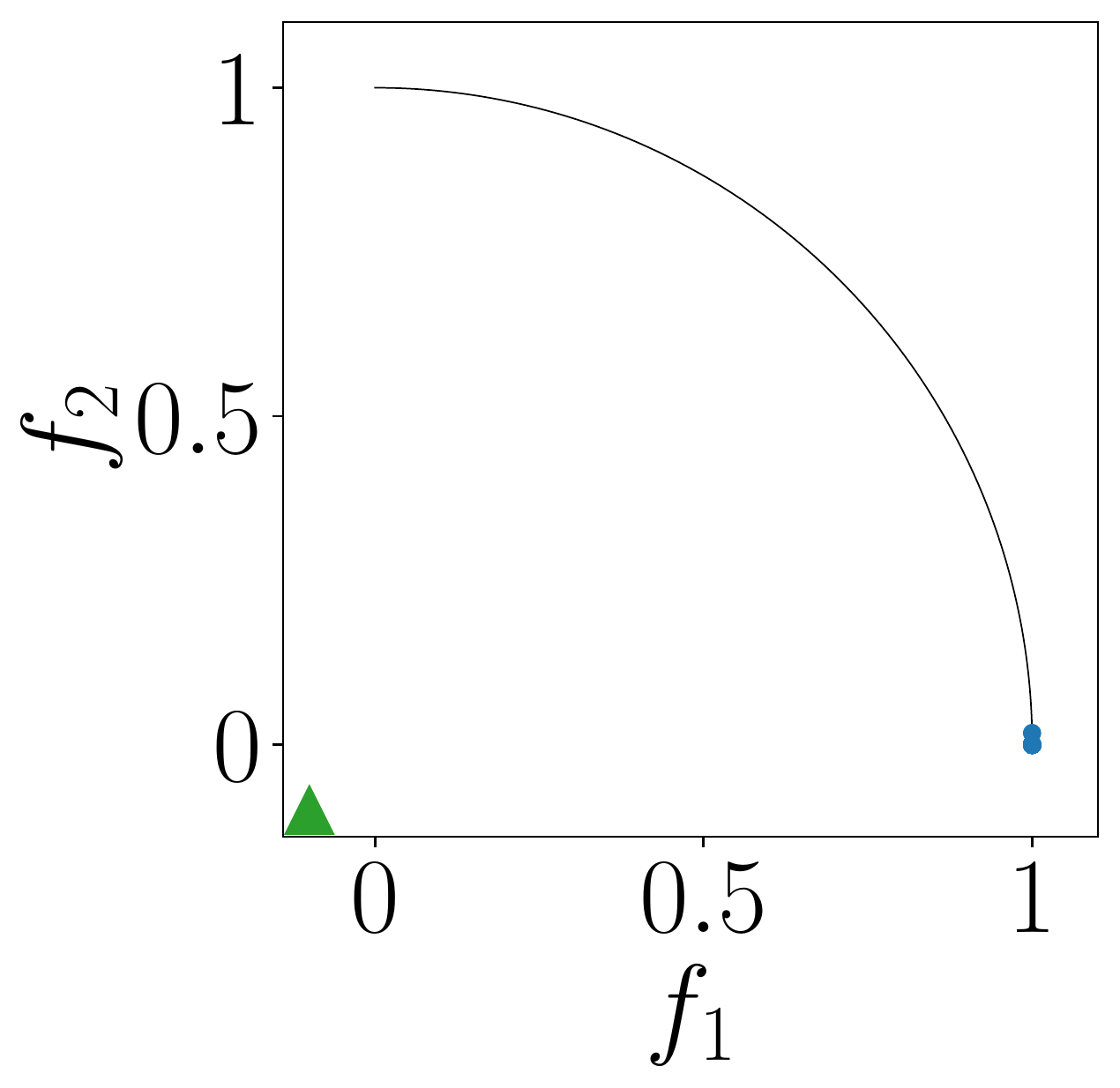}      
     }         
     \subfloat[MOEA/D-NUMS]{  
     \includegraphics[width=0.145\textwidth]{figs/emo_points/MOEADNUMS_mu100_DTLZ2_m2_z-0.1_-0.1.pdf}        
     }                               
     \caption{Distributions of points found by the six EMO algorithms on the DTLZ2 problem when using $\mathbf{z}^{-0.1} =(-0.1, -0.1)^{\top}$.}
   \label{supfig:emo_points_dtlz2_z-0.1}
\end{figure*}

\begin{figure*}[t]
   \centering
   \subfloat[R-NSGA-II]{  
     \includegraphics[width=0.145\textwidth]{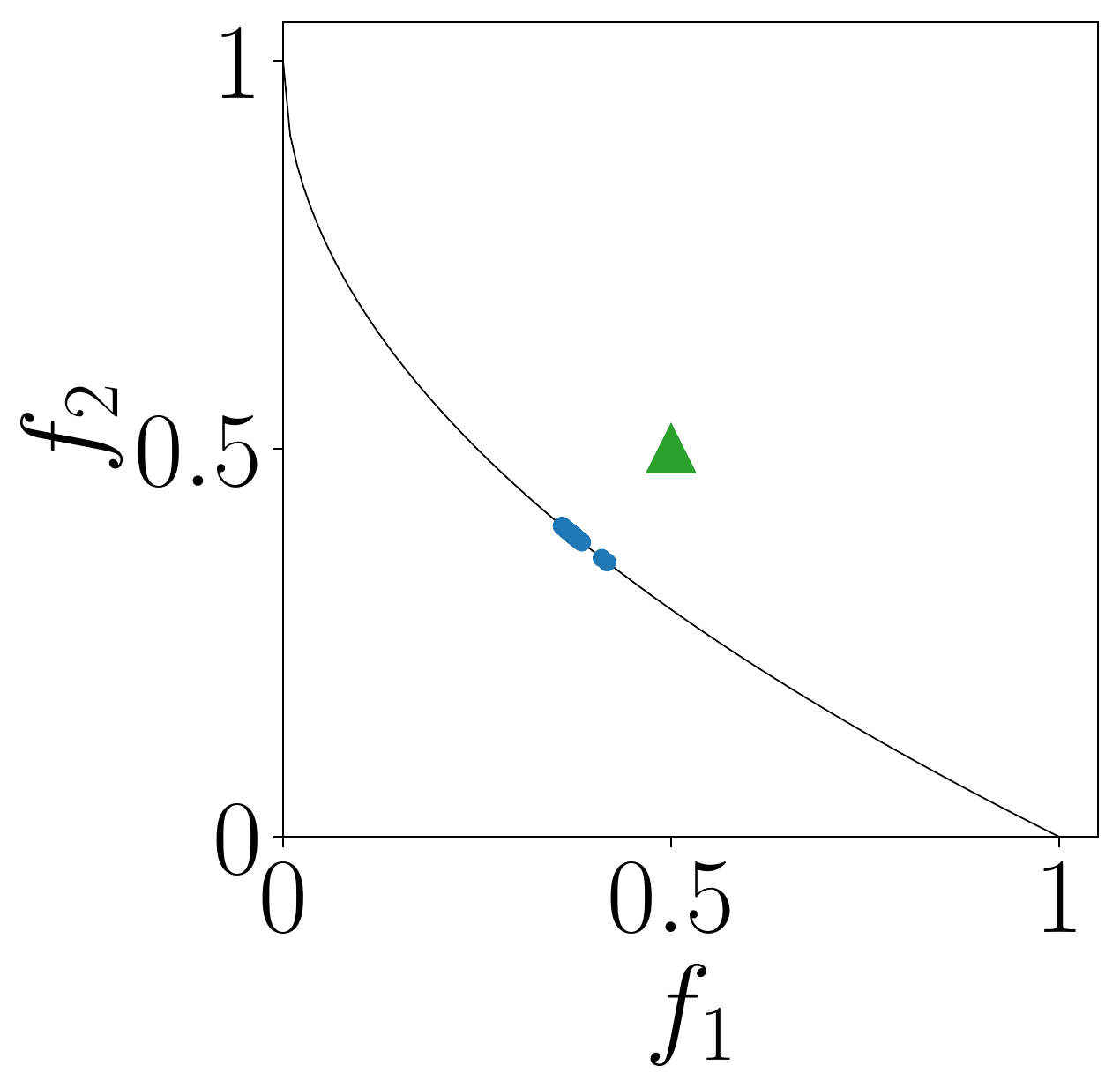}
     }
   \subfloat[r-NSGA-II]{  
     \includegraphics[width=0.145\textwidth]{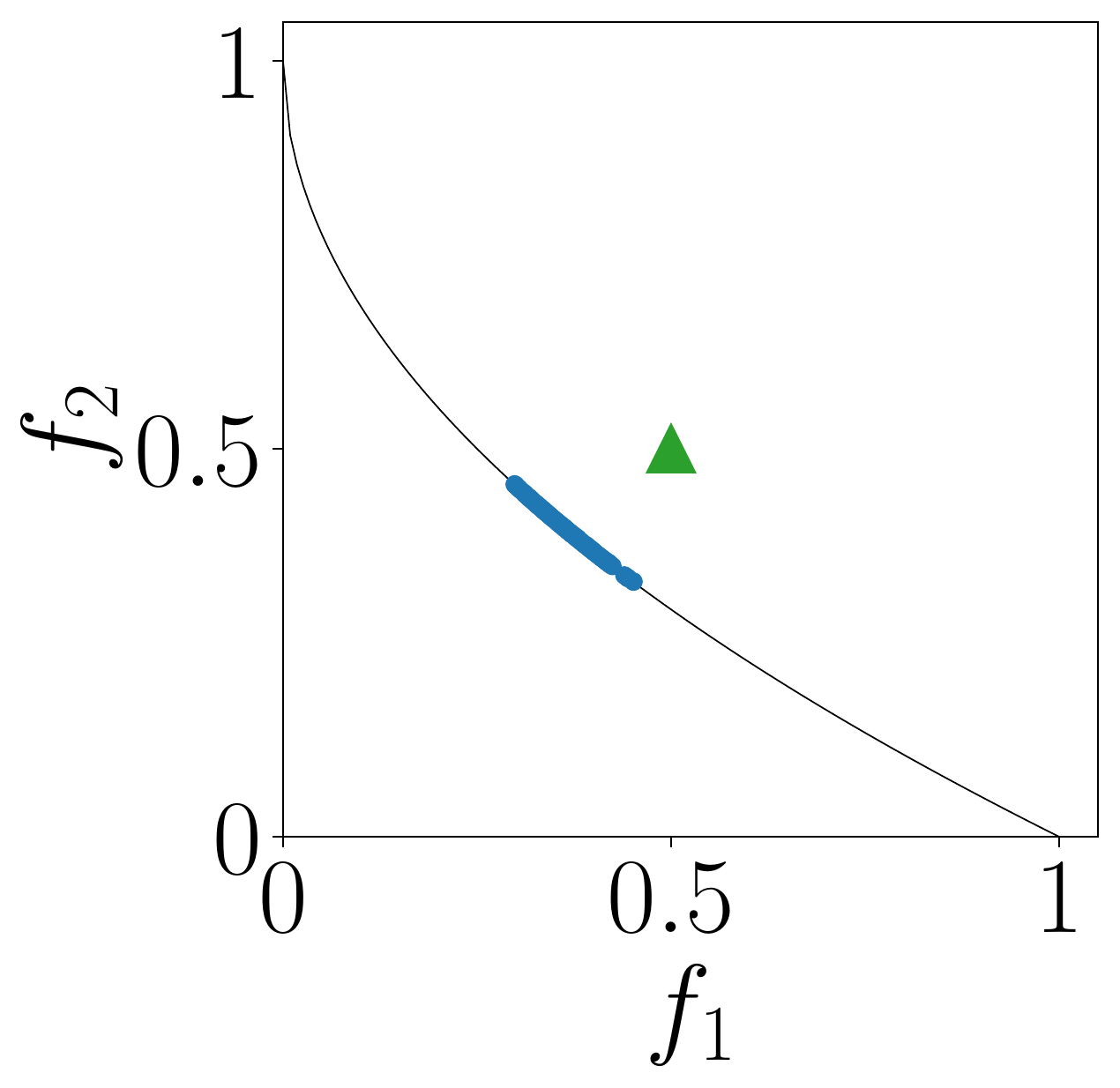}      
     }                                                                                      
   \subfloat[g-NSGA-II]{  
     \includegraphics[width=0.145\textwidth]{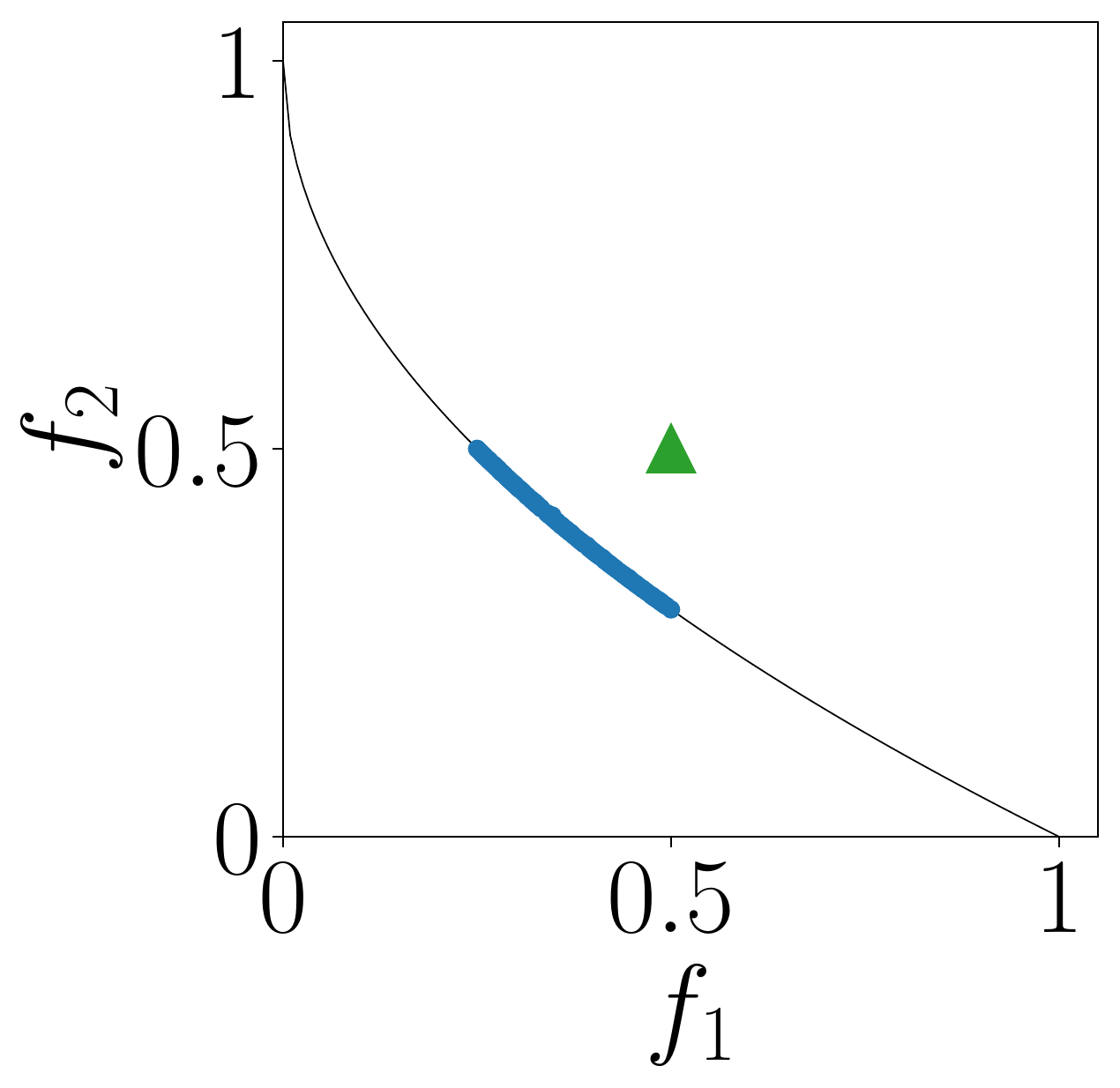}
     }
   \subfloat[PBEA]{  
     \includegraphics[width=0.145\textwidth]{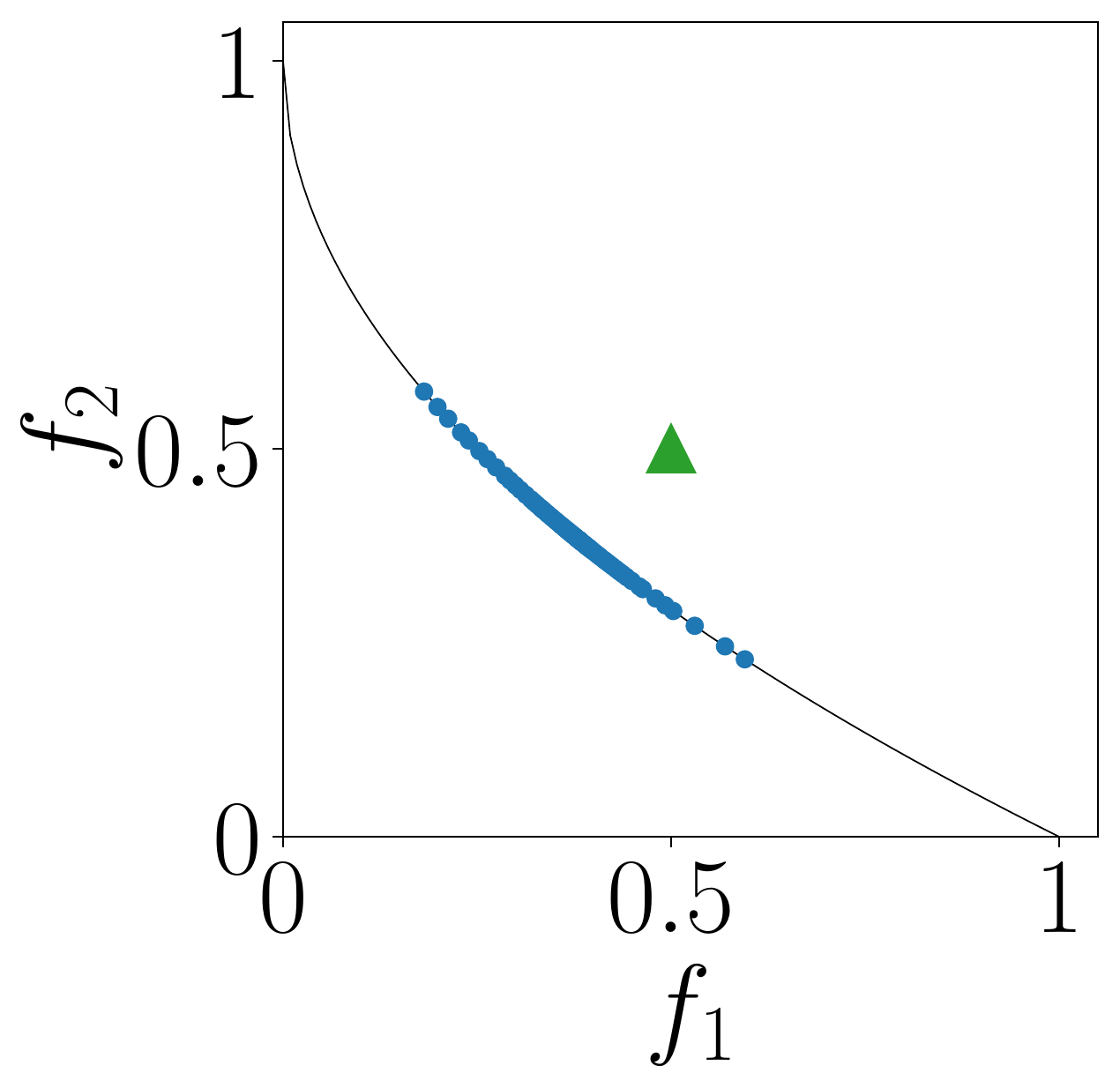}      
     }     
\subfloat[R-MEAD2]{  
     \includegraphics[width=0.145\textwidth]{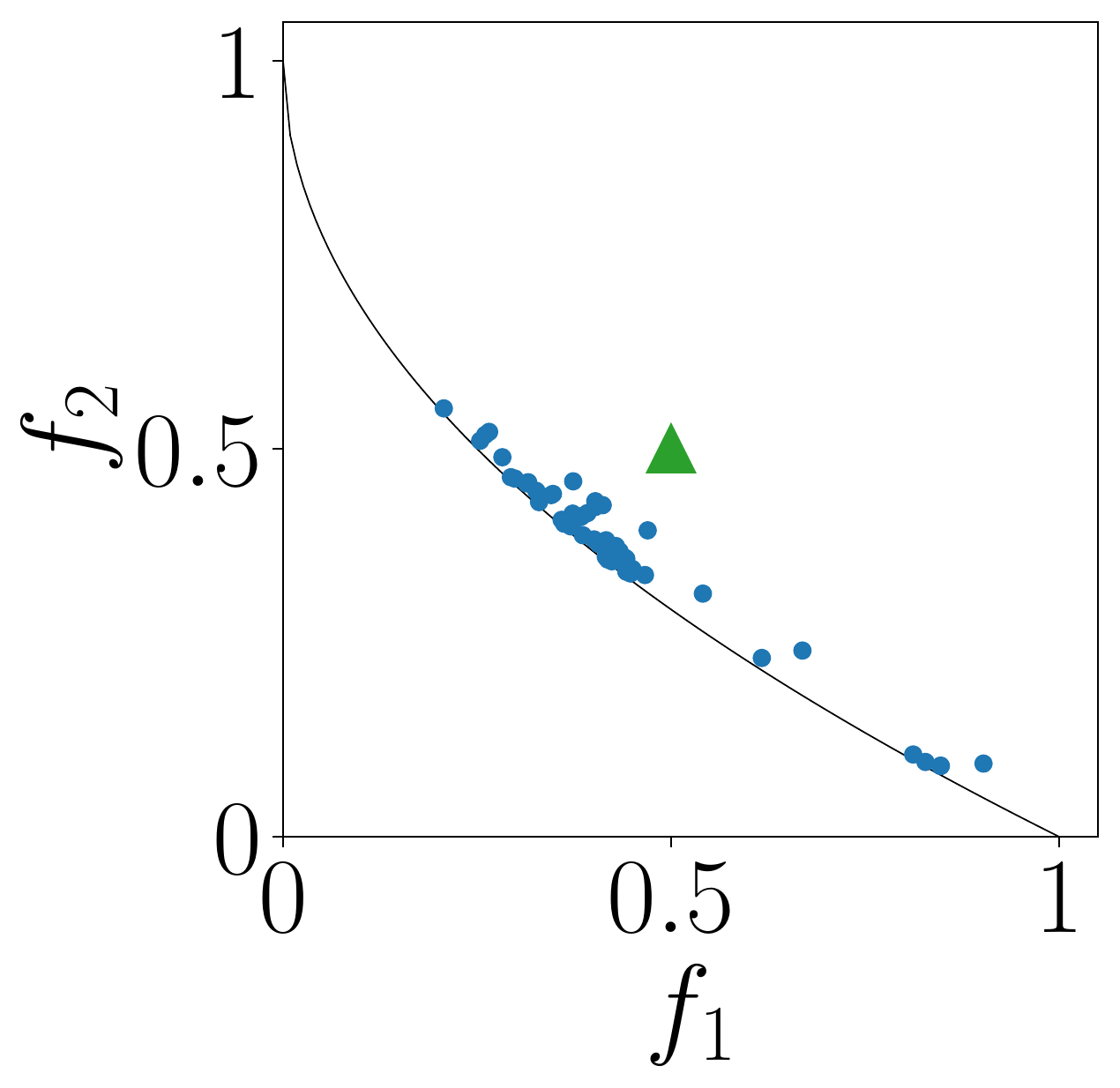}      
     }         
     \subfloat[MOEA/D-NUMS]{  
     \includegraphics[width=0.145\textwidth]{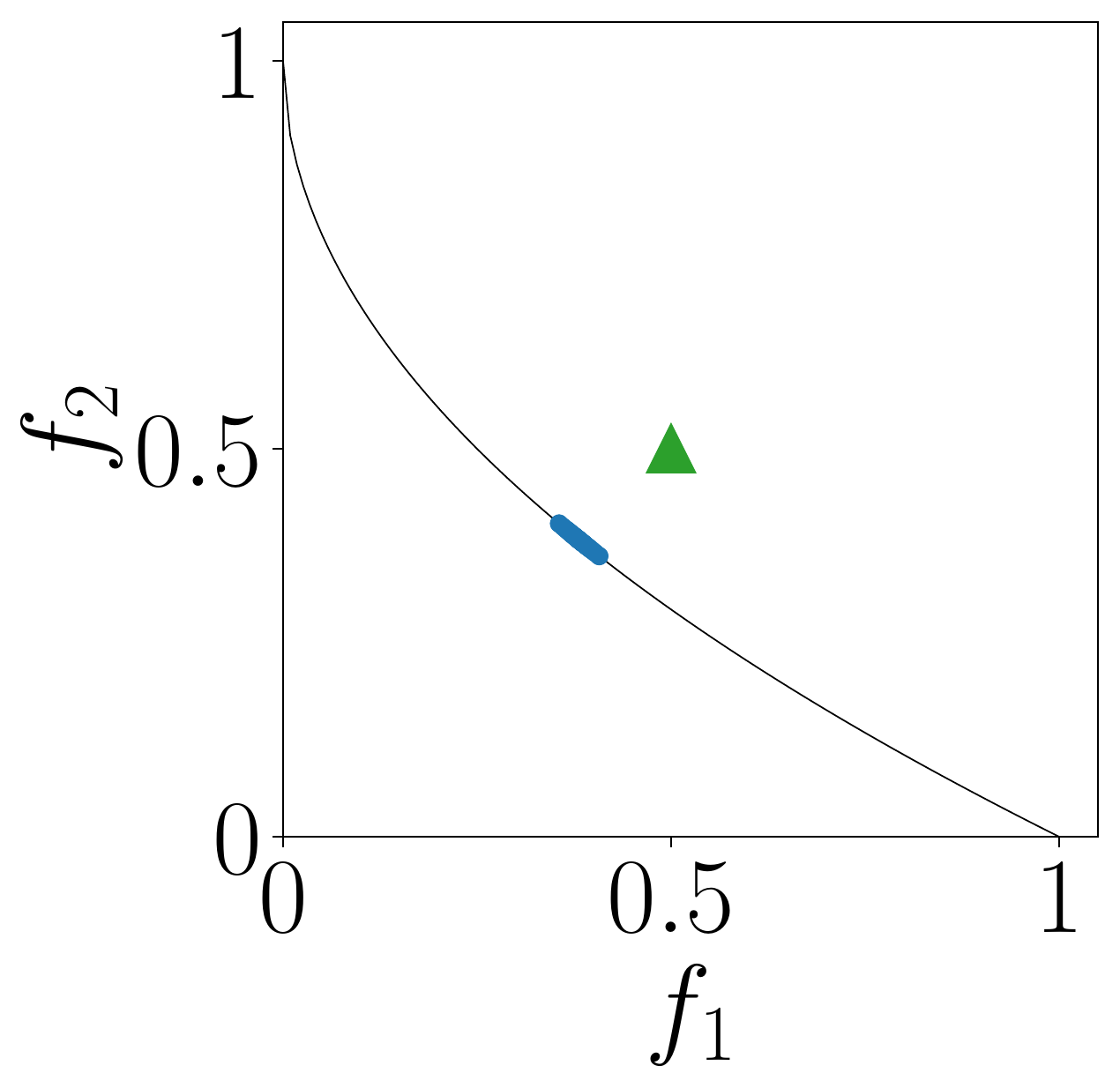}        
     }                               
     \caption{Distributions of points found by the six EMO algorithms on the convDTLZ2 problem when using $\mathbf{z}^{0.5} =(0.5, 0.5)^{\top}$.}
   \label{supfig:emo_points_convdtlz2_z0.5}
   \subfloat[R-NSGA-II]{  
     \includegraphics[width=0.145\textwidth]{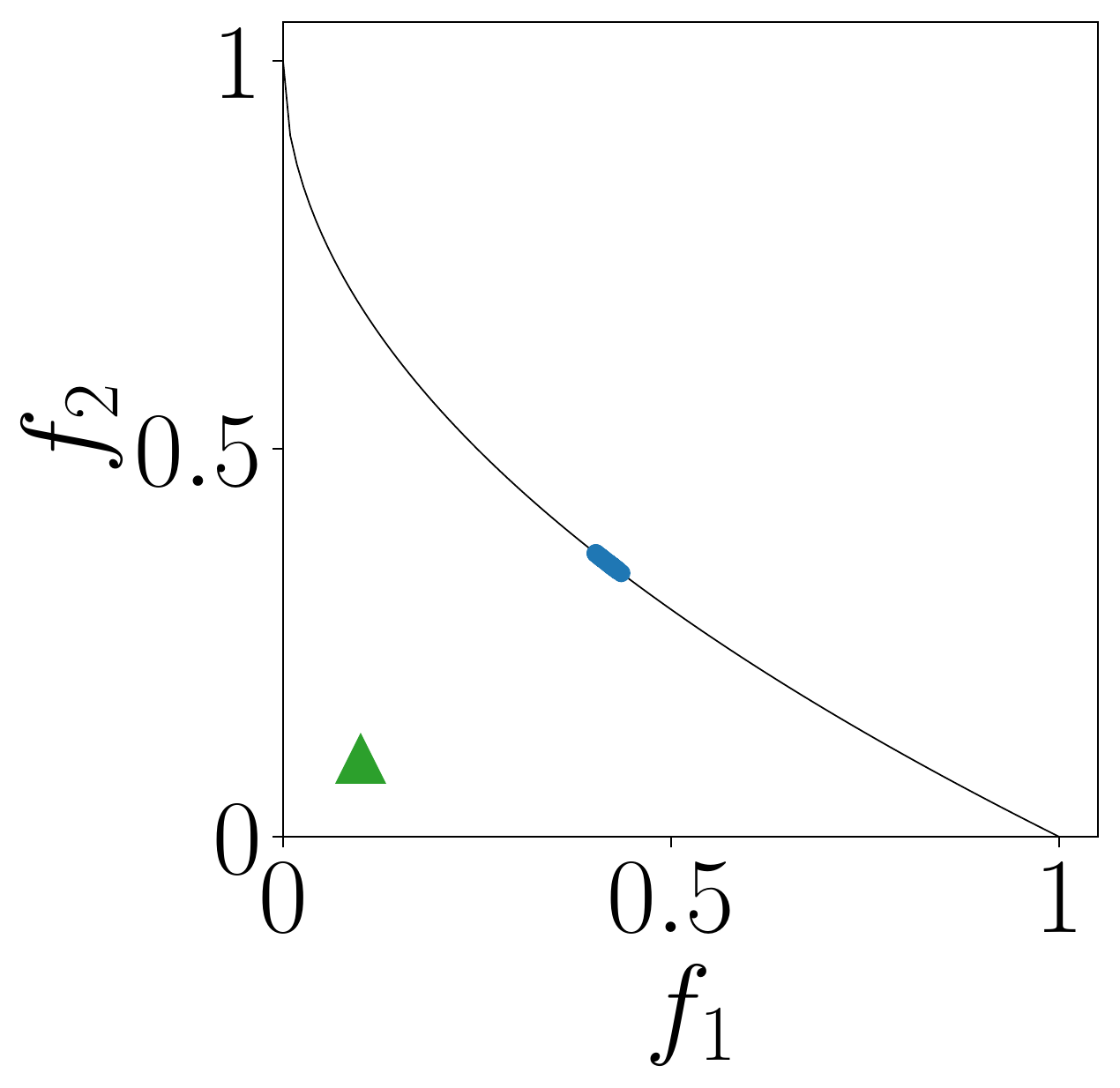}
     }
   \subfloat[r-NSGA-II]{  
     \includegraphics[width=0.145\textwidth]{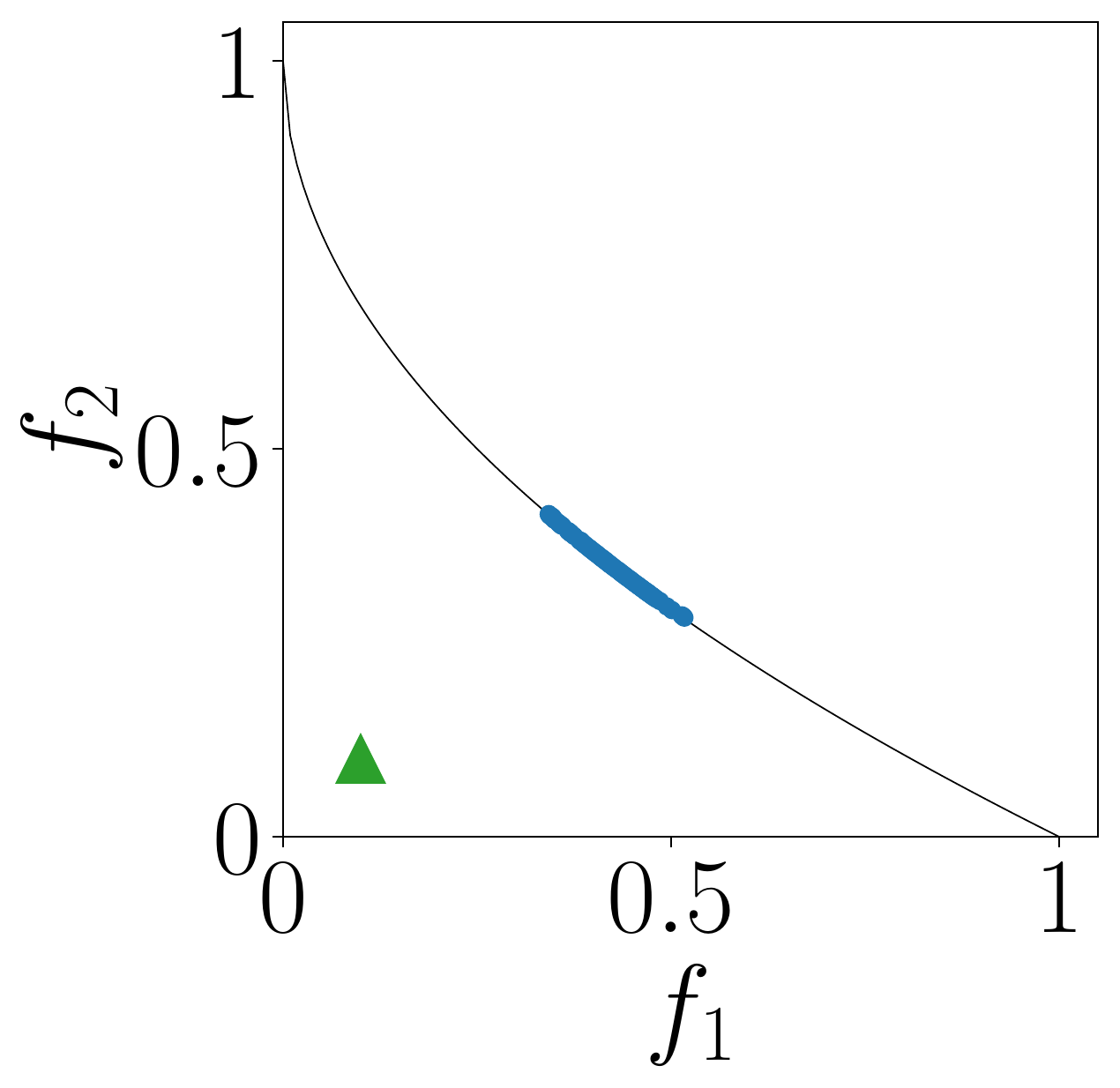}      
     }                                                                                      
   \subfloat[g-NSGA-II]{  
     \includegraphics[width=0.145\textwidth]{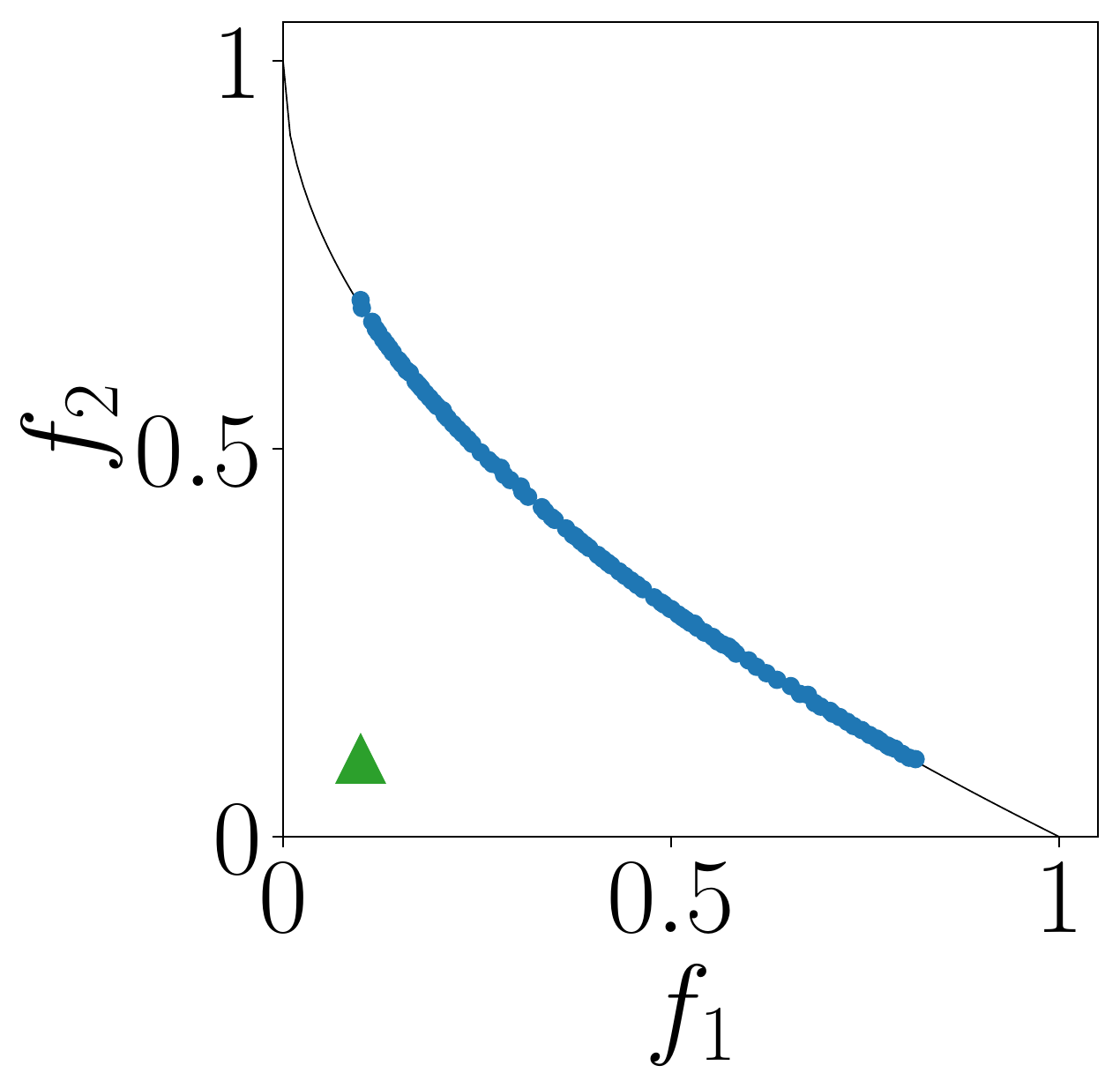}
     }
   \subfloat[PBEA]{  
     \includegraphics[width=0.145\textwidth]{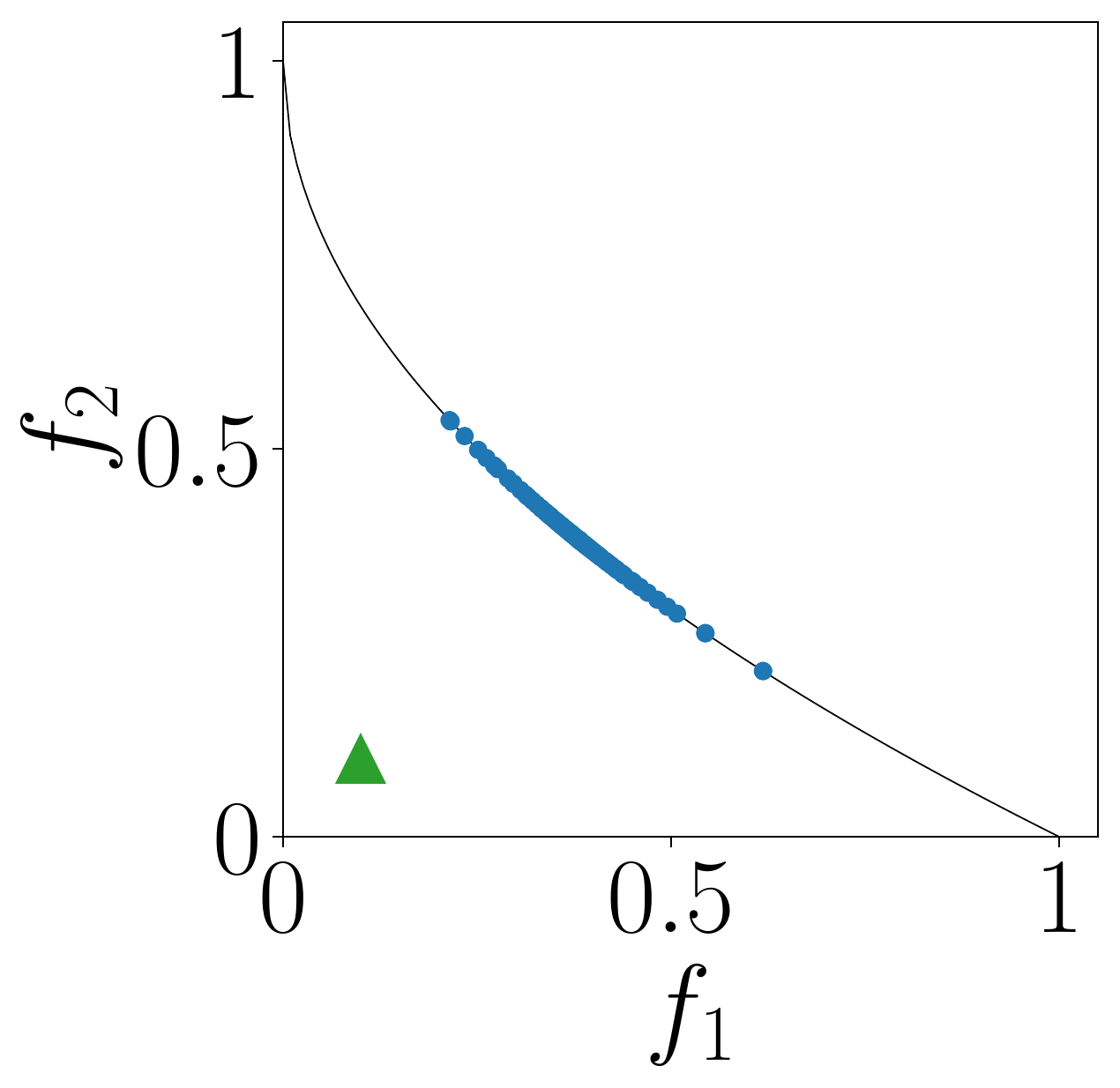}      
     }     
\subfloat[R-MEAD2]{  
     \includegraphics[width=0.145\textwidth]{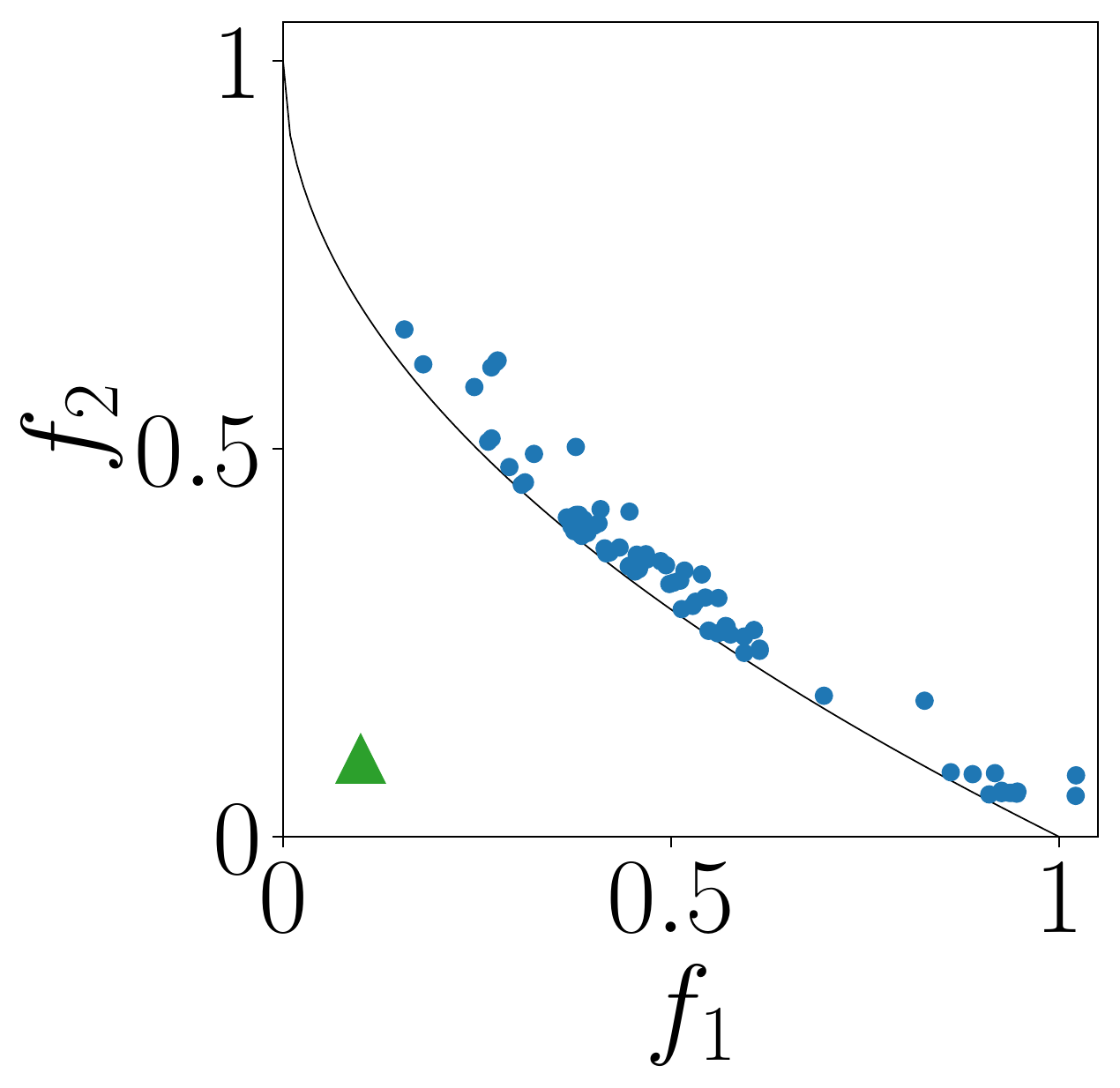}      
     }         
     \subfloat[MOEA/D-NUMS]{  
     \includegraphics[width=0.145\textwidth]{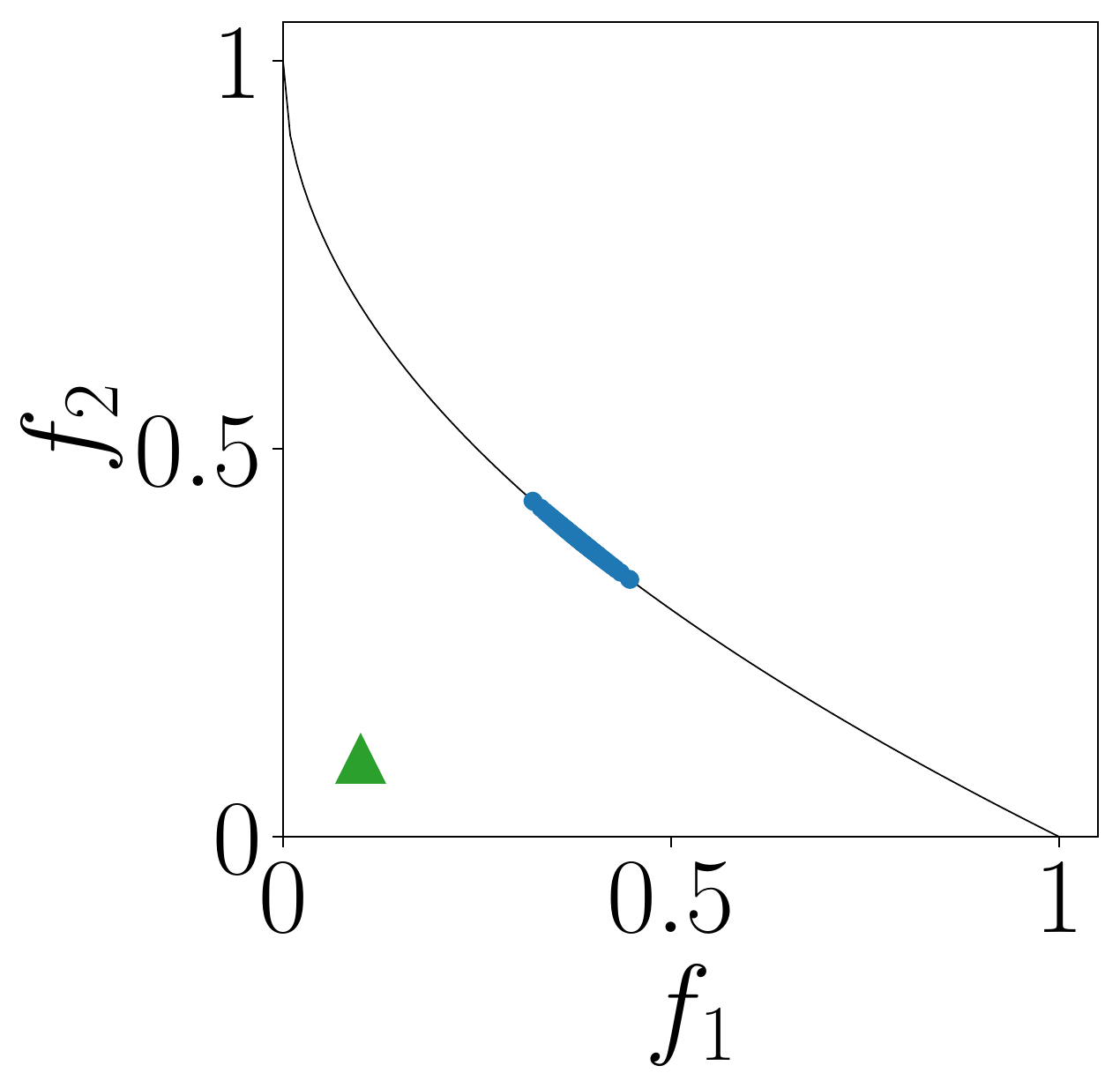}        
     }                               
     \caption{Distributions of points found by the six EMO algorithms on the convDTLZ2 problem when using $\mathbf{z}^{0.1} =(0.1, 0.1)^{\top}$.}
   \label{supfig:emo_points_convdtlz2_z0.1}
\subfloat[R-NSGA-II]{  
     \includegraphics[width=0.145\textwidth]{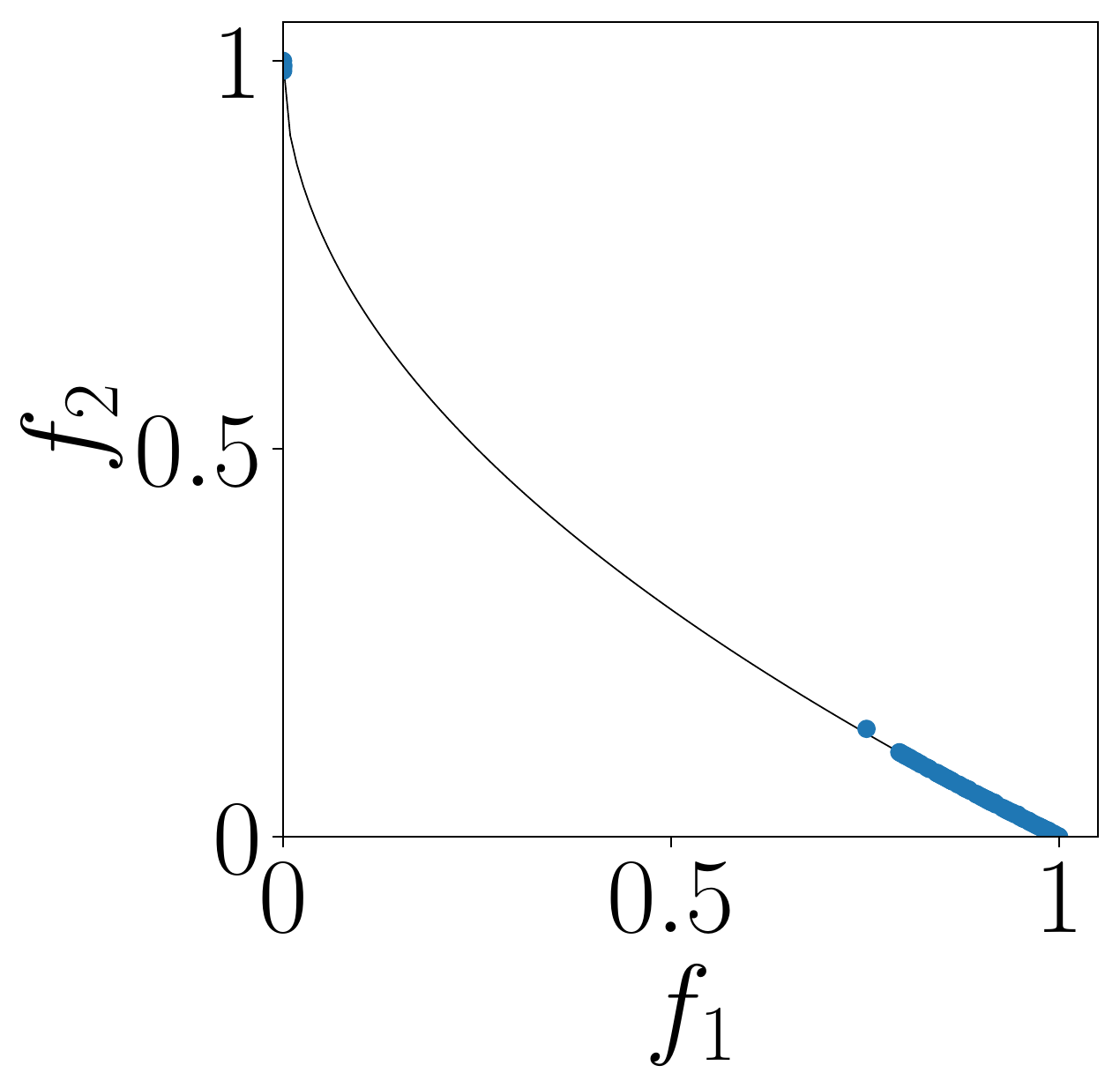}
     }
   \subfloat[r-NSGA-II]{  
     \includegraphics[width=0.145\textwidth]{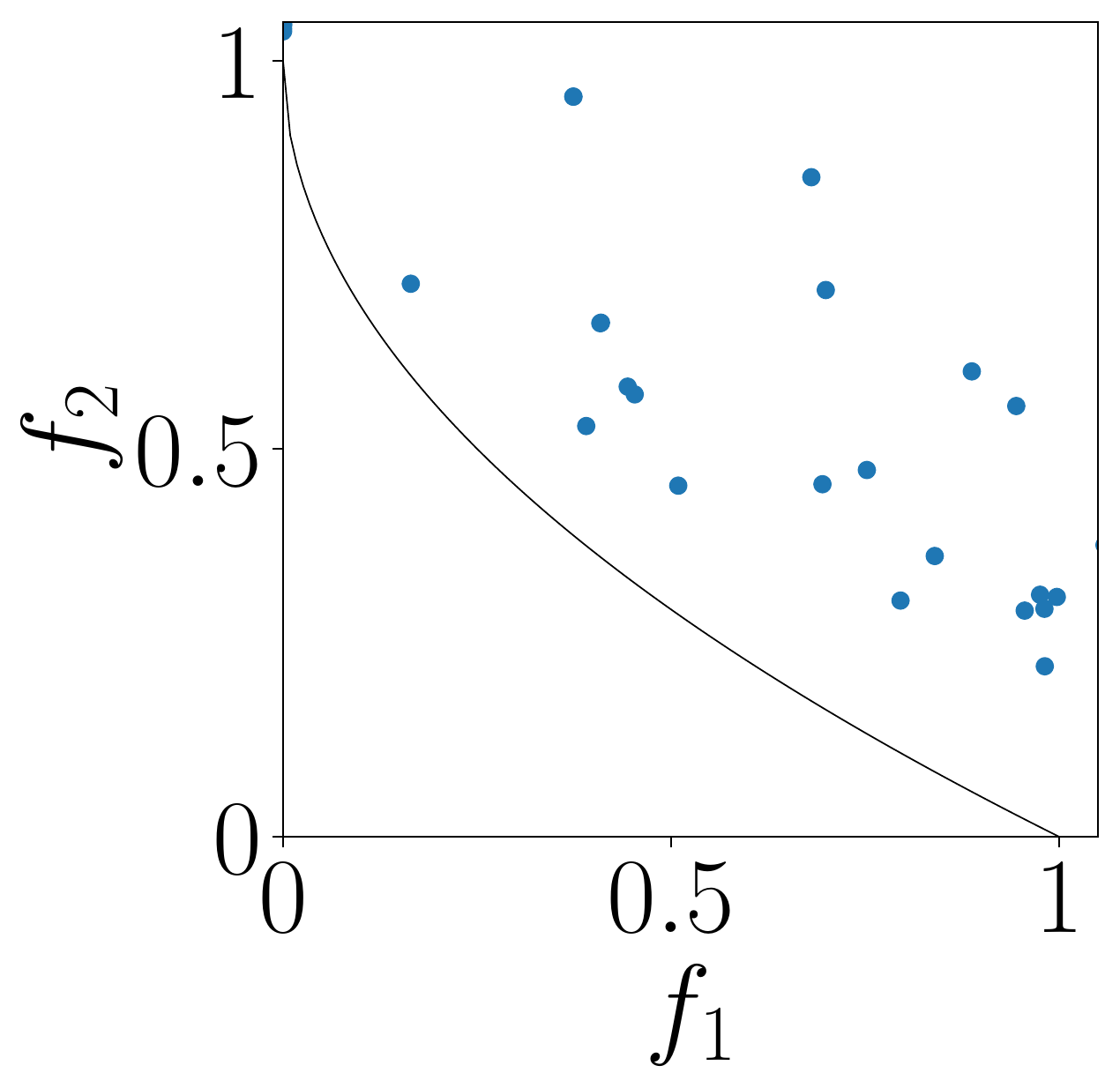}      
     }                                                                                                  
   \subfloat[g-NSGA-II]{  
     \includegraphics[width=0.145\textwidth]{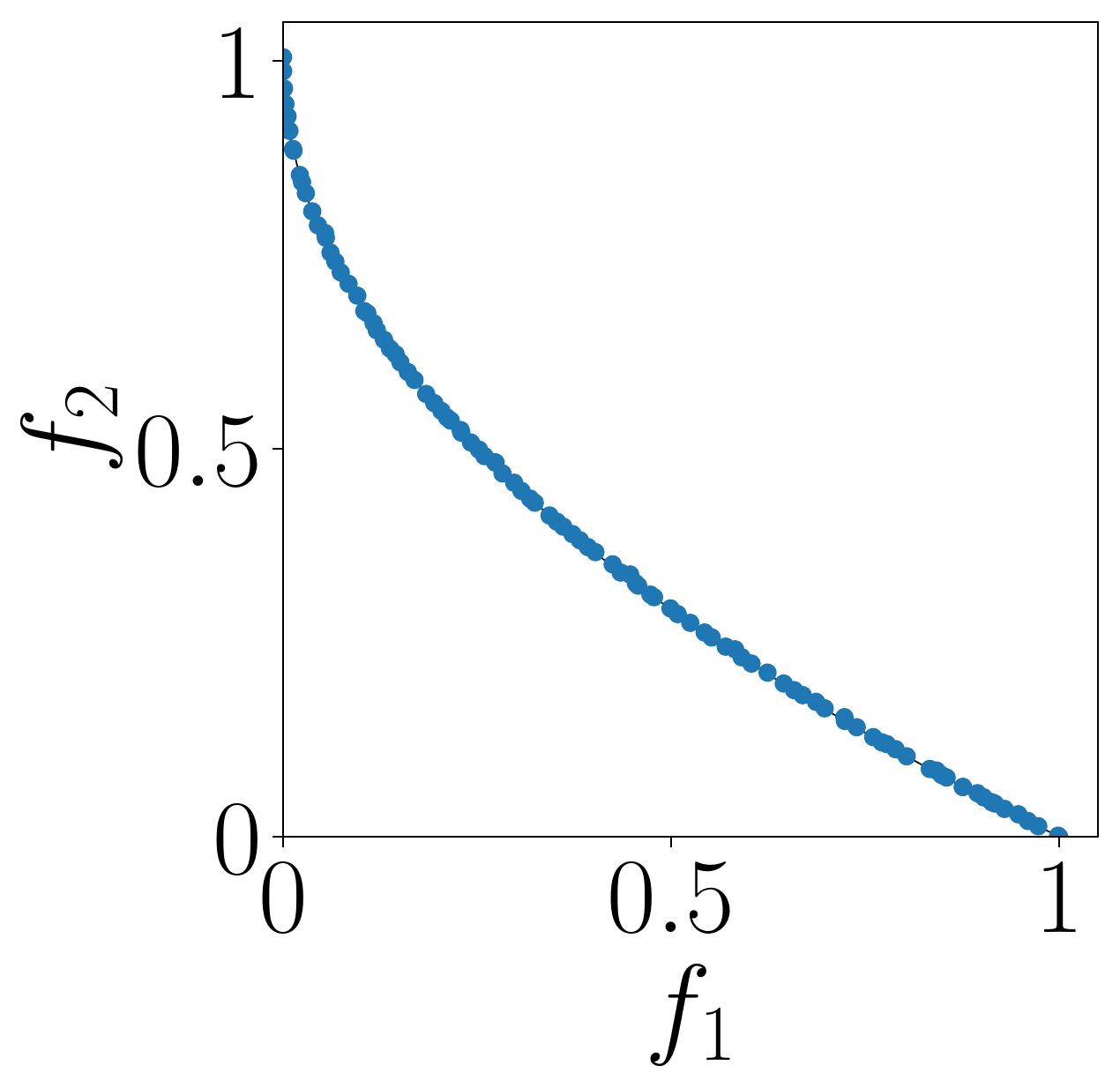}
     }
   \subfloat[PBEA]{  
     \includegraphics[width=0.145\textwidth]{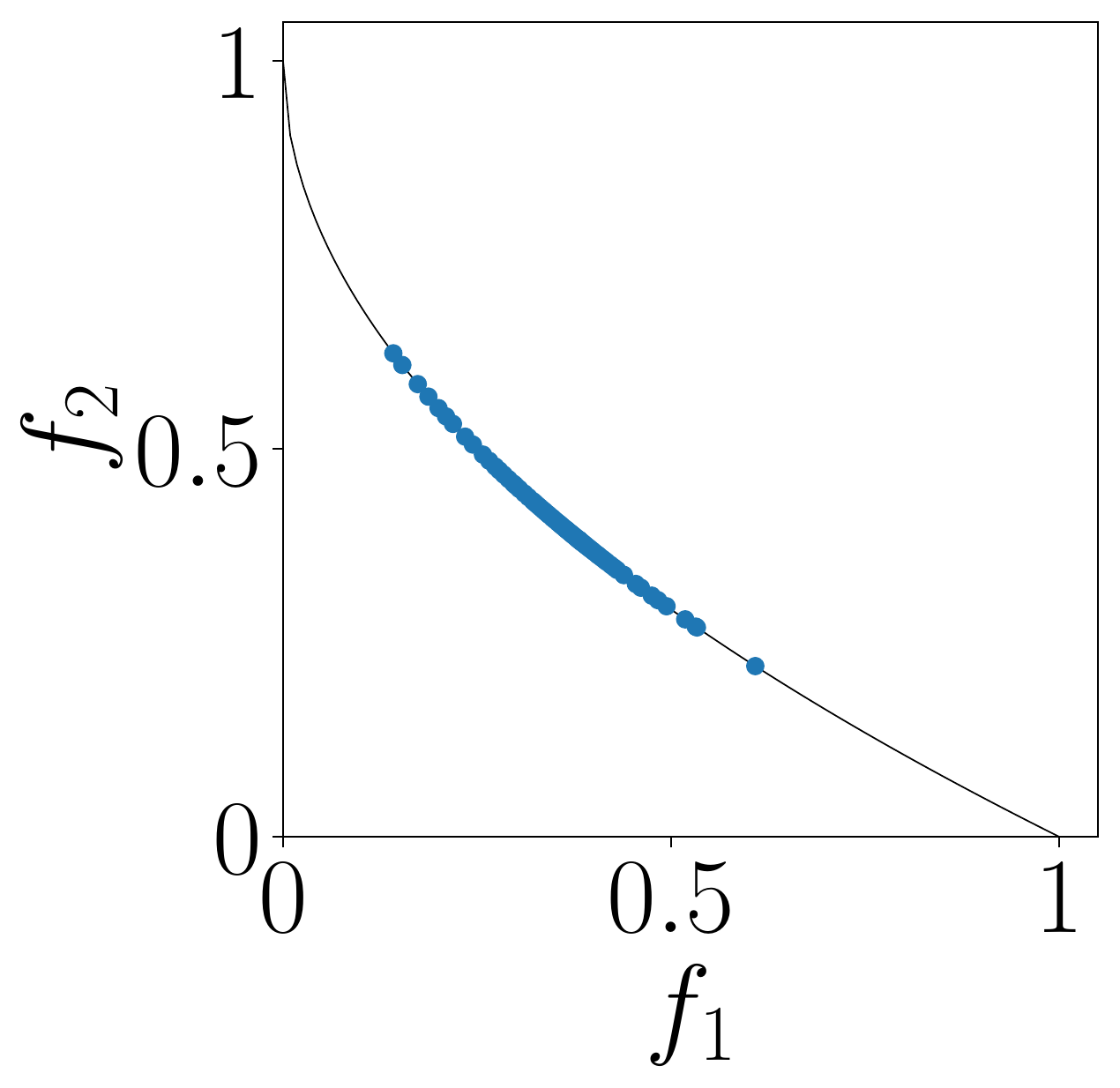}      
     }     
\subfloat[R-MEAD2]{  
     \includegraphics[width=0.145\textwidth]{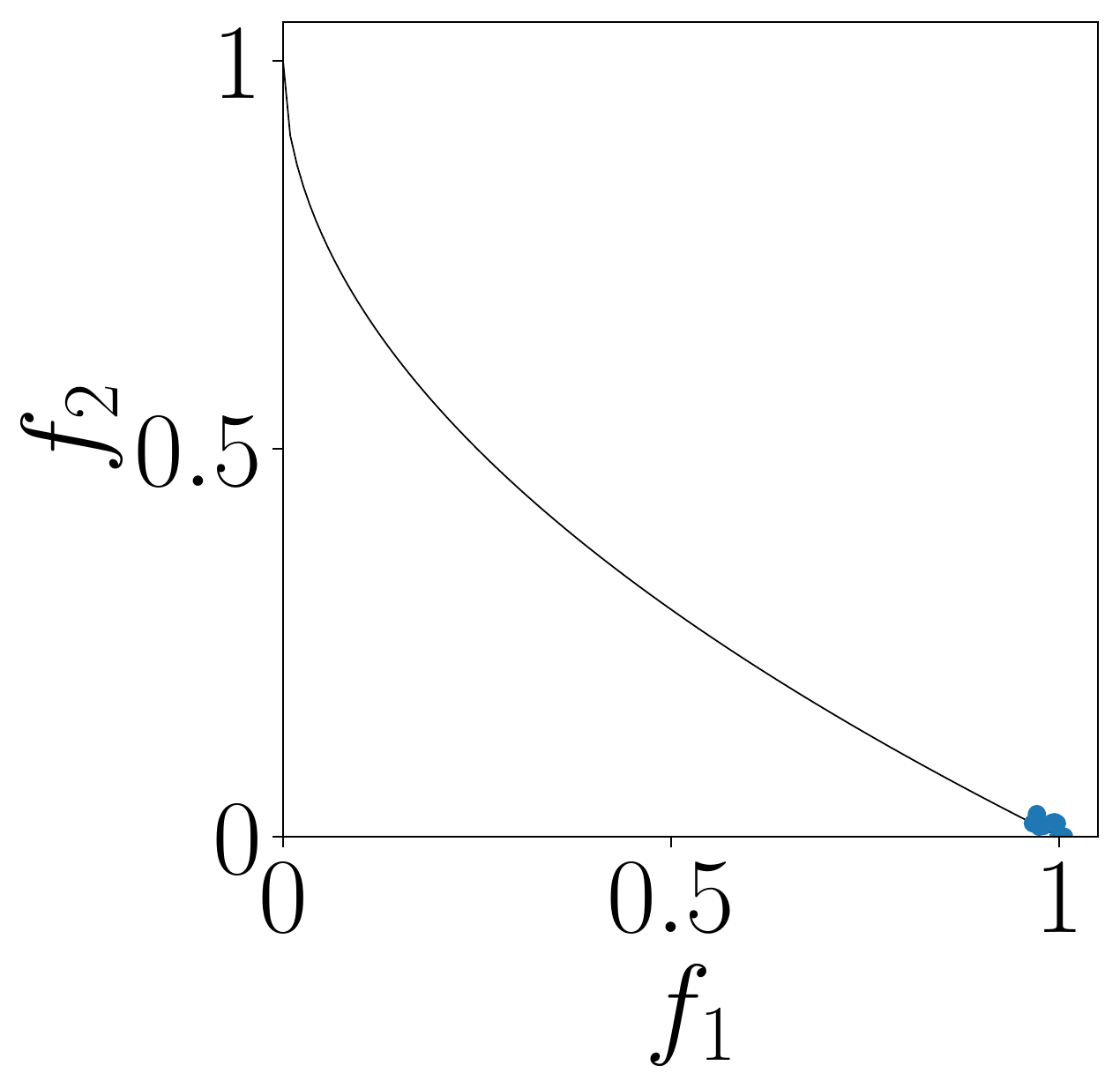}      
     }         
     \subfloat[MOEA/D-NUMS]{  
     \includegraphics[width=0.145\textwidth]{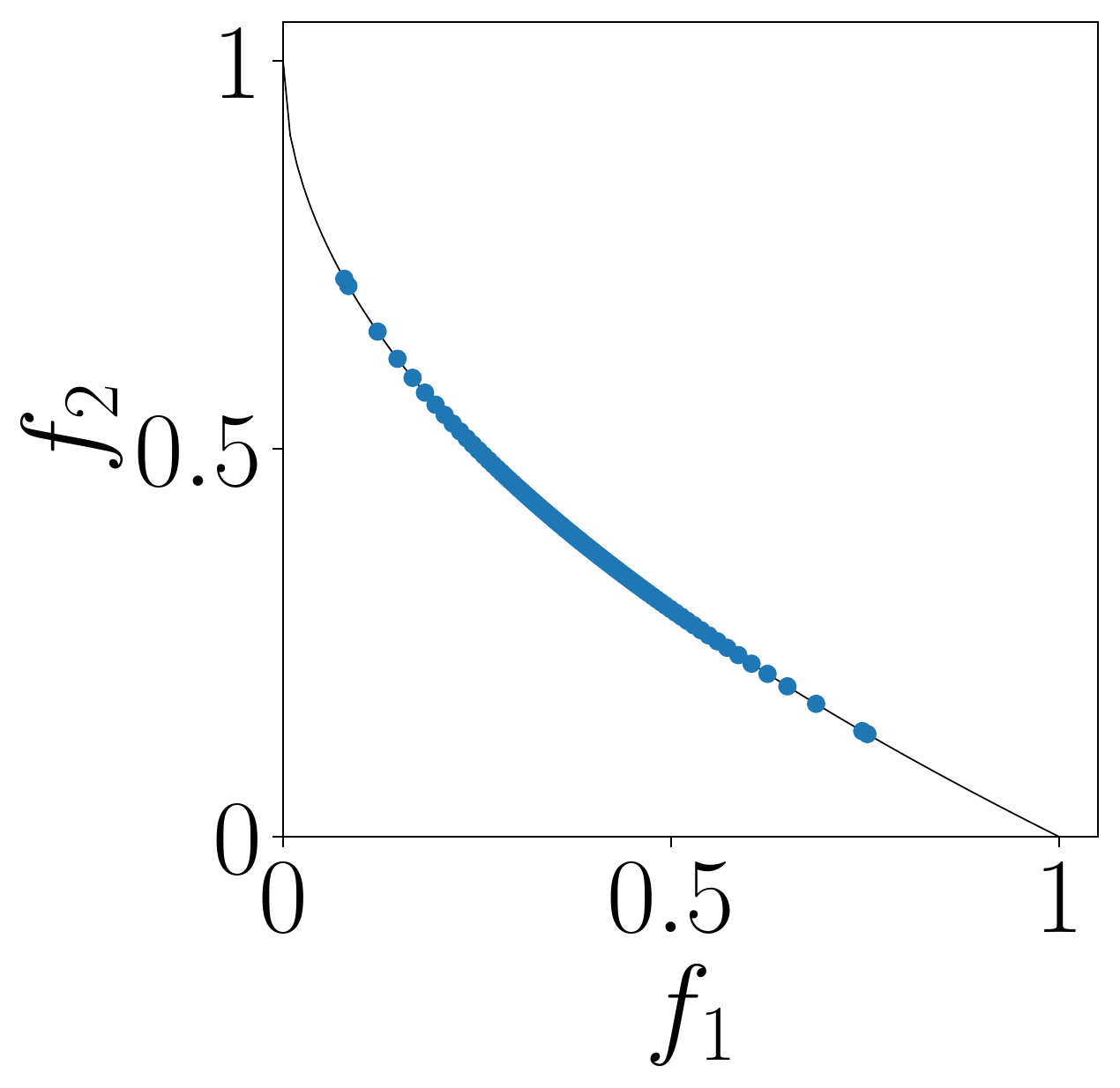}        
     }                         
     \caption{Distributions of points found by the six EMO algorithms on the convDTLZ2 problem when using $\mathbf{z}^{2} =(2, 2)^{\top}$. We found that r-NSGA-II does not work well when $\mathbf{z}$ is significantly worse than the nadir point. This is due to the property of the r-dominance relation, but further analysis is needed. 
     }
   \label{supfig:emo_points_convdtlz2_z-0.1}
\end{figure*}

\clearpage

\begin{figure*}[t]
   \centering
    \subfloat[$1\,000$ fevals.]{  
     \includegraphics[width=0.145\textwidth]{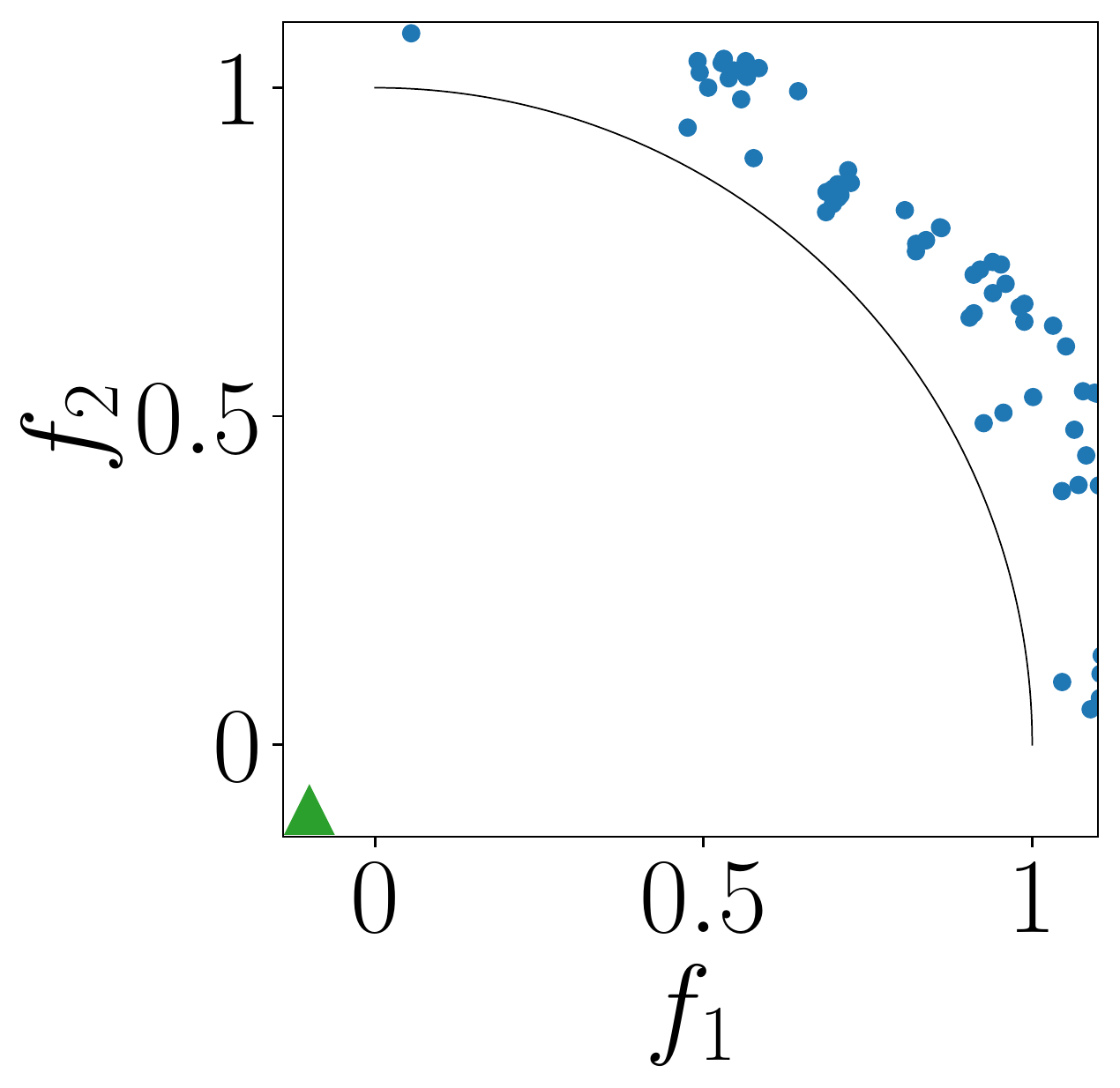}
     }   
     \subfloat[$5\,000$ fevals.]{  
     \includegraphics[width=0.145\textwidth]{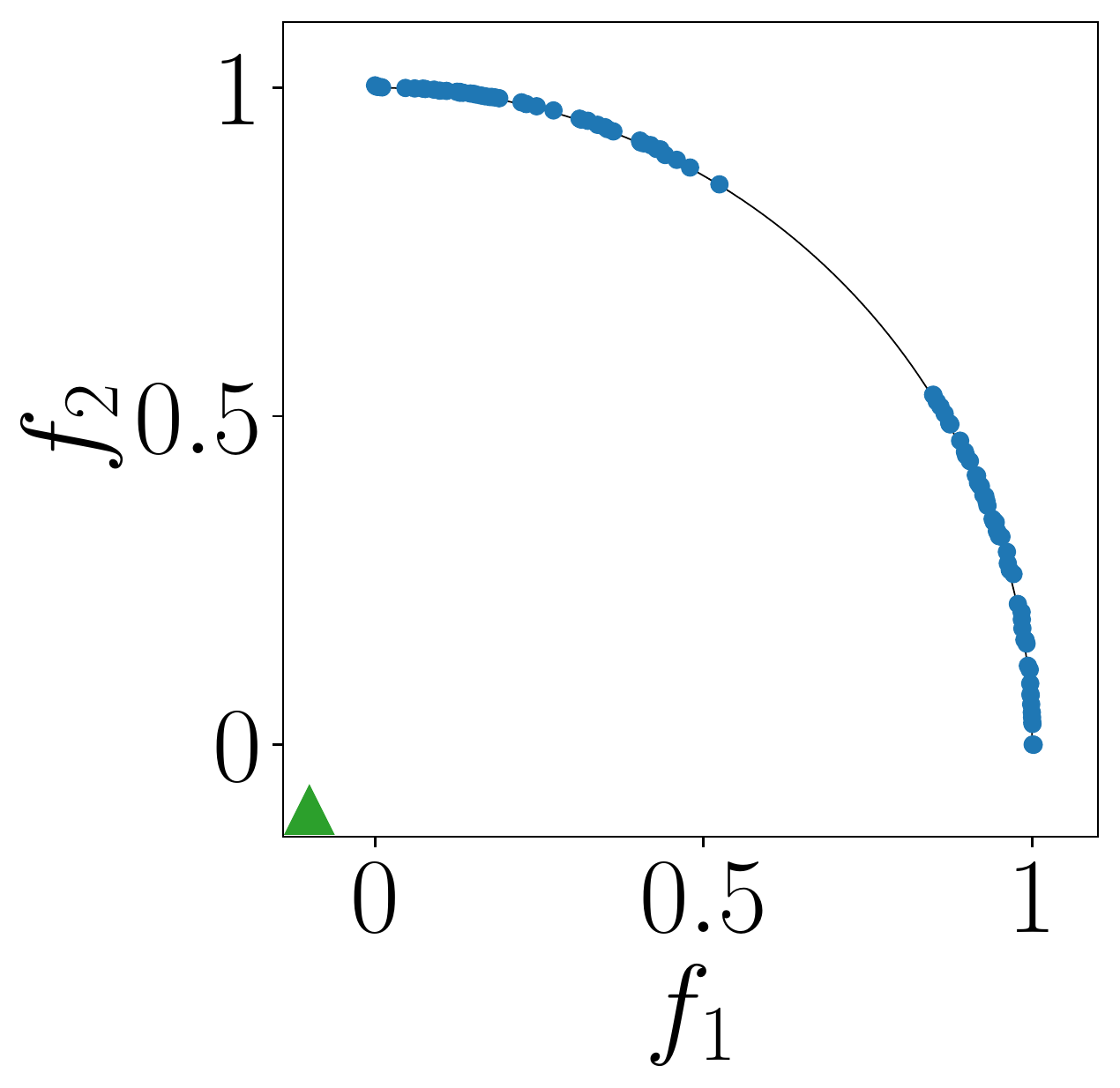}
     } 
     \subfloat[$10\,000$ fevals.]{  
     \includegraphics[width=0.145\textwidth]{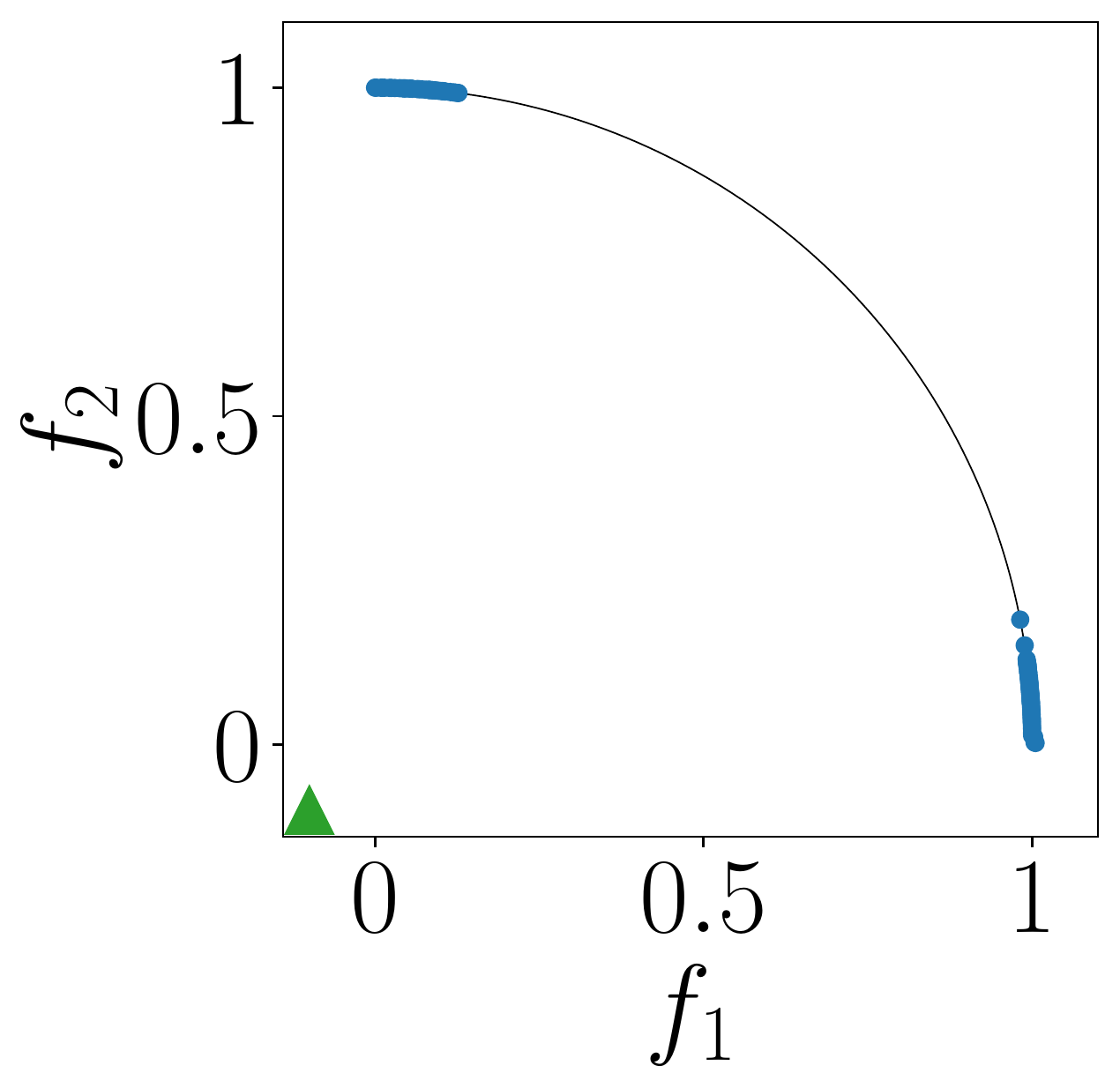}
     }   
     \subfloat[$30\,000$ fevals.]{  
     \includegraphics[width=0.145\textwidth]{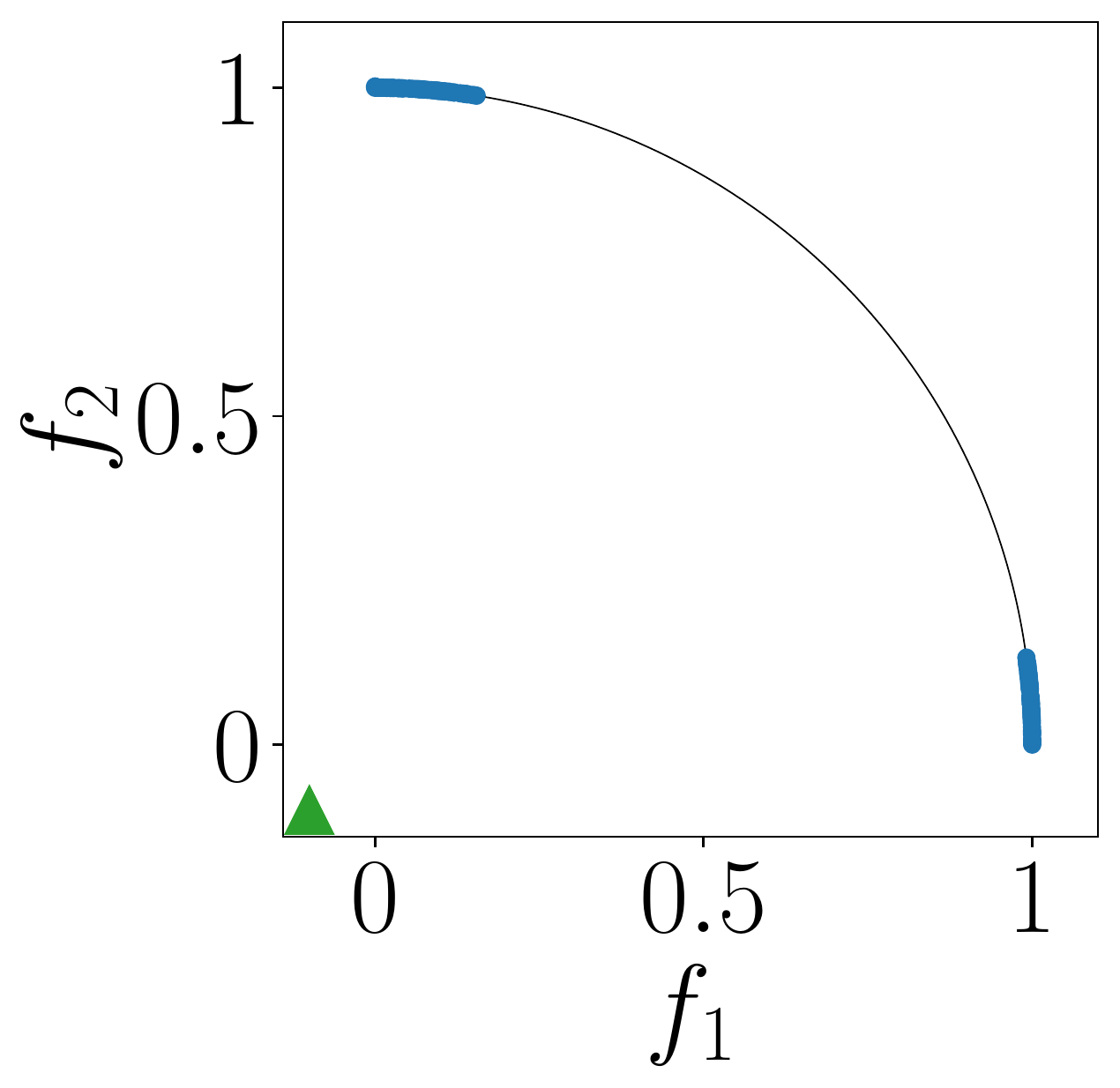}
     }   
     \subfloat[$50\,000$ fevals.]{  
     \includegraphics[width=0.145\textwidth]{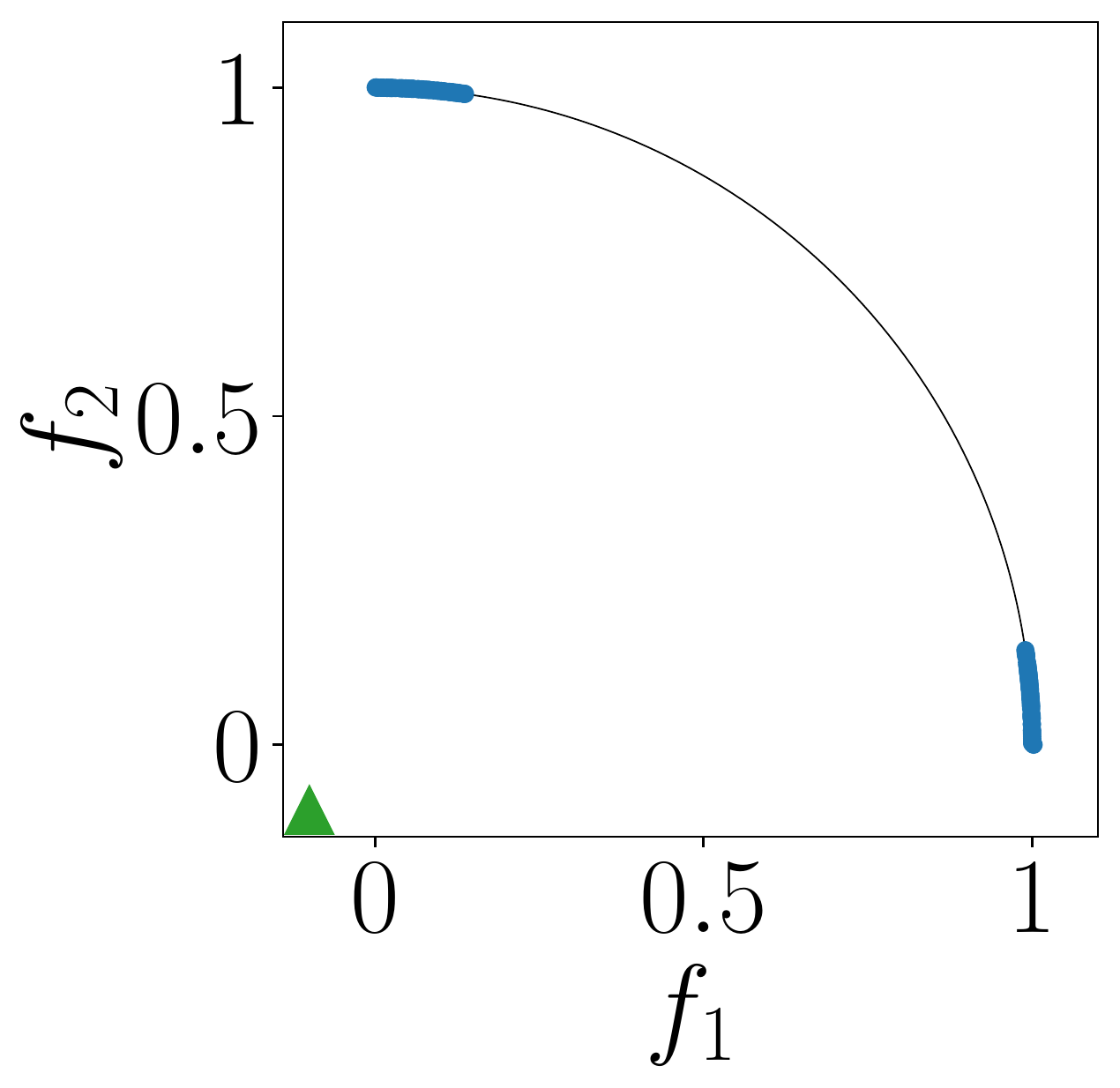}
     }   
     \caption{Distributions of points found by R-NSGA-II on the DTLZ2 problem for $1\,000$, $5\,000$, $10\,000$, $30\,000$, and $50\,000$ function evaluations (fevals) when using $\mathbf{z}^{2} =(-0.1, -0.1)^{\top}$.
     }
   \label{supfig:rnsga2_dtlz2_z-0.1_multi_fevals}
\end{figure*}

\begin{figure*}[t]
   \centering
    \subfloat[$1\,000$ fevals.]{  
     \includegraphics[width=0.145\textwidth]{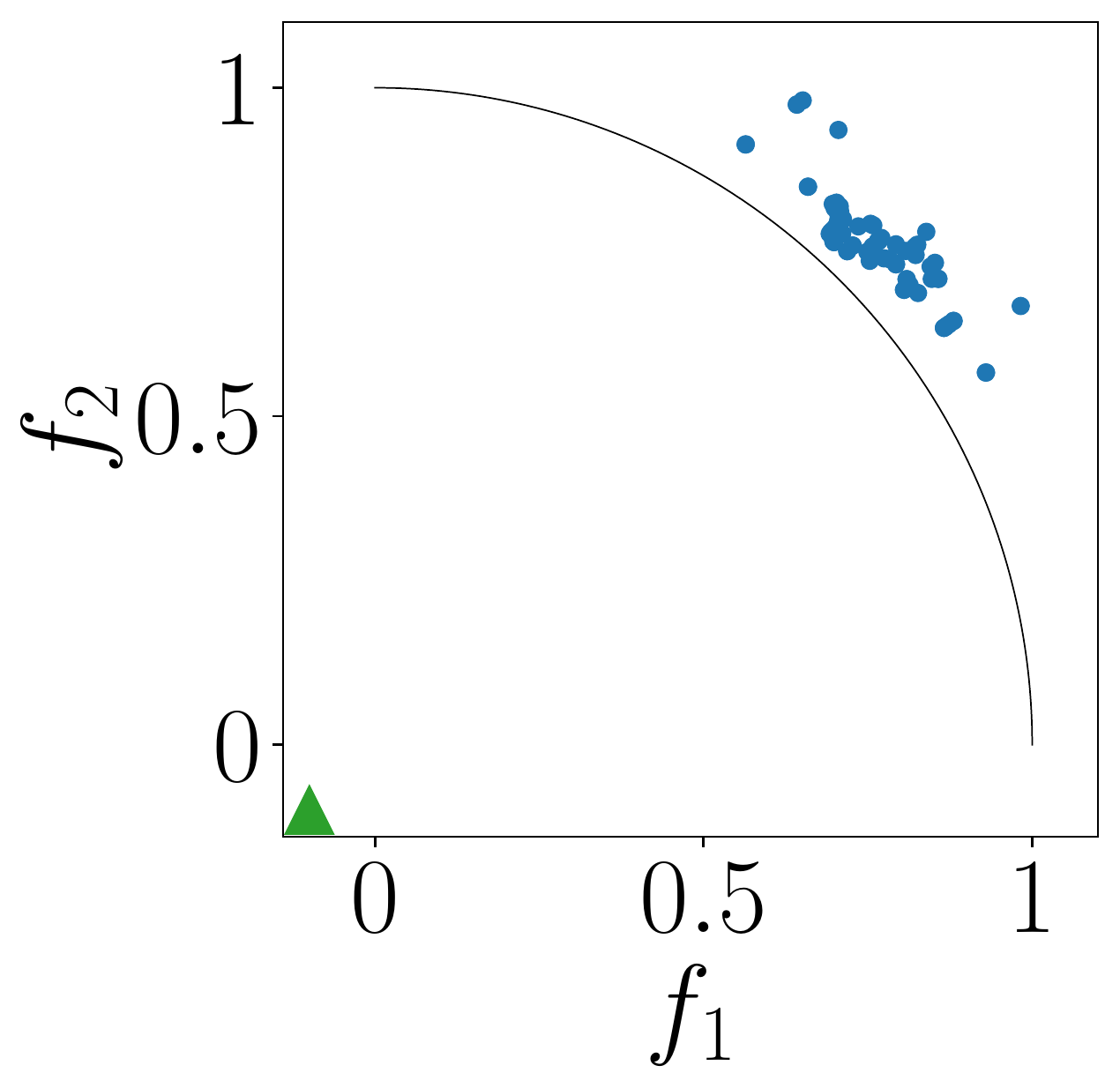}
     }   
     \subfloat[$5\,000$ fevals.]{  
     \includegraphics[width=0.145\textwidth]{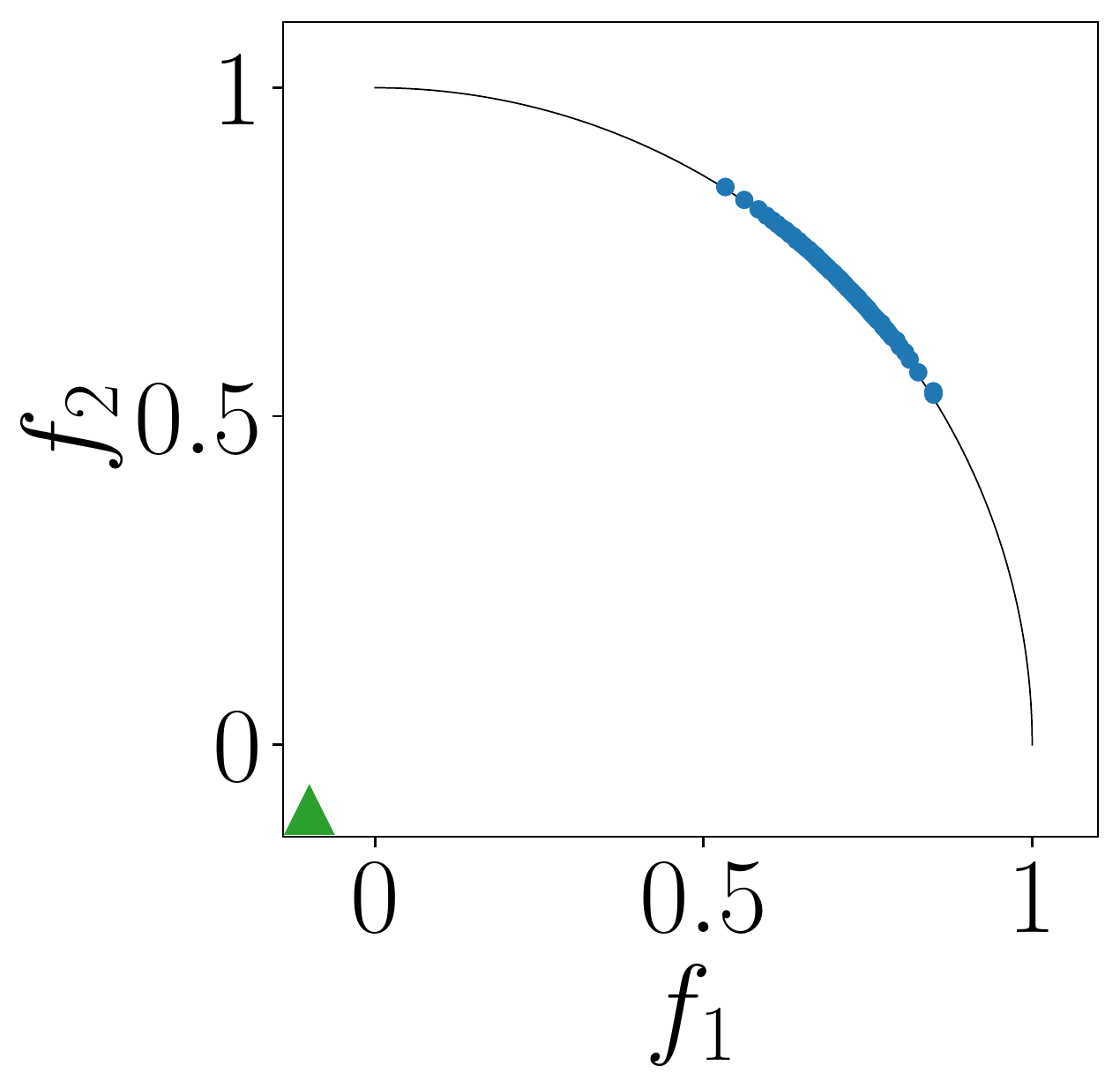}
     } 
     \subfloat[$10\,000$ fevals.]{  
     \includegraphics[width=0.145\textwidth]{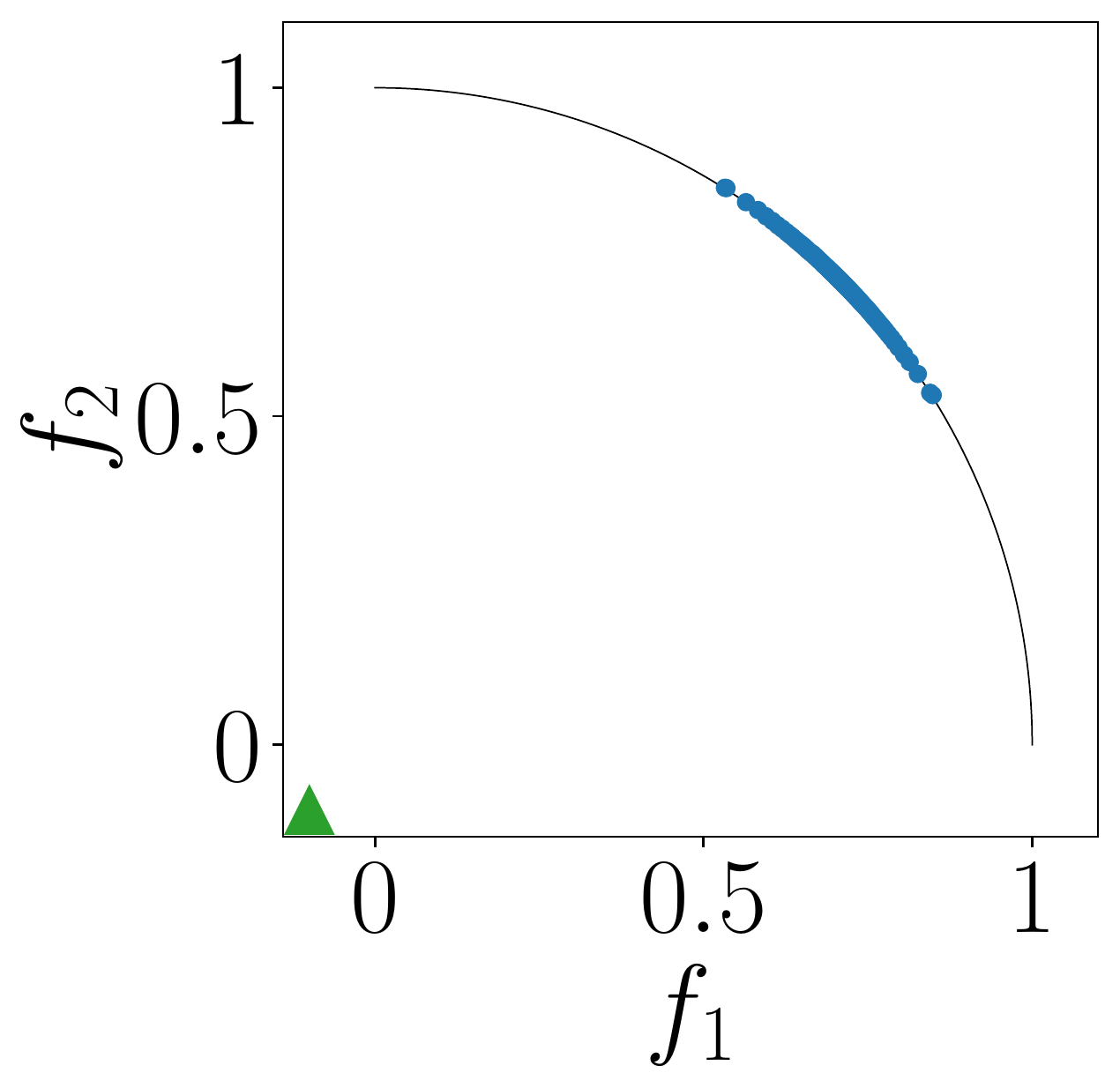}
     }   
     \subfloat[$30\,000$ fevals.]{  
     \includegraphics[width=0.145\textwidth]{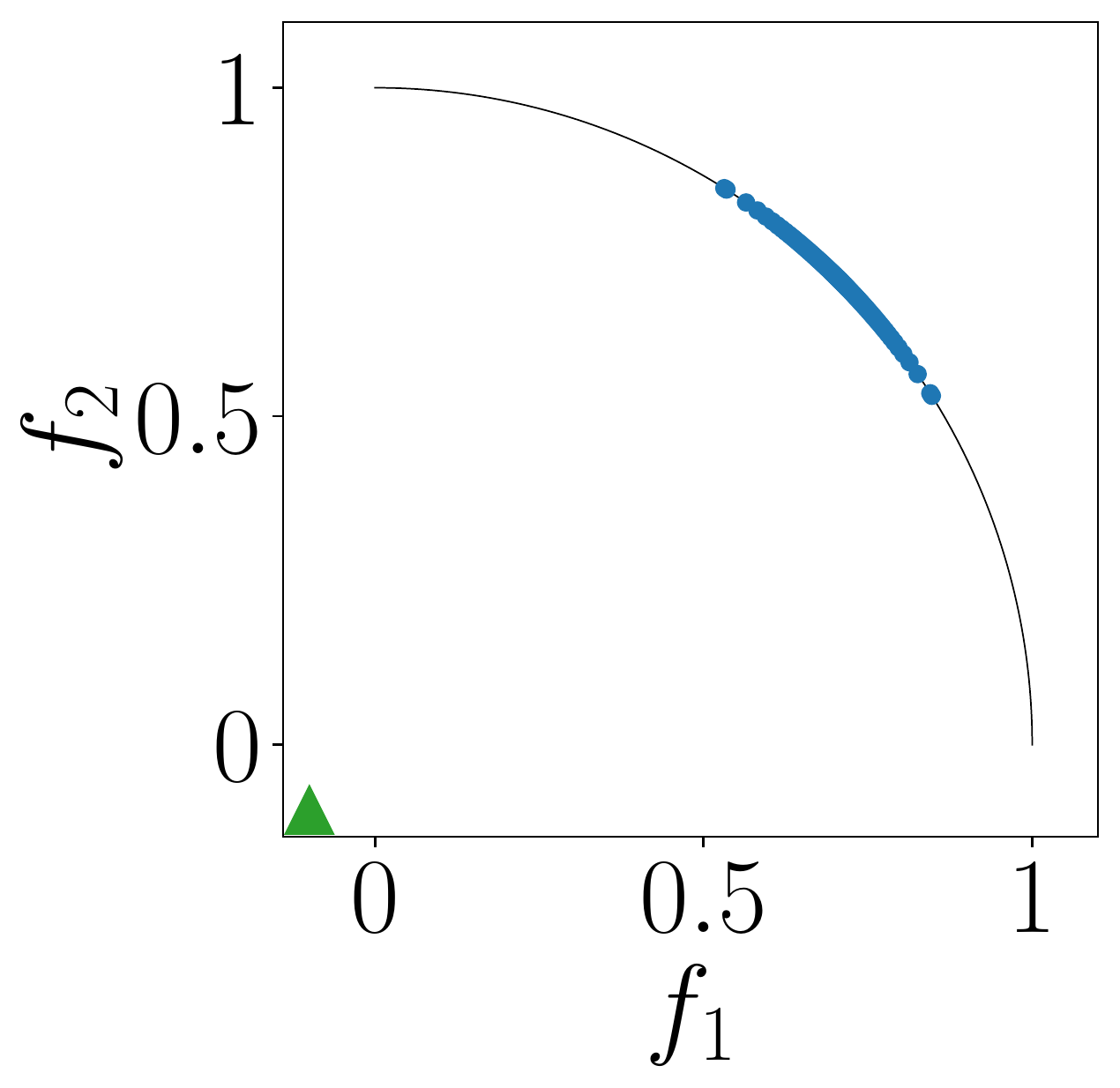}
     }   
     \subfloat[$50\,000$ fevals.]{  
     \includegraphics[width=0.145\textwidth]{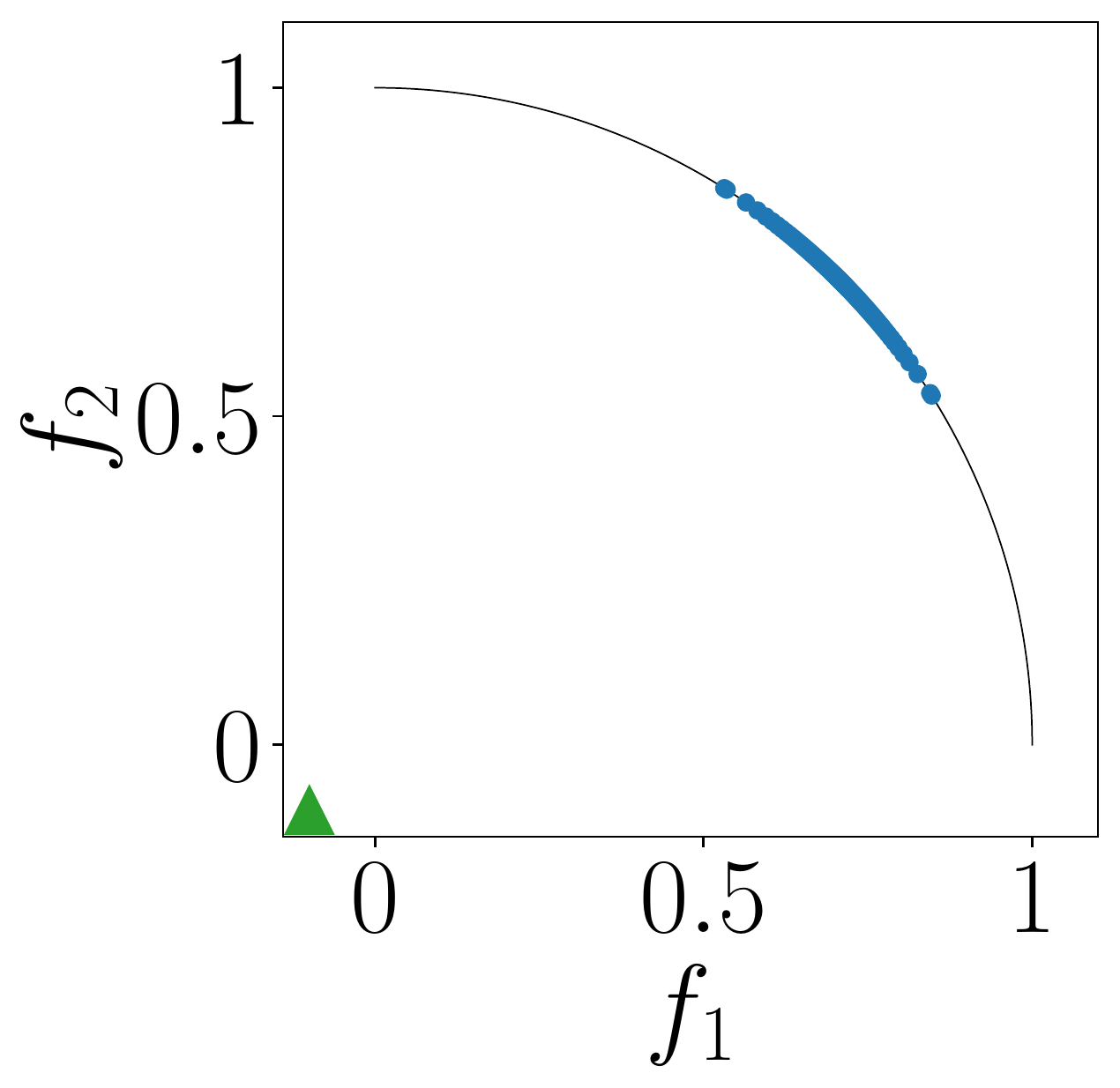}
     }   
     \caption{Distributions of points found by MOEA/D-NUMS on the DTLZ2 problem for $1\,000$, $5\,000$, $10\,000$, $30\,000$, and $50\,000$ function evaluations (fevals) when using $\mathbf{z}^{2} =(-0.1, -0.1)^{\top}$.
     }
   \label{supfig:MOEADNUMS_dtlz2_z-0.1_multi_fevals}
\end{figure*}

\begin{figure*}[t]
   \centering
    \subfloat[$1\,000$ fevals.]{  
     \includegraphics[width=0.145\textwidth]{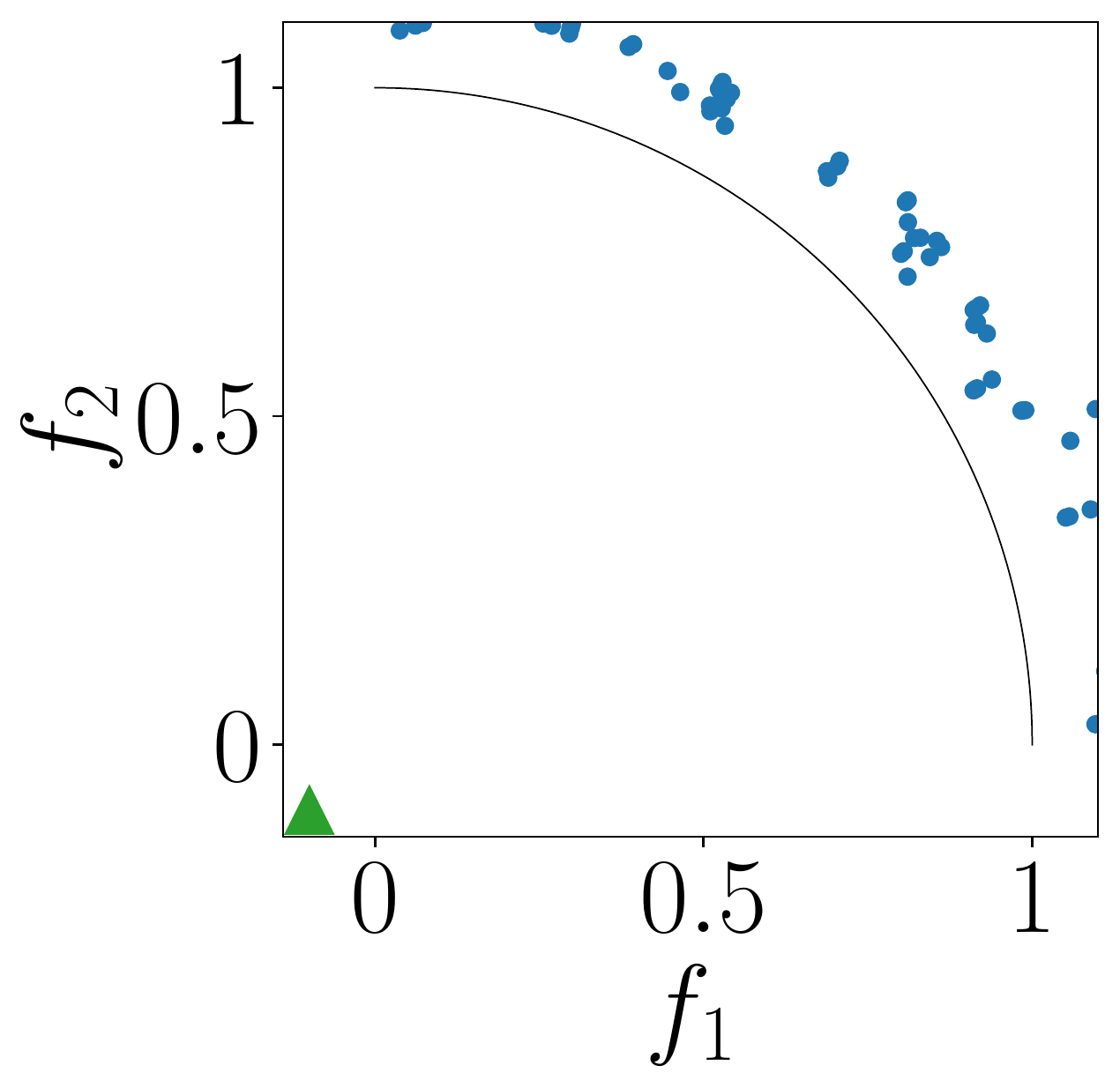}
     }   
     \subfloat[$5\,000$ fevals.]{  
     \includegraphics[width=0.145\textwidth]{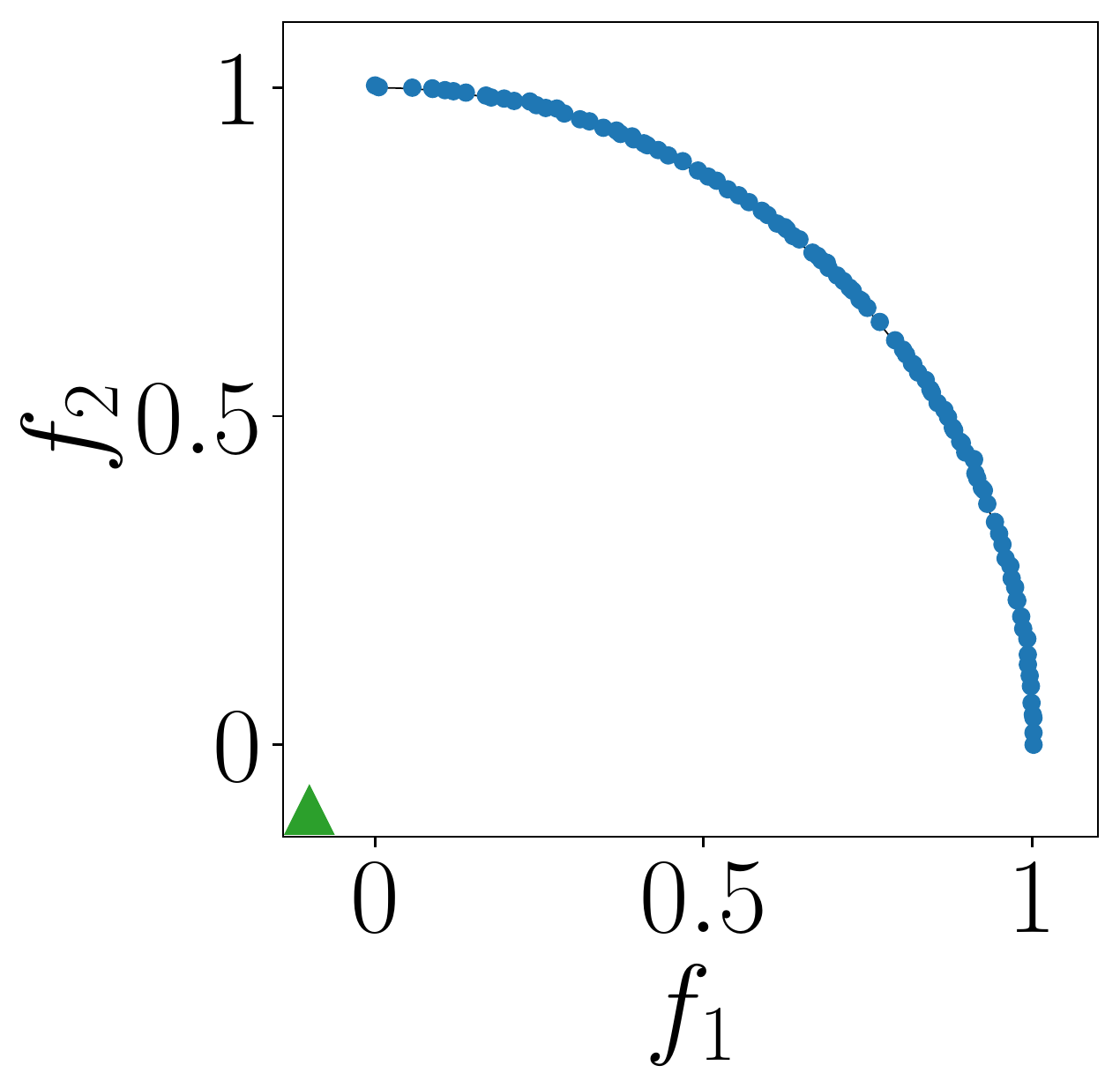}
     } 
     \subfloat[$10\,000$ fevals.]{  
     \includegraphics[width=0.145\textwidth]{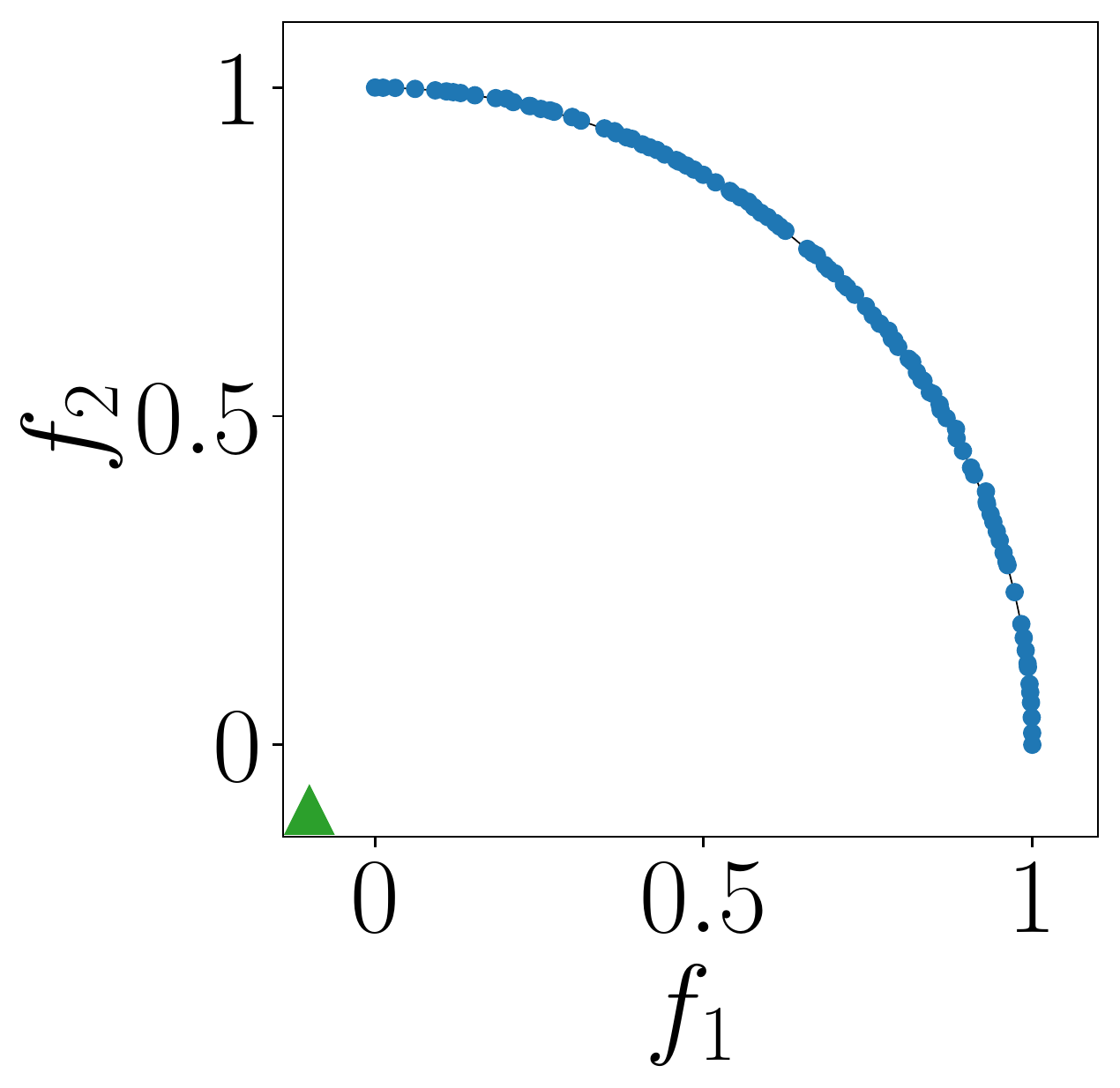}
     }   
     \subfloat[$30\,000$ fevals.]{  
     \includegraphics[width=0.145\textwidth]{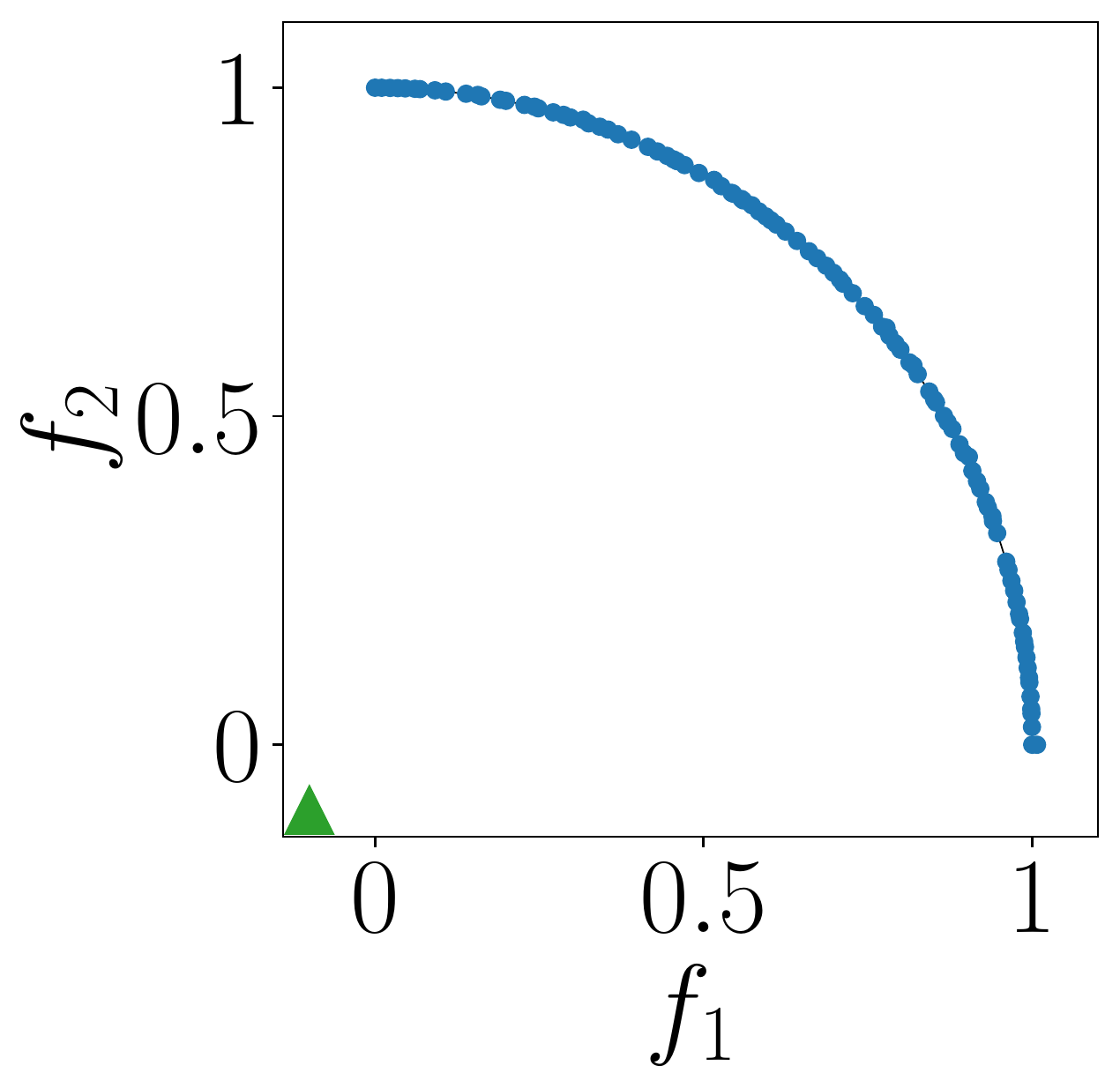}
     }   
     \subfloat[$50\,000$ fevals.]{  
     \includegraphics[width=0.145\textwidth]{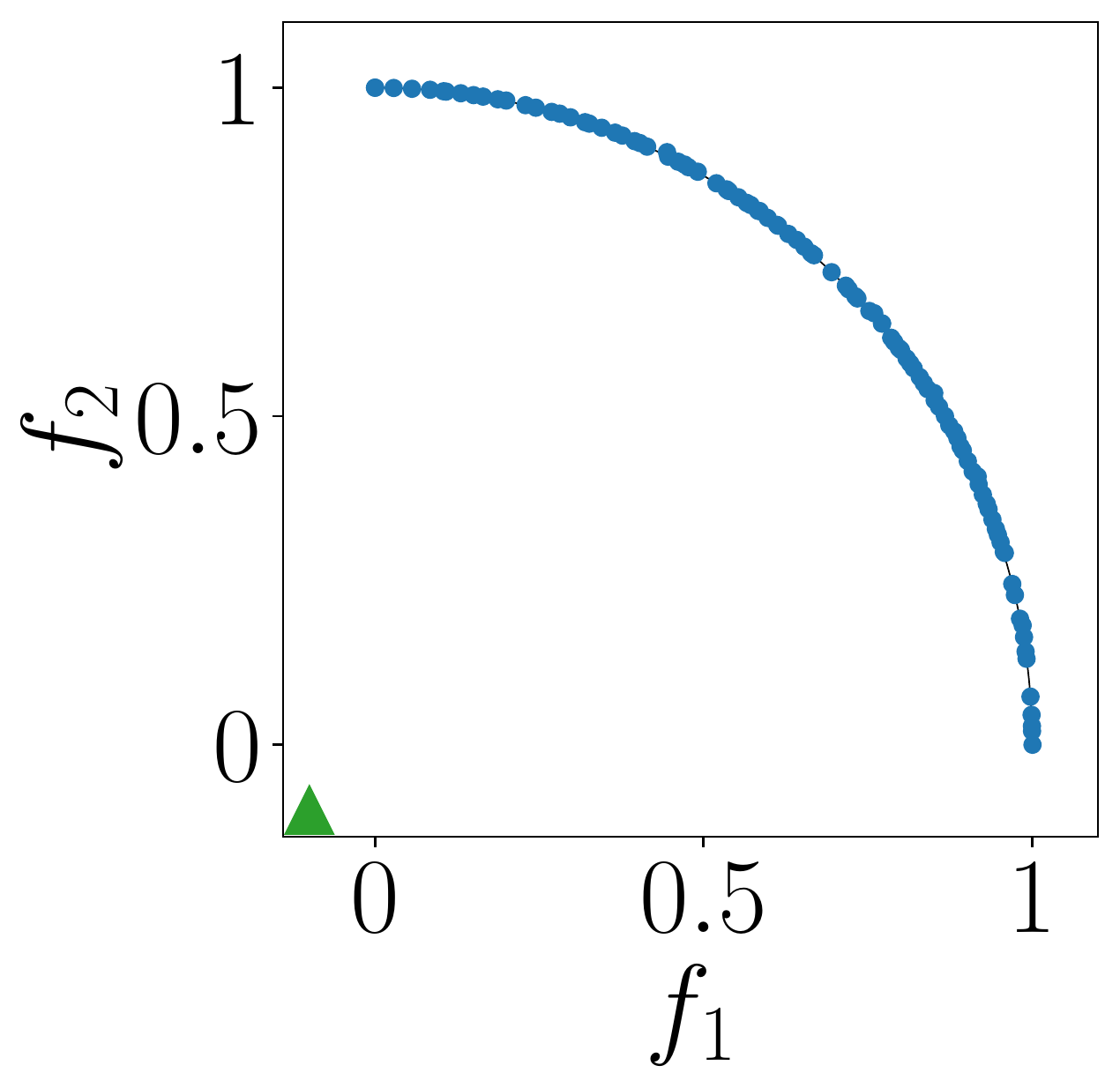}
     }
     \caption{Distributions of points found by g-NSGA-II on the DTLZ2 problem for $1\,000$, $5\,000$, $10\,000$, $30\,000$, and $50\,000$ function evaluations (fevals) when using $\mathbf{z}^{2} =(-0.1, -0.1)^{\top}$.
     }
   \label{supfig:gNSGA2_dtlz2_z-0.1_multi_fevals}
\end{figure*}

\clearpage

 \arrayrulecolor{black}
 \setlength\arrayrulewidth{1.2pt}
 \setlength{\extrarowheight}{1.2pt}
 \newcolumntype{A}{>{\columncolor[rgb]{142, 153, 148}}p{4.8em}}

\newcolumntype{C}{>{\centering\arraybackslash}X}
\newcolumntype{L}{>{\raggedright\arraybackslash}X}
\newcolumntype{R}{>{\raggedleft\arraybackslash}X}


\definecolor{c10}{RGB}{62, 110, 92}
\definecolor{c9}{RGB}{73, 126, 105}
\definecolor{c8}{RGB}{80, 136, 113}
\definecolor{c7}{RGB}{90, 151, 125}
\definecolor{c6}{RGB}{100, 166, 138}
\definecolor{c5}{RGB}{110, 180, 149}
\definecolor{c4}{RGB}{120, 194, 160}
\definecolor{c3}{RGB}{130, 209, 173}
\definecolor{c2}{RGB}{140, 224, 185}
\definecolor{c1}{RGB}{153, 242, 200}

 \newcolumntype{A}{>{\columncolor[rgb]{0.9, 0.9, 0.9}\centering}p{1.8em}}

 \begin{table*}[t]
   \setlength{\tabcolsep}{2pt} 
  \centering
  \caption{\small Rankings of the 10 synthetic point sets on the DTLZ1 problem by the 16 quality indicators when using $\mathbf{z}^{0.51}$ and $\mathbf{z}^{-0.1}$. The quality indicators based on the ROI-P (IGD-P, HV$_{\mathbf{z}}$, PR) do not work when the reference point is on the PF. For this reason, we used $\mathbf{z}^{0.51}=(0.51, 0.51)^{\top}$, instead of $\mathbf{z}^{0.5}=(0.5, 0.5)^{\top}$.}
  \label{suptab:rankings_psets_dtlz1}
        {\scriptsize
\subfloat[$\mathbf{z}^{0.51}=(0.51, 0.51)^{\top}$]{
\begin{tabularx}{72em}{|A|C|C|C|C|C|C|C|C|C|C|C|C|C|C|C|C|}
\hline
\rowcolor[rgb]{0.9, 0.9, 0.9} & MASF & MED & IGD-C & IGD-A & IGD-P & HV$_{\mathbf{z}}$ & PR & PMOD & IGD-CF & HV-CF & PMDA & R-IGD & R-HV & EH & HV & IGD \\
\hline
$\mathcal{P}^{1}$ & \cellcolor{c9}9 & \cellcolor{c10}10 & \cellcolor{c9}9 & \cellcolor{c9}9 & \cellcolor{c9}9 & \cellcolor{c3}3 & \cellcolor{c4}4 & \cellcolor{c8}8 & \cellcolor{c4}4 & \cellcolor{c4}4 & \cellcolor{c10}10 & \cellcolor{c6}6 & \cellcolor{c6}6 & \cellcolor{c6}6 & \cellcolor{c9}9 & \cellcolor{c9}9\\
$\mathcal{P}^{2}$ & \cellcolor{c5}5 & \cellcolor{c4}4 & \cellcolor{c5}5 & \cellcolor{c5}5 & \cellcolor{c6}6 & \cellcolor{c3}3 & \cellcolor{c4}4 & \cellcolor{c1}1 & \cellcolor{c4}4 & \cellcolor{c4}4 & \cellcolor{c4}4 & \cellcolor{c4}4 & \cellcolor{c5}5 & \cellcolor{c3}3 & \cellcolor{c3}3 & \cellcolor{c4}4\\
$\mathcal{P}^{3}$ & \cellcolor{c2}2 & \cellcolor{c2}2 & \cellcolor{c1}1 & \cellcolor{c1}1 & \cellcolor{c2}2 & \cellcolor{c2}2 & \cellcolor{c3}3 & \cellcolor{c4}4 & \cellcolor{c1}1 & \cellcolor{c1}1 & \cellcolor{c2}2 & \cellcolor{c1}1 & \cellcolor{c1}1 & \cellcolor{c2}2 & \cellcolor{c2}2 & \cellcolor{c2}2\\
$\mathcal{P}^{4}$ & \cellcolor{c5}5 & \cellcolor{c5}5 & \cellcolor{c6}6 & \cellcolor{c6}6 & \cellcolor{c5}5 & \cellcolor{c3}3 & \cellcolor{c4}4 & \cellcolor{c5}5 & \cellcolor{c4}4 & \cellcolor{c4}4 & \cellcolor{c5}5 & \cellcolor{c5}5 & \cellcolor{c4}4 & \cellcolor{c3}3 & \cellcolor{c3}3 & \cellcolor{c4}4\\
$\mathcal{P}^{5}$ & \cellcolor{c9}9 & \cellcolor{c9}9 & \cellcolor{c10}10 & \cellcolor{c10}10 & \cellcolor{c9}9 & \cellcolor{c3}3 & \cellcolor{c4}4 & \cellcolor{c9}9 & \cellcolor{c4}4 & \cellcolor{c4}4 & \cellcolor{c9}9 & \cellcolor{c7}7 & \cellcolor{c6}6 & \cellcolor{c6}6 & \cellcolor{c9}9 & \cellcolor{c9}9\\
$\mathcal{P}^{6}$ & \cellcolor{c7}7 & \cellcolor{c6}6 & \cellcolor{c7}7 & \cellcolor{c7}7 & \cellcolor{c7}7 & \cellcolor{c3}3 & \cellcolor{c4}4 & \cellcolor{c3}3 & \cellcolor{c4}4 & \cellcolor{c4}4 & \cellcolor{c7}7 & \cellcolor{c8}8 & \cellcolor{c8}8 & \cellcolor{c8}8 & \cellcolor{c7}7 & \cellcolor{c7}7\\
$\mathcal{P}^{7}$ & \cellcolor{c4}4 & \cellcolor{c3}3 & \cellcolor{c4}4 & \cellcolor{c4}4 & \cellcolor{c4}4 & \cellcolor{c3}3 & \cellcolor{c1}1 & \cellcolor{c7}7 & \cellcolor{c4}4 & \cellcolor{c4}4 & \cellcolor{c3}3 & \cellcolor{c8}8 & \cellcolor{c8}8 & \cellcolor{c8}8 & \cellcolor{c6}6 & \cellcolor{c3}3\\
$\mathcal{P}^{8}$ & \cellcolor{c7}7 & \cellcolor{c7}7 & \cellcolor{c8}8 & \cellcolor{c8}8 & \cellcolor{c7}7 & \cellcolor{c3}3 & \cellcolor{c4}4 & \cellcolor{c6}6 & \cellcolor{c4}4 & \cellcolor{c4}4 & \cellcolor{c8}8 & \cellcolor{c8}8 & \cellcolor{c8}8 & \cellcolor{c8}8 & \cellcolor{c7}7 & \cellcolor{c8}8\\
$\mathcal{P}^{9}$ & \cellcolor{c1}1 & \cellcolor{c1}1 & \cellcolor{c3}3 & \cellcolor{c3}3 & \cellcolor{c1}1 & \cellcolor{c1}1 & \cellcolor{c2}2 & \cellcolor{c2}2 & \cellcolor{c3}3 & \cellcolor{c3}3 & \cellcolor{c1}1 & \cellcolor{c3}3 & \cellcolor{c3}3 & \cellcolor{c1}1 & \cellcolor{c5}5 & \cellcolor{c6}6\\
$\mathcal{P}^{10}$ & \cellcolor{c3}3 & \cellcolor{c8}8 & \cellcolor{c2}2 & \cellcolor{c2}2 & \cellcolor{c3}3 & \cellcolor{c3}3 & \cellcolor{c4}4 & \cellcolor{c10}10 & \cellcolor{c2}2 & \cellcolor{c2}2 & \cellcolor{c6}6 & \cellcolor{c2}2 & \cellcolor{c2}2 & \cellcolor{c5}5 & \cellcolor{c1}1 & \cellcolor{c1}1\\
\hline
\end{tabularx}
}
\\
\subfloat[$\mathbf{z}^{-0.1}=(-0.1, -0.1)^{\top}$]{
\begin{tabularx}{72em}{|A|C|C|C|C|C|C|C|C|C|C|C|C|C|C|C|C|}
\hline
\rowcolor[rgb]{0.9, 0.9, 0.9} & MASF & MED & IGD-C & IGD-A & IGD-P & HV$_{\mathbf{z}}$ & PR & PMOD & IGD-CF & HV-CF & PMDA & R-IGD & R-HV & EH & HV & IGD \\
\hline
$\mathcal{P}^{1}$ & \cellcolor{c9}9 & \cellcolor{c9}9 & \cellcolor{c9}9 & \cellcolor{c9}9 & \cellcolor{c9}9 & \cellcolor{c9}9 & \cellcolor{c1}1 & \cellcolor{c8}8 & \cellcolor{c4}4 & \cellcolor{c4}4 & \cellcolor{c10}10 & \cellcolor{c6}6 & \cellcolor{c7}7 & \cellcolor{c6}6 & \cellcolor{c9}9 & \cellcolor{c9}9\\
$\mathcal{P}^{2}$ & \cellcolor{c5}5 & \cellcolor{c3}3 & \cellcolor{c5}5 & \cellcolor{c5}5 & \cellcolor{c4}4 & \cellcolor{c3}3 & \cellcolor{c1}1 & \cellcolor{c1}1 & \cellcolor{c4}4 & \cellcolor{c4}4 & \cellcolor{c4}4 & \cellcolor{c4}4 & \cellcolor{c5}5 & \cellcolor{c3}3 & \cellcolor{c3}3 & \cellcolor{c4}4\\
$\mathcal{P}^{3}$ & \cellcolor{c2}2 & \cellcolor{c2}2 & \cellcolor{c1}1 & \cellcolor{c1}1 & \cellcolor{c2}2 & \cellcolor{c2}2 & \cellcolor{c1}1 & \cellcolor{c4}4 & \cellcolor{c1}1 & \cellcolor{c1}1 & \cellcolor{c2}2 & \cellcolor{c1}1 & \cellcolor{c1}1 & \cellcolor{c2}2 & \cellcolor{c2}2 & \cellcolor{c2}2\\
$\mathcal{P}^{4}$ & \cellcolor{c5}5 & \cellcolor{c4}4 & \cellcolor{c6}6 & \cellcolor{c6}6 & \cellcolor{c4}4 & \cellcolor{c3}3 & \cellcolor{c1}1 & \cellcolor{c5}5 & \cellcolor{c4}4 & \cellcolor{c4}4 & \cellcolor{c5}5 & \cellcolor{c5}5 & \cellcolor{c4}4 & \cellcolor{c3}3 & \cellcolor{c3}3 & \cellcolor{c4}4\\
$\mathcal{P}^{5}$ & \cellcolor{c9}9 & \cellcolor{c9}9 & \cellcolor{c10}10 & \cellcolor{c10}10 & \cellcolor{c9}9 & \cellcolor{c9}9 & \cellcolor{c1}1 & \cellcolor{c10}10 & \cellcolor{c4}4 & \cellcolor{c4}4 & \cellcolor{c9}9 & \cellcolor{c7}7 & \cellcolor{c6}6 & \cellcolor{c6}6 & \cellcolor{c9}9 & \cellcolor{c9}9\\
$\mathcal{P}^{6}$ & \cellcolor{c7}7 & \cellcolor{c8}8 & \cellcolor{c7}7 & \cellcolor{c7}7 & \cellcolor{c7}7 & \cellcolor{c7}7 & \cellcolor{c1}1 & \cellcolor{c3}3 & \cellcolor{c4}4 & \cellcolor{c4}4 & \cellcolor{c7}7 & \cellcolor{c8}8 & \cellcolor{c8}8 & \cellcolor{c8}8 & \cellcolor{c7}7 & \cellcolor{c7}7\\
$\mathcal{P}^{7}$ & \cellcolor{c4}4 & \cellcolor{c6}6 & \cellcolor{c4}4 & \cellcolor{c4}4 & \cellcolor{c3}3 & \cellcolor{c6}6 & \cellcolor{c1}1 & \cellcolor{c7}7 & \cellcolor{c4}4 & \cellcolor{c4}4 & \cellcolor{c3}3 & \cellcolor{c8}8 & \cellcolor{c8}8 & \cellcolor{c8}8 & \cellcolor{c6}6 & \cellcolor{c3}3\\
$\mathcal{P}^{8}$ & \cellcolor{c7}7 & \cellcolor{c7}7 & \cellcolor{c8}8 & \cellcolor{c8}8 & \cellcolor{c8}8 & \cellcolor{c7}7 & \cellcolor{c1}1 & \cellcolor{c6}6 & \cellcolor{c4}4 & \cellcolor{c4}4 & \cellcolor{c8}8 & \cellcolor{c8}8 & \cellcolor{c8}8 & \cellcolor{c8}8 & \cellcolor{c7}7 & \cellcolor{c8}8\\
$\mathcal{P}^{9}$ & \cellcolor{c1}1 & \cellcolor{c1}1 & \cellcolor{c3}3 & \cellcolor{c3}3 & \cellcolor{c6}6 & \cellcolor{c5}5 & \cellcolor{c1}1 & \cellcolor{c2}2 & \cellcolor{c3}3 & \cellcolor{c3}3 & \cellcolor{c1}1 & \cellcolor{c3}3 & \cellcolor{c3}3 & \cellcolor{c1}1 & \cellcolor{c5}5 & \cellcolor{c6}6\\
$\mathcal{P}^{10}$ & \cellcolor{c3}3 & \cellcolor{c5}5 & \cellcolor{c2}2 & \cellcolor{c2}2 & \cellcolor{c1}1 & \cellcolor{c1}1 & \cellcolor{c1}1 & \cellcolor{c9}9 & \cellcolor{c2}2 & \cellcolor{c2}2 & \cellcolor{c6}6 & \cellcolor{c2}2 & \cellcolor{c2}2 & \cellcolor{c5}5 & \cellcolor{c1}1 & \cellcolor{c1}1\\
\hline
\end{tabularx}
}
}
\end{table*}

 \begin{table*}[t]
   \setlength{\tabcolsep}{2pt} 
  \centering
  \caption{\small Rankings of the 10 synthetic point sets on the convDTLZ2 problem by the 16 quality indicators when using $\mathbf{z}^{0.5}$, $\mathbf{z}^{-0.1}$, and $\mathbf{z}^{2}$.}
  \label{suptab:rankings_psets_convdtlz2}
        {\scriptsize
\subfloat[$\mathbf{z}^{0.5}=(0.5, 0.5)^{\top}$]{
\begin{tabularx}{72em}{|A|C|C|C|C|C|C|C|C|C|C|C|C|C|C|C|C|}
\hline
\rowcolor[rgb]{0.9, 0.9, 0.9} & MASF & MED & IGD-C & IGD-A & IGD-P & HV$_{\mathbf{z}}$ & PR & PMOD & IGD-CF & HV-CF & PMDA & R-IGD & R-HV & EH & HV & IGD \\
\hline
$\mathcal{P}^{1}$ & \cellcolor{c9}9 & \cellcolor{c9}9 & \cellcolor{c9}9 & \cellcolor{c9}9 & \cellcolor{c9}9 & \cellcolor{c6}6 & \cellcolor{c6}6 & \cellcolor{c8}8 & \cellcolor{c4}4 & \cellcolor{c4}4 & \cellcolor{c9}9 & \cellcolor{c6}6 & \cellcolor{c6}6 & \cellcolor{c7}7 & \cellcolor{c6}6 & \cellcolor{c9}9\\
$\mathcal{P}^{2}$ & \cellcolor{c5}5 & \cellcolor{c7}7 & \cellcolor{c6}6 & \cellcolor{c5}5 & \cellcolor{c4}4 & \cellcolor{c4}4 & \cellcolor{c5}5 & \cellcolor{c1}1 & \cellcolor{c4}4 & \cellcolor{c4}4 & \cellcolor{c4}4 & \cellcolor{c4}4 & \cellcolor{c4}4 & \cellcolor{c4}4 & \cellcolor{c3}3 & \cellcolor{c5}5\\
$\mathcal{P}^{3}$ & \cellcolor{c2}2 & \cellcolor{c3}3 & \cellcolor{c1}1 & \cellcolor{c1}1 & \cellcolor{c2}2 & \cellcolor{c1}1 & \cellcolor{c1}1 & \cellcolor{c4}4 & \cellcolor{c1}1 & \cellcolor{c1}1 & \cellcolor{c2}2 & \cellcolor{c1}1 & \cellcolor{c2}2 & \cellcolor{c2}2 & \cellcolor{c2}2 & \cellcolor{c2}2\\
$\mathcal{P}^{4}$ & \cellcolor{c6}6 & \cellcolor{c6}6 & \cellcolor{c5}5 & \cellcolor{c6}6 & \cellcolor{c6}6 & \cellcolor{c6}6 & \cellcolor{c6}6 & \cellcolor{c5}5 & \cellcolor{c4}4 & \cellcolor{c4}4 & \cellcolor{c5}5 & \cellcolor{c5}5 & \cellcolor{c5}5 & \cellcolor{c3}3 & \cellcolor{c4}4 & \cellcolor{c3}3\\
$\mathcal{P}^{5}$ & \cellcolor{c10}10 & \cellcolor{c10}10 & \cellcolor{c10}10 & \cellcolor{c10}10 & \cellcolor{c10}10 & \cellcolor{c6}6 & \cellcolor{c6}6 & \cellcolor{c9}9 & \cellcolor{c4}4 & \cellcolor{c4}4 & \cellcolor{c10}10 & \cellcolor{c7}7 & \cellcolor{c7}7 & \cellcolor{c6}6 & \cellcolor{c10}10 & \cellcolor{c10}10\\
$\mathcal{P}^{6}$ & \cellcolor{c7}7 & \cellcolor{c4}4 & \cellcolor{c7}7 & \cellcolor{c7}7 & \cellcolor{c7}7 & \cellcolor{c6}6 & \cellcolor{c6}6 & \cellcolor{c3}3 & \cellcolor{c4}4 & \cellcolor{c4}4 & \cellcolor{c6}6 & \cellcolor{c8}8 & \cellcolor{c8}8 & \cellcolor{c8}8 & \cellcolor{c8}8 & \cellcolor{c7}7\\
$\mathcal{P}^{7}$ & \cellcolor{c4}4 & \cellcolor{c1}1 & \cellcolor{c4}4 & \cellcolor{c4}4 & \cellcolor{c5}5 & \cellcolor{c5}5 & \cellcolor{c3}3 & \cellcolor{c6}6 & \cellcolor{c4}4 & \cellcolor{c4}4 & \cellcolor{c3}3 & \cellcolor{c8}8 & \cellcolor{c8}8 & \cellcolor{c8}8 & \cellcolor{c7}7 & \cellcolor{c4}4\\
$\mathcal{P}^{8}$ & \cellcolor{c8}8 & \cellcolor{c5}5 & \cellcolor{c8}8 & \cellcolor{c8}8 & \cellcolor{c8}8 & \cellcolor{c6}6 & \cellcolor{c6}6 & \cellcolor{c7}7 & \cellcolor{c4}4 & \cellcolor{c4}4 & \cellcolor{c8}8 & \cellcolor{c8}8 & \cellcolor{c8}8 & \cellcolor{c8}8 & \cellcolor{c9}9 & \cellcolor{c8}8\\
$\mathcal{P}^{9}$ & \cellcolor{c1}1 & \cellcolor{c2}2 & \cellcolor{c3}3 & \cellcolor{c3}3 & \cellcolor{c3}3 & \cellcolor{c3}3 & \cellcolor{c1}1 & \cellcolor{c2}2 & \cellcolor{c3}3 & \cellcolor{c3}3 & \cellcolor{c1}1 & \cellcolor{c3}3 & \cellcolor{c3}3 & \cellcolor{c1}1 & \cellcolor{c5}5 & \cellcolor{c6}6\\
$\mathcal{P}^{10}$ & \cellcolor{c3}3 & \cellcolor{c8}8 & \cellcolor{c2}2 & \cellcolor{c2}2 & \cellcolor{c1}1 & \cellcolor{c2}2 & \cellcolor{c4}4 & \cellcolor{c10}10 & \cellcolor{c2}2 & \cellcolor{c2}2 & \cellcolor{c7}7 & \cellcolor{c2}2 & \cellcolor{c1}1 & \cellcolor{c5}5 & \cellcolor{c1}1 & \cellcolor{c1}1\\
\hline
\end{tabularx}
}
\\
\subfloat[$\mathbf{z}^{-0.1}=(-0.1, -0.1)^{\top}$]{
\begin{tabularx}{72em}{|A|C|C|C|C|C|C|C|C|C|C|C|C|C|C|C|C|}
\hline
\rowcolor[rgb]{0.9, 0.9, 0.9} & MASF & MED & IGD-C & IGD-A & IGD-P & HV$_{\mathbf{z}}$ & PR & PMOD & IGD-CF & HV-CF & PMDA & R-IGD & R-HV & EH & HV & IGD \\
\hline
$\mathcal{P}^{1}$ & \cellcolor{c9}9 & \cellcolor{c8}8 & \cellcolor{c9}9 & \cellcolor{c9}9 & \cellcolor{c9}9 & \cellcolor{c7}7 & \cellcolor{c1}1 & \cellcolor{c8}8 & \cellcolor{c4}4 & \cellcolor{c4}4 & \cellcolor{c9}9 & \cellcolor{c6}6 & \cellcolor{c6}6 & \cellcolor{c6}6 & \cellcolor{c6}6 & \cellcolor{c9}9\\
$\mathcal{P}^{2}$ & \cellcolor{c5}5 & \cellcolor{c3}3 & \cellcolor{c4}4 & \cellcolor{c5}5 & \cellcolor{c5}5 & \cellcolor{c3}3 & \cellcolor{c1}1 & \cellcolor{c1}1 & \cellcolor{c4}4 & \cellcolor{c4}4 & \cellcolor{c4}4 & \cellcolor{c4}4 & \cellcolor{c4}4 & \cellcolor{c3}3 & \cellcolor{c3}3 & \cellcolor{c5}5\\
$\mathcal{P}^{3}$ & \cellcolor{c2}2 & \cellcolor{c2}2 & \cellcolor{c2}2 & \cellcolor{c1}1 & \cellcolor{c2}2 & \cellcolor{c2}2 & \cellcolor{c1}1 & \cellcolor{c4}4 & \cellcolor{c2}2 & \cellcolor{c2}2 & \cellcolor{c2}2 & \cellcolor{c1}1 & \cellcolor{c2}2 & \cellcolor{c2}2 & \cellcolor{c2}2 & \cellcolor{c2}2\\
$\mathcal{P}^{4}$ & \cellcolor{c6}6 & \cellcolor{c4}4 & \cellcolor{c7}7 & \cellcolor{c6}6 & \cellcolor{c3}3 & \cellcolor{c5}5 & \cellcolor{c1}1 & \cellcolor{c5}5 & \cellcolor{c4}4 & \cellcolor{c4}4 & \cellcolor{c5}5 & \cellcolor{c5}5 & \cellcolor{c5}5 & \cellcolor{c4}4 & \cellcolor{c4}4 & \cellcolor{c3}3\\
$\mathcal{P}^{5}$ & \cellcolor{c10}10 & \cellcolor{c10}10 & \cellcolor{c10}10 & \cellcolor{c10}10 & \cellcolor{c10}10 & \cellcolor{c10}10 & \cellcolor{c1}1 & \cellcolor{c9}9 & \cellcolor{c4}4 & \cellcolor{c4}4 & \cellcolor{c10}10 & \cellcolor{c7}7 & \cellcolor{c7}7 & \cellcolor{c7}7 & \cellcolor{c10}10 & \cellcolor{c10}10\\
$\mathcal{P}^{6}$ & \cellcolor{c7}7 & \cellcolor{c7}7 & \cellcolor{c6}6 & \cellcolor{c7}7 & \cellcolor{c7}7 & \cellcolor{c8}8 & \cellcolor{c1}1 & \cellcolor{c3}3 & \cellcolor{c4}4 & \cellcolor{c4}4 & \cellcolor{c6}6 & \cellcolor{c8}8 & \cellcolor{c8}8 & \cellcolor{c8}8 & \cellcolor{c8}8 & \cellcolor{c7}7\\
$\mathcal{P}^{7}$ & \cellcolor{c4}4 & \cellcolor{c6}6 & \cellcolor{c5}5 & \cellcolor{c4}4 & \cellcolor{c4}4 & \cellcolor{c6}6 & \cellcolor{c1}1 & \cellcolor{c6}6 & \cellcolor{c4}4 & \cellcolor{c4}4 & \cellcolor{c3}3 & \cellcolor{c8}8 & \cellcolor{c8}8 & \cellcolor{c8}8 & \cellcolor{c7}7 & \cellcolor{c4}4\\
$\mathcal{P}^{8}$ & \cellcolor{c8}8 & \cellcolor{c9}9 & \cellcolor{c8}8 & \cellcolor{c8}8 & \cellcolor{c8}8 & \cellcolor{c9}9 & \cellcolor{c1}1 & \cellcolor{c7}7 & \cellcolor{c4}4 & \cellcolor{c4}4 & \cellcolor{c8}8 & \cellcolor{c8}8 & \cellcolor{c8}8 & \cellcolor{c8}8 & \cellcolor{c9}9 & \cellcolor{c8}8\\
$\mathcal{P}^{9}$ & \cellcolor{c1}1 & \cellcolor{c1}1 & \cellcolor{c3}3 & \cellcolor{c3}3 & \cellcolor{c6}6 & \cellcolor{c4}4 & \cellcolor{c1}1 & \cellcolor{c2}2 & \cellcolor{c3}3 & \cellcolor{c3}3 & \cellcolor{c1}1 & \cellcolor{c3}3 & \cellcolor{c3}3 & \cellcolor{c1}1 & \cellcolor{c5}5 & \cellcolor{c6}6\\
$\mathcal{P}^{10}$ & \cellcolor{c3}3 & \cellcolor{c5}5 & \cellcolor{c1}1 & \cellcolor{c2}2 & \cellcolor{c1}1 & \cellcolor{c1}1 & \cellcolor{c1}1 & \cellcolor{c10}10 & \cellcolor{c1}1 & \cellcolor{c1}1 & \cellcolor{c7}7 & \cellcolor{c2}2 & \cellcolor{c1}1 & \cellcolor{c5}5 & \cellcolor{c1}1 & \cellcolor{c1}1\\
\hline
\end{tabularx}
}
\\
\subfloat[$\mathbf{z}^{2.0}=(2.0, 2.0)^{\top}$]{
\begin{tabularx}{72em}{|A|C|C|C|C|C|C|C|C|C|C|C|C|C|C|C|C|}
\hline
\rowcolor[rgb]{0.9, 0.9, 0.9} & MASF & MED & IGD-C & IGD-A & IGD-P & HV$_{\mathbf{z}}$ & PR & PMOD & IGD-CF & HV-CF & PMDA & R-IGD & R-HV & EH & HV & IGD \\
\hline
$\mathcal{P}^{1}$ & \cellcolor{c9}9 & \cellcolor{c9}9 & \cellcolor{c1}1 & \cellcolor{c9}9 & \cellcolor{c9}9 & \cellcolor{c6}6 & \cellcolor{c1}1 & \cellcolor{c8}8 & \cellcolor{c1}1 & \cellcolor{c1}1 & \cellcolor{c9}9 & \cellcolor{c6}6 & \cellcolor{c6}6 & \cellcolor{c7}7 & \cellcolor{c6}6 & \cellcolor{c9}9\\
$\mathcal{P}^{2}$ & \cellcolor{c5}5 & \cellcolor{c10}10 & \cellcolor{c4}4 & \cellcolor{c5}5 & \cellcolor{c5}5 & \cellcolor{c3}3 & \cellcolor{c1}1 & \cellcolor{c1}1 & \cellcolor{c3}3 & \cellcolor{c3}3 & \cellcolor{c4}4 & \cellcolor{c4}4 & \cellcolor{c4}4 & \cellcolor{c4}4 & \cellcolor{c3}3 & \cellcolor{c5}5\\
$\mathcal{P}^{3}$ & \cellcolor{c2}2 & \cellcolor{c7}7 & \cellcolor{c6}6 & \cellcolor{c1}1 & \cellcolor{c2}2 & \cellcolor{c2}2 & \cellcolor{c1}1 & \cellcolor{c4}4 & \cellcolor{c3}3 & \cellcolor{c3}3 & \cellcolor{c2}2 & \cellcolor{c1}1 & \cellcolor{c2}2 & \cellcolor{c2}2 & \cellcolor{c2}2 & \cellcolor{c2}2\\
$\mathcal{P}^{4}$ & \cellcolor{c6}6 & \cellcolor{c5}5 & \cellcolor{c9}9 & \cellcolor{c6}6 & \cellcolor{c3}3 & \cellcolor{c4}4 & \cellcolor{c1}1 & \cellcolor{c5}5 & \cellcolor{c3}3 & \cellcolor{c3}3 & \cellcolor{c5}5 & \cellcolor{c5}5 & \cellcolor{c5}5 & \cellcolor{c3}3 & \cellcolor{c4}4 & \cellcolor{c3}3\\
$\mathcal{P}^{5}$ & \cellcolor{c10}10 & \cellcolor{c4}4 & \cellcolor{c10}10 & \cellcolor{c10}10 & \cellcolor{c10}10 & \cellcolor{c9}9 & \cellcolor{c1}1 & \cellcolor{c9}9 & \cellcolor{c3}3 & \cellcolor{c3}3 & \cellcolor{c10}10 & \cellcolor{c7}7 & \cellcolor{c7}7 & \cellcolor{c6}6 & \cellcolor{c10}10 & \cellcolor{c10}10\\
$\mathcal{P}^{6}$ & \cellcolor{c7}7 & \cellcolor{c3}3 & \cellcolor{c3}3 & \cellcolor{c7}7 & \cellcolor{c7}7 & \cellcolor{c8}8 & \cellcolor{c1}1 & \cellcolor{c3}3 & \cellcolor{c3}3 & \cellcolor{c3}3 & \cellcolor{c6}6 & \cellcolor{c8}8 & \cellcolor{c8}8 & \cellcolor{c8}8 & \cellcolor{c8}8 & \cellcolor{c7}7\\
$\mathcal{P}^{7}$ & \cellcolor{c4}4 & \cellcolor{c2}2 & \cellcolor{c5}5 & \cellcolor{c4}4 & \cellcolor{c4}4 & \cellcolor{c7}7 & \cellcolor{c1}1 & \cellcolor{c6}6 & \cellcolor{c3}3 & \cellcolor{c3}3 & \cellcolor{c3}3 & \cellcolor{c8}8 & \cellcolor{c8}8 & \cellcolor{c8}8 & \cellcolor{c7}7 & \cellcolor{c4}4\\
$\mathcal{P}^{8}$ & \cellcolor{c8}8 & \cellcolor{c1}1 & \cellcolor{c8}8 & \cellcolor{c8}8 & \cellcolor{c8}8 & \cellcolor{c10}10 & \cellcolor{c1}1 & \cellcolor{c7}7 & \cellcolor{c3}3 & \cellcolor{c3}3 & \cellcolor{c8}8 & \cellcolor{c8}8 & \cellcolor{c8}8 & \cellcolor{c8}8 & \cellcolor{c9}9 & \cellcolor{c8}8\\
$\mathcal{P}^{9}$ & \cellcolor{c1}1 & \cellcolor{c8}8 & \cellcolor{c7}7 & \cellcolor{c3}3 & \cellcolor{c6}6 & \cellcolor{c5}5 & \cellcolor{c1}1 & \cellcolor{c2}2 & \cellcolor{c3}3 & \cellcolor{c3}3 & \cellcolor{c1}1 & \cellcolor{c3}3 & \cellcolor{c3}3 & \cellcolor{c1}1 & \cellcolor{c5}5 & \cellcolor{c6}6\\
$\mathcal{P}^{10}$ & \cellcolor{c3}3 & \cellcolor{c6}6 & \cellcolor{c2}2 & \cellcolor{c2}2 & \cellcolor{c1}1 & \cellcolor{c1}1 & \cellcolor{c1}1 & \cellcolor{c10}10 & \cellcolor{c2}2 & \cellcolor{c2}2 & \cellcolor{c7}7 & \cellcolor{c2}2 & \cellcolor{c1}1 & \cellcolor{c5}5 & \cellcolor{c1}1 & \cellcolor{c1}1\\
\hline
\end{tabularx}
}
}
\end{table*}

\newcolumntype{A}{>{\columncolor[rgb]{0.9, 0.9, 0.9}\centering}p{5em}}

\begin{table*}[t]
 \setlength{\tabcolsep}{2pt} 
  \centering
  \caption{\small 
  Rankings of the six EMO algorithms on the DTLZ1 problem by the 16 quality indicators when using $\mathbf{z}^{0.51}=(0.51, 0.51)^{\top}$ and $\mathbf{z}^{-0.1}=(-0.1, -0.1)^{\top}$. The quality indicators based on the ROI-P (IGD-P, HV$_{\mathbf{z}}$, PR) do not work when the reference point is on the PF. For this reason, we used $\mathbf{z}^{0.51}=(0.51, 0.51)^{\top}$, instead of $\mathbf{z}^{0.5}=(0.5, 0.5)^{\top}$. \lq\lq NUMS\rq\rq stands for MOEA/D-NUMS.}
  \label{suptab:emo_ranks_dtlz1}
{\scriptsize
\subfloat[$\mathbf{z}=(0.51, 0.51)^{\top}$]{
\begin{tabularx}{72em}{|A|C|C|C|C|C|C|C|C|C|C|C|C|C|C|C|C|}
\hline
\rowcolor[rgb]{0.9, 0.9, 0.9} & MASF & MED & IGD-C & IGD-A & IGD-P & HV$_{\mathbf{z}}$ & PR & PMOD & IGD-CF & HV-CF & PMDA & R-IGD & R-HV & EH & HV & IGD \\
\hline
R-NSGA-II & \cellcolor{c3}3 & \cellcolor{c2}2 & \cellcolor{c3}3 & \cellcolor{c3}3 & \cellcolor{c1}1 & \cellcolor{c1}1 & \cellcolor{c4}4 & \cellcolor{c2}2 & \cellcolor{c2}2 & \cellcolor{c3}3 & \cellcolor{c2}2 & \cellcolor{c5}5 & \cellcolor{c4}4 & \cellcolor{c2}2 & \cellcolor{c4}4 & \cellcolor{c4}4\\
r-NSGA-II & \cellcolor{c4}4 & \cellcolor{c4}4 & \cellcolor{c1}1 & \cellcolor{c1}1 & \cellcolor{c3}3 & \cellcolor{c5}5 & \cellcolor{c6}6 & \cellcolor{c4}4 & \cellcolor{c1}1 & \cellcolor{c1}1 & \cellcolor{c4}4 & \cellcolor{c1}1 & \cellcolor{c1}1 & \cellcolor{c6}6 & \cellcolor{c1}1 & \cellcolor{c1}1\\
g-NSGA-II & \cellcolor{c6}6 & \cellcolor{c6}6 & \cellcolor{c6}6 & \cellcolor{c6}6 & \cellcolor{c6}6 & \cellcolor{c3}3 & \cellcolor{c2}2 & \cellcolor{c5}5 & \cellcolor{c6}6 & \cellcolor{c6}6 & \cellcolor{c6}6 & \cellcolor{c6}6 & \cellcolor{c6}6 & \cellcolor{c5}5 & \cellcolor{c6}6 & \cellcolor{c5}5\\
PBEA & \cellcolor{c2}2 & \cellcolor{c3}3 & \cellcolor{c4}4 & \cellcolor{c4}4 & \cellcolor{c4}4 & \cellcolor{c4}4 & \cellcolor{c3}3 & \cellcolor{c3}3 & \cellcolor{c3}3 & \cellcolor{c4}4 & \cellcolor{c3}3 & \cellcolor{c3}3 & \cellcolor{c5}5 & \cellcolor{c4}4 & \cellcolor{c3}3 & \cellcolor{c3}3\\
R-MEAD2 & \cellcolor{c5}5 & \cellcolor{c5}5 & \cellcolor{c2}2 & \cellcolor{c2}2 & \cellcolor{c5}5 & \cellcolor{c6}6 & \cellcolor{c5}5 & \cellcolor{c6}6 & \cellcolor{c5}5 & \cellcolor{c2}2 & \cellcolor{c5}5 & \cellcolor{c3}3 & \cellcolor{c3}3 & \cellcolor{c3}3 & \cellcolor{c2}2 & \cellcolor{c2}2\\
NUMS & \cellcolor{c1}1 & \cellcolor{c1}1 & \cellcolor{c5}5 & \cellcolor{c5}5 & \cellcolor{c2}2 & \cellcolor{c2}2 & \cellcolor{c1}1 & \cellcolor{c1}1 & \cellcolor{c4}4 & \cellcolor{c5}5 & \cellcolor{c1}1 & \cellcolor{c2}2 & \cellcolor{c2}2 & \cellcolor{c1}1 & \cellcolor{c5}5 & \cellcolor{c6}6\\
\hline
\end{tabularx}
}
\\
\subfloat[$\mathbf{z}=(-0.1, -0.1)^{\top}$]{
\begin{tabularx}{72em}{|A|C|C|C|C|C|C|C|C|C|C|C|C|C|C|C|C|}
\hline
\rowcolor[rgb]{0.9, 0.9, 0.9} & MASF & MED & IGD-C & IGD-A & IGD-P & HV$_{\mathbf{z}}$ & PR & PMOD & IGD-CF & HV-CF & PMDA & R-IGD & R-HV & EH & HV & IGD \\
\hline
R-NSGA-II & \cellcolor{c3}3 & \cellcolor{c1}1 & \cellcolor{c5}5 & \cellcolor{c5}5 & \cellcolor{c6}6 & \cellcolor{c6}6 & \cellcolor{c1}1 & \cellcolor{c1}1 & \cellcolor{c5}5 & \cellcolor{c5}5 & \cellcolor{c1}1 & \cellcolor{c4}4 & \cellcolor{c4}4 & \cellcolor{c2}2 & \cellcolor{c6}6 & \cellcolor{c6}6\\
r-NSGA-II & \cellcolor{c5}5 & \cellcolor{c6}6 & \cellcolor{c2}2 & \cellcolor{c2}2 & \cellcolor{c1}1 & \cellcolor{c1}1 & \cellcolor{c1}1 & \cellcolor{c3}3 & \cellcolor{c2}2 & \cellcolor{c2}2 & \cellcolor{c6}6 & \cellcolor{c2}2 & \cellcolor{c2}2 & \cellcolor{c5}5 & \cellcolor{c1}1 & \cellcolor{c1}1\\
g-NSGA-II & \cellcolor{c4}4 & \cellcolor{c5}5 & \cellcolor{c3}3 & \cellcolor{c3}3 & \cellcolor{c2}2 & \cellcolor{c2}2 & \cellcolor{c1}1 & \cellcolor{c4}4 & \cellcolor{c3}3 & \cellcolor{c3}3 & \cellcolor{c5}5 & \cellcolor{c3}3 & \cellcolor{c3}3 & \cellcolor{c6}6 & \cellcolor{c2}2 & \cellcolor{c2}2\\
PBEA & \cellcolor{c2}2 & \cellcolor{c3}3 & \cellcolor{c6}6 & \cellcolor{c6}6 & \cellcolor{c5}5 & \cellcolor{c5}5 & \cellcolor{c1}1 & \cellcolor{c5}5 & \cellcolor{c6}6 & \cellcolor{c6}6 & \cellcolor{c3}3 & \cellcolor{c4}4 & \cellcolor{c5}5 & \cellcolor{c1}1 & \cellcolor{c5}5 & \cellcolor{c5}5\\
R-MEAD2 & \cellcolor{c6}6 & \cellcolor{c4}4 & \cellcolor{c4}4 & \cellcolor{c4}4 & \cellcolor{c3}3 & \cellcolor{c3}3 & \cellcolor{c1}1 & \cellcolor{c6}6 & \cellcolor{c4}4 & \cellcolor{c4}4 & \cellcolor{c4}4 & \cellcolor{c6}6 & \cellcolor{c6}6 & \cellcolor{c3}3 & \cellcolor{c3}3 & \cellcolor{c3}3\\
NUMS & \cellcolor{c1}1 & \cellcolor{c2}2 & \cellcolor{c1}1 & \cellcolor{c1}1 & \cellcolor{c4}4 & \cellcolor{c4}4 & \cellcolor{c1}1 & \cellcolor{c2}2 & \cellcolor{c1}1 & \cellcolor{c1}1 & \cellcolor{c2}2 & \cellcolor{c1}1 & \cellcolor{c1}1 & \cellcolor{c4}4 & \cellcolor{c4}4 & \cellcolor{c4}4\\
\hline
\end{tabularx}
}
}
\end{table*}

\begin{table*}[t]
 \setlength{\tabcolsep}{2pt} 
  \centering
  \caption{\small 
  Rankings of the six EMO algorithms on the DTLZ2 problem by the 16 quality indicators when using $\mathbf{z}^{0.5}=(0.5, 0.5)^{\top}$ and $\mathbf{z}^{-0.1}=(-0.1, -0.1)^{\top}$. \lq\lq NUMS\rq\rq stands for MOEA/D-NUMS.}
  \label{suptab:emo_ranks_dtlz2}
{\scriptsize
\subfloat[$\mathbf{z}=(0.5, 0.5)^{\top}$]{
\begin{tabularx}{72em}{|A|C|C|C|C|C|C|C|C|C|C|C|C|C|C|C|C|}
\hline
\rowcolor[rgb]{0.9, 0.9, 0.9} & MASF & MED & IGD-C & IGD-A & IGD-P & HV$_{\mathbf{z}}$ & PR & PMOD & IGD-CF & HV-CF & PMDA & R-IGD & R-HV & EH & HV & IGD \\
\hline
R-NSGA-II & \cellcolor{c3}3 & \cellcolor{c2}2 & \cellcolor{c6}6 & \cellcolor{c6}6 & \cellcolor{c5}5 & \cellcolor{c6}6 & \cellcolor{c4}4 & \cellcolor{c4}4 & \cellcolor{c6}6 & \cellcolor{c6}6 & \cellcolor{c3}3 & \cellcolor{c5}5 & \cellcolor{c5}5 & \cellcolor{c1}1 & \cellcolor{c5}5 & \cellcolor{c5}5\\
r-NSGA-II & \cellcolor{c4}4 & \cellcolor{c3}3 & \cellcolor{c3}3 & \cellcolor{c3}3 & \cellcolor{c4}4 & \cellcolor{c4}4 & \cellcolor{c1}1 & \cellcolor{c3}3 & \cellcolor{c3}3 & \cellcolor{c4}4 & \cellcolor{c2}2 & \cellcolor{c3}3 & \cellcolor{c3}3 & \cellcolor{c3}3 & \cellcolor{c4}4 & \cellcolor{c4}4\\
g-NSGA-II & \cellcolor{c5}5 & \cellcolor{c5}5 & \cellcolor{c1}1 & \cellcolor{c1}1 & \cellcolor{c1}1 & \cellcolor{c1}1 & \cellcolor{c1}1 & \cellcolor{c1}1 & \cellcolor{c1}1 & \cellcolor{c1}1 & \cellcolor{c5}5 & \cellcolor{c2}2 & \cellcolor{c1}1 & \cellcolor{c6}6 & \cellcolor{c2}2 & \cellcolor{c2}2\\
PBEA & \cellcolor{c1}1 & \cellcolor{c6}6 & \cellcolor{c2}2 & \cellcolor{c2}2 & \cellcolor{c2}2 & \cellcolor{c2}2 & \cellcolor{c6}6 & \cellcolor{c5}5 & \cellcolor{c2}2 & \cellcolor{c2}2 & \cellcolor{c6}6 & \cellcolor{c1}1 & \cellcolor{c2}2 & \cellcolor{c5}5 & \cellcolor{c1}1 & \cellcolor{c1}1\\
R-MEAD2 & \cellcolor{c6}6 & \cellcolor{c4}4 & \cellcolor{c5}5 & \cellcolor{c5}5 & \cellcolor{c3}3 & \cellcolor{c3}3 & \cellcolor{c5}5 & \cellcolor{c6}6 & \cellcolor{c4}4 & \cellcolor{c3}3 & \cellcolor{c4}4 & \cellcolor{c6}6 & \cellcolor{c6}6 & \cellcolor{c4}4 & \cellcolor{c3}3 & \cellcolor{c3}3\\
NUMS & \cellcolor{c2}2 & \cellcolor{c1}1 & \cellcolor{c4}4 & \cellcolor{c4}4 & \cellcolor{c6}6 & \cellcolor{c5}5 & \cellcolor{c1}1 & \cellcolor{c2}2 & \cellcolor{c5}5 & \cellcolor{c5}5 & \cellcolor{c1}1 & \cellcolor{c4}4 & \cellcolor{c4}4 & \cellcolor{c2}2 & \cellcolor{c6}6 & \cellcolor{c6}6\\
\hline
\end{tabularx}
}
\\
\subfloat[$\mathbf{z}=(-0.1, -0.1)^{\top}$]{
\begin{tabularx}{72em}{|A|C|C|C|C|C|C|C|C|C|C|C|C|C|C|C|C|}
\hline
\rowcolor[rgb]{0.9, 0.9, 0.9} & MASF & MED & IGD-C & IGD-A & IGD-P & HV$_{\mathbf{z}}$ & PR & PMOD & IGD-CF & HV-CF & PMDA & R-IGD & R-HV & EH & HV & IGD \\
\hline
R-NSGA-II & \cellcolor{c6}6 & \cellcolor{c2}2 & \cellcolor{c2}2 & \cellcolor{c6}6 & \cellcolor{c3}3 & \cellcolor{c5}5 & \cellcolor{c1}1 & \cellcolor{c4}4 & \cellcolor{c3}3 & \cellcolor{c2}2 & \cellcolor{c5}5 & \cellcolor{c5}5 & \cellcolor{c5}5 & \cellcolor{c6}6 & \cellcolor{c4}4 & \cellcolor{c3}3\\
r-NSGA-II & \cellcolor{c3}3 & \cellcolor{c6}6 & \cellcolor{c5}5 & \cellcolor{c4}4 & \cellcolor{c5}5 & \cellcolor{c4}4 & \cellcolor{c1}1 & \cellcolor{c3}3 & \cellcolor{c5}5 & \cellcolor{c5}5 & \cellcolor{c1}1 & \cellcolor{c4}4 & \cellcolor{c4}4 & \cellcolor{c1}1 & \cellcolor{c5}5 & \cellcolor{c5}5\\
g-NSGA-II & \cellcolor{c4}4 & \cellcolor{c4}4 & \cellcolor{c1}1 & \cellcolor{c3}3 & \cellcolor{c1}1 & \cellcolor{c1}1 & \cellcolor{c1}1 & \cellcolor{c2}2 & \cellcolor{c1}1 & \cellcolor{c1}1 & \cellcolor{c4}4 & \cellcolor{c3}3 & \cellcolor{c3}3 & \cellcolor{c5}5 & \cellcolor{c1}1 & \cellcolor{c1}1\\
PBEA & \cellcolor{c1}1 & \cellcolor{c3}3 & \cellcolor{c3}3 & \cellcolor{c2}2 & \cellcolor{c2}2 & \cellcolor{c2}2 & \cellcolor{c1}1 & \cellcolor{c6}6 & \cellcolor{c2}2 & \cellcolor{c3}3 & \cellcolor{c3}3 & \cellcolor{c2}2 & \cellcolor{c2}2 & \cellcolor{c3}3 & \cellcolor{c2}2 & \cellcolor{c2}2\\
R-MEAD2 & \cellcolor{c5}5 & \cellcolor{c1}1 & \cellcolor{c6}6 & \cellcolor{c5}5 & \cellcolor{c6}6 & \cellcolor{c6}6 & \cellcolor{c1}1 & \cellcolor{c5}5 & \cellcolor{c4}4 & \cellcolor{c4}4 & \cellcolor{c6}6 & \cellcolor{c6}6 & \cellcolor{c6}6 & \cellcolor{c4}4 & \cellcolor{c6}6 & \cellcolor{c6}6\\
NUMS & \cellcolor{c2}2 & \cellcolor{c5}5 & \cellcolor{c4}4 & \cellcolor{c1}1 & \cellcolor{c4}4 & \cellcolor{c3}3 & \cellcolor{c1}1 & \cellcolor{c1}1 & \cellcolor{c5}5 & \cellcolor{c5}5 & \cellcolor{c2}2 & \cellcolor{c1}1 & \cellcolor{c1}1 & \cellcolor{c2}2 & \cellcolor{c3}3 & \cellcolor{c4}4\\
\hline
\end{tabularx}
}
}
\end{table*}

\begin{table*}[t]
 \setlength{\tabcolsep}{2pt} 
  \centering
  \caption{\small 
  Rankings of the six EMO algorithms on the convDTLZ2 problem by the 16 quality indicators when using $\mathbf{z}^{0.5}=(0.5, 0.5)^{\top}$ and $\mathbf{z}^{2}=(2, 2)^{\top}$. \lq\lq NUMS\rq\rq stands for MOEA/D-NUMS.}
  \label{suptab:emo_ranks_convdtlz2}
{\scriptsize
\subfloat[$\mathbf{z}=(0.5, 0.5)^{\top}$]{
\begin{tabularx}{72em}{|A|C|C|C|C|C|C|C|C|C|C|C|C|C|C|C|C|}
\hline
\rowcolor[rgb]{0.9, 0.9, 0.9} & MASF & MED & IGD-C & IGD-A & IGD-P & HV$_{\mathbf{z}}$ & PR & PMOD & IGD-CF & HV-CF & PMDA & R-IGD & R-HV & EH & HV & IGD \\
\hline
R-NSGA-II & \cellcolor{c3}3 & \cellcolor{c2}2 & \cellcolor{c5}5 & \cellcolor{c5}5 & \cellcolor{c5}5 & \cellcolor{c5}5 & \cellcolor{c4}4 & \cellcolor{c4}4 & \cellcolor{c5}5 & \cellcolor{c5}5 & \cellcolor{c2}2 & \cellcolor{c4}4 & \cellcolor{c4}4 & \cellcolor{c2}2 & \cellcolor{c5}5 & \cellcolor{c5}5\\
r-NSGA-II & \cellcolor{c4}4 & \cellcolor{c3}3 & \cellcolor{c3}3 & \cellcolor{c3}3 & \cellcolor{c3}3 & \cellcolor{c3}3 & \cellcolor{c3}3 & \cellcolor{c3}3 & \cellcolor{c3}3 & \cellcolor{c3}3 & \cellcolor{c3}3 & \cellcolor{c3}3 & \cellcolor{c3}3 & \cellcolor{c3}3 & \cellcolor{c4}4 & \cellcolor{c4}4\\
g-NSGA-II & \cellcolor{c5}5 & \cellcolor{c6}6 & \cellcolor{c1}1 & \cellcolor{c1}1 & \cellcolor{c1}1 & \cellcolor{c1}1 & \cellcolor{c1}1 & \cellcolor{c2}2 & \cellcolor{c1}1 & \cellcolor{c1}1 & \cellcolor{c5}5 & \cellcolor{c2}2 & \cellcolor{c1}1 & \cellcolor{c6}6 & \cellcolor{c3}3 & \cellcolor{c3}3\\
PBEA & \cellcolor{c1}1 & \cellcolor{c5}5 & \cellcolor{c2}2 & \cellcolor{c2}2 & \cellcolor{c2}2 & \cellcolor{c2}2 & \cellcolor{c5}5 & \cellcolor{c5}5 & \cellcolor{c2}2 & \cellcolor{c2}2 & \cellcolor{c4}4 & \cellcolor{c1}1 & \cellcolor{c2}2 & \cellcolor{c4}4 & \cellcolor{c2}2 & \cellcolor{c2}2\\
R-MEAD2 & \cellcolor{c6}6 & \cellcolor{c4}4 & \cellcolor{c4}4 & \cellcolor{c4}4 & \cellcolor{c4}4 & \cellcolor{c4}4 & \cellcolor{c6}6 & \cellcolor{c6}6 & \cellcolor{c4}4 & \cellcolor{c4}4 & \cellcolor{c6}6 & \cellcolor{c6}6 & \cellcolor{c6}6 & \cellcolor{c5}5 & \cellcolor{c1}1 & \cellcolor{c1}1\\
NUMS & \cellcolor{c2}2 & \cellcolor{c1}1 & \cellcolor{c6}6 & \cellcolor{c6}6 & \cellcolor{c6}6 & \cellcolor{c6}6 & \cellcolor{c1}1 & \cellcolor{c1}1 & \cellcolor{c6}6 & \cellcolor{c6}6 & \cellcolor{c1}1 & \cellcolor{c5}5 & \cellcolor{c5}5 & \cellcolor{c1}1 & \cellcolor{c6}6 & \cellcolor{c6}6\\
\hline
\end{tabularx}
}
\\
\subfloat[$\mathbf{z}=(2.0, 2.0)^{\top}$]{
\begin{tabularx}{72em}{|A|C|C|C|C|C|C|C|C|C|C|C|C|C|C|C|C|}
\hline
\rowcolor[rgb]{0.9, 0.9, 0.9} & MASF & MED & IGD-C & IGD-A & IGD-P & HV$_{\mathbf{z}}$ & PR & PMOD & IGD-CF & HV-CF & PMDA & R-IGD & R-HV & EH & HV & IGD \\
\hline
R-NSGA-II & \cellcolor{c4}4 & \cellcolor{c3}3 & \cellcolor{c3}3 & \cellcolor{c4}4 & \cellcolor{c4}4 & \cellcolor{c3}3 & \cellcolor{c1}1 & \cellcolor{c5}5 & \cellcolor{c3}3 & \cellcolor{c3}3 & \cellcolor{c3}3 & \cellcolor{c4}4 & \cellcolor{c4}4 & \cellcolor{c4}4 & \cellcolor{c4}4 & \cellcolor{c4}4\\
r-NSGA-II & \cellcolor{c5}5 & \cellcolor{c1}1 & \cellcolor{c2}2 & \cellcolor{c5}5 & \cellcolor{c5}5 & \cellcolor{c5}5 & \cellcolor{c6}6 & \cellcolor{c6}6 & \cellcolor{c4}4 & \cellcolor{c4}4 & \cellcolor{c5}5 & \cellcolor{c6}6 & \cellcolor{c6}6 & \cellcolor{c6}6 & \cellcolor{c5}5 & \cellcolor{c5}5\\
g-NSGA-II & \cellcolor{c3}3 & \cellcolor{c4}4 & \cellcolor{c1}1 & \cellcolor{c3}3 & \cellcolor{c1}1 & \cellcolor{c1}1 & \cellcolor{c1}1 & \cellcolor{c1}1 & \cellcolor{c1}1 & \cellcolor{c1}1 & \cellcolor{c4}4 & \cellcolor{c3}3 & \cellcolor{c3}3 & \cellcolor{c5}5 & \cellcolor{c1}1 & \cellcolor{c1}1\\
PBEA & \cellcolor{c1}1 & \cellcolor{c6}6 & \cellcolor{c5}5 & \cellcolor{c2}2 & \cellcolor{c3}3 & \cellcolor{c4}4 & \cellcolor{c1}1 & \cellcolor{c2}2 & \cellcolor{c5}5 & \cellcolor{c5}5 & \cellcolor{c1}1 & \cellcolor{c2}2 & \cellcolor{c2}2 & \cellcolor{c2}2 & \cellcolor{c3}3 & \cellcolor{c3}3\\
R-MEAD2 & \cellcolor{c6}6 & \cellcolor{c2}2 & \cellcolor{c6}6 & \cellcolor{c6}6 & \cellcolor{c6}6 & \cellcolor{c6}6 & \cellcolor{c1}1 & \cellcolor{c4}4 & \cellcolor{c2}2 & \cellcolor{c2}2 & \cellcolor{c6}6 & \cellcolor{c5}5 & \cellcolor{c5}5 & \cellcolor{c1}1 & \cellcolor{c6}6 & \cellcolor{c6}6\\
NUMS & \cellcolor{c2}2 & \cellcolor{c5}5 & \cellcolor{c4}4 & \cellcolor{c1}1 & \cellcolor{c2}2 & \cellcolor{c2}2 & \cellcolor{c1}1 & \cellcolor{c3}3 & \cellcolor{c5}5 & \cellcolor{c5}5 & \cellcolor{c2}2 & \cellcolor{c1}1 & \cellcolor{c1}1 & \cellcolor{c3}3 & \cellcolor{c2}2 & \cellcolor{c2}2\\
\hline
\end{tabularx}
}
}
\end{table*}

\clearpage

\begin{figure*}[t]
  \centering
    \subfloat{\includegraphics[width=0.9\textwidth]{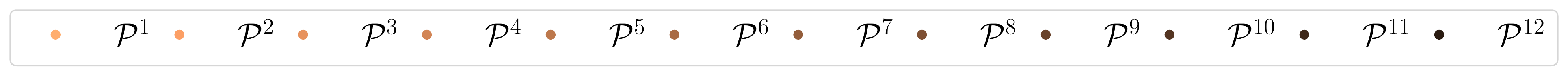}}
  \\
  \subfloat[DTLZ1 ($\mathcal{P}^1, \dots, \mathcal{P}^{12}$)]{  
    \includegraphics[width=0.31\textwidth]{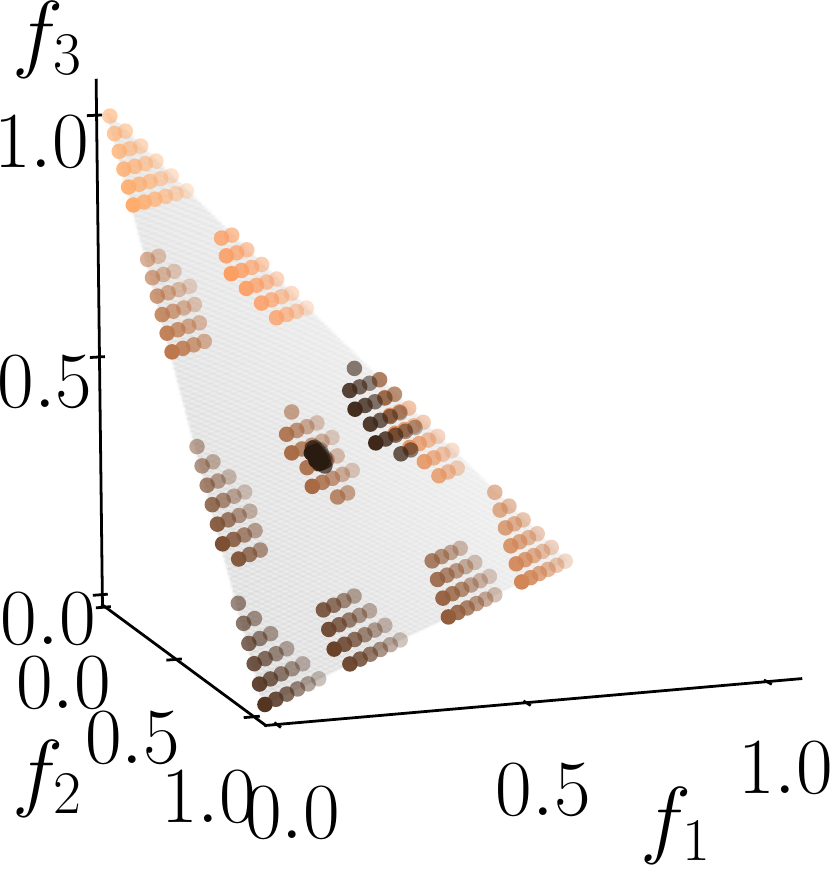}
  }  
  \subfloat[DTLZ2 ($\mathcal{P}^1, \dots, \mathcal{P}^{12}$)]{  
    \includegraphics[width=0.31\textwidth]{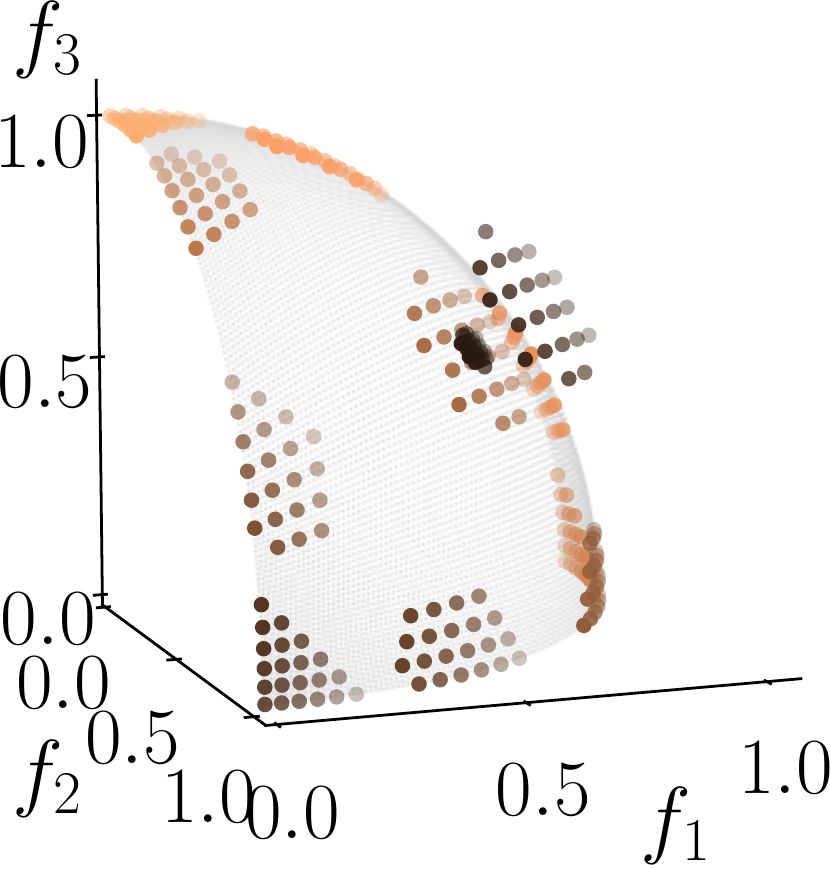}
  }  
  \subfloat[convDTLZ2 ($\mathcal{P}^1, \dots, \mathcal{P}^{12}$)]{  
    \includegraphics[width=0.31\textwidth]{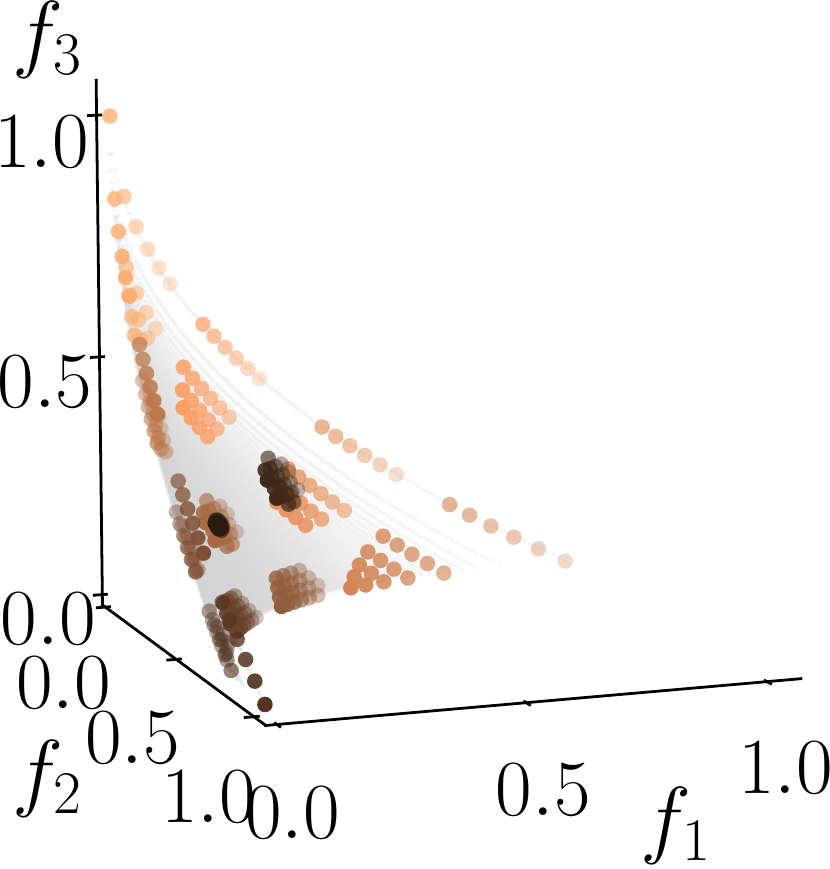}
  }
  \\
  \subfloat[DTLZ1 ($\mathcal{P}^{13}$)]{  
    \includegraphics[width=0.31\textwidth]{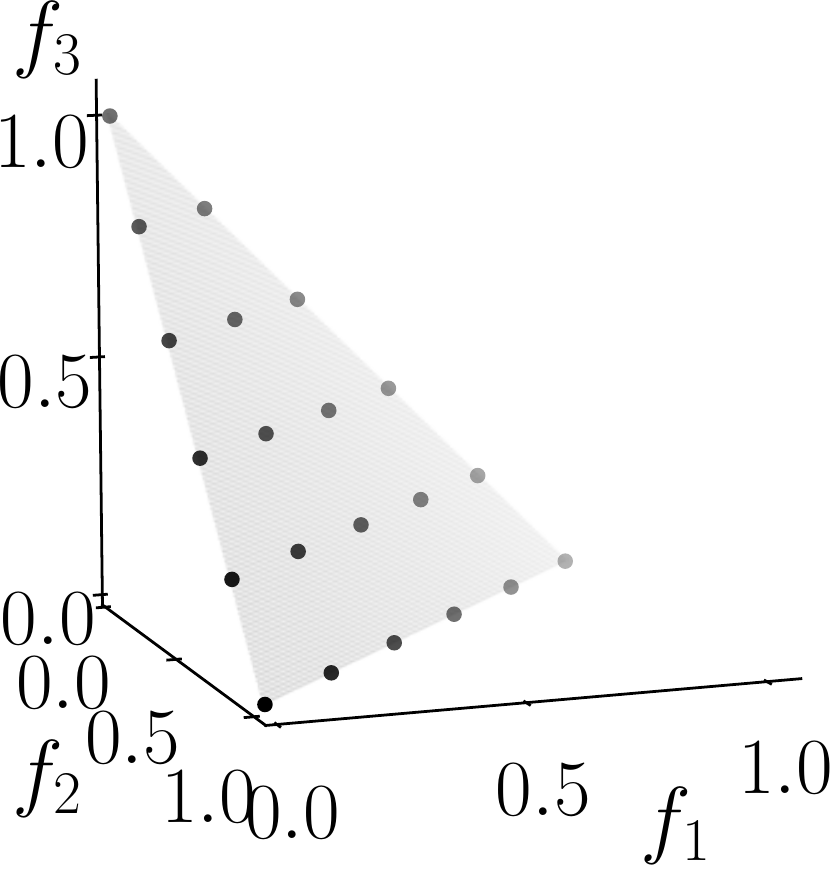}
  }
  \subfloat[DTLZ2 ($\mathcal{P}^{13}$)]{  
    \includegraphics[width=0.31\textwidth]{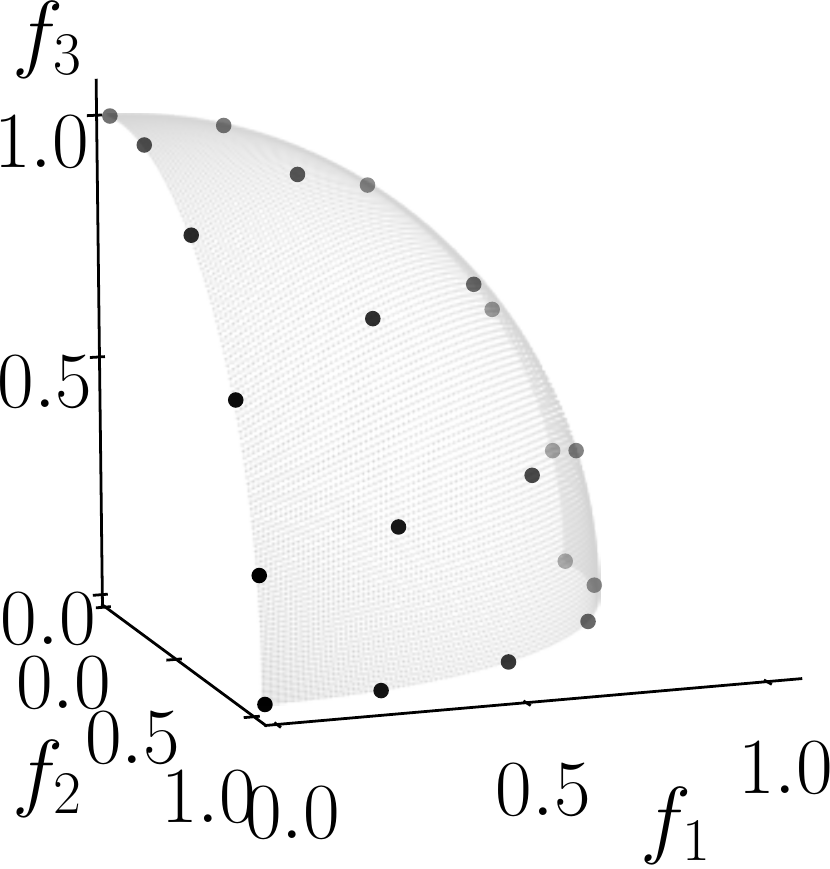}
  }
  \subfloat[convDTLZ2 ($\mathcal{P}^{13}$)]{  
    \includegraphics[width=0.31\textwidth]{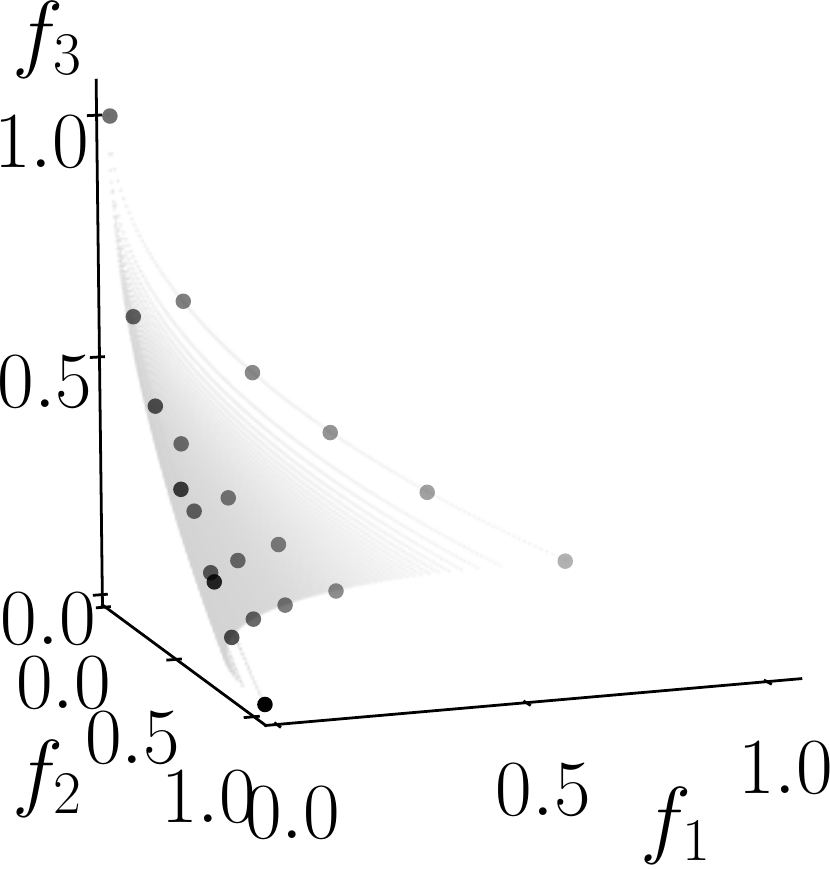}
  }
  \caption{Distributions of the 13 point sets on the PFs of the three-objective DTLZ1, DTLZ2, and convDTLZ2 problems, where the size of each point set is 21. $\mathcal{P}^6$ is on the center of the PF. $\mathcal{P}^1$, $\mathcal{P}^4$, and $\mathcal{P}^{10}$ are on the three extreme regions, respectively. $\mathcal{P}^2$, $\mathcal{P}^3$, $\mathcal{P}^5$, $\mathcal{P}^7$, $\mathcal{P}^8$, and $\mathcal{P}^9$, are on the edge of the PF. $\mathcal{P}^{11}$ is a shifted version of $\mathcal{P}^6$ by adding 0.1 to all elements. Thus, $\mathcal{P}^{11}$ is worse than $\mathcal{P}^6$ in terms of the convergence to the PF. $\mathcal{P}^{12}$ is worse than $\mathcal{P}^6$ in terms of the diversity. The 21 points in $\mathcal{P}^{13}$ are uniformly distributed on the PF.}
   \label{supfig:toy_3d}
\end{figure*}

\definecolor{c13_13}{RGB}{52, 100, 80}
\definecolor{c12_13}{RGB}{62, 110, 92}
\definecolor{c11_13}{RGB}{73, 126, 105}
\definecolor{c10_13}{RGB}{80, 136, 113}
\definecolor{c9_13}{RGB}{90, 151, 125}
\definecolor{c8_13}{RGB}{100, 166, 138}
\definecolor{c7_13}{RGB}{110, 180, 149}
\definecolor{c6_13}{RGB}{120, 194, 160}
\definecolor{c5_13}{RGB}{130, 209, 173}
\definecolor{c4_13}{RGB}{140, 224, 185}
\definecolor{c3_13}{RGB}{153, 242, 200}
\definecolor{c2_13}{RGB}{173, 248, 220}
\definecolor{c1_13}{RGB}{190, 255, 230}

 \begin{table*}[t]
   \setlength{\tabcolsep}{2pt} 
  \centering
  \caption{\small Rankings of the 13 synthetic point sets on the three-objective DTLZ1 problem by the 16 quality indicators when using $\mathbf{z}^{0.5}$ and $\mathbf{z}^{-0.1}$.}
  \label{suptab:rankings_psets_dtlz1_3d}
        {\scriptsize
\subfloat[$\mathbf{z}^{0.5}=(0.5, 0.5, 0.5)^{\top}$]{
\begin{tabularx}{72em}{|A|C|C|C|C|C|C|C|C|C|C|C|C|C|C|C|C|}
\hline
\rowcolor[rgb]{0.9, 0.9, 0.9} & MASF & MED & IGD-C & IGD-A & IGD-P & HV$_{\mathbf{z}}$ & PR & PMOD & IGD-CF & HV-CF & PMDA & R-IGD & R-HV & EH & HV & IGD \\
\hline
$\mathbf{P}_{1}$ & \cellcolor{c11_13}11 & \cellcolor{c11_13}11 & \cellcolor{c11_13}11 & \cellcolor{c11_13}11 & \cellcolor{c11_13}11 & \cellcolor{c5_13}5 & \cellcolor{c5_13}5 & \cellcolor{c10_13}10 & \cellcolor{c3_13}3 & \cellcolor{c3_13}3 & \cellcolor{c11_13}11 & \cellcolor{c11_13}11 & \cellcolor{c12_13}12 & \cellcolor{c10_13}10 & \cellcolor{c13_13}13 & \cellcolor{c11_13}11\\
$\mathbf{P}_{2}$ & \cellcolor{c5_13}5 & \cellcolor{c4_13}4 & \cellcolor{c5_13}5 & \cellcolor{c5_13}5 & \cellcolor{c6_13}6 & \cellcolor{c5_13}5 & \cellcolor{c5_13}5 & \cellcolor{c4_13}4 & \cellcolor{c3_13}3 & \cellcolor{c3_13}3 & \cellcolor{c4_13}4 & \cellcolor{c6_13}6 & \cellcolor{c4_13}4 & \cellcolor{c4_13}4 & \cellcolor{c6_13}6 & \cellcolor{c8_13}8\\
$\mathbf{P}_{3}$ & \cellcolor{c5_13}5 & \cellcolor{c4_13}4 & \cellcolor{c7_13}7 & \cellcolor{c5_13}5 & \cellcolor{c6_13}6 & \cellcolor{c5_13}5 & \cellcolor{c5_13}5 & \cellcolor{c4_13}4 & \cellcolor{c3_13}3 & \cellcolor{c3_13}3 & \cellcolor{c4_13}4 & \cellcolor{c6_13}6 & \cellcolor{c4_13}4 & \cellcolor{c4_13}4 & \cellcolor{c8_13}8 & \cellcolor{c5_13}5\\
$\mathbf{P}_{4}$ & \cellcolor{c11_13}11 & \cellcolor{c11_13}11 & \cellcolor{c13_13}13 & \cellcolor{c11_13}11 & \cellcolor{c11_13}11 & \cellcolor{c5_13}5 & \cellcolor{c5_13}5 & \cellcolor{c10_13}10 & \cellcolor{c3_13}3 & \cellcolor{c3_13}3 & \cellcolor{c11_13}11 & \cellcolor{c10_13}10 & \cellcolor{c10_13}10 & \cellcolor{c10_13}10 & \cellcolor{c11_13}11 & \cellcolor{c11_13}11\\
$\mathbf{P}_{5}$ & \cellcolor{c5_13}5 & \cellcolor{c4_13}4 & \cellcolor{c6_13}6 & \cellcolor{c7_13}7 & \cellcolor{c5_13}5 & \cellcolor{c5_13}5 & \cellcolor{c5_13}5 & \cellcolor{c4_13}4 & \cellcolor{c3_13}3 & \cellcolor{c3_13}3 & \cellcolor{c4_13}4 & \cellcolor{c4_13}4 & \cellcolor{c9_13}9 & \cellcolor{c4_13}4 & \cellcolor{c7_13}7 & \cellcolor{c5_13}5\\
$\mathbf{P}_{6}$ & \cellcolor{c2_13}2 & \cellcolor{c3_13}3 & \cellcolor{c1_13}1 & \cellcolor{c1_13}1 & \cellcolor{c2_13}2 & \cellcolor{c1_13}1 & \cellcolor{c1_13}1 & \cellcolor{c1_13}1 & \cellcolor{c1_13}1 & \cellcolor{c1_13}1 & \cellcolor{c2_13}2 & \cellcolor{c1_13}1 & \cellcolor{c1_13}1 & \cellcolor{c2_13}2 & \cellcolor{c2_13}2 & \cellcolor{c2_13}2\\
$\mathbf{P}_{7}$ & \cellcolor{c5_13}5 & \cellcolor{c4_13}4 & \cellcolor{c10_13}10 & \cellcolor{c7_13}7 & \cellcolor{c6_13}6 & \cellcolor{c5_13}5 & \cellcolor{c5_13}5 & \cellcolor{c9_13}9 & \cellcolor{c3_13}3 & \cellcolor{c3_13}3 & \cellcolor{c4_13}4 & \cellcolor{c5_13}5 & \cellcolor{c4_13}4 & \cellcolor{c4_13}4 & \cellcolor{c3_13}3 & \cellcolor{c8_13}8\\
$\mathbf{P}_{8}$ & \cellcolor{c5_13}5 & \cellcolor{c4_13}4 & \cellcolor{c8_13}8 & \cellcolor{c9_13}9 & \cellcolor{c6_13}6 & \cellcolor{c5_13}5 & \cellcolor{c5_13}5 & \cellcolor{c4_13}4 & \cellcolor{c3_13}3 & \cellcolor{c3_13}3 & \cellcolor{c4_13}4 & \cellcolor{c8_13}8 & \cellcolor{c4_13}4 & \cellcolor{c4_13}4 & \cellcolor{c5_13}5 & \cellcolor{c5_13}5\\
$\mathbf{P}_{9}$ & \cellcolor{c5_13}5 & \cellcolor{c4_13}4 & \cellcolor{c9_13}9 & \cellcolor{c10_13}10 & \cellcolor{c6_13}6 & \cellcolor{c5_13}5 & \cellcolor{c5_13}5 & \cellcolor{c4_13}4 & \cellcolor{c3_13}3 & \cellcolor{c3_13}3 & \cellcolor{c4_13}4 & \cellcolor{c9_13}9 & \cellcolor{c4_13}4 & \cellcolor{c4_13}4 & \cellcolor{c3_13}3 & \cellcolor{c8_13}8\\
$\mathbf{P}_{10}$ & \cellcolor{c11_13}11 & \cellcolor{c11_13}11 & \cellcolor{c12_13}12 & \cellcolor{c13_13}13 & \cellcolor{c13_13}13 & \cellcolor{c5_13}5 & \cellcolor{c5_13}5 & \cellcolor{c10_13}10 & \cellcolor{c3_13}3 & \cellcolor{c3_13}3 & \cellcolor{c11_13}11 & \cellcolor{c12_13}12 & \cellcolor{c10_13}10 & \cellcolor{c10_13}10 & \cellcolor{c11_13}11 & \cellcolor{c11_13}11\\
$\mathbf{P}_{11}$ & \cellcolor{c4_13}4 & \cellcolor{c1_13}1 & \cellcolor{c4_13}4 & \cellcolor{c4_13}4 & \cellcolor{c4_13}4 & \cellcolor{c4_13}4 & \cellcolor{c3_13}3 & \cellcolor{c2_13}2 & \cellcolor{c3_13}3 & \cellcolor{c3_13}3 & \cellcolor{c3_13}3 & \cellcolor{c13_13}13 & \cellcolor{c13_13}13 & \cellcolor{c13_13}13 & \cellcolor{c10_13}10 & \cellcolor{c3_13}3\\
$\mathbf{P}_{12}$ & \cellcolor{c1_13}1 & \cellcolor{c2_13}2 & \cellcolor{c2_13}2 & \cellcolor{c2_13}2 & \cellcolor{c3_13}3 & \cellcolor{c3_13}3 & \cellcolor{c1_13}1 & \cellcolor{c3_13}3 & \cellcolor{c2_13}2 & \cellcolor{c2_13}2 & \cellcolor{c1_13}1 & \cellcolor{c2_13}2 & \cellcolor{c2_13}2 & \cellcolor{c1_13}1 & \cellcolor{c9_13}9 & \cellcolor{c4_13}4\\
$\mathbf{P}_{13}$ & \cellcolor{c3_13}3 & \cellcolor{c10_13}10 & \cellcolor{c3_13}3 & \cellcolor{c3_13}3 & \cellcolor{c1_13}1 & \cellcolor{c2_13}2 & \cellcolor{c4_13}4 & \cellcolor{c13_13}13 & \cellcolor{c3_13}3 & \cellcolor{c3_13}3 & \cellcolor{c10_13}10 & \cellcolor{c3_13}3 & \cellcolor{c3_13}3 & \cellcolor{c3_13}3 & \cellcolor{c1_13}1 & \cellcolor{c1_13}1\\
\hline
\end{tabularx}
}
\\
\subfloat[$\mathbf{z}^{-0.1}=(-0.1, -0.1, -0.1)^{\top}$]{
\begin{tabularx}{72em}{|A|C|C|C|C|C|C|C|C|C|C|C|C|C|C|C|C|}
\hline
\rowcolor[rgb]{0.9, 0.9, 0.9} & MASF & MED & IGD-C & IGD-A & IGD-P & HV$_{\mathbf{z}}$ & PR & PMOD & IGD-CF & HV-CF & PMDA & R-IGD & R-HV & EH & HV & IGD \\
\hline
$\mathbf{P}_{1}$ & \cellcolor{c11_13}11 & \cellcolor{c11_13}11 & \cellcolor{c11_13}11 & \cellcolor{c11_13}11 & \cellcolor{c11_13}11 & \cellcolor{c13_13}13 & \cellcolor{c1_13}1 & \cellcolor{c10_13}10 & \cellcolor{c3_13}3 & \cellcolor{c3_13}3 & \cellcolor{c11_13}11 & \cellcolor{c11_13}11 & \cellcolor{c12_13}12 & \cellcolor{c10_13}10 & \cellcolor{c13_13}13 & \cellcolor{c11_13}11\\
$\mathbf{P}_{2}$ & \cellcolor{c5_13}5 & \cellcolor{c3_13}3 & \cellcolor{c5_13}5 & \cellcolor{c5_13}5 & \cellcolor{c8_13}8 & \cellcolor{c6_13}6 & \cellcolor{c1_13}1 & \cellcolor{c5_13}5 & \cellcolor{c3_13}3 & \cellcolor{c3_13}3 & \cellcolor{c4_13}4 & \cellcolor{c6_13}6 & \cellcolor{c4_13}4 & \cellcolor{c3_13}3 & \cellcolor{c6_13}6 & \cellcolor{c8_13}8\\
$\mathbf{P}_{3}$ & \cellcolor{c5_13}5 & \cellcolor{c3_13}3 & \cellcolor{c7_13}7 & \cellcolor{c5_13}5 & \cellcolor{c5_13}5 & \cellcolor{c7_13}7 & \cellcolor{c1_13}1 & \cellcolor{c5_13}5 & \cellcolor{c3_13}3 & \cellcolor{c3_13}3 & \cellcolor{c4_13}4 & \cellcolor{c6_13}6 & \cellcolor{c6_13}6 & \cellcolor{c3_13}3 & \cellcolor{c8_13}8 & \cellcolor{c5_13}5\\
$\mathbf{P}_{4}$ & \cellcolor{c11_13}11 & \cellcolor{c11_13}11 & \cellcolor{c13_13}13 & \cellcolor{c11_13}11 & \cellcolor{c11_13}11 & \cellcolor{c11_13}11 & \cellcolor{c1_13}1 & \cellcolor{c10_13}10 & \cellcolor{c3_13}3 & \cellcolor{c3_13}3 & \cellcolor{c11_13}11 & \cellcolor{c10_13}10 & \cellcolor{c10_13}10 & \cellcolor{c10_13}10 & \cellcolor{c11_13}11 & \cellcolor{c11_13}11\\
$\mathbf{P}_{5}$ & \cellcolor{c5_13}5 & \cellcolor{c3_13}3 & \cellcolor{c6_13}6 & \cellcolor{c7_13}7 & \cellcolor{c5_13}5 & \cellcolor{c7_13}7 & \cellcolor{c1_13}1 & \cellcolor{c5_13}5 & \cellcolor{c3_13}3 & \cellcolor{c3_13}3 & \cellcolor{c4_13}4 & \cellcolor{c4_13}4 & \cellcolor{c4_13}4 & \cellcolor{c3_13}3 & \cellcolor{c7_13}7 & \cellcolor{c5_13}5\\
$\mathbf{P}_{6}$ & \cellcolor{c2_13}2 & \cellcolor{c2_13}2 & \cellcolor{c1_13}1 & \cellcolor{c1_13}1 & \cellcolor{c2_13}2 & \cellcolor{c2_13}2 & \cellcolor{c1_13}1 & \cellcolor{c1_13}1 & \cellcolor{c1_13}1 & \cellcolor{c1_13}1 & \cellcolor{c2_13}2 & \cellcolor{c1_13}1 & \cellcolor{c1_13}1 & \cellcolor{c2_13}2 & \cellcolor{c2_13}2 & \cellcolor{c2_13}2\\
$\mathbf{P}_{7}$ & \cellcolor{c5_13}5 & \cellcolor{c3_13}3 & \cellcolor{c10_13}10 & \cellcolor{c7_13}7 & \cellcolor{c8_13}8 & \cellcolor{c4_13}4 & \cellcolor{c1_13}1 & \cellcolor{c4_13}4 & \cellcolor{c3_13}3 & \cellcolor{c3_13}3 & \cellcolor{c4_13}4 & \cellcolor{c5_13}5 & \cellcolor{c6_13}6 & \cellcolor{c3_13}3 & \cellcolor{c3_13}3 & \cellcolor{c8_13}8\\
$\mathbf{P}_{8}$ & \cellcolor{c5_13}5 & \cellcolor{c3_13}3 & \cellcolor{c8_13}8 & \cellcolor{c9_13}9 & \cellcolor{c5_13}5 & \cellcolor{c7_13}7 & \cellcolor{c1_13}1 & \cellcolor{c5_13}5 & \cellcolor{c3_13}3 & \cellcolor{c3_13}3 & \cellcolor{c4_13}4 & \cellcolor{c8_13}8 & \cellcolor{c6_13}6 & \cellcolor{c3_13}3 & \cellcolor{c5_13}5 & \cellcolor{c5_13}5\\
$\mathbf{P}_{9}$ & \cellcolor{c5_13}5 & \cellcolor{c3_13}3 & \cellcolor{c9_13}9 & \cellcolor{c10_13}10 & \cellcolor{c8_13}8 & \cellcolor{c4_13}4 & \cellcolor{c1_13}1 & \cellcolor{c5_13}5 & \cellcolor{c3_13}3 & \cellcolor{c3_13}3 & \cellcolor{c4_13}4 & \cellcolor{c9_13}9 & \cellcolor{c6_13}6 & \cellcolor{c3_13}3 & \cellcolor{c3_13}3 & \cellcolor{c8_13}8\\
$\mathbf{P}_{10}$ & \cellcolor{c11_13}11 & \cellcolor{c11_13}11 & \cellcolor{c12_13}12 & \cellcolor{c13_13}13 & \cellcolor{c11_13}11 & \cellcolor{c12_13}12 & \cellcolor{c1_13}1 & \cellcolor{c10_13}10 & \cellcolor{c3_13}3 & \cellcolor{c3_13}3 & \cellcolor{c11_13}11 & \cellcolor{c12_13}12 & \cellcolor{c10_13}10 & \cellcolor{c10_13}10 & \cellcolor{c11_13}11 & \cellcolor{c11_13}11\\
$\mathbf{P}_{11}$ & \cellcolor{c4_13}4 & \cellcolor{c10_13}10 & \cellcolor{c4_13}4 & \cellcolor{c4_13}4 & \cellcolor{c3_13}3 & \cellcolor{c10_13}10 & \cellcolor{c1_13}1 & \cellcolor{c2_13}2 & \cellcolor{c3_13}3 & \cellcolor{c3_13}3 & \cellcolor{c3_13}3 & \cellcolor{c13_13}13 & \cellcolor{c13_13}13 & \cellcolor{c13_13}13 & \cellcolor{c10_13}10 & \cellcolor{c3_13}3\\
$\mathbf{P}_{12}$ & \cellcolor{c1_13}1 & \cellcolor{c1_13}1 & \cellcolor{c2_13}2 & \cellcolor{c2_13}2 & \cellcolor{c4_13}4 & \cellcolor{c3_13}3 & \cellcolor{c1_13}1 & \cellcolor{c3_13}3 & \cellcolor{c2_13}2 & \cellcolor{c2_13}2 & \cellcolor{c1_13}1 & \cellcolor{c2_13}2 & \cellcolor{c2_13}2 & \cellcolor{c1_13}1 & \cellcolor{c9_13}9 & \cellcolor{c4_13}4\\
$\mathbf{P}_{13}$ & \cellcolor{c3_13}3 & \cellcolor{c9_13}9 & \cellcolor{c3_13}3 & \cellcolor{c3_13}3 & \cellcolor{c1_13}1 & \cellcolor{c1_13}1 & \cellcolor{c1_13}1 & \cellcolor{c13_13}13 & \cellcolor{c3_13}3 & \cellcolor{c3_13}3 & \cellcolor{c10_13}10 & \cellcolor{c3_13}3 & \cellcolor{c3_13}3 & \cellcolor{c9_13}9 & \cellcolor{c1_13}1 & \cellcolor{c1_13}1\\
\hline
\end{tabularx}
}
}
\end{table*}

\begin{table*}[t]
   \setlength{\tabcolsep}{2pt} 
  \centering
  \caption{\small Rankings of the 13 synthetic point sets on the three-objective DTLZ2 problem by the 16 quality indicators when using $\mathbf{z}^{0.5}$ and $\mathbf{z}^{-0.1}$.}
  \label{suptab:rankings_psets_dtlz2_3d}
        {\scriptsize
\subfloat[$\mathbf{z}^{0.5}=(0.5, 0.5, 0.5)^{\top}$]{
\begin{tabularx}{72em}{|A|C|C|C|C|C|C|C|C|C|C|C|C|C|C|C|C|}
\hline
\rowcolor[rgb]{0.9, 0.9, 0.9} & MASF & MED & IGD-C & IGD-A & IGD-P & HV$_{\mathbf{z}}$ & PR & PMOD & IGD-CF & HV-CF & PMDA & R-IGD & R-HV & EH & HV & IGD \\
\hline
$\mathbf{P}_{1}$ & \cellcolor{c11_13}11 & \cellcolor{c11_13}11 & \cellcolor{c11_13}11 & \cellcolor{c13_13}13 & \cellcolor{c11_13}11 & \cellcolor{c4_13}4 & \cellcolor{c4_13}4 & \cellcolor{c10_13}10 & \cellcolor{c4_13}4 & \cellcolor{c4_13}4 & \cellcolor{c11_13}11 & \cellcolor{c12_13}12 & \cellcolor{c12_13}12 & \cellcolor{c10_13}10 & \cellcolor{c10_13}10 & \cellcolor{c11_13}11\\
$\mathbf{P}_{2}$ & \cellcolor{c5_13}5 & \cellcolor{c4_13}4 & \cellcolor{c5_13}5 & \cellcolor{c9_13}9 & \cellcolor{c6_13}6 & \cellcolor{c4_13}4 & \cellcolor{c4_13}4 & \cellcolor{c7_13}7 & \cellcolor{c4_13}4 & \cellcolor{c4_13}4 & \cellcolor{c4_13}4 & \cellcolor{c8_13}8 & \cellcolor{c4_13}4 & \cellcolor{c4_13}4 & \cellcolor{c6_13}6 & \cellcolor{c5_13}5\\
$\mathbf{P}_{3}$ & \cellcolor{c5_13}5 & \cellcolor{c4_13}4 & \cellcolor{c7_13}7 & \cellcolor{c8_13}8 & \cellcolor{c6_13}6 & \cellcolor{c4_13}4 & \cellcolor{c4_13}4 & \cellcolor{c6_13}6 & \cellcolor{c4_13}4 & \cellcolor{c4_13}4 & \cellcolor{c4_13}4 & \cellcolor{c4_13}4 & \cellcolor{c4_13}4 & \cellcolor{c4_13}4 & \cellcolor{c6_13}6 & \cellcolor{c10_13}10\\
$\mathbf{P}_{4}$ & \cellcolor{c11_13}11 & \cellcolor{c11_13}11 & \cellcolor{c12_13}12 & \cellcolor{c11_13}11 & \cellcolor{c13_13}13 & \cellcolor{c4_13}4 & \cellcolor{c4_13}4 & \cellcolor{c9_13}9 & \cellcolor{c4_13}4 & \cellcolor{c4_13}4 & \cellcolor{c11_13}11 & \cellcolor{c10_13}10 & \cellcolor{c10_13}10 & \cellcolor{c11_13}11 & \cellcolor{c12_13}12 & \cellcolor{c12_13}12\\
$\mathbf{P}_{5}$ & \cellcolor{c5_13}5 & \cellcolor{c4_13}4 & \cellcolor{c5_13}5 & \cellcolor{c9_13}9 & \cellcolor{c5_13}5 & \cellcolor{c4_13}4 & \cellcolor{c4_13}4 & \cellcolor{c2_13}2 & \cellcolor{c4_13}4 & \cellcolor{c4_13}4 & \cellcolor{c4_13}4 & \cellcolor{c8_13}8 & \cellcolor{c4_13}4 & \cellcolor{c4_13}4 & \cellcolor{c4_13}4 & \cellcolor{c5_13}5\\
$\mathbf{P}_{6}$ & \cellcolor{c2_13}2 & \cellcolor{c2_13}2 & \cellcolor{c1_13}1 & \cellcolor{c1_13}1 & \cellcolor{c1_13}1 & \cellcolor{c1_13}1 & \cellcolor{c3_13}3 & \cellcolor{c8_13}8 & \cellcolor{c1_13}1 & \cellcolor{c1_13}1 & \cellcolor{c2_13}2 & \cellcolor{c1_13}1 & \cellcolor{c1_13}1 & \cellcolor{c2_13}2 & \cellcolor{c8_13}8 & \cellcolor{c2_13}2\\
$\mathbf{P}_{7}$ & \cellcolor{c5_13}5 & \cellcolor{c4_13}4 & \cellcolor{c9_13}9 & \cellcolor{c5_13}5 & \cellcolor{c6_13}6 & \cellcolor{c4_13}4 & \cellcolor{c4_13}4 & \cellcolor{c2_13}2 & \cellcolor{c4_13}4 & \cellcolor{c4_13}4 & \cellcolor{c4_13}4 & \cellcolor{c6_13}6 & \cellcolor{c4_13}4 & \cellcolor{c4_13}4 & \cellcolor{c2_13}2 & \cellcolor{c5_13}5\\
$\mathbf{P}_{8}$ & \cellcolor{c5_13}5 & \cellcolor{c4_13}4 & \cellcolor{c7_13}7 & \cellcolor{c7_13}7 & \cellcolor{c6_13}6 & \cellcolor{c4_13}4 & \cellcolor{c4_13}4 & \cellcolor{c2_13}2 & \cellcolor{c4_13}4 & \cellcolor{c4_13}4 & \cellcolor{c4_13}4 & \cellcolor{c4_13}4 & \cellcolor{c4_13}4 & \cellcolor{c4_13}4 & \cellcolor{c4_13}4 & \cellcolor{c5_13}5\\
$\mathbf{P}_{9}$ & \cellcolor{c5_13}5 & \cellcolor{c4_13}4 & \cellcolor{c9_13}9 & \cellcolor{c5_13}5 & \cellcolor{c6_13}6 & \cellcolor{c4_13}4 & \cellcolor{c4_13}4 & \cellcolor{c2_13}2 & \cellcolor{c4_13}4 & \cellcolor{c4_13}4 & \cellcolor{c4_13}4 & \cellcolor{c6_13}6 & \cellcolor{c4_13}4 & \cellcolor{c4_13}4 & \cellcolor{c2_13}2 & \cellcolor{c5_13}5\\
$\mathbf{P}_{10}$ & \cellcolor{c11_13}11 & \cellcolor{c11_13}11 & \cellcolor{c12_13}12 & \cellcolor{c12_13}12 & \cellcolor{c11_13}11 & \cellcolor{c4_13}4 & \cellcolor{c4_13}4 & \cellcolor{c10_13}10 & \cellcolor{c4_13}4 & \cellcolor{c4_13}4 & \cellcolor{c11_13}11 & \cellcolor{c11_13}11 & \cellcolor{c10_13}10 & \cellcolor{c11_13}11 & \cellcolor{c11_13}11 & \cellcolor{c12_13}12\\
$\mathbf{P}_{11}$ & \cellcolor{c4_13}4 & \cellcolor{c3_13}3 & \cellcolor{c4_13}4 & \cellcolor{c4_13}4 & \cellcolor{c3_13}3 & \cellcolor{c4_13}4 & \cellcolor{c1_13}1 & \cellcolor{c12_13}12 & \cellcolor{c3_13}3 & \cellcolor{c3_13}3 & \cellcolor{c3_13}3 & \cellcolor{c13_13}13 & \cellcolor{c13_13}13 & \cellcolor{c13_13}13 & \cellcolor{c13_13}13 & \cellcolor{c4_13}4\\
$\mathbf{P}_{12}$ & \cellcolor{c1_13}1 & \cellcolor{c1_13}1 & \cellcolor{c2_13}2 & \cellcolor{c2_13}2 & \cellcolor{c2_13}2 & \cellcolor{c2_13}2 & \cellcolor{c1_13}1 & \cellcolor{c1_13}1 & \cellcolor{c2_13}2 & \cellcolor{c2_13}2 & \cellcolor{c1_13}1 & \cellcolor{c2_13}2 & \cellcolor{c2_13}2 & \cellcolor{c1_13}1 & \cellcolor{c9_13}9 & \cellcolor{c3_13}3\\
$\mathbf{P}_{13}$ & \cellcolor{c3_13}3 & \cellcolor{c10_13}10 & \cellcolor{c3_13}3 & \cellcolor{c3_13}3 & \cellcolor{c4_13}4 & \cellcolor{c3_13}3 & \cellcolor{c4_13}4 & \cellcolor{c13_13}13 & \cellcolor{c4_13}4 & \cellcolor{c4_13}4 & \cellcolor{c10_13}10 & \cellcolor{c3_13}3 & \cellcolor{c3_13}3 & \cellcolor{c3_13}3 & \cellcolor{c1_13}1 & \cellcolor{c1_13}1\\
\hline
\end{tabularx}
}
\\
\subfloat[$\mathbf{z}^{-0.1}=(-0.1, -0.1, -0.1)^{\top}$]{
\begin{tabularx}{72em}{|A|C|C|C|C|C|C|C|C|C|C|C|C|C|C|C|C|}
\hline
\rowcolor[rgb]{0.9, 0.9, 0.9} & MASF & MED & IGD-C & IGD-A & IGD-P & HV$_{\mathbf{z}}$ & PR & PMOD & IGD-CF & HV-CF & PMDA & R-IGD & R-HV & EH & HV & IGD \\
\hline
$\mathbf{P}_{1}$ & \cellcolor{c11_13}11 & \cellcolor{c3_13}3 & \cellcolor{c1_13}1 & \cellcolor{c13_13}13 & \cellcolor{c11_13}11 & \cellcolor{c11_13}11 & \cellcolor{c1_13}1 & \cellcolor{c11_13}11 & \cellcolor{c2_13}2 & \cellcolor{c2_13}2 & \cellcolor{c11_13}11 & \cellcolor{c12_13}12 & \cellcolor{c11_13}11 & \cellcolor{c10_13}10 & \cellcolor{c10_13}10 & \cellcolor{c11_13}11\\
$\mathbf{P}_{2}$ & \cellcolor{c5_13}5 & \cellcolor{c5_13}5 & \cellcolor{c3_13}3 & \cellcolor{c9_13}9 & \cellcolor{c5_13}5 & \cellcolor{c6_13}6 & \cellcolor{c1_13}1 & \cellcolor{c6_13}6 & \cellcolor{c3_13}3 & \cellcolor{c3_13}3 & \cellcolor{c4_13}4 & \cellcolor{c8_13}8 & \cellcolor{c4_13}4 & \cellcolor{c8_13}8 & \cellcolor{c6_13}6 & \cellcolor{c5_13}5\\
$\mathbf{P}_{3}$ & \cellcolor{c5_13}5 & \cellcolor{c5_13}5 & \cellcolor{c9_13}9 & \cellcolor{c8_13}8 & \cellcolor{c10_13}10 & \cellcolor{c8_13}8 & \cellcolor{c1_13}1 & \cellcolor{c4_13}4 & \cellcolor{c3_13}3 & \cellcolor{c3_13}3 & \cellcolor{c4_13}4 & \cellcolor{c4_13}4 & \cellcolor{c4_13}4 & \cellcolor{c4_13}4 & \cellcolor{c6_13}6 & \cellcolor{c10_13}10\\
$\mathbf{P}_{4}$ & \cellcolor{c11_13}11 & \cellcolor{c1_13}1 & \cellcolor{c10_13}10 & \cellcolor{c11_13}11 & \cellcolor{c12_13}12 & \cellcolor{c13_13}13 & \cellcolor{c1_13}1 & \cellcolor{c9_13}9 & \cellcolor{c3_13}3 & \cellcolor{c3_13}3 & \cellcolor{c11_13}11 & \cellcolor{c10_13}10 & \cellcolor{c10_13}10 & \cellcolor{c11_13}11 & \cellcolor{c12_13}12 & \cellcolor{c12_13}12\\
$\mathbf{P}_{5}$ & \cellcolor{c5_13}5 & \cellcolor{c5_13}5 & \cellcolor{c3_13}3 & \cellcolor{c9_13}9 & \cellcolor{c5_13}5 & \cellcolor{c6_13}6 & \cellcolor{c1_13}1 & \cellcolor{c6_13}6 & \cellcolor{c3_13}3 & \cellcolor{c3_13}3 & \cellcolor{c4_13}4 & \cellcolor{c8_13}8 & \cellcolor{c4_13}4 & \cellcolor{c8_13}8 & \cellcolor{c4_13}4 & \cellcolor{c5_13}5\\
$\mathbf{P}_{6}$ & \cellcolor{c2_13}2 & \cellcolor{c11_13}11 & \cellcolor{c5_13}5 & \cellcolor{c1_13}1 & \cellcolor{c2_13}2 & \cellcolor{c2_13}2 & \cellcolor{c1_13}1 & \cellcolor{c8_13}8 & \cellcolor{c3_13}3 & \cellcolor{c3_13}3 & \cellcolor{c2_13}2 & \cellcolor{c1_13}1 & \cellcolor{c1_13}1 & \cellcolor{c2_13}2 & \cellcolor{c8_13}8 & \cellcolor{c2_13}2\\
$\mathbf{P}_{7}$ & \cellcolor{c5_13}5 & \cellcolor{c5_13}5 & \cellcolor{c12_13}12 & \cellcolor{c5_13}5 & \cellcolor{c5_13}5 & \cellcolor{c3_13}3 & \cellcolor{c1_13}1 & \cellcolor{c2_13}2 & \cellcolor{c3_13}3 & \cellcolor{c3_13}3 & \cellcolor{c4_13}4 & \cellcolor{c6_13}6 & \cellcolor{c4_13}4 & \cellcolor{c4_13}4 & \cellcolor{c2_13}2 & \cellcolor{c5_13}5\\
$\mathbf{P}_{8}$ & \cellcolor{c5_13}5 & \cellcolor{c5_13}5 & \cellcolor{c8_13}8 & \cellcolor{c7_13}7 & \cellcolor{c5_13}5 & \cellcolor{c5_13}5 & \cellcolor{c1_13}1 & \cellcolor{c4_13}4 & \cellcolor{c3_13}3 & \cellcolor{c3_13}3 & \cellcolor{c4_13}4 & \cellcolor{c4_13}4 & \cellcolor{c4_13}4 & \cellcolor{c4_13}4 & \cellcolor{c4_13}4 & \cellcolor{c5_13}5\\
$\mathbf{P}_{9}$ & \cellcolor{c5_13}5 & \cellcolor{c5_13}5 & \cellcolor{c12_13}12 & \cellcolor{c5_13}5 & \cellcolor{c5_13}5 & \cellcolor{c3_13}3 & \cellcolor{c1_13}1 & \cellcolor{c3_13}3 & \cellcolor{c3_13}3 & \cellcolor{c3_13}3 & \cellcolor{c4_13}4 & \cellcolor{c6_13}6 & \cellcolor{c4_13}4 & \cellcolor{c4_13}4 & \cellcolor{c2_13}2 & \cellcolor{c5_13}5\\
$\mathbf{P}_{10}$ & \cellcolor{c11_13}11 & \cellcolor{c1_13}1 & \cellcolor{c10_13}10 & \cellcolor{c12_13}12 & \cellcolor{c12_13}12 & \cellcolor{c11_13}11 & \cellcolor{c1_13}1 & \cellcolor{c9_13}9 & \cellcolor{c3_13}3 & \cellcolor{c3_13}3 & \cellcolor{c11_13}11 & \cellcolor{c11_13}11 & \cellcolor{c11_13}11 & \cellcolor{c11_13}11 & \cellcolor{c11_13}11 & \cellcolor{c12_13}12\\
$\mathbf{P}_{11}$ & \cellcolor{c4_13}4 & \cellcolor{c13_13}13 & \cellcolor{c7_13}7 & \cellcolor{c4_13}4 & \cellcolor{c4_13}4 & \cellcolor{c10_13}10 & \cellcolor{c1_13}1 & \cellcolor{c12_13}12 & \cellcolor{c3_13}3 & \cellcolor{c3_13}3 & \cellcolor{c3_13}3 & \cellcolor{c13_13}13 & \cellcolor{c13_13}13 & \cellcolor{c13_13}13 & \cellcolor{c13_13}13 & \cellcolor{c4_13}4\\
$\mathbf{P}_{12}$ & \cellcolor{c1_13}1 & \cellcolor{c12_13}12 & \cellcolor{c6_13}6 & \cellcolor{c2_13}2 & \cellcolor{c3_13}3 & \cellcolor{c9_13}9 & \cellcolor{c1_13}1 & \cellcolor{c1_13}1 & \cellcolor{c3_13}3 & \cellcolor{c3_13}3 & \cellcolor{c1_13}1 & \cellcolor{c2_13}2 & \cellcolor{c2_13}2 & \cellcolor{c1_13}1 & \cellcolor{c9_13}9 & \cellcolor{c3_13}3\\
$\mathbf{P}_{13}$ & \cellcolor{c3_13}3 & \cellcolor{c4_13}4 & \cellcolor{c2_13}2 & \cellcolor{c3_13}3 & \cellcolor{c1_13}1 & \cellcolor{c1_13}1 & \cellcolor{c1_13}1 & \cellcolor{c13_13}13 & \cellcolor{c1_13}1 & \cellcolor{c1_13}1 & \cellcolor{c10_13}10 & \cellcolor{c3_13}3 & \cellcolor{c3_13}3 & \cellcolor{c3_13}3 & \cellcolor{c1_13}1 & \cellcolor{c1_13}1\\
\hline
\end{tabularx}
}
}
\end{table*}

\begin{table*}[t]
   \setlength{\tabcolsep}{2pt} 
  \centering
  \caption{\small Rankings of the 13 synthetic point sets on the three-objective convDTLZ2 problem by the 16 quality indicators when using $\mathbf{z}^{0.2}$ and $\mathbf{z}^{2.0}$.}
  \label{suptab:rankings_psets_convdtlz2_3d}
        {\scriptsize
\subfloat[$\mathbf{z}^{0.2}=(0.2, 0.2, 0.2)^{\top}$]{
\begin{tabularx}{72em}{|A|C|C|C|C|C|C|C|C|C|C|C|C|C|C|C|C|}
\hline
\rowcolor[rgb]{0.9, 0.9, 0.9} & MASF & MED & IGD-C & IGD-A & IGD-P & HV$_{\mathbf{z}}$ & PR & PMOD & IGD-CF & HV-CF & PMDA & R-IGD & R-HV & EH & HV & IGD \\
\hline
$\mathbf{P}_{1}$ & \cellcolor{c13_13}13 & \cellcolor{c13_13}13 & \cellcolor{c13_13}13 & \cellcolor{c13_13}13 & \cellcolor{c13_13}13 & \cellcolor{c3_13}3 & \cellcolor{c5_13}5 & \cellcolor{c12_13}12 & \cellcolor{c4_13}4 & \cellcolor{c4_13}4 & \cellcolor{c13_13}13 & \cellcolor{c11_13}11 & \cellcolor{c11_13}11 & \cellcolor{c11_13}11 & \cellcolor{c12_13}12 & \cellcolor{c13_13}13\\
$\mathbf{P}_{2}$ & \cellcolor{c7_13}7 & \cellcolor{c6_13}6 & \cellcolor{c7_13}7 & \cellcolor{c9_13}9 & \cellcolor{c9_13}9 & \cellcolor{c3_13}3 & \cellcolor{c5_13}5 & \cellcolor{c5_13}5 & \cellcolor{c4_13}4 & \cellcolor{c4_13}4 & \cellcolor{c6_13}6 & \cellcolor{c7_13}7 & \cellcolor{c5_13}5 & \cellcolor{c4_13}4 & \cellcolor{c5_13}5 & \cellcolor{c7_13}7\\
$\mathbf{P}_{3}$ & \cellcolor{c9_13}9 & \cellcolor{c8_13}8 & \cellcolor{c9_13}9 & \cellcolor{c7_13}7 & \cellcolor{c7_13}7 & \cellcolor{c3_13}3 & \cellcolor{c5_13}5 & \cellcolor{c8_13}8 & \cellcolor{c4_13}4 & \cellcolor{c4_13}4 & \cellcolor{c8_13}8 & \cellcolor{c5_13}5 & \cellcolor{c8_13}8 & \cellcolor{c6_13}6 & \cellcolor{c7_13}7 & \cellcolor{c9_13}9\\
$\mathbf{P}_{4}$ & \cellcolor{c11_13}11 & \cellcolor{c11_13}11 & \cellcolor{c11_13}11 & \cellcolor{c11_13}11 & \cellcolor{c11_13}11 & \cellcolor{c3_13}3 & \cellcolor{c5_13}5 & \cellcolor{c10_13}10 & \cellcolor{c4_13}4 & \cellcolor{c4_13}4 & \cellcolor{c12_13}12 & \cellcolor{c9_13}9 & \cellcolor{c9_13}9 & \cellcolor{c9_13}9 & \cellcolor{c9_13}9 & \cellcolor{c11_13}11\\
$\mathbf{P}_{5}$ & \cellcolor{c7_13}7 & \cellcolor{c6_13}6 & \cellcolor{c7_13}7 & \cellcolor{c9_13}9 & \cellcolor{c9_13}9 & \cellcolor{c3_13}3 & \cellcolor{c5_13}5 & \cellcolor{c5_13}5 & \cellcolor{c4_13}4 & \cellcolor{c4_13}4 & \cellcolor{c6_13}6 & \cellcolor{c8_13}8 & \cellcolor{c5_13}5 & \cellcolor{c4_13}4 & \cellcolor{c5_13}5 & \cellcolor{c7_13}7\\
$\mathbf{P}_{6}$ & \cellcolor{c2_13}2 & \cellcolor{c2_13}2 & \cellcolor{c1_13}1 & \cellcolor{c1_13}1 & \cellcolor{c1_13}1 & \cellcolor{c2_13}2 & \cellcolor{c4_13}4 & \cellcolor{c2_13}2 & \cellcolor{c1_13}1 & \cellcolor{c1_13}1 & \cellcolor{c2_13}2 & \cellcolor{c1_13}1 & \cellcolor{c1_13}1 & \cellcolor{c2_13}2 & \cellcolor{c2_13}2 & \cellcolor{c2_13}2\\
$\mathbf{P}_{7}$ & \cellcolor{c4_13}4 & \cellcolor{c4_13}4 & \cellcolor{c4_13}4 & \cellcolor{c5_13}5 & \cellcolor{c5_13}5 & \cellcolor{c3_13}3 & \cellcolor{c3_13}3 & \cellcolor{c4_13}4 & \cellcolor{c4_13}4 & \cellcolor{c4_13}4 & \cellcolor{c4_13}4 & \cellcolor{c12_13}12 & \cellcolor{c12_13}12 & \cellcolor{c12_13}12 & \cellcolor{c11_13}11 & \cellcolor{c3_13}3\\
$\mathbf{P}_{8}$ & \cellcolor{c9_13}9 & \cellcolor{c8_13}8 & \cellcolor{c9_13}9 & \cellcolor{c8_13}8 & \cellcolor{c7_13}7 & \cellcolor{c3_13}3 & \cellcolor{c5_13}5 & \cellcolor{c8_13}8 & \cellcolor{c4_13}4 & \cellcolor{c4_13}4 & \cellcolor{c8_13}8 & \cellcolor{c5_13}5 & \cellcolor{c7_13}7 & \cellcolor{c6_13}6 & \cellcolor{c7_13}7 & \cellcolor{c9_13}9\\
$\mathbf{P}_{9}$ & \cellcolor{c6_13}6 & \cellcolor{c5_13}5 & \cellcolor{c6_13}6 & \cellcolor{c4_13}4 & \cellcolor{c6_13}6 & \cellcolor{c3_13}3 & \cellcolor{c5_13}5 & \cellcolor{c3_13}3 & \cellcolor{c4_13}4 & \cellcolor{c4_13}4 & \cellcolor{c5_13}5 & \cellcolor{c4_13}4 & \cellcolor{c4_13}4 & \cellcolor{c3_13}3 & \cellcolor{c3_13}3 & \cellcolor{c6_13}6\\
$\mathbf{P}_{10}$ & \cellcolor{c11_13}11 & \cellcolor{c11_13}11 & \cellcolor{c11_13}11 & \cellcolor{c11_13}11 & \cellcolor{c11_13}11 & \cellcolor{c3_13}3 & \cellcolor{c5_13}5 & \cellcolor{c11_13}11 & \cellcolor{c4_13}4 & \cellcolor{c4_13}4 & \cellcolor{c11_13}11 & \cellcolor{c9_13}9 & \cellcolor{c10_13}10 & \cellcolor{c9_13}9 & \cellcolor{c10_13}10 & \cellcolor{c12_13}12\\
$\mathbf{P}_{11}$ & \cellcolor{c5_13}5 & \cellcolor{c3_13}3 & \cellcolor{c5_13}5 & \cellcolor{c6_13}6 & \cellcolor{c4_13}4 & \cellcolor{c3_13}3 & \cellcolor{c1_13}1 & \cellcolor{c7_13}7 & \cellcolor{c4_13}4 & \cellcolor{c4_13}4 & \cellcolor{c3_13}3 & \cellcolor{c12_13}12 & \cellcolor{c12_13}12 & \cellcolor{c12_13}12 & \cellcolor{c13_13}13 & \cellcolor{c5_13}5\\
$\mathbf{P}_{12}$ & \cellcolor{c1_13}1 & \cellcolor{c1_13}1 & \cellcolor{c3_13}3 & \cellcolor{c3_13}3 & \cellcolor{c2_13}2 & \cellcolor{c1_13}1 & \cellcolor{c1_13}1 & \cellcolor{c1_13}1 & \cellcolor{c3_13}3 & \cellcolor{c3_13}3 & \cellcolor{c1_13}1 & \cellcolor{c2_13}2 & \cellcolor{c2_13}2 & \cellcolor{c1_13}1 & \cellcolor{c4_13}4 & \cellcolor{c4_13}4\\
$\mathbf{P}_{13}$ & \cellcolor{c3_13}3 & \cellcolor{c10_13}10 & \cellcolor{c2_13}2 & \cellcolor{c2_13}2 & \cellcolor{c3_13}3 & \cellcolor{c3_13}3 & \cellcolor{c5_13}5 & \cellcolor{c13_13}13 & \cellcolor{c2_13}2 & \cellcolor{c2_13}2 & \cellcolor{c10_13}10 & \cellcolor{c3_13}3 & \cellcolor{c3_13}3 & \cellcolor{c8_13}8 & \cellcolor{c1_13}1 & \cellcolor{c1_13}1\\
\hline
\end{tabularx}
}
\\
\subfloat[$\mathbf{z}^{2.0}=(2.0, 2.0, 2.0)^{\top}$]{
\begin{tabularx}{72em}{|A|C|C|C|C|C|C|C|C|C|C|C|C|C|C|C|C|}
\hline
\rowcolor[rgb]{0.9, 0.9, 0.9} & MASF & MED & IGD-C & IGD-A & IGD-P & HV$_{\mathbf{z}}$ & PR & PMOD & IGD-CF & HV-CF & PMDA & R-IGD & R-HV & EH & HV & IGD \\
\hline
$\mathbf{P}_{1}$ & \cellcolor{c13_13}13 & \cellcolor{c3_13}3 & \cellcolor{c1_13}1 & \cellcolor{c13_13}13 & \cellcolor{c13_13}13 & \cellcolor{c12_13}12 & \cellcolor{c1_13}1 & \cellcolor{c12_13}12 & \cellcolor{c1_13}1 & \cellcolor{c1_13}1 & \cellcolor{c13_13}13 & \cellcolor{c11_13}11 & \cellcolor{c11_13}11 & \cellcolor{c9_13}9 & \cellcolor{c12_13}12 & \cellcolor{c13_13}13\\
$\mathbf{P}_{2}$ & \cellcolor{c7_13}7 & \cellcolor{c9_13}9 & \cellcolor{c3_13}3 & \cellcolor{c9_13}9 & \cellcolor{c7_13}7 & \cellcolor{c5_13}5 & \cellcolor{c1_13}1 & \cellcolor{c6_13}6 & \cellcolor{c3_13}3 & \cellcolor{c3_13}3 & \cellcolor{c6_13}6 & \cellcolor{c7_13}7 & \cellcolor{c5_13}5 & \cellcolor{c6_13}6 & \cellcolor{c5_13}5 & \cellcolor{c7_13}7\\
$\mathbf{P}_{3}$ & \cellcolor{c9_13}9 & \cellcolor{c6_13}6 & \cellcolor{c8_13}8 & \cellcolor{c7_13}7 & \cellcolor{c9_13}9 & \cellcolor{c7_13}7 & \cellcolor{c1_13}1 & \cellcolor{c8_13}8 & \cellcolor{c3_13}3 & \cellcolor{c3_13}3 & \cellcolor{c8_13}8 & \cellcolor{c6_13}6 & \cellcolor{c7_13}7 & \cellcolor{c4_13}4 & \cellcolor{c7_13}7 & \cellcolor{c9_13}9\\
$\mathbf{P}_{4}$ & \cellcolor{c11_13}11 & \cellcolor{c4_13}4 & \cellcolor{c12_13}12 & \cellcolor{c11_13}11 & \cellcolor{c11_13}11 & \cellcolor{c9_13}9 & \cellcolor{c1_13}1 & \cellcolor{c10_13}10 & \cellcolor{c3_13}3 & \cellcolor{c3_13}3 & \cellcolor{c12_13}12 & \cellcolor{c9_13}9 & \cellcolor{c9_13}9 & \cellcolor{c10_13}10 & \cellcolor{c9_13}9 & \cellcolor{c11_13}11\\
$\mathbf{P}_{5}$ & \cellcolor{c7_13}7 & \cellcolor{c9_13}9 & \cellcolor{c3_13}3 & \cellcolor{c9_13}9 & \cellcolor{c7_13}7 & \cellcolor{c4_13}4 & \cellcolor{c1_13}1 & \cellcolor{c5_13}5 & \cellcolor{c3_13}3 & \cellcolor{c3_13}3 & \cellcolor{c6_13}6 & \cellcolor{c8_13}8 & \cellcolor{c6_13}6 & \cellcolor{c6_13}6 & \cellcolor{c5_13}5 & \cellcolor{c7_13}7\\
$\mathbf{P}_{6}$ & \cellcolor{c2_13}2 & \cellcolor{c11_13}11 & \cellcolor{c6_13}6 & \cellcolor{c1_13}1 & \cellcolor{c2_13}2 & \cellcolor{c2_13}2 & \cellcolor{c1_13}1 & \cellcolor{c2_13}2 & \cellcolor{c3_13}3 & \cellcolor{c3_13}3 & \cellcolor{c2_13}2 & \cellcolor{c1_13}1 & \cellcolor{c1_13}1 & \cellcolor{c2_13}2 & \cellcolor{c2_13}2 & \cellcolor{c2_13}2\\
$\mathbf{P}_{7}$ & \cellcolor{c4_13}4 & \cellcolor{c1_13}1 & \cellcolor{c5_13}5 & \cellcolor{c5_13}5 & \cellcolor{c3_13}3 & \cellcolor{c11_13}11 & \cellcolor{c1_13}1 & \cellcolor{c4_13}4 & \cellcolor{c3_13}3 & \cellcolor{c3_13}3 & \cellcolor{c4_13}4 & \cellcolor{c12_13}12 & \cellcolor{c12_13}12 & \cellcolor{c12_13}12 & \cellcolor{c11_13}11 & \cellcolor{c3_13}3\\
$\mathbf{P}_{8}$ & \cellcolor{c9_13}9 & \cellcolor{c6_13}6 & \cellcolor{c8_13}8 & \cellcolor{c8_13}8 & \cellcolor{c9_13}9 & \cellcolor{c7_13}7 & \cellcolor{c1_13}1 & \cellcolor{c9_13}9 & \cellcolor{c3_13}3 & \cellcolor{c3_13}3 & \cellcolor{c8_13}8 & \cellcolor{c5_13}5 & \cellcolor{c7_13}7 & \cellcolor{c4_13}4 & \cellcolor{c7_13}7 & \cellcolor{c9_13}9\\
$\mathbf{P}_{9}$ & \cellcolor{c6_13}6 & \cellcolor{c13_13}13 & \cellcolor{c11_13}11 & \cellcolor{c4_13}4 & \cellcolor{c6_13}6 & \cellcolor{c3_13}3 & \cellcolor{c1_13}1 & \cellcolor{c3_13}3 & \cellcolor{c3_13}3 & \cellcolor{c3_13}3 & \cellcolor{c5_13}5 & \cellcolor{c4_13}4 & \cellcolor{c4_13}4 & \cellcolor{c8_13}8 & \cellcolor{c3_13}3 & \cellcolor{c6_13}6\\
$\mathbf{P}_{10}$ & \cellcolor{c11_13}11 & \cellcolor{c4_13}4 & \cellcolor{c12_13}12 & \cellcolor{c11_13}11 & \cellcolor{c12_13}12 & \cellcolor{c9_13}9 & \cellcolor{c1_13}1 & \cellcolor{c11_13}11 & \cellcolor{c3_13}3 & \cellcolor{c3_13}3 & \cellcolor{c11_13}11 & \cellcolor{c9_13}9 & \cellcolor{c9_13}9 & \cellcolor{c10_13}10 & \cellcolor{c10_13}10 & \cellcolor{c12_13}12\\
$\mathbf{P}_{11}$ & \cellcolor{c5_13}5 & \cellcolor{c2_13}2 & \cellcolor{c7_13}7 & \cellcolor{c6_13}6 & \cellcolor{c5_13}5 & \cellcolor{c13_13}13 & \cellcolor{c1_13}1 & \cellcolor{c7_13}7 & \cellcolor{c3_13}3 & \cellcolor{c3_13}3 & \cellcolor{c3_13}3 & \cellcolor{c12_13}12 & \cellcolor{c12_13}12 & \cellcolor{c12_13}12 & \cellcolor{c13_13}13 & \cellcolor{c5_13}5\\
$\mathbf{P}_{12}$ & \cellcolor{c1_13}1 & \cellcolor{c12_13}12 & \cellcolor{c10_13}10 & \cellcolor{c3_13}3 & \cellcolor{c4_13}4 & \cellcolor{c6_13}6 & \cellcolor{c1_13}1 & \cellcolor{c1_13}1 & \cellcolor{c3_13}3 & \cellcolor{c3_13}3 & \cellcolor{c1_13}1 & \cellcolor{c2_13}2 & \cellcolor{c2_13}2 & \cellcolor{c1_13}1 & \cellcolor{c4_13}4 & \cellcolor{c4_13}4\\
$\mathbf{P}_{13}$ & \cellcolor{c3_13}3 & \cellcolor{c8_13}8 & \cellcolor{c1_13}1 & \cellcolor{c2_13}2 & \cellcolor{c1_13}1 & \cellcolor{c1_13}1 & \cellcolor{c1_13}1 & \cellcolor{c13_13}13 & \cellcolor{c1_13}1 & \cellcolor{c1_13}1 & \cellcolor{c10_13}10 & \cellcolor{c3_13}3 & \cellcolor{c3_13}3 & \cellcolor{c3_13}3 & \cellcolor{c1_13}1 & \cellcolor{c1_13}1\\
\hline
\end{tabularx}
}

}
\end{table*}

\clearpage

\begin{figure*}[t]
  \centering
    \includegraphics[width=0.31\textwidth]{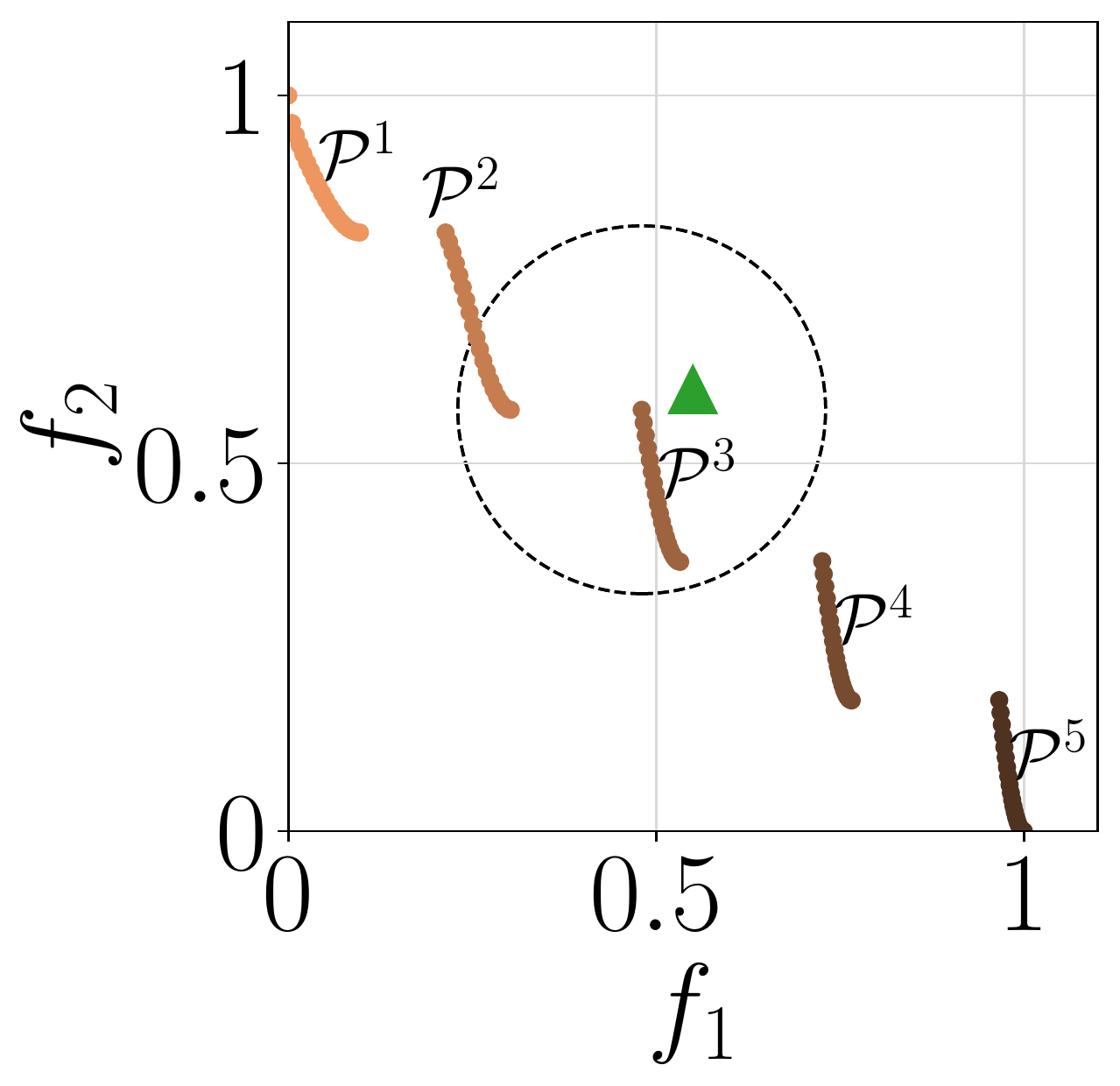}
    \includegraphics[width=0.31\textwidth]{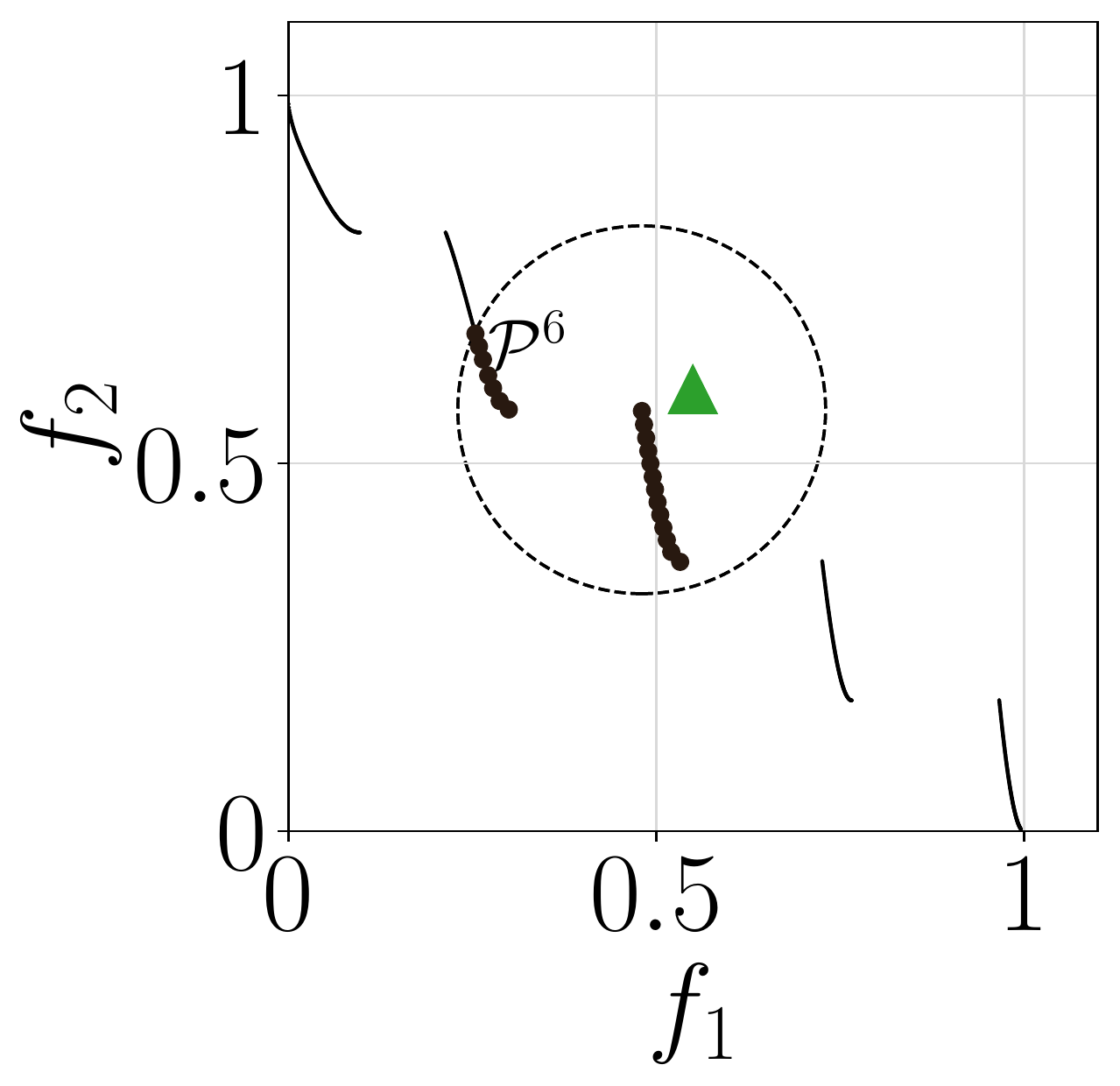}
    \includegraphics[width=0.31\textwidth]{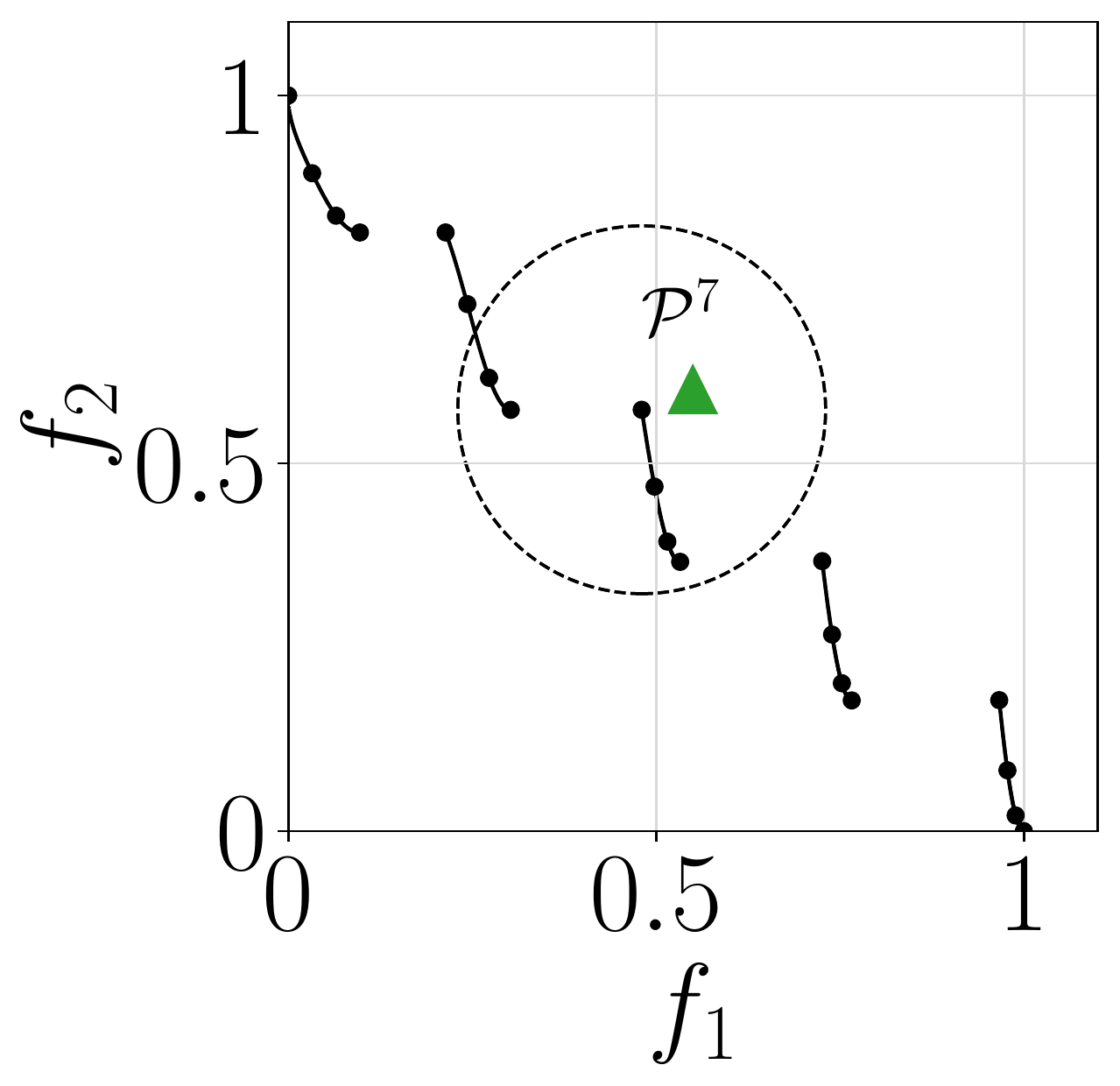}
  \caption{Distributions of the seven point sets on the PF of the ZDT3 problem, where the size of each point set is 20. \tabgreen{$\blacktriangle$} is the reference point $\mathbf{z}=(0.55,0.6)^\top$. Only in this experiment, the radius of the ROI $r$ is set to $0.25$ so that the ROI is disconnected. $\mathcal{P}^1, \ldots, \mathcal{P}^5$ are on the five subsets of the PF, respectively. Since all the 20 points in $\mathcal{P}^6$ are on the ROI, $\mathcal{P}^6$ is the best approximation. The 20 points in $\mathcal{P}^7$ are uniformly distributed on the whole PF.}
   \label{supfig:toy_zdt3}
\end{figure*}


\begin{table*}[t]
   \setlength{\tabcolsep}{2pt} 
  \centering
  \caption{\small Rankings of the 7 synthetic point sets on the two-objective ZDT3 problem by the 16 quality indicators when using $\mathbf{z}=(0.55, 0.6)^{\top}$ and $r=0.25$.}
  \label{suptab:rankings_psets_zdt3}
        {\scriptsize
\begin{tabularx}{72em}{|A|C|C|C|C|C|C|C|C|C|C|C|C|C|C|C|C|}
\hline
\rowcolor[rgb]{0.9, 0.9, 0.9} & MASF & MED & IGD-C & IGD-A & IGD-P & HV$_{\mathbf{z}}$ & PR & PMOD & IGD-CF & HV-CF & PMDA & R-IGD & R-HV & EH & HV & IGD \\
\hline
$\mathbf{P}_{1}$ & \cellcolor{c6}6 & \cellcolor{c6}6 & \cellcolor{c6}6 & \cellcolor{c6}6 & \cellcolor{c6}6 & \cellcolor{c5}5 & \cellcolor{c5}5 & \cellcolor{c5}5 & \cellcolor{c5}5 & \cellcolor{c5}5 & \cellcolor{c6}6 & \cellcolor{c6}6 & \cellcolor{c6}6 & \cellcolor{c6}6 & \cellcolor{c6}6 & \cellcolor{c7}7\\
$\mathbf{P}_{2}$ & \cellcolor{c4}4 & \cellcolor{c3}3 & \cellcolor{c4}4 & \cellcolor{c4}4 & \cellcolor{c4}4 & \cellcolor{c4}4 & \cellcolor{c3}3 & \cellcolor{c3}3 & \cellcolor{c4}4 & \cellcolor{c4}4 & \cellcolor{c3}3 & \cellcolor{c4}4 & \cellcolor{c4}4 & \cellcolor{c3}3 & \cellcolor{c3}3 & \cellcolor{c4}4\\
$\mathbf{P}_{3}$ & \cellcolor{c2}2 & \cellcolor{c1}1 & \cellcolor{c3}3 & \cellcolor{c3}3 & \cellcolor{c3}3 & \cellcolor{c3}3 & \cellcolor{c1}1 & \cellcolor{c2}2 & \cellcolor{c3}3 & \cellcolor{c3}3 & \cellcolor{c1}1 & \cellcolor{c2}2 & \cellcolor{c2}2 & \cellcolor{c1}1 & \cellcolor{c4}4 & \cellcolor{c3}3\\
$\mathbf{P}_{4}$ & \cellcolor{c5}5 & \cellcolor{c4}4 & \cellcolor{c5}5 & \cellcolor{c5}5 & \cellcolor{c5}5 & \cellcolor{c5}5 & \cellcolor{c5}5 & \cellcolor{c4}4 & \cellcolor{c5}5 & \cellcolor{c5}5 & \cellcolor{c4}4 & \cellcolor{c5}5 & \cellcolor{c5}5 & \cellcolor{c4}4 & \cellcolor{c5}5 & \cellcolor{c5}5\\
$\mathbf{P}_{5}$ & \cellcolor{c7}7 & \cellcolor{c7}7 & \cellcolor{c7}7 & \cellcolor{c7}7 & \cellcolor{c7}7 & \cellcolor{c5}5 & \cellcolor{c5}5 & \cellcolor{c6}6 & \cellcolor{c5}5 & \cellcolor{c5}5 & \cellcolor{c7}7 & \cellcolor{c7}7 & \cellcolor{c7}7 & \cellcolor{c7}7 & \cellcolor{c7}7 & \cellcolor{c6}6\\
$\mathbf{P}_{6}$ & \cellcolor{c1}1 & \cellcolor{c2}2 & \cellcolor{c1}1 & \cellcolor{c1}1 & \cellcolor{c1}1 & \cellcolor{c1}1 & \cellcolor{c2}2 & \cellcolor{c1}1 & \cellcolor{c1}1 & \cellcolor{c1}1 & \cellcolor{c2}2 & \cellcolor{c1}1 & \cellcolor{c1}1 & \cellcolor{c2}2 & \cellcolor{c2}2 & \cellcolor{c2}2\\
$\mathbf{P}_{7}$ & \cellcolor{c3}3 & \cellcolor{c5}5 & \cellcolor{c2}2 & \cellcolor{c2}2 & \cellcolor{c2}2 & \cellcolor{c2}2 & \cellcolor{c3}3 & \cellcolor{c7}7 & \cellcolor{c2}2 & \cellcolor{c2}2 & \cellcolor{c5}5 & \cellcolor{c3}3 & \cellcolor{c3}3 & \cellcolor{c5}5 & \cellcolor{c1}1 & \cellcolor{c1}1\\
\hline
\end{tabularx}
}
\end{table*}